%% file: main.tex
\documentclass[10pt,logo,copyright]{nvidiatechreport}
\input{packages.tex}
\input{common.tex}

\usepackage{tikz}
\usetikzlibrary{arrows.meta,positioning}
\usepackage{pifont}
\usepackage{diagbox}
\usepackage{pgfplots}
\pgfplotsset{compat=1.18}
\usepackage{colortbl}
\usepackage{graphicx}
\usepackage{wrapfig}

\newcommand{\cmark}{\ding{51}}

\usepackage[nameinlink]{cleveref}
\crefname{equation}{Eq.}{Eqs.}
\crefname{figure}{Fig.}{Figs.}
\crefname{section}{Sec.}{Sec.}
\crefname{appendix}{App.}{App.}
\crefname{table}{Tab.}{Tabs.}
\crefname{algorithm}{Algo}{Algo}
\crefname{thm}{Thm}{Thm}
\Crefname{thm}{Thm}{Thm}
\crefname{prop}{Prop}{Prop}

\definecolor{darkred}{rgb}{0.7, 0.0, 0.0}
\definecolor{rowours}{HTML}{E8F4E0}
\definecolor{columnours}{HTML}{E8F4E0}
\newcolumntype{Y}{>{\raggedright\arraybackslash}X}
\newcolumntype{C}{>{\centering\arraybackslash}X}
\newcolumntype{L}[1]{>{\raggedright\arraybackslash}p{#1}}
\input{report_support.tex}

\AddToHook{cmd/appendix/after}{%
  \counterwithin{figure}{section}%
  \counterwithin{table}{section}%
  \counterwithin{equation}{section}%
  \setcounter{figure}{0}%
  \setcounter{table}{0}%
  \setcounter{equation}{0}%
}

\newcommand{\crefnames}[3]{%
  \@for\next:=#1\do{%
    \expandafter\crefname\expandafter{\next}{#2}{#3}%
  }%
}

\title{Interactive World Simulation with Cosmos for Physical AI}

\author{NVIDIA\footnote{Contributors and acknowledgments are listed in Appendix~\ref{sec::contributors}.}}

\hypersetup{pdftitle={Interactive World Simulation with Cosmos for Physical AI},pdfauthor={NVIDIA},bookmarksdepth=3}

\input{sections/0__abstract.tex}

\begin{document}
\maketitle
\abscontent
\par\vspace{\baselineskip}

\begingroup{}
\begin{tcolorbox}[
  title={Open-Source Code},
  colframe=nvidiagreen!75,
  colback=black!2,
  colbacktitle=nvidiagreen!75,
  coltitle=white,
  fonttitle=\normalfont,
  boxrule=0.4pt,
  arc=0pt,
  left=8pt,right=8pt,top=4pt,bottom=4pt]
\small
\begin{tabularx}{\linewidth}{@{}L{4.7cm}Y@{}}
Cosmos & {\footnotesize\urlstyle{same}\href{https://github.com/nvidia/cosmos}{\nolinkurl{github.com/nvidia/cosmos}}}\\
Cosmos-Framework & {\footnotesize\urlstyle{same}\href{https://github.com/nvidia/cosmos-framework}{\nolinkurl{github.com/nvidia/cosmos-framework}}}\\
\end{tabularx}
\end{tcolorbox}
\ifhmode\par\fi\endgroup

\begingroup{}
\begin{tcolorbox}[
  title={Open-Weight Model Checkpoint},
  colframe=nvidiagreen!75,
  colback=black!2,
  colbacktitle=nvidiagreen!75,
  coltitle=white,
  fonttitle=\normalfont,
  boxrule=0.4pt,
  arc=0pt,
  left=8pt,right=8pt,top=4pt,bottom=4pt]
\small
\begin{tabularx}{\linewidth}{@{}L{4.7cm}Y@{}}
Cosmos3-Nano-Sim-Bimanual & {\footnotesize\urlstyle{same}\href{https://huggingface.co/nvidia/Cosmos3-Nano-Sim-Bimanual}{\nolinkurl{huggingface.co/nvidia/Cosmos3-Nano-Sim-Bimanual}}}\\
Cosmos3-Nano-Sim-Transfer & {\footnotesize\urlstyle{same}\href{https://huggingface.co/nvidia/Cosmos3-Nano-Sim-Transfer}{\nolinkurl{huggingface.co/nvidia/Cosmos3-Nano-Sim-Transfer}}}\\
Cosmos3-Nano-Sim-Auto & {\footnotesize\urlstyle{same}\href{https://huggingface.co/nvidia/Cosmos3-Nano-Sim-Auto}{\nolinkurl{huggingface.co/nvidia/Cosmos3-Nano-Sim-Auto}}}\\
Cosmos3-Nano-Transfer-Auto & {\footnotesize\urlstyle{same}\href{https://huggingface.co/nvidia/Cosmos3-Nano-Transfer-Auto}{\nolinkurl{huggingface.co/nvidia/Cosmos3-Nano-Transfer-Auto}}}\\
\end{tabularx}
\end{tcolorbox}
\ifhmode\par\fi\endgroup

\newpage
\setcounter{tocdepth}{2}
\begingroup
\makeatletter
\let\cosmostocsection\l@section
\renewcommand{\l@section}[2]{%
  \Needspace{4\baselineskip}%
  \cosmostocsection{#1}{#2}%
}
\makeatother
\tableofcontents
\endgroup
\newpage

\input{sections/1__introduction.tex}

\input{sections/2__architecture.tex}

\input{sections/3__training.tex}

\input{sections/4__inference.tex}
\input{sections/5__examples.tex}

\input{sections/6__related_work.tex}

\input{sections/7__conclusion.tex}

\clearpage
\appendix

\input{appendix/contributors.tex}

\input{appendix/bimanual_supplement.tex}

\input{appendix/transfer_supplement.tex}
\input{appendix/auto_appendix.tex}

\clearpage
\setcitestyle{numbers}
\renewcommand{\natexlab}[1]{#1}
\bibliographystyle{plainnat}

\input{main.bbl}
\end{document}

%% file: packages.tex
\usepackage[authoryear,round]{natbib}

\usepackage[utf8]{inputenc} %
\usepackage[T1]{fontenc}    %

\usepackage{parskip}        %
\usepackage{url}            %
\usepackage{booktabs}       %
\usepackage{amsfonts}       %
\usepackage{nicefrac}       %
\usepackage{microtype}      %
\usepackage{xcolor}         %
\usepackage[dvipsnames]{xcolor} %
\usepackage{graphicx}
\usepackage{animate}        %
\usepackage{subcaption}
\usepackage{tabularx}
\usepackage{makecell}
\usepackage{adjustbox}
\usepackage{setspace}
\newcolumntype{M}[1]{>{\centering\arraybackslash}m{#1}}
\usepackage{float}
\usepackage{placeins}
\renewcommand{\topfraction}{0.85}
\renewcommand{\bottomfraction}{0.7}
\renewcommand{\textfraction}{0.12}
\usepackage{tikz}
\usetikzlibrary{positioning,shapes,shapes.geometric,arrows, decorations.pathreplacing, backgrounds}
\usepackage{amsmath,amsfonts,bm, bbm,leftindex}
\usepackage{multirow}
\usepackage{comment}
\usepackage{gensymb}
\usepackage{lipsum}
\usetikzlibrary{arrows.meta, positioning, fit}
\usepackage[para]{threeparttable}
\usetikzlibrary{tikzmark}
\usepackage{tasks}

\usepackage[most]{tcolorbox}
\usepackage{fancyvrb}
\usepackage{fvextra}
\usepackage{dashrule}
\usepackage{nicematrix}

\usepackage{multicol}

\titleformat{\paragraph}[runin]{\normalfont\normalsize\bfseries}{\theparagraph}{1em}{#1}
\titlespacing*{\paragraph}{0pt}{2ex plus2pt minus1pt}{0.5em}

%% file: common.tex
\def\eqref#1{equation~\ref{#1}}

\def\1{\bm{1}}

\def\x{\mathbf{x}}
\def\y{\mathbf{y}}
\newcommand{\txtvar}{\mathbf{t}}

\newcommand{\txttok}[1]{\txtvar_{#1}}

\DeclareMathAlphabet{\mathsfit}{\encodingdefault}{\sfdefault}{m}{sl}
\SetMathAlphabet{\mathsfit}{bold}{\encodingdefault}{\sfdefault}{bx}{n}

\makeatletter
\let\save@mathaccent\mathaccent

\newcommand*\if@single[3]{%
  \setbox0\hbox{${\mathaccent"0362{#1}}^H$}%
  \setbox2\hbox{${\mathaccent"0362{\kern0pt#1}}^H$}%
  \ifdim\ht0=\ht2 #3\else #2\fi
  }
\newcommand*\rel@kern[1]{\kern#1\dimexpr\macc@kerna}
\newcommand*\widebar[1]{\@ifnextchar^{{\wide@bar{#1}{0}}}{\wide@bar{#1}{1}}}
\newcommand*\wide@bar[2]{\if@single{#1}{\wide@bar@{#1}{#2}{1}}{\wide@bar@{#1}{#2}{2}}}

\newcommand*\wide@bar@[3]{%
  \begingroup
  \def\mathaccent##1##2{%
    \let\mathaccent\save@mathaccent
    \if#32 \let\macc@nucleus\first@char \fi
    \setbox\z@\hbox{$\macc@style{\macc@nucleus}_{}$}%
    \setbox\tw@\hbox{$\macc@style{\macc@nucleus}{}_{}$}%
    \dimen@\wd\tw@
    \advance\dimen@-\wd\z@
    \divide\dimen@ 3
    \@tempdima\wd\tw@
    \advance\@tempdima-\scriptspace
    \divide\@tempdima 10
    \advance\dimen@-\@tempdima
    \ifdim\dimen@>\z@ \dimen@0pt\fi
    \rel@kern{0.6}\kern-\dimen@
    \if#31
      \overline{\rel@kern{-0.6}\kern\dimen@\macc@nucleus\rel@kern{0.4}\kern\dimen@}%
      \advance\dimen@0.4\dimexpr\macc@kerna
      \let\final@kern#2%
      \ifdim\dimen@<\z@ \let\final@kern1\fi
      \if\final@kern1 \kern-\dimen@\fi
    \else
      \overline{\rel@kern{-0.6}\kern\dimen@#1}%
    \fi
  }%
  \macc@depth\@ne
  \let\math@bgroup\@empty \let\math@egroup\macc@set@skewchar
  \mathsurround\z@ \frozen@everymath{\mathgroup\macc@group\relax}%
  \macc@set@skewchar\relax
  \let\mathaccentV\macc@nested@a
  \if#31
    \macc@nested@a\relax111{#1}%
  \else
    \def\gobble@till@marker##1\endmarker{}%
    \futurelet\first@char\gobble@till@marker#1\endmarker
    \ifcat\noexpand\first@char A\else
      \def\first@char{}%
    \fi
    \macc@nested@a\relax111{\first@char}%
  \fi
  \endgroup
}
\makeatother

\DeclareMathOperator{\sign}{sign}

%% file: report_support.tex
\definecolor{placeholderred}{HTML}{FF0000}
\definecolor{internalorange}{HTML}{FF8000}
\newcommand{\reporttablestyle}[1][6pt]{%
  \small
  \setlength{\tabcolsep}{#1}%
  \renewcommand{\arraystretch}{1.2}%
  \captionsetup{skip=8pt}%
}
\newsavebox{\reportdiagrambox}

\newcommand{\fitdiagram}[1]{%
  \sbox{\reportdiagrambox}{#1}%
  \ifdim\wd\reportdiagrambox>\linewidth
    \resizebox{\linewidth}{!}{\usebox{\reportdiagrambox}}%
  \else
    \usebox{\reportdiagrambox}%
  \fi
}
\newif\ifshowinternalnotes

  \showinternalnotesfalse



\makeatletter
\renewcommand{\@pnumwidth}{1.8em}

\makeatother

%% file: sections/0__abstract.tex
\begin{abstract}{
We present a shared framework that adapts Cosmos world foundation models for real-time interactive world simulation in Physical AI.
By streaming visual observations in response to actions and scene controls, the framework enables users and embodied agents to interact with evolving virtual environments.
It supports both predicting the visual consequences of actions and generating observations from physical state maintained by an external simulator.
We demonstrate this approach across camera navigation and robot manipulation, controllable video transfer, and multi-view autonomous vehicle (AV) simulation.
These examples provide a shared foundation for developing interactive virtual environments for Physical AI training and evaluation.
Our code and model checkpoints are available at \href{https://github.com/nvidia/cosmos}{{\urlstyle{same}\nolinkurl{github.com/nvidia/cosmos}}} and \href{https://huggingface.co/collections/nvidia/cosmos3-examples}{{\urlstyle{same}\nolinkurl{huggingface.co/collections/nvidia/cosmos3-examples}}}.
The project website is available at \href{https://research.nvidia.com/labs/cosmos-lab/cosmos-simulation}{{\urlstyle{same}\nolinkurl{research.nvidia.com/labs/cosmos-lab/cosmos-simulation}}}.
\hbox{}}\end{abstract}

%% file: sections/1__introduction.tex
\section{Introduction}
\label{sec::introduction}

\begingroup{}
Physical AI enables autonomous machines, such as robots and vehicles, to extend perception and reasoning into the physical world~\citep{cosmos_v1}.
Since their actions influence future observations, these systems must continually adapt to changing contexts: a robot adjusts its manipulation strategy for different object poses, while a vehicle responds to road and traffic conditions.
This continuous feedback makes perception, prediction, and action interdependent~\citep{cosmos3}, making Physical AI consequential for tasks like object handling, spatial navigation, and vehicle operation.
Building these capabilities requires exposing systems to diverse environments and interventions, while also testing how they respond when conditions change.
\ifhmode\par\fi\endgroup

\begingroup{}
Gathering all such experience solely in the real world is costly and can entail risks to people and equipment~\citep{cosmos_v1}.
Simulation offers a safe and controllable environment for training, testing, and debugging before deployment.
Cosmos world foundation models provide synthetic training data and generate observations for learning in virtual environments~\citep{cosmos_v1,cosmos3}.
Learned visual priors expand the range of appearances and scene conditions in these environments, while structured controls retain grounding in the external physical state.
As a result, world simulation bridges the gap between data available for learning and the environments needed to evaluate policy behavior.
\ifhmode\par\fi\endgroup

\begingroup{}
Interactive simulation generates observations incrementally as new controls arrive from a user, a policy, or an external simulator~\citep{gao2026dreamdojo,omnidreams2026}.
Evaluating this behavior requires testing how changed controls affect subsequent observations and how well scene changes persist over time.
To enable live policy or human interaction at real-world timescales, observations must be returned within the required response time.
Achieving this in neural world simulation demands coupling visual generation with control fidelity, retention of state changes, and responsive execution.
These requirements motivate transforming a world foundation model into an interactive simulator, while showing that it meets downstream needs still requires independent validation.
\ifhmode\par\fi\endgroup

\begingroup{}
We introduce a shared post-training framework for converting Cosmos world foundation models into interactive simulators.
Built on Cosmos 3~\citep{cosmos3}, the framework adapts broad visual knowledge to predicting how scenes evolve under supplied actions or scene controls.
Its purpose is to make that knowledge usable throughout an ongoing interaction, so a user, policy, or external simulator can shape the experience as it unfolds.
This supports both learning visual dynamics from recorded experience and generating observations guided by an external simulator's physical state.
\ifhmode\par\fi\endgroup

\begingroup{}
The framework combines causal training and distillation with streaming execution in a shared recipe for interactive simulation~\citep{zheng2026causalrcm}.
Post-training teaches the model to follow domain-specific controls and continue from its own generated observations~\citep{huang2025selfforcing}.
Distillation reduces the computation needed to produce those observations, while streaming execution supports repeated interaction as new inputs become available.
Together, these components address visual quality, control fidelity, and timely responses within a common framework.
The shared recipe provides a reusable starting point for specialized simulators, with each domain contributing its own controls, data, and evaluation requirements.
\ifhmode\par\fi\endgroup

\begingroup{}
We ground this approach in three post-trained Cosmos3-Nano model examples:
\begin{enumerate}
\item Cosmos3-Nano-Sim-Bimanual predicts future visual dynamics from a streaming sequence of camera navigation and robot manipulation actions.
\item Cosmos3-Nano-Sim-Transfer supports single-view video transfer based on spatial controls, while the state can be maintained externally by a simulator.
\item Cosmos3-Nano-Sim-Auto supports synchronized multi-view AV simulation with LiDAR.
\end{enumerate}
All three examples follow the shared procedure, each adapting control representations, data preparation, and modeling choices for its domain.
They connect robot interaction, simulator-guided video transfer, and AV scene generation through one post-training framework.
Quality, control fidelity, and runtime are evaluated under each example's stated protocols, with policy-driven simulator utility requiring separate closed-loop validation.
\Cref{fig::simulation_overview} summarizes this shared foundation and its three post-trained specializations.
\ifhmode\par\fi\endgroup

\input{figures/introduction/simulation_overview.tex}

\begingroup{}
The technical report is organized as follows.
\Cref{sec::shared_architecture} introduces the shared model architecture and its support for simulation controls and causal generation.
\Cref{sec::training} presents the post-training recipe for causal training and distillation on the foundation model.
\Cref{sec::inference_optimization} describes streaming inference optimizations for efficient execution.
\Cref{sec::bimanual,sec::transfer,sec::auto} describe the three model examples, covering their respective architectures, data preparation, training procedures, and evaluation results.
\Cref{sec::related_work} discusses related work and \cref{sec::conclusion} concludes the paper.
\ifhmode\par\fi\endgroup

%% file: figures/introduction/simulation_overview.tex
\begingroup{}
\begin{figure}[!t]
\centering
\input{figures/diagrams/simulation_overview.tex}
\caption{\textbf{From Cosmos 3 to interactive world simulation.}
A shared post-training recipe adapts Cosmos3-Nano from the Cosmos 3 family~\citep{cosmos3} into three example models for camera navigation and robot manipulation, single-view video transfer, and multi-view autonomous vehicle (AV) simulation.
Each specialized model incorporates domain-specific controls tailored to its application and trained on datasets reflecting the types of interactions and physical dynamics relevant to that domain.}
\label{fig::simulation_overview}
\end{figure}
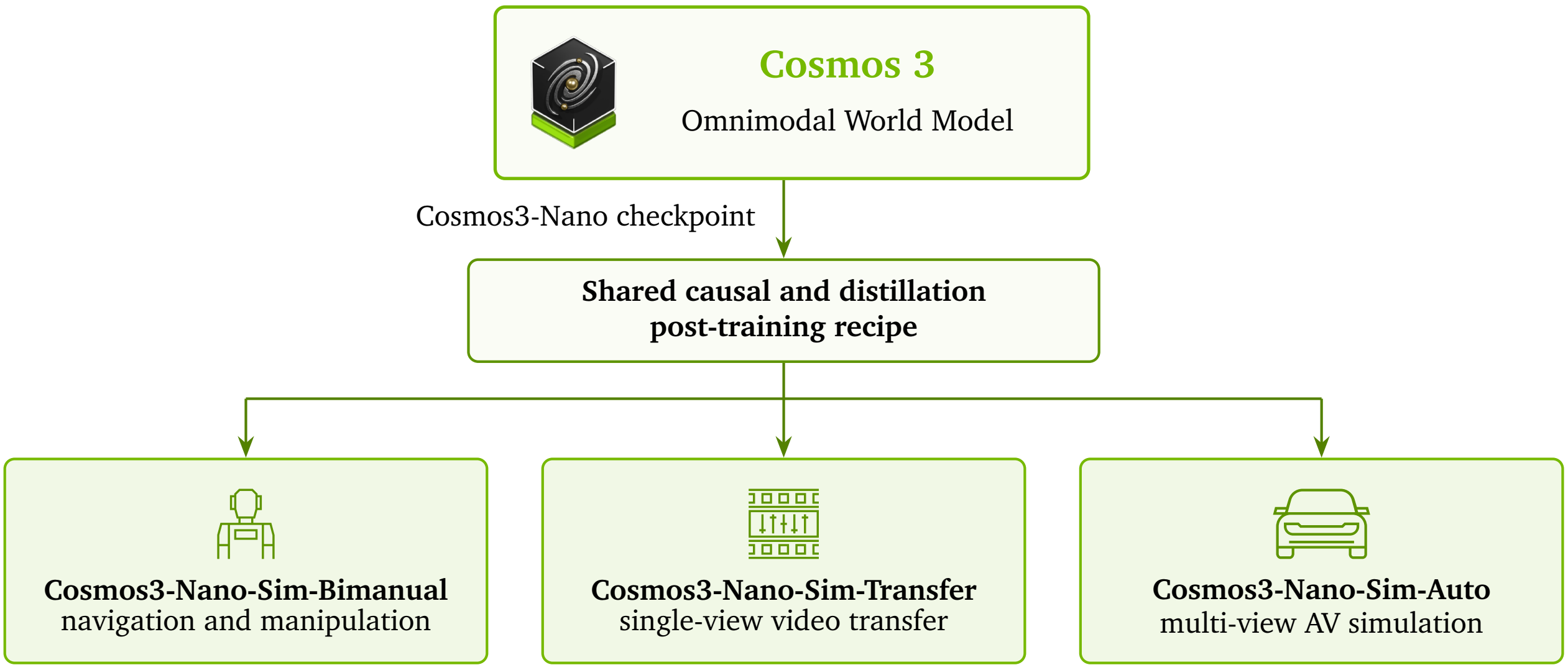
\ifhmode\par\fi\endgroup

%% file: figures/diagrams/simulation_overview.tex
\begingroup
\input{figures/diagrams/architecture_styles.tex}
\definecolor{overviewgreen}{HTML}{76B900}
\colorlet{simactionedge}{overviewgreen!80!black}
\input{figures/introduction/icons/nvidia_icons.tex}
\fitdiagram{%
\begin{tikzpicture}[
  font=\fontfamily{bch}\selectfont\small,
  >=Stealth,
  every node/.append style={text=.},
  overview recipe/.style={draw=overviewgreen!80!black,fill=overviewgreen!4,
    rounded corners=3pt,line width=0.8pt,
    minimum width=6.8cm,minimum height=1.10cm,align=center,
    inner xsep=0.30cm,inner ysep=0.18cm,
    font=\fontfamily{bch}\selectfont\small\bfseries},
  overview example/.style={draw=overviewgreen,fill=overviewgreen!10,
    rounded corners=3pt,line width=0.8pt,
    minimum width=5.2cm,minimum height=2.20cm,inner sep=0pt},
  overview title/.style={font=\fontfamily{bch}\selectfont\small\bfseries,
    align=center,inner sep=0pt},
  overview detail/.style={font=\fontfamily{bch}\selectfont\small,
    align=center,inner sep=0pt},
  overview connector/.style={draw=overviewgreen!70!black,line width=0.8pt},
  overview arrow/.style={overview connector,->}
]

\node[inner sep=0pt,outer sep=0pt] (foundation) at (0,0)
  {\input{figures/diagrams/cosmos3_model_box.tex}};

\node[overview recipe] (recipe) at (0,-2.35)
  {Shared causal and distillation\\post-training recipe};
\draw[overview arrow] (foundation.south) -- (recipe.north);
\node[overview detail,anchor=east] at (-0.30,-1.35) {Cosmos3-Nano checkpoint};
\draw[overview connector] (recipe.south) -- (0,-3.30);
\draw[overview connector] (-5.8,-3.30) -- (5.8,-3.30);

\node[overview example] (bimanual) at (-5.8,-5.05) {};
\pic at (-5.8,-4.65) {overview manipulation};
\node[overview title] at (-5.8,-5.35) {Cosmos3-Nano-Sim-Bimanual};
\node[overview detail] at (-5.8,-5.71) {navigation and manipulation};

\node[overview example] (transfer) at (0,-5.05) {};
\pic at (0,-4.65) {overview transfer};
\node[overview title] at (0,-5.35) {Cosmos3-Nano-Sim-Transfer};
\node[overview detail] at (0,-5.71) {single-view video transfer};

\node[overview example] (auto) at (5.8,-5.05) {};
\pic[scale=1.2222222222] at (5.8,-4.65) {overview av};
\node[overview title] at (5.8,-5.35) {Cosmos3-Nano-Sim-Auto};
\node[overview detail] at (5.8,-5.71) {multi-view AV simulation};

\draw[overview arrow] (-5.8,-3.30) -- (bimanual.north);
\draw[overview arrow] (0,-3.30) -- (transfer.north);
\draw[overview arrow] (5.8,-3.30) -- (auto.north);
\end{tikzpicture}}
\endgroup

%% file: figures/diagrams/cosmos3_model_box.tex
\begingroup
\definecolor{cosmoscardgreen}{HTML}{76B900}
\colorlet{cosmoscardfill}{cosmoscardgreen!4}
\begin{tikzpicture}
\node[draw=cosmoscardgreen,fill=cosmoscardfill,
  rounded corners=3pt,line width=1.0pt,
  minimum width=6.4cm,minimum height=1.85cm,inner sep=0pt] at (0,0) {};
\node[inner sep=0pt] at (-2.35,0)
  {\includegraphics[height=1.45cm]{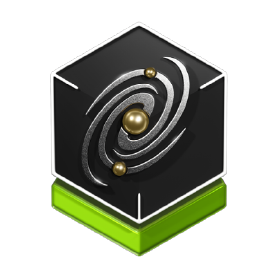}};
\node[font=\fontfamily{bch}\fontsize{12.2}{14}\selectfont\bfseries,
  text=cosmoscardgreen,inner sep=0pt] at (0.60,0.31) {Cosmos 3};
\node[font=\fontfamily{bch}\fontsize{9.2}{11}\selectfont,
  text=black,inner sep=0pt] at (0.60,-0.29) {Omnimodal World Model};
\end{tikzpicture}
\endgroup

%% file: sections/2__architecture.tex
\FloatBarrier

\section{Model Architecture}
\label{sec::shared_architecture}

\begingroup{}
Our framework builds directly on Cosmos 3, retaining its reasoner and generator towers, unified representation space, and positional embedding formulation~\citep{cosmos3}.
To support interactive simulation, the architecture must generate observations incrementally as conditioning inputs arrive.
The primary shared architectural modification is the introduction of temporal-causal attention across video chunks.
This temporal-causal attention ensures that, as each new control becomes available, the simulator predicts subsequent observations autoregressively, without access to future target chunks.
During generation, each chunk is produced through iterative denoising, conditioned on both the available observation history and any provided control signals, including language, actions, source video, or structured visual controls.
\ifhmode\par\fi\endgroup

\subsection{Cosmos 3 Backbone}
\begingroup{}
All three example models build directly on Cosmos 3 checkpoints, preserving their multimodal backbone, representation space, and architectural foundations.
At the core of Cosmos 3 is a Mixture-of-Transformers (MoT) design, which organizes processing into two complementary towers: the reasoner and the generator~\citep{deng2025bagel,cosmos3}.
Modality-specific encoders and projections first embed each input type (\eg language, videos, actions) into a shared representation space.
Within this architecture, the reasoner tower specializes in autoregressive next-token prediction for language, while the generator tower handles non-language modalities through iterative denoising.
\ifhmode\par\fi\endgroup

\paragraph{MoT towers and joint attention.}
\begingroup{}
In each MoT layer, the reasoner and generator towers maintain separate sets of parameters for layer normalization, query/key/value projections, and feed-forward networks.
The simulator architecture routes the text prompt into the reasoner tower as the autoregressive subsequence.
Visual, control, and denoising-related tokens (including clean conditions, denoised history, and noisy prediction targets) are routed into the generator tower as the diffusion subsequence.
Generator queries attend to the entire diffusion subsequence and, via dual-stream joint attention, can also attend to keys and values from the reasoner, while language tokens in the reasoner retain standard causal self-attention.
During causal conversion, the generator tower replaces full temporal visibility with temporal-causal attention masking, governed by token role, physical time, and the availability of conditions (\mbox{\cref{sec::token_packing,sec::model_architecture}}).
\ifhmode\par\fi\endgroup

\paragraph{Visual encoding and generation.}
{
Visual features are extracted by a vision transformer (ViT)~\citep{dosovitskiy2021vit} and fed to the reasoner tower, while the generator tower receives latent representations from a causal video variational autoencoder (VAE)~\citep{cosmos3}.
A causal RGB tokenizer achieves $4\times$ temporal compression, mapping $1+4n$ input frames to $1+n$ latent time steps~\citep{wan2025}.
The tokenizer's first latent time step represents the initial RGB frame, and each subsequent step adds 4 RGB frames.
Learned projections merge latent patches into the unified representation space.
Sampling integrates the output projection's rectified flow velocities into the latents~\citep{liu2023rectifiedflow}, and the tokenizer decoder reconstructs RGB observations.
\hbox{}}
{For autonomous driving, we extend this design to jointly generate synchronized multi-view RGB videos and LiDAR; see \cref{sec::multiview_attention} for details.}

\paragraph{Multimodal positional embedding.}
\begingroup{}
Cosmos 3 applies 3D multimodal RoPE (MRoPE)~\citep{bai2025qwen3} to its embeddings, dividing the rotary channels of each attention head among temporal, height, and width coordinates $(t^{\text{pos}},h,w)$~\citep{cosmos3}.
Within each latent time, both source and target visual tokens share the temporal coordinate $t^{\text{pos}}$, with their spatial coordinates tiling a grid that resets at every frame.
Language tokens are encoded with increasing, identical values along all three axes, while action tokens carry temporal positional information and set $h=w=0$.
A large temporal gap (15,000) separates the autoregressive and diffusion subsequences, and both conditioning and generated visual streams initiate from a common temporal origin.
These instantiations inherit the foundation model's MRoPE formulation and maintain the same coordinate offsets throughout causal conversion.
\ifhmode\par\fi\endgroup

\paragraph{Conditioning and control.}
\begingroup{}
Cosmos 3 accommodates both language and control streams as direct conditions for video generation~\citep{cosmos3}.
Control tokens are concatenated alongside noisy target video tokens and remain clean throughout the denoising of targets.
For video transfer, the video tokenizer can process source video or structured visual controls (\eg depth, segmentation, edges, or blurred videos) to steer generation.
In action-conditioned video generation, action tokens express relative camera, end-effector motions, or other transformations.
Domain-specific projections adapt action representations of varying dimension into the unified representation~\citep{zheng2026xvla}.
\ifhmode\par\fi\endgroup

\subsection{Multimodal Token Arrangement}
\label{sec::token_packing}

\begingroup{}
The framework groups visual targets and aligned clean controls into \emph{chunks} for temporal-causal attention and autoregressive video generation.
A chunk contains the spatial tokens at one or more consecutive latent video time steps together with their corresponding control tokens.
For multi-view generation, it includes the aligned visual groups from the participating views.
This prediction unit is distinct from the tokenizer's internal units for encoding or decoding video.
\ifhmode\par\fi\endgroup

\begingroup{}
Each prediction uses generation history and the provided control streams.
For action-conditioned video generation, controls specify the desired actions and transformations that determine the upcoming predicted visual outcomes.
For video transfer tasks, observations for the target interval are guided by either a source video or structured visual controls.
Both configurations may include language tokens as conditions, follow the same next-chunk prediction structure, and maintain distinct token roles and attention rules.
\ifhmode\par\fi\endgroup

\paragraph{Multimodal inputs for chunk prediction.}
\begingroup{}
Partition the $L$ target latent time steps into ordered, disjoint, contiguous index sets $\mathcal{B}_1,\ldots,\mathcal{B}_J$.
Chunk $j$ spans $N_j=|\mathcal{B}_j|\geq1$ target latent time steps, with $\sum_{j=1}^{J}N_j=L$.
After the initial tokenizer step, these $N_j$ latent time steps correspond to $4N_j$ new RGB frames per view.
Its visual part $\mathcal{V}_j=\mathcal{V}_{\mathcal{B}_j}$ contains all spatial tokens at those latent time steps, and all of its target visual tokens are jointly denoised.
At diffusion step $\tau$, the noisy tokens for chunk $j$ are denoted $\widetilde{\mathcal{V}}_j^\tau$.\footnote{Plain letters denote scalars and indices.
Bold lowercase letters are vectors, bold uppercase letters are matrices, and calligraphic uppercase letters are collections.
Token sequences are ordered and may include repeated entries.
Discrete token IDs use scalar labels.}
The aligned controls $\mathcal{U}_j$ belong to the same chunk, and their token types and lengths depend on the post-trained model.
The current generator chunk and its logical prediction input are
\begin{equation}
 \begin{aligned}
 \mathcal{G}_j^\tau&=[\mathcal{U}_j,\widetilde{\mathcal{V}}_j^\tau],\\
 \mathcal{S}_j(\tau)&=[\mathcal{T}\mid\mathcal{H}_j,\mathcal{G}_j^\tau],
 \end{aligned}
 \label{eq::token_layouts}
\end{equation}
where $\mathcal{T}$ denotes available language tokens and $\mathcal{H}_j$ denotes retained clean generator history.
Only $\widetilde{\mathcal{V}}_j^\tau$ is noisy and predicted.
The controls remain clean and are excluded from the target loss, even though they are processed inside the same chunk.
\ifhmode\par\fi\endgroup

\begingroup{}
Autoregression between video chunks routes visual history and targets through the generator, with new chunks produced by denoising.
The notation describes which token sets are included in prediction, but not their memory layout.
At inference, the clean history $\mathcal{H}_j$ retained before predicting chunk $j$ is represented by its K/V cache $\mathcal{H}^{\text{KV}}_j$.
These history tokens do not need to be physically concatenated with the current input.
\ifhmode\par\fi\endgroup

\begingroup{}
Within the input, only the current target chunk is noisy.
The clean generator history $\mathcal{H}_j$ may include an initial prefix $\mathcal{H}_0$, previously used clean control groups $\mathcal{U}_k$, and denoised vision chunks $\widehat{\mathcal{V}}_k$ for $k<j$.
When sampling starts from provided controls, the initial prefix can be empty or may contain initial observations from the target sequence.
Cached controls remain as conditioning information, not part of the generative targets.
\ifhmode\par\fi\endgroup

\begingroup{}
Once denoising produces $\widehat{\mathcal{V}}_j$, the completed clean chunk is $\widehat{\mathcal{G}}_j=[\mathcal{U}_j,\widehat{\mathcal{V}}_j]$.
The runtime constructs its clean layerwise K/V and commits the completed history before predicting the next chunk (\cref{eq::streaming_clean_commit}).
If language tokens are unchanged and no history eviction or reselection occurs, the next logical prediction input is given by
\begin{equation}
 \begin{aligned}
  \mathcal{S}_{j+1}(\tau)
  &= [\mathcal{T}\mid\mathcal{H}_{j+1},\mathcal{U}_{j+1},\widetilde{\mathcal{V}}_{j+1}^{\tau}]\\
  &= [\mathcal{T}\mid
      \underbrace{\mathcal{H}_j,\mathcal{U}_j,\widehat{\mathcal{V}}_j}_{\mathcal{H}_{j+1}},
      \mathcal{U}_{j+1},\widetilde{\mathcal{V}}_{j+1}^{\tau}].
 \end{aligned}
 \label{eq::next_chunk_token_layouts}
\end{equation}
With a limited-size K/V cache, adding new entries may require evicting older history tokens to keep the retained history within capacity.
For the single-view action-conditioned path, clean K/V are refreshed in latent-time order within the completed chunk.
Multi-view execution follows the clean-history visibility rules configured in~\cref{sec::multiview_attention}.
\ifhmode\par\fi\endgroup

\begingroup{}
\begin{figure}[!t]
\centering
\input{figures/diagrams/token_packing.tex}
\caption{\textbf{Autoregressive video generation with streaming controls.}
(a) Each generator chunk groups clean controls $\mathcal{U}_k$ with the corresponding visual tokens.
The history $\mathcal{H}_j$ comprises previously denoised chunks $\widehat{\mathcal{V}}_k$ and the retained controls.
The current chunk $\widetilde{\mathcal{V}}_j^\tau$ is progressively denoised with clean controls held constant.
(b) The action-based example shows 4 individual clean action tokens per latent interval.
For video transfer, aligned source latent frames form the clean control group $\mathcal{U}_j$ for the target chunk's interval.
Source tokens entering the cache keep their conditioning role for future predictions.}
\label{fig::token_packing}
\end{figure}
\ifhmode\par\fi\endgroup

\paragraph{Streaming control inputs.}
\begingroup{}
In autoregressive video generation, each target chunk $j$ pairs with a group of clean control tokens $\mathcal{U}_j$ matched to the chunk's physical interval.
Each generator chunk $\mathcal{G}_j^\tau$ contains both its clean controls $\mathcal{U}_j$ and noisy visual targets $\widetilde{\mathcal{V}}_j^\tau$.
The next control group belongs to the following chunk and is introduced for that prediction.
Within a chunk, controls remain fixed and clean while denoising proceeds, and they continue serving as conditions once committed to history.
All required controls must be available before sampling the chunk starts.
During deployment, a policy or external simulator must therefore supply the corresponding action or control group before the model jointly denoises its visual part.
If these inputs arrive incrementally, larger chunks can require a longer wait to collect the controls for their physical interval.
The logical meaning of token roles and temporal association is unaffected by how data is laid out in memory.
\ifhmode\par\fi\endgroup

\begingroup{}
Figure~\ref{fig::token_packing} illustrates the control-group arrangement for both actions and source video.
For action-conditioned generation, $\mathcal{U}_j$ contains the transition actions corresponding to the current predicted chunk.
For video transfer, clean source tokens $\mathcal{V}^{\text{src}}$ supply observations or structured controls, with alignment to the target's timestamp enforced, even if source and target use the same tokenizer and projection.
This scheme can be extended to extra input signals, such as audio.
Source tokens may be pre-encoded or provided incrementally, as long as all tokens for the interval are available before target denoising.
After each prediction, the corresponding generated chunk and any new control tokens are appended to the history for subsequent predictions.
Section~\ref{sec::training} details the specific training targets and roles for each group.
\ifhmode\par\fi\endgroup

\paragraph{Temporal alignment.}
\label{sec::streaming_positions}
\begingroup{}
Each control group is temporally aligned with its associated target chunk: actions carry their transition times, source video matches target timestamps, and spatial controls are matched to their target frames.
FPS modulation transforms physical time to rotary temporal coordinates.
For video transfer, source video is encoded at the target's frame rate and temporal compression, ensuring matched temporal increments.
Since actions are recorded at 1 per RGB frame with a $4\times$ temporal compression, every step between latent vision frames covers 4 transition actions.
Actions thus use $1/4$ of the rotary temporal increment used for vision tokens, resulting in the last action sample sharing a temporal coordinate with the corresponding vision latent frame.
Action groups for chunks spanning multiple latent frames are collected accordingly.
The first observation in the sequence lacks a preceding real action transition.
\ifhmode\par\fi\endgroup

\paragraph{Sequence packing.}
\begingroup{}
Cosmos 3 reduces padding by packing recordings with different durations and resolutions into shared token buffers~\citep{cosmos3}.
Each recording always retains its own temporal origin, token roles, and attention domain, regardless of where tokens are packed within shared buffers.
To ensure recording isolation, a boolean attention mask is applied for every query and key index.
This mask ensures that only tokens belonging to the same recording session, at valid (non-padding) positions, and allowed by cross-tower, temporal, and token-role rules are able to attend to each other.
Padding is excluded from both attention and loss computation.
Device partitions do not define attention boundaries: tokens from the same recording may exchange information across devices, and tokens from different recordings remain isolated—even if packed on the same device.
A retained prefix must preserve its source recording's identity and temporal coordinates.
\ifhmode\par\fi\endgroup

\subsection{Temporal-causal Attention}
\label{sec::model_architecture}

\begingroup{}
Cosmos 3 uses full self-attention in the generator tower to denoise entire video clips.
Our post-training recipe replaces the generator's bidirectional temporal attention with a block-causal pattern to enable autoregressive video generation.
Each attention block corresponds to one generator chunk $\mathcal{G}_j^\tau$, containing aligned clean controls and the visual targets for $N_j\geq1$ latent time steps.
The temporal mask operates on chunks, and the visual targets within the current chunk are jointly denoised.
The action-conditioned layout permits mutual attention between the current clean actions and visual targets.
Video-transfer layouts can impose additional restrictions on control queries while retaining the controls inside the same logical chunk.
Teacher forcing additionally separates the clean-reference and noisy-target branches as described in \cref{sec::teacher_forcing}.
\ifhmode\par\fi\endgroup

\begingroup{}
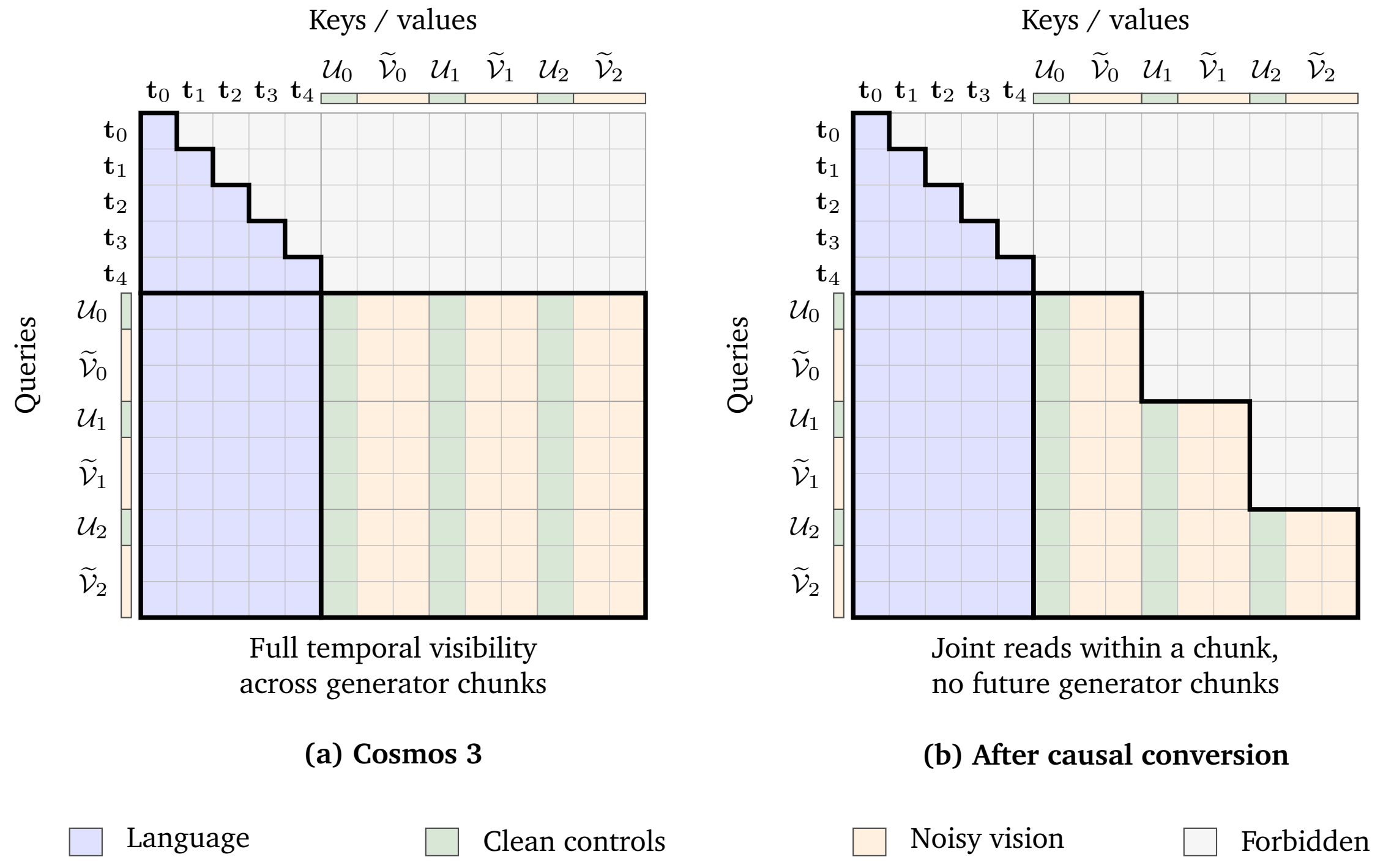
\begin{figure}[!t]
\centering
\input{figures/diagrams/attention_masks.tex}
\caption{\textbf{Temporal-causal attention for action-conditioned video generation.}
Rows represent queries and columns stand for keys/values.
Each generator chunk contains clean controls $\mathcal{U}_k$ and visual tokens $\widetilde{\mathcal{V}}_k$, with chunk indices starting at $k=0$.
When conditioning on actions, $\mathcal{U}_0$ is a null group and the first vision group is the clean frame $\mathcal{V}_0$.
Language uses causal self-attention, while generator queries access reasoner keys/values, but not vice versa.
Token counts are schematic only.
(a) In Cosmos 3, every generator chunk can read from every other.
(b) After causal conversion, reads are allowed only within the current block and to earlier ones.
Video-transfer and multi-view layouts can further restrict token-role visibility.}
\label{fig::attention_masks}
\end{figure}
\ifhmode\par\fi\endgroup

\paragraph{Generator visibility.}
\begingroup{}
The generator's attention mask specifies which tokens each query is allowed to access while generating video.
Within the current noisy target chunk $\widetilde{\mathcal{V}}_j^\tau$, queries may attend to all tokens in that chunk, enabling joint denoising.
Queries can also attend to the retained clean history $\mathcal{H}_j$, which contains previously denoised frames and other finalized context.
No attention is allowed to tokens from future target chunks.
Supplied source video that coincides with the current interval remains accessible, but captions and cached conditioning must restrict visibility to prevent leaking information from future targets.
\Cref{fig::attention_masks} visually outlines which reads are allowed and which are blocked.
In some cases, more specialized token-role masks are enforced, such as for cross-view attention (\cref{sec::multiview_attention}).
Details on constructing these masks during training, particularly for teacher forcing, are given in \cref{sec::teacher_forcing}.
\ifhmode\par\fi\endgroup

\paragraph{Next-chunk factorization.}
\begingroup{}
Let $\mathbf{z}_{1:L}$ represent the $L$ target latents for generated vision frames, omitting the initial supplied observations.
Divide the frame indices into $J$ prediction chunk sets $\mathcal{B}_1,\ldots,\mathcal{B}_J$.
Define $\mathcal{I}_j$ as the conditioning accepted before generation begins for chunk $j$.
Only this conditioning is exposed to its queries via the attention mask.
The conditioning $\mathcal{I}_j$ comprises the accepted control groups, any supplied initial observations, initialization metadata, physical timestamps, and available language.
With a predetermined sequence of such conditions, autoregressive generation factorizes as:
\begin{equation}
 p_\theta(\mathbf{z}_{1:L}\mid\mathcal{I}_{1:J})
 =\prod_{j=1}^{J}p_\theta\!\left(\mathbf{z}_{\mathcal{B}_j}\mid
 \mathbf{z}_{\mathcal{B}_{<j}},\mathcal{I}_j\right).
 \label{eq::causal_factorization}
\end{equation}
Here, $\mathbf{z}_{\mathcal{B}_{<j}}$ indicates all previously generated targets.
The initial observed frame establishes the clean history $\mathcal{H}_0$ and counts as a conditioning input, not a generated output.
If new inputs arrive during denoising, they only impact subsequent chunks unless the current prediction is interrupted or redone.
Once a chunk prediction completes, its observations are appended to the clean history for the prediction of chunks that follow (\cref{sec::streaming_execution}).
Temporal causality is thus enforced between chunks, but the frames within a chunk interact freely in joint denoising.
The chunk size sets the granularity of this interaction: smaller chunks permit the model to respond to new interactive controls more often, while larger chunks capture longer visual intervals during joint modeling and may enable higher visual quality.
\ifhmode\par\fi\endgroup

\begingroup{}
This generative factorization assumes conditioning inputs are predetermined and presented from outside the system, not arising from closed-loop feedback.
Under closed-loop simulation, a policy would select new controls based on the model's previous outputs, so the conditioning sequence would depend on the simulated rollout.
Running with a prerecorded sequence of controls covers conditional generation but does not exercise feedback from policy-driven interaction.
\ifhmode\par\fi\endgroup

%% file: figures/diagrams/token_packing.tex
\begingroup
\input{figures/diagrams/architecture_styles.tex}

\newcommand{\simpackinggrid}[4]{%
  \begin{scope}[shift={(#1,#2)}]
    \filldraw[#3] (0,0) rectangle (1.5,0.75);
    \foreach \division in {1,2} {
      \draw[sim grid] ({0.5*\division},0) -- ({0.5*\division},0.75);
      \draw[sim grid] (0,{0.25*\division}) -- (1.5,{0.25*\division});
    }
    \node[anchor=base,inner sep=0pt] at (0.75,1.0) {#4};
  \end{scope}%
}
\newcommand{\simpackingrgb}[3]{%
  \begin{scope}[shift={(#1,#2)}]
    \draw[#3] (-0.48,-0.22) -- (-0.48,-0.33) -- (0.63,-0.33)
      -- (0.63,0.22) -- (0.52,0.22);
    \draw[#3,fill=white] (-0.58,-0.22) rectangle (0.52,0.33);
    \node at (-0.03,0.055) {RGB};
  \end{scope}%
}
\fitdiagram{%
\begin{tikzpicture}[font=\fontfamily{bch}\fontsize{9}{10.5}\selectfont,>=Stealth,
  every node/.append style={text=.}]
\node[anchor=north,align=center,font=\fontfamily{bch}\fontsize{9}{10.5}\selectfont\bfseries]
  at (7.6,-2.8) {(a) Autoregressive video generation at inference step $j$};

\draw[sim flow] (4.7,3.5) -- node[above] {Advance across chunks} (15.15,3.5);
\node[align=center] at (2.75,3.35) {Optional initial\\observation};
\node at (7.8,3.05) {Decoded RGB outputs};
\node at (13.65,3.05) {RGB once complete};
\simpackingrgb{2.75}{2.4}{draw=simcleanedge}
\simpackingrgb{5.75}{2.4}{draw=simcleanedge}
\simpackingrgb{9.9}{2.4}{draw=simcleanedge}
\simpackingrgb{13.65}{2.4}{draw=simcleanedge,dashed}

\node[sim text,minimum width=1.4cm,minimum height=0.75cm]
  at (0.8,0.3) {$\mathcal{T}$};
\node at (0.8,-0.45) {Language};
\draw[semithick] (1.75,-0.15) -- (1.75,1.05);
\simpackinggrid{2.0}{-0.075}{fill=simcleanfill,draw=simcleanedge}{$\mathcal{H}_0$}
\node[sim action,minimum width=1.0cm,minimum height=0.75cm]
  at (4.25,0.3) {$\mathcal{U}_1$};
\simpackinggrid{5.0}{-0.075}{fill=simcleanfill,draw=simcleanedge}{$\widehat{\mathcal{V}}_1$}
\node at (7.1,0.3) {$\cdots$};
\node[sim action,minimum width=1.15cm,minimum height=0.75cm]
  at (8.35,0.3) {$\mathcal{U}_{j-1}$};
\simpackinggrid{9.15}{-0.075}{fill=simcleanfill,draw=simcleanedge}{$\widehat{\mathcal{V}}_{j-1}$}
\node[sim action,minimum width=1.0cm,minimum height=0.75cm]
  at (11.75,0.3) {$\mathcal{U}_j$};
\simpackinggrid{12.9}{-0.075}{fill=simnoisyfill,draw=simnoisyedge}{$\widetilde{\mathcal{V}}_j^\tau$}

\draw[draw=simgrayedge,rounded corners=2pt] (3.6,-0.25) rectangle (6.65,1.35);
\draw[draw=simgrayedge,rounded corners=2pt] (7.6,-0.25) rectangle (10.85,1.35);
\draw[draw=simgrayedge,rounded corners=2pt] (11.1,-0.25) rectangle (14.75,1.35);
\node[anchor=base,inner sep=0pt] at (4.25,1.55) {$\widehat{\mathcal{G}}_1$};
\node[anchor=base,inner sep=0pt] at (8.3,1.55) {$\widehat{\mathcal{G}}_{j-1}$};
\node[anchor=base,inner sep=0pt] at (11.75,1.55) {$\mathcal{G}_j^\tau$};

\draw[sim flow,draw=simcleanedge] (2.75,2.02) -- (2.75,1.3);
\foreach \visionx in {5.75,9.9} {
  \draw[sim flow,draw=simcleanedge] (\visionx,1.35) -- (\visionx,2.02);
}
\draw[sim flow,draw=simcleanedge,dashed] (13.65,1.35) -- (13.65,2.02);

\draw[draw=simgrayedge] (2.0,-0.45) -- (2.0,-0.65)
  -- (10.65,-0.65) -- (10.65,-0.45);
\node[align=center] at (6.3,-1.12)
  {Retained clean generator\\history $\mathcal{H}_j$ (cached)};

\draw[sim flow,draw=simactionedge] (3.65,-2.1) -- (12.3,-2.1);
\foreach \controlx in {4.25,8.35,11.75} {
  \draw[sim flow,draw=simactionedge,
    preaction={draw=white,line width=2.5pt}]
    (\controlx,-2.1) -- (\controlx,-0.14);
  \filldraw[fill=simactionfill,draw=simactionedge]
    ({\controlx-0.08},-2.18) rectangle ({\controlx+0.08},-2.02);
}
\node[below] at (7.9,-2.2) {Supplied control stream};

\draw[sim flow,draw=simnoisyedge] (13.88,-0.45)
  arc[start angle=45,end angle=-255,radius=0.3];
\node[align=center] at (13.65,-1.7) {Denoise within chunk $j$\\Controls stay clean};

\node[anchor=north,align=center,font=\fontfamily{bch}\fontsize{9}{10.5}\selectfont\bfseries]
  at (7.6,-6.7) {(b) Two example control streams for $\mathcal{U}_j$};
\node[anchor=north,align=center] at (3.3,-6.05) {Action controls};
\node[anchor=north,align=center] at (11.5,-6.05) {Source video or spatial controls};
\draw[draw=simgrayedge] (7.4,-3.6) -- (7.4,-5.85);

\filldraw[fill=simactionfill,draw=simactionedge]
  (0.3,-5.125) rectangle (3.3,-4.375);
\foreach \actiontokenx in {0.5,1.2,1.9,2.6} {
  \draw[draw=simactionedge,fill=simactionfill!45!white]
    (\actiontokenx,-4.975) rectangle ++(0.5,0.45);
}
\node[anchor=base,inner sep=0pt] at (1.8,-4.125) {$\mathcal{U}_j$};
\draw[sim flow] (3.6,-4.75) -- (4.75,-4.75);
\simpackinggrid{5.0}{-5.125}{fill=simnoisyfill,draw=simnoisyedge}{$\widetilde{\mathcal{V}}_j^\tau$}
\node[below] at (1.8,-5.35) {Clean action group};

\simpackinggrid{8.8}{-5.125}{fill=simactionfill,draw=simactionedge}{$\mathcal{U}_j$}
\node[below] at (9.55,-5.35) {Clean source latents};
\draw[sim flow] (10.6,-4.75) -- (12.65,-4.75);
\simpackinggrid{12.9}{-5.125}{fill=simnoisyfill,draw=simnoisyedge}{$\widetilde{\mathcal{V}}_j^\tau$}

\end{tikzpicture}}
\endgroup

%% file: figures/diagrams/attention_masks.tex
\begingroup
\input{figures/diagrams/architecture_styles.tex}
\pgfmathsetmacro{\maskcell}{4/7}
\pgfmathsetmacro{\masklanguage}{5*\maskcell}
\pgfmathsetmacro{\maskblock}{3*\maskcell}
\fitdiagram{%
\begin{tikzpicture}[scale=1.1,font=\fontfamily{bch}\selectfont\small,>=Stealth,
  mask token/.style={font=\fontfamily{bch}\fontsize{8}{9.2}\selectfont},
  mask partition/.style={draw=black!35,line width=0.4pt},
  mask boundary/.style={draw=black,line width=1.4pt}]
\foreach \paneloffset/\panelmode/\paneltitle in {0/0/{(a) Cosmos 3},7.0/1/{(b) After causal conversion}} {
  \begin{scope}[shift={(\paneloffset,0)},x=0.62cm,y=0.62cm]
  \node[anchor=north,align=center,font=\fontfamily{bch}\selectfont\small\bfseries] at (4,-9.8) {\paneltitle};

  \node at (4,1.45) {Keys / values};

  \node[rotate=90] at (-1.8,-4) {Queries};

  \foreach \rowindex in {0,...,13} {
    \foreach \columnindex in {0,...,13} {
      \pgfmathsetmacro{\rowcoordinate}{\maskcell*\rowindex}
      \pgfmathsetmacro{\columncoordinate}{\maskcell*\columnindex}
      \fill[simgrayfill!45] (\columncoordinate,-\rowcoordinate) rectangle ++(\maskcell,-\maskcell);
      \ifnum\rowindex<5
        \ifnum\columnindex<5
          \ifnum\columnindex>\rowindex\else
            \fill[simtextfill] (\columncoordinate,-\rowcoordinate) rectangle ++(\maskcell,-\maskcell);
          \fi
        \fi
      \else
        \ifnum\columnindex<5
          \fill[simtextfill] (\columncoordinate,-\rowcoordinate) rectangle ++(\maskcell,-\maskcell);
        \else
          \pgfmathtruncatemacro{\querychunk}{floor((\rowindex-5)/3)}
          \pgfmathtruncatemacro{\keychunk}{floor((\columnindex-5)/3)}
          \pgfmathtruncatemacro{\keyrole}{mod(\columnindex-5,3)}
          \def\readallowed{0}
          \ifnum\panelmode=0
            \def\readallowed{1}
          \else
            \ifnum\keychunk>\querychunk\else
              \def\readallowed{1}
            \fi
          \fi
          \ifnum\readallowed=1
            \ifnum\keyrole=0
              \fill[simactionfill] (\columncoordinate,-\rowcoordinate) rectangle ++(\maskcell,-\maskcell);
            \else
              \fill[simnoisyfill] (\columncoordinate,-\rowcoordinate) rectangle ++(\maskcell,-\maskcell);
            \fi
          \fi
        \fi
      \fi
    }
  }
  \foreach \subdivision in {0,...,14} {
    \pgfmathsetmacro{\gridcoordinate}{\maskcell*\subdivision}
    \draw[sim grid] (\gridcoordinate,0) -- (\gridcoordinate,-8);
    \draw[sim grid] (0,-\gridcoordinate) -- (8,-\gridcoordinate);
  }
  \foreach \tokenindex in {0,...,4} {
    \pgfmathsetmacro{\tokenposition}{\maskcell*(\tokenindex+0.5)}
    \node[above,mask token] at (\tokenposition,0) {$\txttok{\tokenindex}$};

    \node[left,mask token] at (0,-\tokenposition) {$\txttok{\tokenindex}$};
  }
  \foreach \chunkindex in {0,1,2} {
    \pgfmathsetmacro{\chunkstart}{\masklanguage+\maskblock*\chunkindex}
    \fill[simactionfill] (\chunkstart,0.15) rectangle ++(\maskcell,0.16);
    \fill[simnoisyfill] ({\chunkstart+\maskcell},0.15) rectangle ++({2*\maskcell},0.16);
    \draw[simgrayedge] (\chunkstart,0.15) rectangle ++(\maskblock,0.16);
    \draw[simgrayedge] ({\chunkstart+\maskcell},0.15) -- ++(0,0.16);
    \node[above,mask token] at ({\chunkstart+0.5*\maskcell},0.32) {$\mathcal{U}_{\chunkindex}$};
    \node[above,mask token] at ({\chunkstart+2*\maskcell},0.32) {$\widetilde{\mathcal{V}}_{\chunkindex}$};

    \fill[simactionfill] (-0.31,-\chunkstart) rectangle ++(0.16,-\maskcell);
    \fill[simnoisyfill] (-0.31,{-\chunkstart-\maskcell}) rectangle ++(0.16,{-2*\maskcell});
    \draw[simgrayedge] (-0.31,-\chunkstart) rectangle ++(0.16,-\maskblock);
    \draw[simgrayedge] (-0.31,{-\chunkstart-\maskcell}) -- ++(0.16,0);
    \node[left,mask token] at (-0.32,{-\chunkstart-0.5*\maskcell}) {$\mathcal{U}_{\chunkindex}$};
    \node[left,mask token] at (-0.32,{-\chunkstart-2*\maskcell}) {$\widetilde{\mathcal{V}}_{\chunkindex}$};
  }
  \draw[mask partition] (0,0) rectangle (8,-8);
  \draw[mask partition] (\masklanguage,0) -- (\masklanguage,-8);
  \draw[mask partition] (0,-\masklanguage) -- (8,-\masklanguage);

  \foreach \boundaryindex in {1,2} {
    \pgfmathsetmacro{\chunkboundary}{\masklanguage+\maskblock*\boundaryindex}
    \draw[mask partition] (\chunkboundary,-\masklanguage) -- (\chunkboundary,-8);
    \draw[mask partition] (\masklanguage,-\chunkboundary) -- (8,-\chunkboundary);
  }
  \draw[mask boundary] (0,0)
    \foreach \stepindex in {1,...,5} {
      -- ({\stepindex*\maskcell},{-(\stepindex-1)*\maskcell})
      -- ({\stepindex*\maskcell},{-\stepindex*\maskcell})
    }
    -- (0,-\masklanguage) -- cycle;
  \draw[mask boundary]
    (0,-\masklanguage) rectangle (\masklanguage,-8);
  \ifnum\panelmode=0
    \draw[mask boundary]
      (\masklanguage,-\masklanguage) rectangle (8,-8);
  \else
    \draw[mask boundary]
      (\masklanguage,-\masklanguage)
      -- ++(\maskblock,0) -- ++(0,-\maskblock)
      -- ++(\maskblock,0) -- ++(0,-\maskblock)
      -- ++(\maskblock,0) -- (8,-8) -- (\masklanguage,-8) -- cycle;
  \fi
  \end{scope}
}
\node[align=center] at (2.48,-5.45) {Full temporal visibility\\across generator chunks};

\node[align=center] at (9.48,-5.45) {Joint reads within a chunk,\\no future generator chunks};

\foreach \legendposition/\legendcolor/\legendlabel in {
  -0.7/simtextfill/Language,
  2.8/simactionfill/Clean controls,
  7.0/simnoisyfill/Noisy vision,
  10.25/{simgrayfill!45}/Forbidden} {
  \filldraw[fill=\legendcolor,draw=simgrayedge] (\legendposition,-7.3) rectangle ++(0.32,0.27);

  \node[anchor=west] at ({\legendposition+0.45},-7.15) {\legendlabel};
}
\end{tikzpicture}}
\endgroup

%% file: sections/3__training.tex
\section{Post-training Recipe}
\label{sec::training}

\begingroup{}
\suppressfloats[t]
Our post-training recipe converts the base Cosmos 3 architecture into a generator capable of continuing from its own predictions.
This post-training process unfolds in four sequential stages, summarized in \cref{fig::training_pipeline}:
\begin{itemize}
    \item \textbf{Stage 0:} Bidirectional adaptation trains on reference clips with full temporal visibility to improve domain and control fidelity over long contexts.
    The adapted teacher initializes causal conversion and remains the frozen bidirectional reference for stage 3.
    \item \textbf{Stage 1:} Teacher Forcing (\textbf{TF}) learns causal chunk prediction from clean reference history.
    Its multi-step checkpoint initializes the stage-2 student and supplies its frozen causal teacher.
    \item \textbf{Stage 2:} Teacher Forcing with discrete-time Consistency Models (\textbf{TF-dCM}) retains clean reference history while training a few-step student to reduce denoising evaluations and preserve conditional behavior.
    \item \textbf{Stage 3:} Self Forcing with Distribution Matching Distillation (\textbf{SF-DMD}) trains on autoregressive student rollouts to address exposure bias and refine perceptual quality.
    The stage-0 bidirectional teacher and a trainable auxiliary score model evaluate the generated clips.
\end{itemize}
The causal distillation curriculum employs the same recipe as Causal-rCM~\citep{zheng2026causalrcm}.
Conditioning tokens, such as actions, source video tokens, structured visual controls, and initial observations, are treated as clean conditions and are excluded from the loss.
\ifhmode\par\fi\endgroup

\begingroup{}
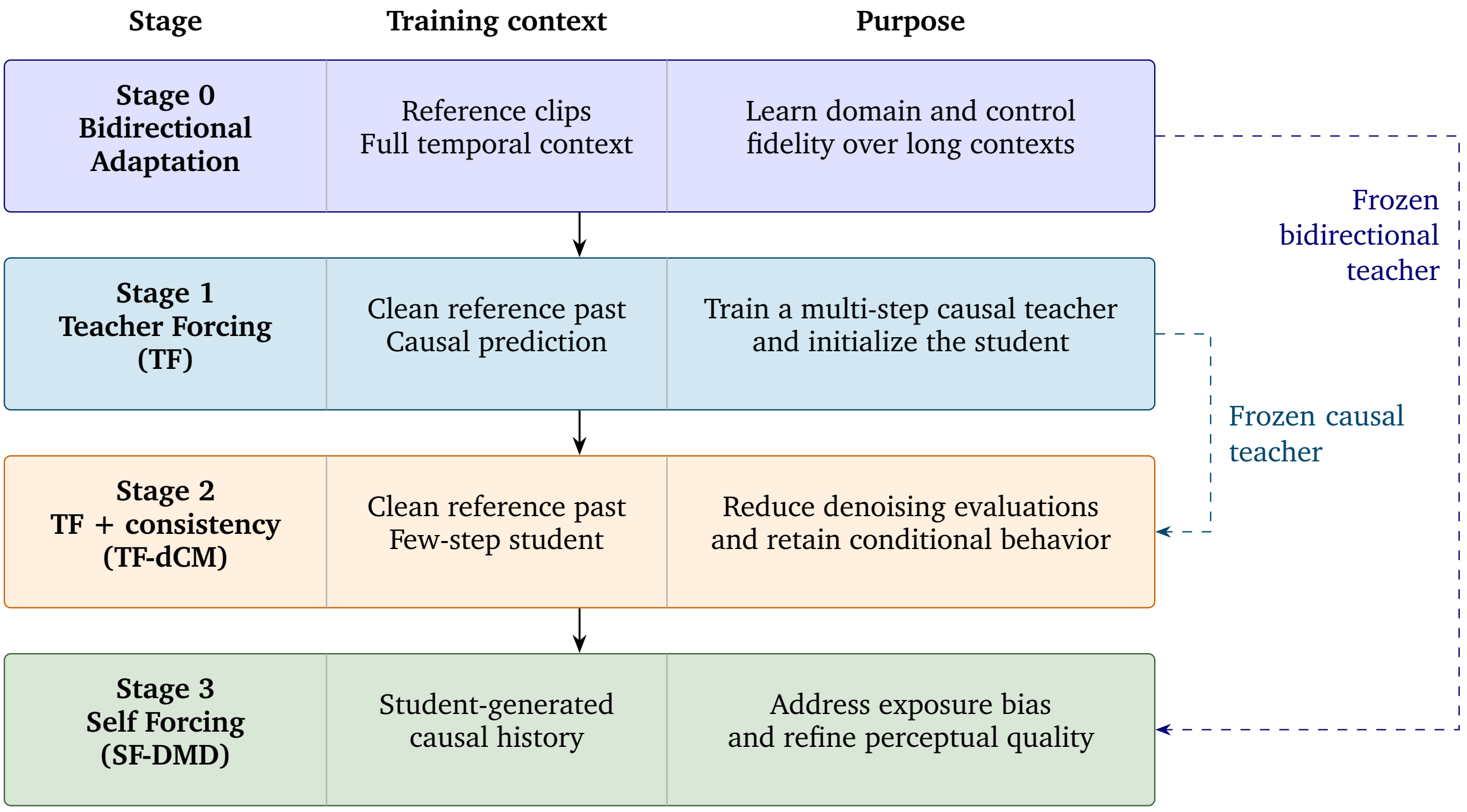
\begin{figure}[!t]
\centering
\input{figures/diagrams/training_pipeline.tex}
\caption{\textbf{Shared post-training stages for interactive world simulation.}
Stages follow Causal-rCM~\citep{zheng2026causalrcm} from top to bottom, with context and purpose alongside.
Solid arrows initialize the next stage.
Dashed paths supply the frozen stage-1 causal teacher for TF-dCM and the frozen stage-0 bidirectional teacher for SF-DMD.
SF-DMD also uses a trainable auxiliary score model to evaluate student rollouts.}
\label{fig::training_pipeline}
\end{figure}
\ifhmode\par\fi\endgroup

\subsection{Bidirectional Adaptation}
\label{sec::bidirectional_adaptation}

\begingroup{}
Stage 0 adapts the bidirectional Cosmos 3 generator to the observation domain and conditioning interface used by each example.
Action-conditioned records pair transition actions with the observations they affect, while video-transfer records align source controls and target observations over the same physical interval.
The teacher retains full temporal attention while learning these domain-specific relationships.
Long-context adaptation extends the temporal span of contiguous condition/observation records presented with their aligned captions.
It exposes the teacher to longer visual histories before the subsequent stages restrict temporal visibility and train autoregressive generation.
The adapted weights initialize causal conversion and supply the frozen bidirectional teacher for SF-DMD.
\Cref{sec::bimanual,sec::transfer,sec::auto} describe the example-specific data, sequence lengths, and adaptation schedules.
\ifhmode\par\fi\endgroup

\begingroup{}
For one latent sequence, let $\mathbf{z}$ stack the clean latent targets over the complete training clip, excluding supplied target observations.
The conditioning $\mathcal{C}$ contains language, aligned controls, and any supplied observations, all held clean during denoising.
We train the bidirectional predictor with rectified flow~\citep{liu2023rectifiedflow}:
\begin{equation}
 \begin{aligned}
  \mathbf{x}_{\tau}
  &=(1-\tau)\mathbf{z}+\tau\boldsymbol\epsilon,
  \qquad \boldsymbol\epsilon\sim\mathcal{N}(\mathbf{0},\mathbf{I}),\\
  L_{\text{bidir}}
  &=\mathbb{E}_{\mathbf{z},\mathcal{C},\boldsymbol\epsilon,\tau}\!\left[
  w_{\text{bidir}}(\tau)
  \left\|\mathbf{v}_{\theta}^{\text{bidir}}(\mathbf{x}_{\tau},\tau;\mathcal{C})
  -(\boldsymbol\epsilon-\mathbf{z})\right\|_2^2\right].
 \end{aligned}
 \label{eq::bidirectional_loss}
\end{equation}
The expectation jointly samples each clip and its conditions, independent Gaussian noise, and a diffusion time $\tau\in[0,1]$ shared across the clip's generated targets.
For the loss definitions below, let $C,T,H,W$ denote the channel count, number of latent frames, and latent height and width of the full latent sequence for a clip, including any conditioning frames.
We set $w_{\text{bidir}}(\tau)=1/(CTHW)$, normalizing the summed squared velocity error over generated frames by the full sequence size.
We write the loss weights as functions of diffusion time for generality; in this work, they are independent of $\tau$.
Full temporal attention lets each target location read all noisy target times and the supplied clean conditions.
Only generated target coordinates are noised and contribute to the loss.

Bidirectional adaptation and TF use nonuniform noise-time sampling followed by a resolution-dependent shift, as specified in \cref{app::noise_time_sampling}.
\ifhmode\par\fi\endgroup

\subsection{Causal Training with Teacher Forcing}
\label{sec::teacher_forcing}

\begingroup{}
The number of latent time steps in each chunk defines both the prediction unit and the temporal context available for modeling.
Each example selects a nominal prediction chunk size for its causal training stages.
Chunks also contain the aligned clean controls, which are held fixed while their visual targets are denoised.
Example-specific choices for chunk size are summarized in \cref{sec::bimanual,sec::transfer,sec::auto}.
The RGB tokenizer processes $1+4n$ input frames and produces $1+n$ latent time steps~\citep{wan2025}.
The initial latent time step is a singleton at startup, followed by chunks of at least 1 latent time step representing 4 new RGB frames per step.
An observed initial frame provides clean conditioning and is excluded from the generated targets.
When no initial frame is supplied, the model also generates that startup singleton.
\ifhmode\par\fi\endgroup

\begingroup{}
A sampled training example specifies its target latents, supplied conditions, and chunk partition.
Using the latent-time partition $\mathcal{B}_j$ from \cref{sec::token_packing}, let $\mathbf{z}^{j}$ flatten the generated visual coordinates at those times, excluding controls and supplied observations.
The noisy target $\mathbf{x}_{\tau}^{j}$ and noise $\boldsymbol\epsilon^{j}$ are the corresponding chunk restrictions of the rectified flow variables in \cref{eq::bidirectional_loss}.
Chunks without generated coordinates contribute no loss.
Under teacher forcing, the retained clean generator history $\mathcal{H}_j$ contains reference targets, previously supplied controls, and any initial target observations (\cref{sec::token_packing}).
The additional conditioning $\mathcal{C}_j=(\mathcal{T},\mathcal{U}_j)$ comprises the available language and the clean controls grouped with the current visual targets.
A single diffusion time $\tau$ is sampled per training example and shared across its noisy target chunks.
Physical-time chunk indices and diffusion times remain distinct even when several targets are evaluated together.
\ifhmode\par\fi\endgroup

\paragraph{Causal denoising objective.}
\begingroup{}
Let $\mathbf{v}_{\theta}^{\text{TF}}$ denote the velocity prediction for the noisy target chunk, conditioned on the clean reference history $\mathcal{H}_j$ and supplied inputs $\mathcal{C}_j$.
The denoising loss is
\begin{equation}
 L_{\text{TF}}=
 \mathbb{E}_{\mathbf{z},\mathcal{C},\boldsymbol\epsilon,\tau}
 \sum_j w_j^{\text{TF}}(\tau)
 \left\|\mathbf{v}_{\theta}^{\text{TF}}(\mathbf{x}_{\tau}^{j},\tau;
 \mathcal{H}_{j},\mathcal{C}_{j})-
 (\boldsymbol\epsilon^{j}-\mathbf{z}^{j})\right\|_2^2.
 \label{eq::tf_loss}
\end{equation}
The expectation samples each example jointly with its conditions, independent standard Gaussian target noise, and a diffusion time using the mode sampling and shift in \cref{sec::bidirectional_adaptation}.
TF uses the same full-sequence normalization as bidirectional training, applying it to each chunk's squared error before summing over chunks.
Clean reference history and supplied conditions, including any initial image, provide context and are excluded from the supervised coordinates.
\ifhmode\par\fi\endgroup

\paragraph{Clean and noisy teacher forcing masks.}
\begingroup{}
Teacher forcing separates clean reference inputs from noisy prediction targets, as illustrated in \cref{fig::replay_teacher_forcing}(a).
Clean-reference queries attend only to causally available clean inputs.
Noisy queries read clean history from strictly earlier chunks and attend jointly to controls and noisy visual tokens within their own prediction chunk.
Controls remain clean throughout denoising.
The masks prevent noisy queries from accessing the current clean visual target or future visual tokens, including indirectly through conditioning states or cached K/V.
\ifhmode\par\fi\endgroup

\paragraph{Replay-based teacher forcing.}
\begingroup{}
Replay-based teacher forcing implements this visibility pattern in two passes, avoiding the complexity of a single attention mask that jointly handles clean and noisy tokens.
The first pass processes clean references with the selected clean-history mask and stores their layerwise K/V.
The second pass evaluates noisy targets in parallel using joint attention within each prediction chunk and cached clean history from strictly earlier chunks.
Action controls are reevaluated with their branch's visual tokens, while the transfer path routes the clean control K/V directly to its noisy target queries.
Matching a packed forward also requires retaining or recomputing gradients through the clean cache.
\Cref{fig::replay_teacher_forcing} illustrates the action-conditioned layout with 1 latent time step per schematic chunk.
\ifhmode\par\fi\endgroup

\begingroup{}
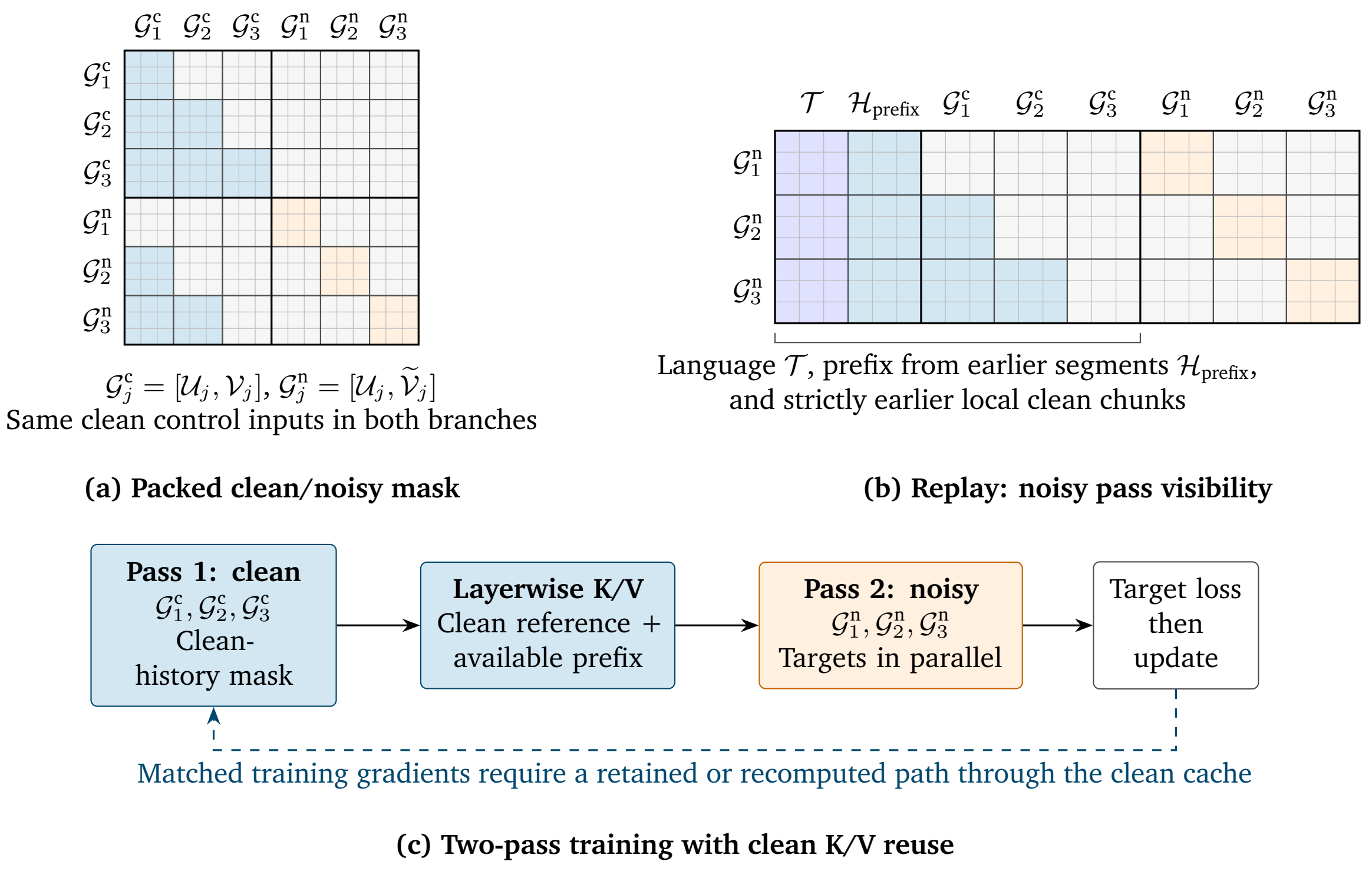
\begin{figure}[!t]
\centering
\input{figures/diagrams/replay_teacher_forcing.tex}
\caption{\textbf{Packed and replay-based teacher forcing.}
(a) The action-conditioned schematic shows 1 latent time step per chunk, with the same clean controls in both branches.
Color identifies the clean-reference or noisy-target branch, while controls remain clean in both.
The mask permits clean-to-clean attention, strictly earlier clean history for noisy queries, and joint reads within each noisy chunk.
For larger single-view chunks, clean queries can read only their own latent interval and earlier clean intervals.
(b) The replay noisy pass reads language $\mathcal{T}$, an available prefix $\mathcal{H}_{\text{prefix}}$ from earlier segments, and strictly earlier clean chunks.
(c) The clean pass supplies layerwise K/V for parallel noisy-target evaluation, with shared parameters and a single optimizer update.
The dashed arrow marks the retained or recomputed gradient path through the cache.}
\label{fig::replay_teacher_forcing}
\end{figure}
\ifhmode\par\fi\endgroup

\Needspace{8\baselineskip}
\subsection{Causal Consistency Distillation With Teacher Forcing}
\label{sec::consistency_distillation}

\begingroup{}
Stage 2 distills the causal flow model into a few-step generator to reduce denoising evaluations per chunk.
With the clean reference history and controls fixed, the student learns a consistency map~\citep{song2023consistency} that predicts the same clean endpoint from different states along a teacher denoising trajectory:
\begin{equation}
 f_{\theta}(\mathbf{x}_{\tau}^{j},\tau;\mathcal{H}_{j},\mathcal{C}_{j})
 =\mathbf{x}_{\tau}^{j}-\tau
 \mathbf{v}_{\theta}^{\text{TF}}(\mathbf{x}_{\tau}^{j},\tau;\mathcal{H}_{j},\mathcal{C}_{j}),
\end{equation}
\begin{equation}
 f_{\theta}(\mathbf{x},0;\mathcal{H}_{j},\mathcal{C}_{j})=\mathbf{x}.
 \label{eq::consistency_map}
\end{equation}
The velocity-parameterized network is retrained by the consistency objective, so $f_{\theta}$ estimates a trajectory endpoint rather than the posterior mean implied by an optimal RF regression model.
The boundary condition holds by construction for finite network outputs.
\ifhmode\par\fi\endgroup

\begingroup{}
We sample noise-time pairs $(\tau,s)$ with $0\leq s<\tau\leq1$, sharing each pair across target chunks.
The sampling rule and resolution-dependent shift are specified in \cref{app::noise_time_sampling}.
The initial $\mathbf{x}_{\tau}^{j}$ is drawn from the rectified flow corruption in \cref{eq::bidirectional_loss}, restricted to chunk $j$.
The frozen causal teacher $\psi_{\text{AR}}$ numerically advances this state toward lower noise:
\begin{equation}
 \widehat{\mathbf{x}}^{j}_{s}
 =\operatorname{ODESolve}\!\left(\mathbf{v}_{\psi_{\text{AR}}}^{\text{CFG}},
 \mathbf{x}_{\tau}^{j},\tau\!\rightarrow\!s;\mathcal{H}_{j},\mathcal{C}_{j}\right).
\end{equation}
We implement $\operatorname{ODESolve}$ with a single Euler step, keeping history and controls fixed.
The field $\mathbf{v}_{\psi_{\text{AR}}}^{\text{CFG}}$ includes the configured classifier-free guidance (CFG)~\citep{ho2022classifierfree}.
For conditional and reference-branch velocities $\mathbf{v}_{\text{cond}}$ and $\mathbf{v}_{\text{base}}$, our guidance convention is $\mathbf{v}^{\text{CFG}}=\mathbf{v}_{\text{base}}+\gamma(\mathbf{v}_{\text{cond}}-\mathbf{v}_{\text{base}})$, with $\gamma=1$ giving the conditional field.
Guidance scales and reference-branch conditioning are described in the model-specific training appendices.
\ifhmode\par\fi\endgroup

\begingroup{}
The student matches endpoint predictions at the two noise levels:
\begin{equation}
 L_{\text{TF-dCM}}
 =\mathbb{E}_{\mathbf{z},\mathcal{C},\boldsymbol\epsilon,\tau,s}
 \sum_j w_j^{\text{dCM}}(\tau)
 \left\|f_{\theta}(\mathbf{x}_{\tau}^{j},\tau;\mathcal{H}_{j},\mathcal{C}_{j})-
 \operatorname{sg}\!\left[f_{\theta}(\widehat{\mathbf{x}}^{j}_{s},s;
 \mathcal{H}_{j},\mathcal{C}_{j})\right]\right\|_2^2.
 \label{eq::tfdcm_loss}
\end{equation}
The expectation samples training examples, target noise, and the diffusion-time pairs defined above.
We set $w_j^{\text{dCM}}(\tau)=1/(CHW)$, averaging the squared error over channels and spatial locations and summing over generated latent frames.
We then average these losses across latent sequences.
Both evaluations use the current student parameters, with gradients stopped through the entire target evaluation, including its clean-history cache.
The teacher and student branches share the same reference history and controls, with targets constructed online.
\ifhmode\par\fi\endgroup

\begingroup{}
TF-dCM evaluates the consistency map using the two-pass replay layout in \cref{sec::teacher_forcing}.
A clean-reference pass constructs layerwise K/V, followed by a noisy-target pass that reads strictly earlier clean prefixes to predict the endpoint at each noise level.
Each network and teacher CFG branch constructs its own clean K/V using its own parameters and conditioning.
The student reuses its clean-reference cache for its two endpoint evaluations within the same update.
Gradients through that cache are retained or recomputed for the online prediction, while the complete target branch is detached.
The student learns the guided teacher trajectories and uses only a conditional forward pass per denoising step at inference.
\ifhmode\par\fi\endgroup

\subsection{Distribution Matching Distillation with Self Forcing}
\label{sec::self_forcing}

\begingroup{}
Stage 2 learns few-step denoising with clean reference history, whereas autoregressive inference conditions on previously generated chunks.
Starting from the stage-2 student, stage 3 applies Self Forcing Distribution Matching Distillation (SF-DMD) to optimize rollouts conditioned on the student's own generated history~\citep{huang2025selfforcing,yin2024improved}.
The retained cache $\mathcal{H}^{\text{KV}}_j$ represents the history $\mathcal{H}_j$ before chunk $j$, including previous control tokens and generated visual chunks.
Within chunk $j$, $K_j$ denoising evaluations follow the selected schedule $1=\tau_{j,K_j}>\ldots>\tau_{j,0}=0$, alternating endpoint prediction and Gaussian re-noising:
\begin{equation}
 \widetilde{\mathbf{x}}^{j}_{\tau_{j,n-1}}
 =(1-\tau_{j,n-1})f_{\theta}
 (\widetilde{\mathbf{x}}^{j}_{\tau_{j,n}},\tau_{j,n};
 \mathcal{H}^{\text{KV}}_j,\mathcal{C}_{j})
 +\tau_{j,n-1}\boldsymbol\epsilon_{j,n},
\end{equation}
\begin{equation}
 \widetilde{\mathbf{x}}^{j}_{\tau_{j,K_j}}\sim\mathcal{N}(\mathbf{0},\mathbf{I}).
 \label{eq::sf_transition}
\end{equation}
The transition is applied for $n=K_j,\ldots,1$, with independent standard Gaussian draws $\boldsymbol\epsilon_{j,n}$.
The selected schedule depends on the exit budget $K_j$.
The recurrence specifies forward sample values, with completed endpoint $\widehat{\mathbf{z}}^{j}=\widetilde{\mathbf{x}}^{j}_{0}$ corresponding to visual chunk $\widehat{\mathcal{V}}_j$.
Before advancing, the runtime completes the configured clean-history refresh and retention operations from \cref{sec::streaming_execution}, with gradients detached.
The single-view action path refreshes latent frames in temporal order within the completed chunk.
\ifhmode\par\fi\endgroup

\begingroup{}
Training differentiates only the final denoising evaluation of each chunk:
\begin{equation}
 \widehat{\mathbf{z}}^{j}
 =f_{\theta}\!\left(
 \operatorname{sg}[\widetilde{\mathbf{x}}^{j}_{\tau_{j,1}}],\tau_{j,1};
 \operatorname{sg}[\mathcal{H}^{\text{KV}}_j],\mathcal{C}_{j}\right).
 \label{eq::sf_truncated_endpoint}
\end{equation}
This expression has the same forward value as the rollout endpoint, but stops gradients through earlier denoising evaluations and generated history~\citep{huang2025selfforcing}.
\ifhmode\par\fi\endgroup

\paragraph{Teacher and auxiliary score model.}
\begingroup{}
The frozen stage-0 teacher $\psi_{\text{bidir}}$ supplies a reference score estimate, while a trainable auxiliary model $\phi$ estimates the score of the current student distribution~\citep{yin2024improved,zheng2026causalrcm}.
Both score models use bidirectional temporal attention over the rollout during training.
\ifhmode\par\fi\endgroup

\begingroup{}
Let $\widehat{\mathbf{z}}_i$ stack the nonempty set of generated target coordinates of rollout $i$, excluding supplied observations and controls.
Here $\mathcal{C}_i$ denotes the full supplied conditioning sequence, initial observations, and target layout, rather than the local chunk conditioning $\mathcal{C}_j$.
Let $\boldsymbol\xi_i$ collect the student rollout's sampling noise and exit-policy choices.
For fixed $\mathcal{C}_i$, let $p_{\theta}^{0}$ be the conditional law of these outputs under draws of $\boldsymbol\xi_i$.
For either the student law or a fixed reference law $p_{\text{ref}}^{0}$, define $p_a^{\tau}$ as the law of $(1-\tau)\mathbf{Z}+\tau\boldsymbol\epsilon$ with $\mathbf{Z}\sim p_a^{0}$ and independent standard Gaussian $\boldsymbol\epsilon$.
The ideal distribution-matching objective is the noisy reverse KL~\citep{yin2024improved}:
\begin{equation}
 L_{\text{DMD,raw}} =
 \mathbb{E}_{\mathcal{C}_i,\tau}\!\left[
 D_{\text{KL}}\!\left(p_{\theta}^{\tau}(\cdot\mid\mathcal{C}_i)
 \|\; p_{\text{ref}}^{\tau}(\cdot\mid\mathcal{C}_i)\right)\right].
 \label{eq::dmd_divergence}
\end{equation}
The expectation samples conditioning examples and a configured score time with $0<\tau<1$, with finite expected KL assumed.
The rollout law includes the selected exit and history-retention policies, which need not coincide with those used for evaluation.
\ifhmode\par\fi\endgroup

\begingroup{}
To estimate the student score, the auxiliary model is trained on re-noised, detached student outputs:
\begin{equation}
 \mathbf{y}_{i,\tau}=(1-\tau)\operatorname{sg}[\widehat{\mathbf{z}}_i]
 + \tau\boldsymbol\epsilon_i,
 \qquad \boldsymbol\epsilon_i\sim\mathcal{N}(\mathbf{0},\mathbf{I}),
\end{equation}
\begin{equation}
 L_{\text{aux}}(\phi)
 =\mathbb{E}_{\mathcal{C}_i,\boldsymbol\xi_i,\tau,\boldsymbol\epsilon_i}\!\left[
 \lambda_{\text{RF}}(\tau)\operatorname{mean}\!\left[
 \left|\mathbf{v}_{\phi}(\mathbf{y}_{i,\tau},\tau;\mathcal{C}_i)
 -(\boldsymbol\epsilon_i-\operatorname{sg}[\widehat{\mathbf{z}}_i])\right|^2\right]\right].
\end{equation}
The expectation samples conditioning examples, student rollouts, score times, and corruption noise.
Score times are sampled uniformly before applying the resolution-dependent noise shift in \cref{app::noise_time_sampling}.
The RF time weight is $\lambda_{\text{RF}}(\tau)=1$.
Here and below, $\operatorname{mean}$ averages over all generated coordinates of the scored stream, including its temporal extent, with absolute values and squares applied elementwise.
For $\tau>0$, the squared velocity error equals squared clean-endpoint error divided by $\tau^2$, using $f_a(\mathbf{y},\tau)=\mathbf{y}-\tau\mathbf{v}_a(\mathbf{y},\tau)$ for each score model $a$.
\ifhmode\par\fi\endgroup

\begingroup{}
For a clean law $p_a^0$, let $\mathbf{m}_{a,\tau}(\mathbf{y};\mathcal{C})=\mathbb{E}_a[\mathbf{Z}\mid\mathbf{Y}_{\tau}=\mathbf{y},\mathcal{C}]$ be its posterior endpoint mean.
Gaussian corruption relates this mean to the exact score~\citep{yin2024improved}:
\begin{equation}
 \mathbf{s}_{a,\tau}(\mathbf{y}\mid\mathcal{C})
 =\nabla_{\mathbf{y}}\log p_a^{\tau}(\mathbf{y}\mid\mathcal{C})
 =\frac{(1-\tau)\mathbf{m}_{a,\tau}(\mathbf{y};\mathcal{C})-\mathbf{y}}{\tau^2},
 \qquad 0<\tau<1.
 \label{eq::rf_score_endpoint}
\end{equation}
An exact RF regression model satisfies $f_a=\mathbf{m}_{a,\tau}$.
The teacher and auxiliary networks approximate these posterior means, whereas the consistency student's endpoint map has a different training objective.
\ifhmode\par\fi\endgroup

\begingroup{}
With fixed reference, conditioning and sampling laws, exact scores, and the regularity needed to interchange differentiation and expectation, the ideal pathwise gradient is
\begin{equation}
 \nabla_{\theta}L_{\text{DMD,raw}}
 =\mathbb{E}_{\mathcal{C}_i,\boldsymbol\xi_i,\tau,\boldsymbol\epsilon_i}\!\left[
 \left(\frac{\partial\widehat{\mathbf{z}}_i}{\partial\theta}\right)^{\!\mathsf T}
 \frac{(1-\tau)^2}{\tau^2}
 \bigl(\mathbf{m}_{\theta,\tau}(\mathbf{y}_{i,\tau};\mathcal{C}_i)
 -\mathbf{m}_{\text{ref},\tau}(\mathbf{y}_{i,\tau};\mathcal{C}_i)\bigr)\right].
 \label{eq::dmd_ideal_gradient}
\end{equation}
The expectation uses the same draws as the auxiliary loss.
The derivative follows the full rollout with sampled noise and exit choices held fixed, before the truncation in \cref{eq::sf_truncated_endpoint}.
One factor of $1-\tau$ comes from the RF corruption Jacobian and the other from the score-to-endpoint conversion.
\ifhmode\par\fi\endgroup

\begingroup{}
The practical update replaces exact posterior means by score-model predictions and uses adaptive normalization~\citep{yin2024improved,zheng2026causalrcm}:
\begin{equation}
 \begin{aligned}
 \rho_{i,\tau}
 &=\operatorname{mean}\!\left[
 \left|
 \operatorname{sg}[\widehat{\mathbf{z}}_i]
 -f_{\psi_{\text{bidir}}}(\mathbf{y}_{i,\tau},\tau;\mathcal{C}_i)
 \right|\right],\\
 \mathbf{g}_{i,\tau}
 &=\frac{
 f_{\phi}(\mathbf{y}_{i,\tau},\tau;\mathcal{C}_i)
 -f_{\psi_{\text{bidir}}}(\mathbf{y}_{i,\tau},\tau;\mathcal{C}_i)
 }{\rho_{i,\tau}+\delta},\qquad\delta>0.
 \end{aligned}
 \label{eq::dmd_normalized_direction}
\end{equation}
We use the additive stabilizer $\delta=10^{-6}$ in all three models.
The generator fits a detached endpoint offset:
\begin{equation}
 L_{\text{DMD}}(\theta)
 =\mathbb{E}_{\mathcal{C}_i,\boldsymbol\xi_i,\tau,\boldsymbol\epsilon_i}\!\left[
 \frac{1}{2a_i}\left\|\widehat{\mathbf{z}}_i-
 \operatorname{sg}[\widehat{\mathbf{z}}_i-\mathbf{g}_{i,\tau}]\right\|_2^2\right].
 \label{eq::dmd_surrogate}
\end{equation}
The expectation uses the same conditioning, rollout, time, and noise sampling as the auxiliary loss.
Here $a_i>0$ specifies the loss reduction separately from the adaptive normalizer $\rho_{i,\tau}$.
Auxiliary updates optimize $L_{\text{aux}}$ over $\phi$ using detached student outputs.
Generator updates optimize $L_{\text{DMD}}$ over $\theta$ through \cref{eq::sf_truncated_endpoint}, with both score models and the complete pseudo-target held fixed.
\ifhmode\par\fi\endgroup

\begingroup{}
For the single-view RGB models, $a_i=CHW$, giving a sum of per-frame means over the rollout.
For multimodal generation, such as joint RGB--LiDAR generation in Cosmos3-Nano-Sim-Auto, we apply this objective separately to each modality and combine their losses, as detailed in \cref{sec::auto_training_objectives}.
Holding the entire pseudo-target fixed gives derivative $\mathbf{g}_{i,\tau}/a_i$ with respect to the generated endpoint.
Adaptive normalization replaces the ideal diffusion-time factor, and final-call differentiation replaces the full rollout Jacobian by its truncated counterpart.
Together with approximate score models, these choices define a practical surrogate motivated by reverse-KL matching, rather than an exact gradient of $L_{\text{DMD,raw}}$.
Teacher CFG, when used, also replaces the reference prediction by a guided prediction, which need not correspond to the Gaussian corruption of one fixed clean reference law.
Teacher guidance and the alternating update schedule are specified for each example in \cref{sec::bimanual,sec::transfer,sec::auto}.

\ifhmode\par\fi\endgroup

\paragraph{Replayed backpropagation.}
\begingroup{}
Even with gradient truncation, differentiating the full rollout together retains final-step activations for every chunk.
Replayed backpropagation reduces this peak activation memory.
A rollout generated without gradients stores each chunk's final noisy input, diffusion time, detached cached prefix, and detached DMD target.
We then recompute and backpropagate the final prediction one chunk at a time, accumulating gradients before a single optimizer update.
With model parameters, saved inputs, pseudo-targets, conditioning, and stochastic choices fixed, replay preserves the selected truncated gradient estimator at the cost of recomputation.
Each chunk retains the same rollout-level loss reduction, so gradient accumulation differentiates the same sum of losses.
This equivalence does not restore gradients through earlier denoising evaluations or generated history.
\ifhmode\par\fi\endgroup

\paragraph{Exit policy.}
\begingroup{}
The exit policy selects the number of student denoising evaluations for each chunk.
The \texttt{max} rule uses the configured per-chunk maximum, matching inference with the same sampling budget.
The SF-DMD recipes for the three reported final models use \texttt{max} with 4 denoising evaluations per chunk.
Under final-step gradient truncation, this concentrates direct supervision at the schedule's last, lowest-noise input.
The \texttt{cycle} rule cycles through the available counts across updates, subject to each chunk's maximum.
Changing the exit depth allows different denoising evaluations to receive direct DMD supervision.
Cycling reduces the average rollout computation and broadens gradient coverage, but spreads updates across exit depths.
For fixed $\theta$, $p_{\theta}^{0}$ under a cycle denotes the mixture of rollout laws induced by the cycle's frequencies, which the auxiliary model is trained to estimate.
That mixture generally differs from the fixed-budget law used by \texttt{max}, even when their history policies agree.
\ifhmode\par\fi\endgroup

%% file: figures/diagrams/training_pipeline.tex
\begingroup
\input{figures/diagrams/architecture_styles.tex}
\newcommand{\trainingstagerow}[8]{%
  \node[stage,stage color #2] (stage#2) at (0,#1) {};
  \draw[simgrayedge!45] (-2.75,#1-0.825) -- (-2.75,#1+0.825);
  \draw[simgrayedge!45] (0.95,#1-0.825) -- (0.95,#1+0.825);
  \node[stage heading] at (-4.50,#1+0.34) {Stage #2};
  \node[stage heading] at (-4.50,#1-0.02) {#3};
  \node[stage heading] at (-4.50,#1-0.38) {#4};
  \node[stage text] at (-0.90,#1+0.17) {#5};
  \node[stage text] at (-0.90,#1-0.19) {#6};
  \node[stage text] at (3.60,#1+0.17) {#7};
  \node[stage text] at (3.60,#1-0.19) {#8};
}
\fitdiagram{%
\begin{tikzpicture}[font=\fontfamily{bch}\selectfont\small,>=Stealth,
  stage/.style={sim stage,fill=simgrayfill,minimum width=12.5cm,
    minimum height=1.65cm,inner sep=0pt},
  stage color 0/.style={fill=simtextfill,draw=simtextedge},
  stage color 1/.style={fill=simcleanfill,draw=simcleanedge},
  stage color 2/.style={fill=simnoisyfill,draw=simnoisyedge},
  stage color 3/.style={fill=simactionfill,draw=simactionedge},
  stage text/.style={anchor=base,align=center,inner sep=0pt,
    text height=1.6ex,text depth=0.4ex},
  stage heading/.style={stage text,font=\fontfamily{bch}\selectfont\small\bfseries},
  teacher label/.style={align=left,anchor=west,text width=2.2cm,
    inner sep=0pt,text=simgrayedge}]

\node[stage heading] at (-4.50,1.15) {Stage};
\node[stage heading] at (-0.90,1.15) {Training context};
\node[stage heading] at (3.60,1.15) {Purpose};

\trainingstagerow{0}{0}{Bidirectional}{Adaptation}
  {Reference clips}{Full temporal context}
  {Learn domain and control}{fidelity over long contexts}
\trainingstagerow{-2.15}{1}{Teacher Forcing}{(TF)}
  {Clean reference past}{Causal prediction}
  {Train a multi-step causal teacher}{and initialize the student}
\trainingstagerow{-4.30}{2}{TF + consistency}{(TF-dCM)}
  {Clean reference past}{Few-step student}
  {Reduce denoising evaluations}{and retain conditional behavior}
\trainingstagerow{-6.45}{3}{Self Forcing}{(SF-DMD)}
  {Student-generated}{causal history}
  {Address exposure bias}{and refine perceptual quality}

\draw[sim flow] (stage0.south) -- (stage1.north);
\draw[sim flow] (stage1.south) -- (stage2.north);
\draw[sim flow] (stage2.south) -- (stage3.north);

\draw[->,dashed,simcleanedge] (stage1.east) -- ++(0.60,0)
  |- (stage2.east);
\node[teacher label,text=simcleanedge] at (7.05,-3.225) {Frozen causal\\teacher};
\draw[->,dashed,simtextedge] (stage0.east) -- ++(3.30,0)
  |- (stage3.east);
\node[teacher label,anchor=east,align=right,text=simtextedge] at (9.35,-1.075) {Frozen\\bidirectional\\teacher};
\end{tikzpicture}}
\endgroup

%% file: figures/diagrams/replay_teacher_forcing.tex
\begingroup
\input{figures/diagrams/architecture_styles.tex}
\fitdiagram{%
\begin{tikzpicture}[font=\fontfamily{bch}\selectfont\small,>=Stealth]
\node[anchor=north,align=center,font=\fontfamily{bch}\selectfont\small\bfseries] at (1.65,-4.65) {(a) Packed clean/noisy mask};
\begin{scope}[x=0.55cm,y=0.55cm]
\foreach \rowindex in {0,...,5} {
  \foreach \columnindex in {0,...,5} {
    \fill[simgrayfill!45] (\columnindex,-\rowindex) rectangle ++(1,-1);
    \ifnum\rowindex<3
      \ifnum\columnindex>\rowindex\else
        \fill[simcleanfill] (\columnindex,-\rowindex) rectangle ++(1,-1);
      \fi
    \else
      \pgfmathtruncatemacro{\querychunk}{\rowindex-3}
      \ifnum\columnindex<\querychunk
        \fill[simcleanfill] (\columnindex,-\rowindex) rectangle ++(1,-1);
      \fi
      \ifnum\columnindex=\rowindex
        \fill[simnoisyfill] (\columnindex,-\rowindex) rectangle ++(1,-1);
      \fi
    \fi
    \foreach \subdivision in {1,2} {
      \draw[sim grid] ({\columnindex+\subdivision/3},-\rowindex) -- ++(0,-1);
      \draw[sim grid] (\columnindex,{-\rowindex-\subdivision/3}) -- ++(1,0);
    }
    \draw[simgrayedge,line width=0.4pt] (\columnindex,-\rowindex) rectangle ++(1,-1);
  }
}
\foreach \tokenposition/\tokenlabel in {0.5/$\mathcal{G}_1^{\text{c}}$,1.5/$\mathcal{G}_2^{\text{c}}$,2.5/$\mathcal{G}_3^{\text{c}}$,3.5/$\mathcal{G}_1^{\text{n}}$,4.5/$\mathcal{G}_2^{\text{n}}$,5.5/$\mathcal{G}_3^{\text{n}}$} {
  \node[above] at (\tokenposition,0) {\tokenlabel};

  \node[left] at (0,-\tokenposition) {\tokenlabel};
}
\draw[semithick] (0,0) rectangle (6,-6);
\draw[semithick] (3,0) -- (3,-6);
\draw[semithick] (0,-3) -- (6,-3);

\node[align=center] at (3,-7.15) {$\mathcal{G}_j^{\text{c}}=[\mathcal{U}_j,\mathcal{V}_j]$, $\mathcal{G}_j^{\text{n}}=[\mathcal{U}_j,\widetilde{\mathcal{V}}_j]$\\Same clean control inputs in both branches};
\end{scope}
\node[anchor=north,align=center,font=\fontfamily{bch}\selectfont\small\bfseries] at (10.58,-4.65) {(b) Replay: noisy pass visibility};
\begin{scope}[shift={(7.3,-0.9)},x=0.82cm,y=0.72cm]
\foreach \rowindex in {0,...,2} {
  \foreach \columnindex in {0,...,7} {
    \fill[simgrayfill!45] (\columnindex,-\rowindex) rectangle ++(1,-1);
    \ifnum\columnindex=0
      \fill[simtextfill] (\columnindex,-\rowindex) rectangle ++(1,-1);
    \fi
    \ifnum\columnindex=1
      \fill[simcleanfill] (\columnindex,-\rowindex) rectangle ++(1,-1);
    \fi
    \ifnum\columnindex>1
      \ifnum\columnindex<5
        \pgfmathtruncatemacro{\localclean}{\columnindex-2}
        \ifnum\localclean<\rowindex
          \fill[simcleanfill] (\columnindex,-\rowindex) rectangle ++(1,-1);
        \fi
      \fi
    \fi
    \pgfmathtruncatemacro{\selfcolumn}{\rowindex+5}
    \ifnum\columnindex=\selfcolumn
      \fill[simnoisyfill] (\columnindex,-\rowindex) rectangle ++(1,-1);
    \fi
    \foreach \subdivision in {1,2} {
      \draw[sim grid] ({\columnindex+\subdivision/3},-\rowindex) -- ++(0,-1);
      \draw[sim grid] (\columnindex,{-\rowindex-\subdivision/3}) -- ++(1,0);
    }
    \draw[simgrayedge,line width=0.4pt] (\columnindex,-\rowindex) rectangle ++(1,-1);
  }
}
\foreach \tokenposition/\tokenlabel in {0.5/$\mathcal{T}$,1.5/$\mathcal{H}_{\text{prefix}}$,2.5/$\mathcal{G}_1^{\text{c}}$,3.5/$\mathcal{G}_2^{\text{c}}$,4.5/$\mathcal{G}_3^{\text{c}}$,5.5/$\mathcal{G}_1^{\text{n}}$,6.5/$\mathcal{G}_2^{\text{n}}$,7.5/$\mathcal{G}_3^{\text{n}}$} {
  \node[anchor=base,inner sep=0pt] at (\tokenposition,0.3) {\tokenlabel};
}
\foreach \rowcenter/\rowlabel in {0.5/$\mathcal{G}_1^{\text{n}}$,1.5/$\mathcal{G}_2^{\text{n}}$,2.5/$\mathcal{G}_3^{\text{n}}$} {
  \node[left] at (0,-\rowcenter) {\rowlabel};
}
\draw[semithick] (0,0) rectangle (8,-3);

\foreach \boundary in {2,5} {\draw[semithick] (\boundary,0) -- (\boundary,-3);}
\draw[simgrayedge] (0,-3.15) -- (0,-3.3) -- (5,-3.3) -- (5,-3.15);

\node[below,align=center] at (2.5,-3.3) {Language $\mathcal{T}$, prefix from earlier segments $\mathcal{H}_{\text{prefix}}$,\\and strictly earlier local clean chunks};
\end{scope}
\node[anchor=north,align=center,font=\fontfamily{bch}\selectfont\small\bfseries] at (6.175,-8.65) {(c) Two-pass training with clean K/V reuse};

\node[sim stage,fill=simcleanfill,draw=simcleanedge,text width=2.4cm] (cleanpass) at (1.0,-6.45) {\textbf{Pass 1: clean}\\$\mathcal{G}_1^{\text{c}},\mathcal{G}_2^{\text{c}},\mathcal{G}_3^{\text{c}}$\\Clean-history mask};

\node[sim stage,fill=simcleanfill,draw=simcleanedge,text width=2.5cm] (trainingcache) at (4.75,-6.45) {\textbf{Layerwise K/V}\\Clean reference +\\available prefix};

\node[sim stage,fill=simnoisyfill,draw=simnoisyedge,text width=2.6cm] (noisypass) at (8.6,-6.45) {\textbf{Pass 2: noisy}\\$\mathcal{G}_1^{\text{n}},\mathcal{G}_2^{\text{n}},\mathcal{G}_3^{\text{n}}$\\Targets in parallel};

\node[sim stage,text width=1.5cm] (targetloss) at (11.8,-6.45) {Target loss\\then update};

\draw[sim flow] (cleanpass) -- (trainingcache);
\draw[sim flow] (trainingcache) -- (noisypass);
\draw[sim flow] (noisypass) -- (targetloss);
\draw[->,dashed,simcleanedge,semithick] (targetloss.south) -- (11.8,-7.85) --
  node[midway,below,align=center] {Matched training gradients require a retained or recomputed path through the clean cache} (1.0,-7.85) -- (cleanpass.south);

\end{tikzpicture}}
\endgroup

%% file: sections/4__inference.tex
\section{Streaming Inference}
\label{sec::inference_optimization}

\begingroup{}
Few-step distillation reduces the number of generator evaluations per chunk, but each prediction still requires condition preparation, history access and updates, and decoding.
Streaming inference seeks to optimize per-chunk work by reusing valid state from earlier computations.
We describe the generation loop and its memory requirements, the runtime methods that this state enables, and the timing boundaries used to evaluate performance.
The examples in \cref{sec::bimanual,sec::transfer,sec::auto} apply these methods to their conditioning interfaces.
\ifhmode\par\fi\endgroup

\subsection{Streaming State and Memory}
\label{sec::streaming_execution}

\begingroup{}
Streaming inference organizes execution into discrete chunks, each dictated by the temporal-causal attention described in \cref{sec::shared_architecture}.
The initialization step establishes any available clean history from provided target observations.
As the model progresses from one chunk to the next, it continually reuses language representations, clean generator keys and values (K/V), and the causal decoder state between predictions.
\ifhmode\par\fi\endgroup

\paragraph{Next-chunk prediction.}
\begingroup{}
Autoregressive inference samples one chunk at a time, using generated observations in the history rather than the clean reference targets used by TF training (\cref{sec::teacher_forcing}).
For chunk $j$, the supplied conditions $\mathcal{U}_j$ encode the available actions, source video, or spatial controls.
The sampler processes $\mathcal{G}_j^\tau=[\mathcal{U}_j,\widetilde{\mathcal{V}}_j^\tau]$, jointly denoising its visual targets with language and retained clean history as context.
During sampling, the conditions are fixed and history remains unchanged across the denoising steps.
New conditions can influence the following unsampled chunk.
Sampling requires all controls for the current chunk to be available.
Larger chunks can amortize execution overhead, but also extend the physical interval covered before the next condition update is accepted.
Once sampling is complete, the resulting latents $\widehat{\mathcal{V}}_j$ feed into two branches: the K/V cache update prepares for the next prediction, and causal decoding reconstructs observations (\cref{fig::streaming_step}).
Decoder state maintains temporal information across chunks.
Both branches can execute concurrently on separate GPUs.
\ifhmode\par\fi\endgroup

\paragraph{Append clean K/V to history.}
\begingroup{}
History $\mathcal{H}_j$ tracks clean conditions and completed observations.
After denoising, the runtime refreshes clean K/V from the completed chunk, including its controls and generated latents:
\begin{equation}
  \Delta\mathcal{H}^{\text{KV}}_j =
  \operatorname{RefreshKV}_{\theta}
  (\widehat{\mathcal{G}}_j; \mathcal{H}^{\text{KV}}_j, \mathcal{T}^{\text{KV}}_j),
  \label{eq::streaming_delta_kv}
\end{equation}
The generator appends the resulting layerwise K/V and applies the configured history-retention rule:
\begin{equation}
  \mathcal{H}^{\text{KV}}_{j+1} =
  \operatorname{Retain}_{j+1}\!\left(
  \mathcal{H}^{\text{KV}}_j \oplus \Delta\mathcal{H}^{\text{KV}}_j\right).
  \label{eq::streaming_clean_commit}
\end{equation}
Here, $\widehat{\mathcal{G}}_j=[\mathcal{U}_j,\widehat{\mathcal{V}}_j]$ is the completed clean chunk, and $\Delta\mathcal{H}^{\text{KV}}_j$ contains its new clean K/V.
The operator $\operatorname{RefreshKV}_{\theta}$ denotes the complete history-refresh procedure for the chunk and can contain several internal forward passes.
The cache $\mathcal{T}^{\text{KV}}_j$ represents the available language $\mathcal{T}$.
The operation $\oplus$ appends entries with their associated token roles and timestamps.
The operator $\operatorname{Retain}_{j+1}$ selects the entries kept for the next prediction, and is the identity when full history is retained.
The resulting $\mathcal{H}^{\text{KV}}_{j+1}$ represents the retained history $\mathcal{H}_{j+1}$ for the subsequent prediction.

Chunk progression waits until the history refresh is complete.
The runtime processes the completed latent frames in temporal order, each with its aligned clean actions and allowed earlier history.
Therefore, later clean frames in the chunk can read refreshed earlier frames, matching the frame-causal clean branch of TF (\cref{sec::teacher_forcing}).
Joint denoising within the prediction chunk and ordered clean-history construction serve different roles.
Multi-view backends can instead construct clean K/V with their configured chunk-causal mask.

For $K_j$ denoising evaluations and $R_j$ clean-history evaluations, the continuing chunk costs $K_j+R_j$ generator evaluations.
The reference single-view path can refresh all $N_j$ latent frames separately.
A backend that batches the required refresh into one forward pass has $R_j=1$.
\ifhmode\par\fi\endgroup

\begingroup{}
\begin{figure}[!t]
\centering
\input{figures/diagrams/streaming_step.tex}
\caption{\textbf{State and dependencies in one streaming generation step.}
Initial context supplies language and any available observed history, while streaming inputs provide the current conditions $\mathcal{U}_j$.
Denoising reads these inputs without modifying retained K/V.
Completed latents feed the cache update and stateful decoding, which can overlap on separate GPUs.
The generator refreshes clean K/V for the completed chunk and appends the retained entries before the next prediction (\cref{eq::streaming_delta_kv,eq::streaming_clean_commit}).
For single-view action conditioning, this refresh processes latent frames in temporal order within the chunk.}
\label{fig::streaming_step}
\end{figure}
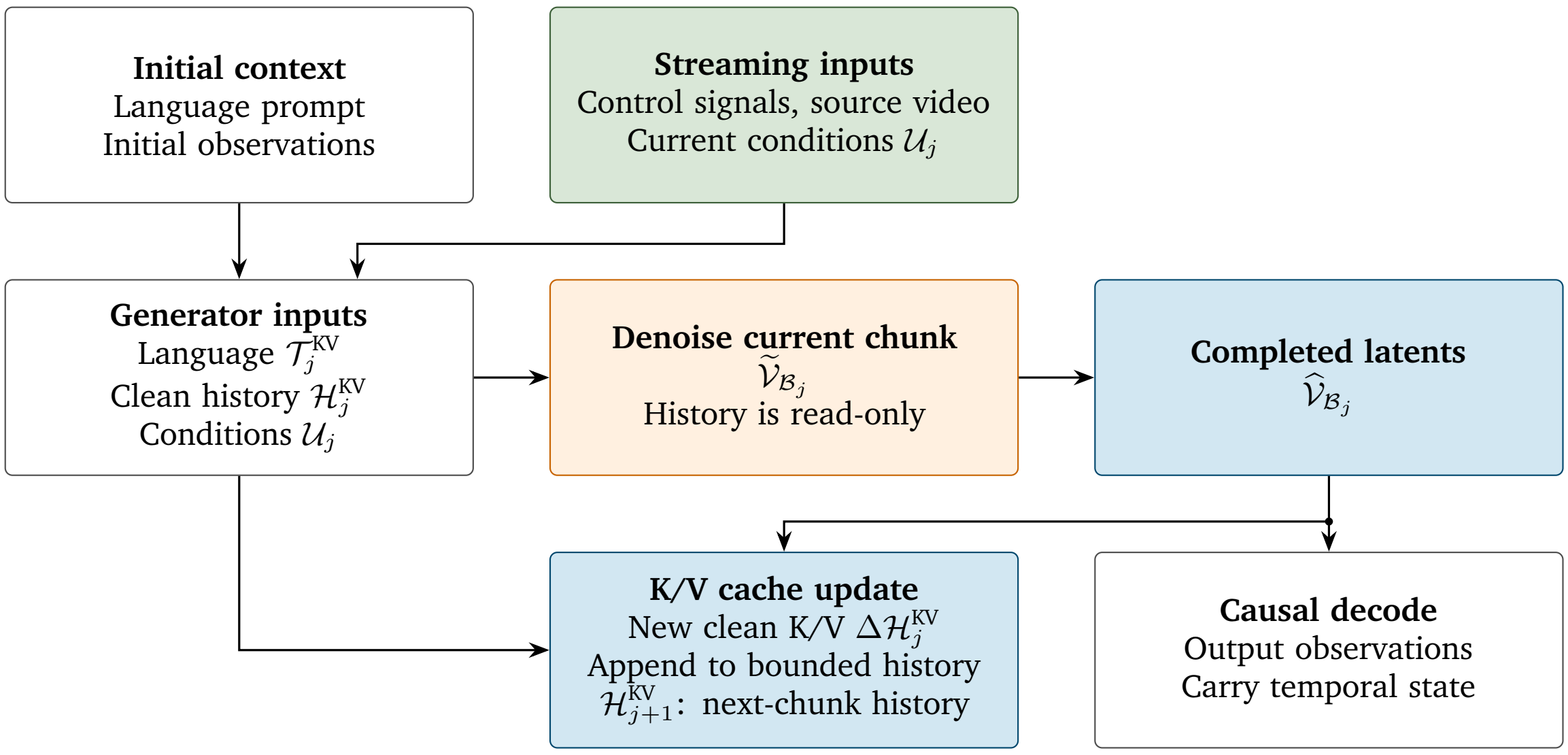
\ifhmode\par\fi\endgroup

\paragraph{Lifetime of context in the K/V caches.}
\begingroup{}
State in this loop persists for different durations depending on its type.
Language representations can be reused as long as their inputs and attention dependencies remain unchanged.
Generator history is stable within each target block and only updates when new entries are committed, enabling all denoising steps for a block to share the same clean K/V~\citep{zheng2026causalrcm}.
Supplied conditions retain their representations if they are independent of noisy targets.
During denoising, the read path leverages attention buffers until history updates, avoiding extra concatenation, allocation, or padding in each iteration.
\ifhmode\par\fi\endgroup

\Needspace{10\baselineskip}
\paragraph{Bounded history with initial sinks.}
\begingroup{}
Reusing history avoids redundant computation, but retaining all past observations would still increase storage and attention cost as the rollout grows.
To bound history, the cache maintains a fixed initial sink region together with a rolling window of recent clean entries, as in StreamingLLM~\citep{streamingllm2023}.
Sink entries keep initial visual context, while the rolling window retains recent motion.
When new history is committed, the cache replaces older non-sink entries.
Storage locations keep logical timestamps and token roles, so eviction does not renumber observations.
Valid-length or role metadata can exclude unused capacity from attention.
After the rolling window fills, both the stored K/V and active attention context stay bounded.
This limits access to older observations but ensures a fixed history capacity for the runtime methods described below.
\ifhmode\par\fi\endgroup

\begingroup{}
History retention is a separate choice from whether history is recorded or generated.
A model may retain the full history during causal training, but employ a fixed-size window with an attention sink during evaluation (\cref{sec::bimanual_training,sec::bimanual_results}).
SF-DMD exposes the student to its own generated observations, but this alone does not remove a difference between training and evaluation retention policies.
The selected history policy therefore remains part of the evaluated operating point.
\ifhmode\par\fi\endgroup

\Needspace{8\baselineskip}
\subsection{Runtime Optimization}
\label{sec::runtime_optimization}

\begingroup{}
We optimize autoregressive inference by reducing the cost of each generator call and scheduling work across the available GPUs.
The objective is to minimize control-to-observation latency while sustaining the required throughput.
Communication, synchronization, and duplicated work can offset the savings from parallelism.
\ifhmode\par\fi\endgroup

\Needspace{8\baselineskip}
\paragraph{Optimization metrics and execution cost.}
\phantomsection
\label{sec::latency_accounting}
\begingroup{}
Let $t_{\text{in}}$ be the time the runtime receives an input control assigned to a future chunk, and $t_{\text{obs}}$ the time its first affected generated observation is ready after decoding.
The control-to-observation latency is defined as
\begin{equation}
 T_{\text{control}\to\text{obs}} = t_{\text{obs}}-t_{\text{in}},
 \label{eq::control_observation_latency}
\end{equation}
This interval includes waiting for the rest of the chunk's required controls, condition preparation, generation, decoding, and any synchronization or bookkeeping that delays the corresponding output.
Conditions remain fixed during denoising, so a newly received control can affect only a subsequent unsampled chunk (\cref{sec::streaming_execution}).
Let $t_{\text{controls},j}$ be the time all required inputs for chunk $j$ are available and $t_{\text{history},j}$ the time its clean-history cache is ready.
The start of denoising satisfies
\begin{equation}
 t_{\text{start},j}\geq
 \max\!\left(t_{\text{controls},j},t_{\text{history},j}\right).
 \label{eq::chunk_readiness}
\end{equation}
Collecting inputs over a longer chunk interval can therefore increase response latency even when the model sustains the required output throughput.
Controls supplied as a complete plan or prerecorded stream need not incur this input-collection delay.
The timing boundary here is decoded-output readiness, with display or transport delay counted separately when relevant to deployment.
\ifhmode\par\fi\endgroup

\begingroup{}
Let $F$ count output timestamps produced in elapsed generation time $T$, with physical duration $D$.
We use
\begin{equation}
 \mathrm{FPS}_{\text{sync}}=\frac{F}{T},\qquad
 \mathrm{RTF}=\frac{T}{D}.
 \label{eq::runtime_definitions}
\end{equation}
The measurement states the execution boundary for $T$.
At a fixed rate $r$, a sequence including its initial frame spans $D=(F-1)/r$, while $F$ new steady-state outputs span $D=F/r$.
Real-time throughput requires $\mathrm{RTF}\leq1$ for the specified workload and timing boundary.
Throughput measures generation capacity, while control-to-observation latency measures responsiveness.
\ifhmode\par\fi\endgroup

\begingroup{}
\Needspace{6\baselineskip}
To identify optimization targets, the sequential per-chunk execution budget is
\begin{equation}
 T_{\text{chunk}}=
 T_{\text{condition}}+T_{\text{denoise}}+
 T_{\text{cache}}+T_{\text{decode}}+T_{\text{bookkeeping}},
 \label{eq::chunk_latency}
\end{equation}
where $T_{\text{cache}}$ is the duration of the clean K/V update required before the next prediction.
With overlap, measured wall time follows the schedule's dependency critical path, including transfer and synchronization costs.
The methods below reduce repeated execution overhead, distribute transformer work, and overlap independent reconstruction and history updates.
\ifhmode\par\fi\endgroup

\paragraph{Minimizing compilation and launch overhead.}
\begingroup{}
Every denoising pass reuses the same transformer layers.
Reusing the compilation artifacts and reducing the launch overhead improves total efficiency.
Once the rolling attention window reaches its configured capacity, the generation component operates with a fixed history context size even as new entries are committed (\cref{sec::streaming_execution}).
Padding the language tokens to a fixed capacity further stabilizes the cache shapes, while valid-length metadata avoids attending to unused rows.
Fixed shapes, preallocated metadata, and persistent storage allow static transformer compilation~\citep{pytorch_aot_inductor} and CUDA-graph replay~\citep{nguyen2021cudagraphs}.
\ifhmode\par\fi\endgroup

\paragraph{Maskless attention implementation.}
\begingroup{}
Each generator call involves attention over the language, cached history, and the current chunk, following the temporal-causal structure in \cref{sec::shared_architecture}.
Constructing large pairwise masks adds computation overhead.
Maskless attention partitions the allowed attention targets into groups and computes each group separately, reducing the computational overhead.
Furthermore, log-sum-exp statistics merge these outputs to ensure that every valid key is counted just once~\citep{ye2025flashinfer}.
\ifhmode\par\fi\endgroup

\paragraph{Context parallelism.}
\begingroup{}
Dividing the transformer computation across multiple GPUs allows for parallel workloads but introduces communication at each attention stage.
The design is based on the Ulysses-style context-parallel approach from Cosmos 3~\citep{cosmos3,deepspeed_ulysses}.
Tokens are distributed across the GPUs.
During each attention layer, the first all-to-all exchange groups tokens by attention head for local computation.
The second exchange restores the original token order.
\ifhmode\par\fi\endgroup

\begingroup{}
The system can also use replicated input and output to change the communication pattern.
Each GPU keeps a full copy of the current chunk's hidden state while only splitting the cached K/V among attention heads.
Every GPU computes local projections and then combines results using all-reduce.
This replaces the corresponding all-to-all exchanges with an all-reduce while repeating projection and feed-forward computation.
The resulting communication volume and runtime depend on the tensor layout, collective implementation, and GPU count.
Replicated computation trades against communication, so this layout does not guarantee lower latency than the original layout.
\ifhmode\par\fi\endgroup

\paragraph[\kern0pt Distribute camera work.]{{Distribute camera work.}}
\begingroup{}
Multi-view execution additionally distributes camera loading, eligible encoding, decoding, and saving across the inference replica.
Decoded outputs are collected for continuation, which retains communication and host-storage costs.
This reduces redundant media processing around the transformer without changing the model's attention visibility.
As camera count grows, distributing sensor processing becomes useful when loading, encoding, or decoding limits throughput.
Joint RGB and LiDAR generation uses modality-specific reconstruction and returns temporally aligned sensor observations.
\ifhmode\par\fi\endgroup

\paragraph[\kern0pt Compiled encoding and decoding.]{{Compiled encoding and decoding.}}
\begingroup{}
Compilation also applies to repeated condition-encoding and observation-decoding operations.
Stable input shapes allow these operations to reuse compiled execution while preserving any temporal state carried between chunks.
These operations have compilation requirements independent of whether the generator history window is full.
\ifhmode\par\fi\endgroup

\paragraph[\kern0pt Decoder performance--quality tradeoff.]{{Decoder performance--quality tradeoff.}}
\begingroup{}
The pipeline supports compiled Wan 2.2 VAE~\citep{wan2025} and TAEHV Super~\citep{BoerBohan2025TAEHV} as selectable RGB decoders.
TAEHV Super uses a lighter reconstruction network while retaining the Wan encoder and generator latent representation.
Users can select either decoder according to their application's reconstruction quality and performance requirements.
\ifhmode\par\fi\endgroup

\paragraph{Asynchronous decoding.}
\begingroup{}
Once a chunk's denoising is complete, the latents are sent to a decoder that runs on a separate GPU.
The decoder transforms these latents into RGB observations, producing the outputs seen by downstream tasks or users.
At the same time, the generator remains active on its original GPU, using the just-completed latents and conditions to compute and append clean K/V to history.
Performing decoding and history updates in parallel allows the system to overlap the two computation branches.
The overlap can shorten the critical path relative to sequential execution when its savings exceed the added transfer and synchronization costs.
The next prediction waits for its clean-history update, while downstream consumers wait for the required decoded observations.
Measured throughput and control-to-observation latency must account for these distinct dependencies under the selected schedule.
\ifhmode\par\fi\endgroup

%% file: figures/diagrams/streaming_step.tex
\begingroup
\hyphenpenalty=10000
\input{figures/diagrams/architecture_styles.tex}
\fitdiagram{%
\begin{tikzpicture}[font=\fontfamily{bch}\fontsize{9}{10.5}\selectfont,>=Stealth,
  every node/.append style={text=.},
  stream stage/.style={sim stage,text width=4.2cm,minimum height=1.9cm}]
\node[stream stage]
  (initial) at (0,2.65) {\textbf{Initial context}\\
    Language prompt\\Initial observations};
\node[stream stage,fill=simactionfill,draw=simactionedge]
  (controls) at (5.3,2.65) {\textbf{Streaming inputs}\\
    Control signals, source video\\
    Current conditions $\mathcal{U}_j$};
\node[stream stage]
  (inputs) at (0,0) {\textbf{Generator inputs}\\
    Language $\mathcal{T}^{\text{KV}}_j$\\
    Clean history $\mathcal{H}^{\text{KV}}_j$\\
    Conditions $\mathcal{U}_j$};
\node[stream stage,fill=simnoisyfill,draw=simnoisyedge]
  (sample) at (5.3,0) {\textbf{Denoise current chunk}\\
    $\widetilde{\mathcal{V}}_{\mathcal{B}_j}$\\History is read-only};
\node[stream stage,fill=simcleanfill,draw=simcleanedge]
  (latent) at (10.6,0) {\textbf{Completed latents}\\
    $\widehat{\mathcal{V}}_{\mathcal{B}_j}$};
\node[stream stage,fill=simcleanfill,draw=simcleanedge]
  (cacheupdate) at (5.3,-2.65) {\textbf{K/V cache update}\\
    New clean K/V $\Delta\mathcal{H}^{\text{KV}}_j$\\
    Append to bounded history\\
    $\mathcal{H}^{\text{KV}}_{j+1}$: next-chunk history};
\node[stream stage]
  (decode) at (10.6,-2.65) {\textbf{Causal decode}\\
    Output observations\\Carry temporal state};
\draw[sim flow] (initial.south) -- (inputs.north);
\draw[sim flow] (controls.south) -- (5.3,1.3) -- (1.15,1.3)
  -- ([xshift=1.15cm]inputs.north);
\draw[sim flow] (inputs.east) -- (sample.west);
\draw[sim flow] (sample.east) -- (latent.west);
\draw[sim flow] (inputs.south) -- (0,-2.65) -- (cacheupdate.west);
\coordinate (completed) at (10.6,-1.4);
\draw[semithick] (latent.south) -- (completed);
\draw[sim flow] (completed) -- (decode.north);
\draw[sim flow] (completed) -- (5.3,-1.4) -- (cacheupdate.north);
\fill (completed) circle (1.1pt);
\end{tikzpicture}}
\endgroup

%% file: sections/5__examples.tex
\FloatBarrier

\section{Post-training Example: Cosmos3-Nano-Sim-Bimanual}
\label{sec::bimanual}

\begingroup{}
We demonstrate three models that use the shared causal and distillation recipe, each initialized from Cosmos3-Nano~\citep{cosmos3}.
Cosmos3-Nano-Sim-Bimanual predicts visual dynamics for camera navigation and manipulation action streams.
Cosmos3-Nano-Sim-Transfer generates single-view observations from spatial control streams.
Cosmos3-Nano-Sim-Auto extends video transfer to multiple views with synchronized cameras and LiDAR streams for AV simulation.
All three examples use the same training recipe described in \cref{sec::training} with different interfaces, supervision, and temporal layouts.
For each example model, we explain the architectural choices, data preparation, post-training steps, inference optimization, and results.
\ifhmode\par\fi\endgroup

\begingroup{}
Cosmos3-Nano-Sim-Bimanual uses a shared action interface to predict observations for camera navigation and robot manipulation.
Camera motion changes which parts of a scene are visible, while manipulation also changes the configuration of the robot and nearby objects.
Both require a generator that follows supplied actions while preserving a coherent scene as its own predictions become history.
We specialize Cosmos3-Nano with a common action representation and a mixture of camera, robot, and human videos, then apply the shared causal and distillation curriculum.
We evaluate the 4-step student's visual quality, action-conditioned observation agreement, and streaming cost across these domains.
\ifhmode\par\fi\endgroup

\input{sections/5__examples/bimanual/architecture.tex}
\input{sections/5__examples/bimanual/data.tex}

\input{sections/5__examples/bimanual/post_training.tex}
\input{sections/5__examples/bimanual/optimization.tex}
\input{sections/5__examples/bimanual/results.tex}

\FloatBarrier

\section{Post-training Example: Cosmos3-Nano-Sim-Transfer}
\label{sec::transfer}

\begingroup{}
Closed-loop simulation provides diverse scenarios for training robot policies through repeated interaction.
The simulator updates the physical state in response to policy actions, and rendered observations guide the next action.
Simplified simulator scenes lack the visual richness and diversity of real environments.
\ifhmode\par\fi\endgroup

\begingroup{}
Cosmos3-Nano-Sim-Transfer converts simple simulator outputs into realistic RGB observations using language and spatial controls.
Depth maps, segmentation maps, edges, or blurred control videos encode scene structure, while language specifies appearance.
The generator synthesizes detailed textures, materials, lighting, and backgrounds.
This combination couples simulated physical interactions with realistic visual variation for training robust, generalizable robot policies.
\ifhmode\par\fi\endgroup

\begingroup{}
Autoregressive generation incorporates updated simulator controls and generated RGB history to render successive chunks as the environment changes.
Sampling with 4 denoising steps reduces the denoising work per chunk.
We combine these mechanisms for real-time closed-loop simulation and online robot policy learning.
\ifhmode\par\fi\endgroup

\input{sections/5__examples/transfer/architecture.tex}

\input{sections/5__examples/transfer/data.tex}

\input{sections/5__examples/transfer/post_training.tex}

\input{sections/5__examples/transfer/optimization.tex}
\input{sections/5__examples/transfer/results.tex}

\FloatBarrier

\section{Post-training Example: Cosmos3-Nano-Sim-Auto}
\label{sec::auto}

\input{sections/5__examples/auto/overview.tex}
\input{sections/5__examples/auto/architecture.tex}
\input{sections/5__examples/auto/data.tex}
\input{sections/5__examples/auto/post_training.tex}
\input{sections/5__examples/auto/optimization.tex}
\input{sections/5__examples/auto/results.tex}
\FloatBarrier

%% file: sections/5__examples/bimanual/architecture.tex
\subsection{Architecture}

\begingroup{}
Cosmos3-Nano-Sim-Bimanual predicts RGB observations from an initial image, language description, and a sequence of time-indexed actions.
A single generator covers camera navigation and robot manipulation through the shared action space inherited from Cosmos 3~\citep{cosmos3}.
Domain-specific preprocessing and learned projections accommodate differences in coordinates and effectors, while the visual generator models their consequences.
The generator predicts each observation using the projected command and shared scene history.
\ifhmode\par\fi\endgroup

\paragraph{Action representation.}
\begingroup{}
A single 59-dimensional action vector unifies all control commands.
This vector includes a 9D ego pose, a 24D right wrist and fingertip encoding along with right gripper width, and a 24D left wrist and fingertip encoding along with left gripper width.
Each pose representation consists of 3 translation values and a 6D rotation code~\citep{zhou2019continuity}.
Each 24D hand structure combines a 9D wrist pose with positions of 5 fingertips, each specified by a 3D coordinate relative to the wrist.
\ifhmode\par\fi\endgroup

\begingroup{}
The action pipeline first maps each source's native fields into this 59D layout, converting coordinates, units, and gripper conventions as required.
A validity mask records which fields are observed, so missing measurements can be identified during preprocessing rather than interpreted as physical commands.
The domain-specific normalization applies an inverse hyperbolic sine transform before a learned projection maps the vector into the model's feature space.
The embedded action at time $t$ is
\begin{equation}
 \mathbf{h}^{\text{act}}_t
 = \operatorname{Proj}_{d_t}\!\left(\mathbf{m}_t\odot\operatorname{Norm}_{d_t}(\mathbf{a}_t)\right),
 \label{eq::action_embedding}
\end{equation}
where $\mathbf{a}_t$ is the unified action vector, $d_t$ specifies its domain, $\mathbf{m}_t$ is the validity mask, and $\operatorname{Norm}_{d_t}$ encodes coordinate conversion and normalization.
The domain selects the appropriate projection after the common layout has been constructed.
\ifhmode\par\fi\endgroup

\paragraph{Temporal alignment.}
\begingroup{}
Pose actions use anchors that update every 16 RGB frames.
For an observed pose $\mathbf{T}_t$ in anchor interval $j$, the action encodes $\mathbf{T}_{b(j)}^{-1}\mathbf{T}_t$, where $b(j)$ is the observation immediately preceding that interval.
The corresponding global and frame-relative alternatives encode $\mathbf{T}_0^{-1}\mathbf{T}_t$ and $\mathbf{T}_{t-1}^{-1}\mathbf{T}_t$, respectively.
The interval anchor expresses motion in a common local frame without allowing its displacement to grow over the entire trajectory.
Its 16-frame span defines the pose reference and is independent of the generator's prediction chunk size.
\ifhmode\par\fi\endgroup

\begingroup{}
Groups of 4 consecutive per-frame actions are paired with each latent visual interval (\cref{fig::bimanual_action_timing}).
The causal model predicts chunks of 2 latent frames, corresponding to 8 new RGB frames after startup.
Thus an anchor interval spans 4 latent intervals, or 2 prediction chunks, rather than a single model step.
Actions remain supplied conditions, and the prediction loss applies only to RGB targets.
The initial observation precedes physical motion and is paired with null actions.
This startup condition is semantically different from a later valid zero-motion command, which asks the model to preserve the commanded pose.
\ifhmode\par\fi\endgroup

\begingroup{}
\begin{figure}[!t]
\centering
\input{figures/diagrams/bimanual_action_timing.tex}
\caption{\textbf{Cosmos3-Nano-Sim-Bimanual temporal alignment.}
The leftmost clean latent encodes the initial RGB frame, before any physical action.
After startup, 4 per-frame actions align with each latent visual interval, and 2 latent frames form an 8-RGB-frame prediction chunk.
The 16-RGB-frame pose-anchor interval spans the 4 subsequent latent frames and excludes the initial observation.}
\label{fig::bimanual_action_timing}
\end{figure}
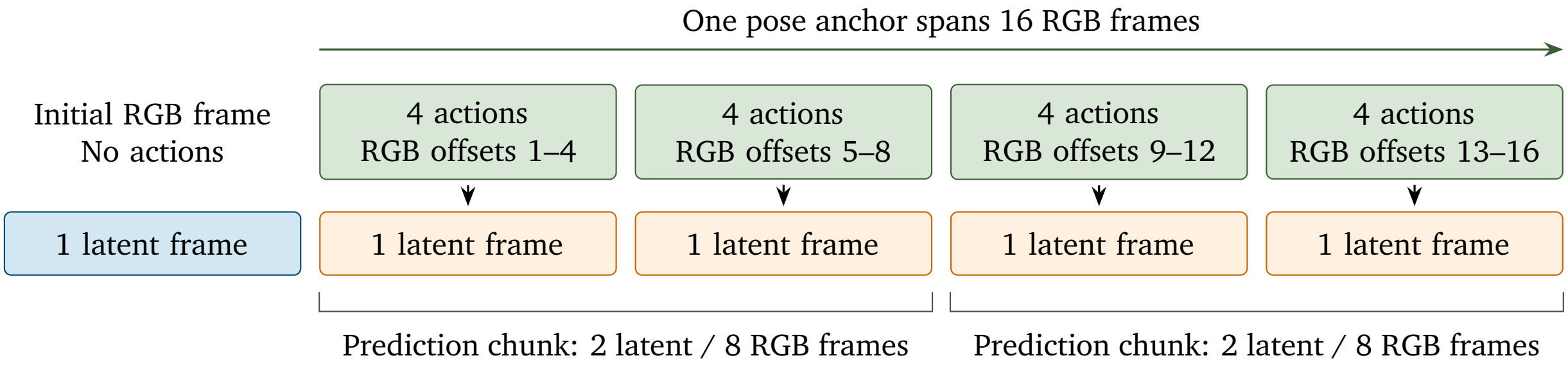
\ifhmode\par\fi\endgroup

\paragraph{Mixed-action sequences.}
\begingroup{}
The interface also permits contiguous navigation and manipulation spans in one action sequence.
Each transition retains its domain tag and validity mask, and the corresponding preprocessing and projection map it into the shared generator's input space.
When the domain changes, generation continues from the retained visual history without loading another model or restarting the rollout.
This implements a persistent session across control modes.

\ifhmode\par\fi\endgroup

%% file: figures/diagrams/bimanual_action_timing.tex
\begingroup
\input{figures/diagrams/architecture_styles.tex}
\fitdiagram{%
\begin{tikzpicture}[font=\fontfamily{bch}\selectfont\small,>=Stealth,
  every node/.append style={text=.},
  bimanual flow/.style={sim flow,shorten <=1.5pt,shorten >=1.5pt}]
\node[align=center,text width=2.7cm] at (0,0) {Initial RGB frame\\No actions};
\node[sim stage,fill=simcleanfill,draw=simcleanedge,text width=2.7cm,minimum height=0.65cm]
  (initial) at (0,-1.15) {1 latent frame};
\foreach \i/\x/\rgb in {1/3.25/1--4,2/6.5/5--8,3/9.75/9--12,4/13.0/13--16} {
  \node[sim stage,fill=simactionfill,draw=simactionedge,text width=2.7cm]
    (action\i) at (\x,0) {4 actions\\\mbox{RGB offsets \rgb}};
  \node[sim stage,fill=simnoisyfill,draw=simnoisyedge,text width=2.7cm,minimum height=0.65cm]
    (latent\i) at (\x,-1.15) {1 latent frame};
  \draw[bimanual flow] (action\i.south) -- (latent\i.north);
}
\foreach \first/\last/\chunk in {1/2/1,3/4/2} {
  \draw[draw=simgrayedge]
    (latent\first.west |- 0,-1.65) -- (latent\first.west |- 0,-1.85)
    -- (latent\last.east |- 0,-1.85) -- (latent\last.east |- 0,-1.65);
  \path (latent\first) -- (latent\last) coordinate[midway] (chunk\chunk);
  \node[align=center] at (chunk\chunk |- 0,-2.2)
    {Prediction chunk: 2 latent / 8 RGB frames};
}
\draw[sim flow,draw=simactionedge]
  (action1.west |- 0,0.85) -- node[above] {One pose anchor spans 16 RGB frames}
  (action4.east |- 0,0.85);
\end{tikzpicture}}
\endgroup

%% file: sections/5__examples/bimanual/data.tex
\subsection{Data Preparation}
\label{sec::bimanual_data}

\paragraph{Sources and sampling.}
\begingroup{}
Training combines camera trajectories, recorded robot demonstrations, and human egocentric interactions (\cref{tab::bimanual_data_sources}).
Camera videos expose the generator to scene changes caused by viewpoint motion, while robot and human videos show effectors interacting with objects.
Egocentric human interaction data broadens the interaction coverage and contributes to robot generalization.
We match RGB observations, actions, and captions by their source video and physical interval.
Each training window aligns the observed motion with the commands supplied for prediction.
\ifhmode\par\fi\endgroup

\begingroup{}
\begin{table}[!t]
\centering\reporttablestyle
\caption{\textbf{Cosmos3-Nano-Sim-Bimanual data sources and conditioning roles.}
All records pair RGB observations with aligned actions and language.
}
\label{tab::bimanual_data_sources}
\begin{tabularx}{\linewidth}{L{4.3cm}YY}
\toprule
\textbf{Source} & \textbf{Action information} & \textbf{Language condition}\\
\midrule
Camera videos, mostly drone and egocentric & Camera poses, supplied or estimated & Environment description independent of camera motion\\
\addlinespace[3pt]
AgiBot~\citep{bu2025agibot} and YAM~\citep{i2rt_yam} & Embodiment-specific robot fields & Rewritten initial-state description\\
\addlinespace[3pt]
Mecka~\citep{mecka} & Human wrist/fingertip and available ego fields & Rewritten initial-state description\\
\addlinespace[3pt]
Behavior-1K~\citep{li2023behavior1k} & Simulated embodiment actions & Supplied task description\\
\bottomrule
\end{tabularx}
\end{table}
\ifhmode\par\fi\endgroup

\paragraph{Filtering and alignment.}
\begingroup{}
For in-the-wild videos without raw pose labels, we use ViPE~\citep{huang2025vipe} to estimate camera trajectories.
We filter out trajectories with low confidence or high jitter by checking pose continuity and smoothness with an empirical threshold.
For robot and human egocentric sequences, we extract long continuous episodes and normalize embodiment-specific fields.
This normalization covers coordinates, moving-base states, and gripper conventions, transforming them into standard actions and validity masks.
Unavailable fields are masked or excluded from the corresponding task.
These checks directly assess whether each supplied action matches the observed transition.
\ifhmode\par\fi\endgroup

\paragraph{Captioning.}
\label{sec::bimanual_captioning}
\begingroup{}
Language describes the scene in which an action takes place, while the action stream specifies the controlled motion.
Captions should supply appearance, spatial relationships, and relevant context without giving a second, potentially conflicting account of the commanded trajectory.
A teacher model first generates a structured scene breakdown and a detailed general caption, then rewrites the caption for the control domain.
The prompts prioritize accuracy and coverage of visible content, instructing the model to generalize or omit uncertain details.
The scene breakdowns and rewritten captions supervise separate in-house captioners for camera navigation, robotics, and human egocentric video.
These captioners annotate the corresponding collections at scale.
\ifhmode\par\fi\endgroup

\begingroup{}
For camera navigation, the caption describes the environment independently of the camera trajectory.
It may describe scene content outside the initial view, but excludes camera motion and apparent scene motion caused by viewpoint changes.
Object appearance, spatial relationships, and independently moving subjects remain useful conditions.
This provides a description of the scene without duplicating the camera trajectory supplied through the action stream.
\ifhmode\par\fi\endgroup

\begingroup{}
For robot and human egocentric training clips, captions must avoid revealing the consequences of future actions.
Removing a sentence about the agent's motion is insufficient if the remaining description gives an object's eventual location or state.
We therefore rewrite the first temporal segment as an initial-state description, retaining the setting, appearance, configuration, and existing contacts while removing controlled actions and their outcomes.
An object already held at the start may be described, while its later displacement or placement is excluded.
The rewrite receives neither later segments nor the video and must not infer an initial state from a removed outcome.
Because the source description was generated from the full video, the rewrite is an initial-state approximation rather than a frame-level visibility annotation.
\Cref{fig::bimanual_caption_rewrite} illustrates this rewrite on an AgiBot-GEAR benchmark example: it retains the slippers' initial location but removes their subsequent grasp and lift.
\ifhmode\par\fi\endgroup

\begingroup{}
\begin{figure}[!t]
\centering
\input{figures/bimanual/caption_initial_state_example.tex}
\caption{\textbf{Initial-state caption rewriting on an AgiBot-GEAR example.}
The initial frame shows the patterned slippers on the floor beside the right cabinet wall.
The excerpts compare the first caption segment with its final rewrite.
Pink shading marks the removed grasp and lift, while the initial location is retained.
Ellipses omit other caption text.
The image is shown for comparison only: the rewriter receives the first-segment text, without the video or later segments.}
\label{fig::bimanual_caption_rewrite}
\end{figure}
\ifhmode\par\fi\endgroup

\begingroup{}
The generator receives the final caption without the intermediate structured scene breakdown.
All four stages use the same final captions for camera, robot, and human videos, preserving the language condition through causal conversion and distillation.
Robot benchmark captions use the training initial-state rewrite (\cref{sec::bimanual_results}).
Behavior-1K uses separately supplied task descriptions, which are available before rollout and express the intended activity without summarizing future observations.
We retain this source-native condition because the caption rewrite is calibrated for recorded robot and human egocentric video.
\ifhmode\par\fi\endgroup

%% file: figures/bimanual/caption_initial_state_example.tex
\begingroup
\setlength{\fboxsep}{5pt}
\begin{minipage}[t]{0.45\linewidth}
\vspace{0pt}
\includegraphics[width=\linewidth]{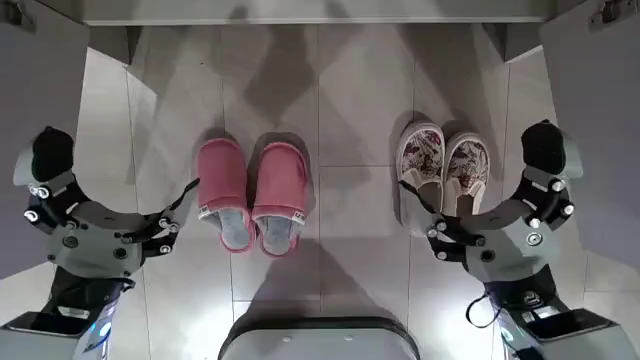}
\par\smallskip
\centering\small Initial frame ($t=0$)
\end{minipage}\hfill
\begin{minipage}[t]{0.52\linewidth}
\vspace{0pt}\small\raggedright
\textbf{Source caption (excerpt)}\par\smallskip
``\ldots a white pair with rounded fronts, white soles, tan interiors, and busy multicolored floral or illustrated uppers sits near the right cabinet wall. \ldots''
\par\smallskip
\colorbox{red!8}{\parbox{\dimexpr\linewidth-2\fboxsep\relax}{\raggedright
``The right gripper moves inward from frame-right, closes on the patterned pair near their upper openings, lifts both together from the tile, \ldots''}}
\par\medskip
\textbf{Final caption (excerpt)}\par\smallskip
``\ldots a white pair with rounded fronts, white soles, tan interiors, and busy multicolored floral or illustrated uppers sits near the right cabinet wall. \ldots''
\end{minipage}
\endgroup

%% file: sections/5__examples/bimanual/post_training.tex
\subsection{Post-training}
\label{sec::bimanual_training}

\paragraph{Training setup.}
\begingroup{}
All four stages use camera, robot, and human interaction data with $2:2:1$ source weights.
Each GPU processes one clip using the action-processing pipeline for its source, and gradients from all GPUs are combined before one optimizer step.
The first three stages sample 256p and 480p clips at a $2:1$ ratio, whereas SF-DMD uses only 480p clips.
Cosmos3-Nano-Sim-Bimanual training clips contain at most 901 RGB frames.
These settings expose the same model to both viewpoint motion and object interaction throughout the curriculum, while the final stage concentrates on the higher resolution tier.
\ifhmode\par\fi\endgroup

\begingroup{}
The causal stages predict 2 latent frames per chunk and retain the full available sequence history without attention sinks.
This chunk size matches the final student's prediction unit, but training history is longer than the bounded K/V window used in the reported streaming benchmarks.
Consequently, the curriculum and final inference configuration differ in both the source of visual history and how much of that history remains visible.
\Cref{tab::bimanual_training_summary} summarizes the selected stage checkpoints.
\ifhmode\par\fi\endgroup

\paragraph{Training curriculum.}
\begingroup{}
Stage 0 initializes the bidirectional teacher from the public Cosmos3-Nano model and adapts it to the action-conditioned data before imposing causal attention (\cref{sec::bidirectional_adaptation}).
The action projection and action-modality embedding use $5\times$ the base learning rate, allowing these action-specific parameters to adapt at a different rate from the inherited generator.
The teacher provides the domain-adapted starting point for causal conversion.
Its selected checkpoint is iteration 35,000.
\ifhmode\par\fi\endgroup

\begingroup{}
Stage 1 applies causal teacher forcing with the shared replay procedure (\cref{sec::teacher_forcing,eq::tf_loss}).
Each prediction uses the supplied actions and earlier recorded RGB history, and the rectified flow loss supervises RGB targets rather than actions.
This establishes the causal prediction problem while keeping the visual context clean.
Using recorded history keeps this stage's context clean, while the subsequent SF-DMD stage addresses stability under generated history.
\ifhmode\par\fi\endgroup

\begingroup{}
Stage 2 uses TF-dCM to distill the causal model to 4 denoising steps while retaining recorded history (\cref{sec::consistency_distillation,eq::tfdcm_loss}).
The teacher uses classifier-free guidance (CFG) with scale 3.
This stage reduces the denoising work within each chunk without changing the source of its preceding RGB context.
It therefore addresses sampling cost separately from the mismatch between recorded training history and generated inference history.
\ifhmode\par\fi\endgroup

\begingroup{}
Stage 3 applies SF-DMD so that the 4-step student learns while continuing from its own generated chunks (\cref{sec::self_forcing}).
Completed predictions become the visual history for subsequent chunks, exposing training to the consequences of earlier prediction errors.
We alternate 1 student update with 5 auxiliary score-model updates, using a teacher with CFG disabled and only 480p training clips.
The selected checkpoint at iteration 1,380 uses the combined phase-update counter, comprising 230 student updates and 1,150 auxiliary score-model updates.
\ifhmode\par\fi\endgroup

\begingroup{}
\begin{table}[!t]
\centering\reporttablestyle
\caption{\textbf{Cosmos3-Nano-Sim-Bimanual post-training settings.}
Iterations identify selected stage checkpoints rather than a shared cumulative counter.
RGB frame limits include the initial frame.
The first three stages mix 256p/480p at $2:1$, and SF-DMD uses 480p only.
SF-DMD learning rates are student / auxiliary score model.
Teacher CFG lists the distillation guidance scale, with scale 1 denoting no CFG.}
\label{tab::bimanual_training_summary}
\begin{tabular}{lccccc}
\toprule
\textbf{Stage} & \textbf{Checkpoint iteration} & \textbf{RGB frame limit} & \textbf{Learning rate} & \textbf{Resolution} & \textbf{Teacher CFG}\\
\midrule
Teacher & 35,000 & 901 & $10^{-5}$ & 256p / 480p & ---\\
TF & 8,000 & 901 & $10^{-5}$ & 256p / 480p & ---\\
TF-dCM & 4,000 & 901 & $2\times10^{-6}$ & 256p / 480p & 3\\
SF-DMD & 1,380 & 901 & $10^{-6}$ / $2\times10^{-7}$ & 480p & 1\\
\bottomrule
\end{tabular}
\end{table}
\ifhmode\par\fi\endgroup

%% file: sections/5__examples/bimanual/optimization.tex
\subsection{Inference Optimization}
\label{sec::bimanual_inference_optimization}

\paragraph{Execution configuration.}
\begingroup{}
Cosmos3-Nano-Sim-Bimanual combines compiled denoising, CUDA-graph replay, attention bounded by a fixed K/V history window, and TAEHV Super decoding using the mechanisms in \cref{sec::runtime_optimization}.
The measured configuration runs on 1 GB300 GPU with 4 denoising steps, a history window of 64 latent frames, and chunks of 2 latent frames decoded into 8 RGB frames at $640\times352$ resolution.
Bounding the visible history controls the attention workload as the session grows, while the lightweight decoder reduces the cost of turning completed latents into observations.
These choices complement few-step sampling because clean-history refresh and decoding remain necessary for continued generation.
\ifhmode\par\fi\endgroup

\paragraph{Measured performance.}
\begingroup{}
\Cref{tab::latency_bimanual} compares two decoders while holding the recorded-action workload and compiled execution preset fixed.
TAEHV Super achieves 48.1 FPS with median and 95th-percentile chunk times of 166 and 169 ms, compared with 35.0 FPS and 229/231 ms for Wan 2.2.
The median reduction is 63 ms, or 27.5\%, and the measured capacity increases by approximately $1.37\times$.
At a 30-FPS playback rate, an 8-frame chunk represents 267 ms of physical time, so both rows have steady generation capacity above the playback rate on this workload.
These measurements include the recorded generation call through decoded output and exclude the first 10 warm-up chunks of each run.
\ifhmode\par\fi\endgroup

\begingroup{}
\begin{table}[!t]
\centering\reporttablestyle
\caption{\textbf{Cosmos3-Nano-Sim-Bimanual decoder runtime on 1 GB300.}
Chunks are 8 decoded frames at 30 FPS (267\,ms of physical time) from 2 latent frames with 4 denoising steps and a K/V window of 64 latent frames.
Chunk wall time covers the recorded generation call, and capacity is 8 frames divided by that time.
All rows use the same recorded-action workload.
Counts pool steady-state chunks after 10 warm-up chunks across the listed runs.
Nsight runs are excluded.}
\label{tab::latency_bimanual}
\begin{tabularx}{\linewidth}{L{6.0cm}CCCCC}
\toprule
\textbf{Configuration} & \multicolumn{2}{c}{\textbf{Chunk wall time (ms)}} & \textbf{Capacity} & \textbf{Chunks} & \textbf{Runs}\\
\cmidrule(lr){2-3}
 & p50 & p95 & (FPS) & & \\
\midrule
\input{figures/latency/data/latency_bimanual_rows.tex}\\
\bottomrule
\end{tabularx}
\end{table}
\ifhmode\par\fi\endgroup

\begingroup{}
The separate instrumented run in \cref{fig::latency_bimanual_stages} attributes approximately 89 ms to denoising and 69 ms to clean-history K/V updates, compared with 4.4 ms for TAEHV Super decoding.
The substantial history-update cost explains why reducing the number of denoising evaluations alone cannot determine the end-to-end speedup.
This profile has a 167-ms median chunk time and is separate from the pooled 166-ms result.
Component medians attribute time to the major stages, while the synchronized chunk timer provides the execution-level measurement.
\ifhmode\par\fi\endgroup

\begingroup{}
\begin{figure}[!t]
\centering
\input{figures/latency/latency_bimanual_compiled_stages.tex}
\caption{\textbf{Cosmos3-Nano-Sim-Bimanual compiled inference on 1 GB300.}
The bar shows denoising, clean-history K/V updates, decoding, and residual time for the compiled configuration with CUDA graphs and TAEHV Super on the recorded-action workload.
Stage medians come from 1 instrumented run with a median chunk time of 167 ms, separate from the pooled 166 ms result in \cref{tab::latency_bimanual}.
Host timers surround device synchronizations.
The stacked components summarize timing attribution rather than an execution timeline.}
\label{fig::latency_bimanual_stages}
\end{figure}
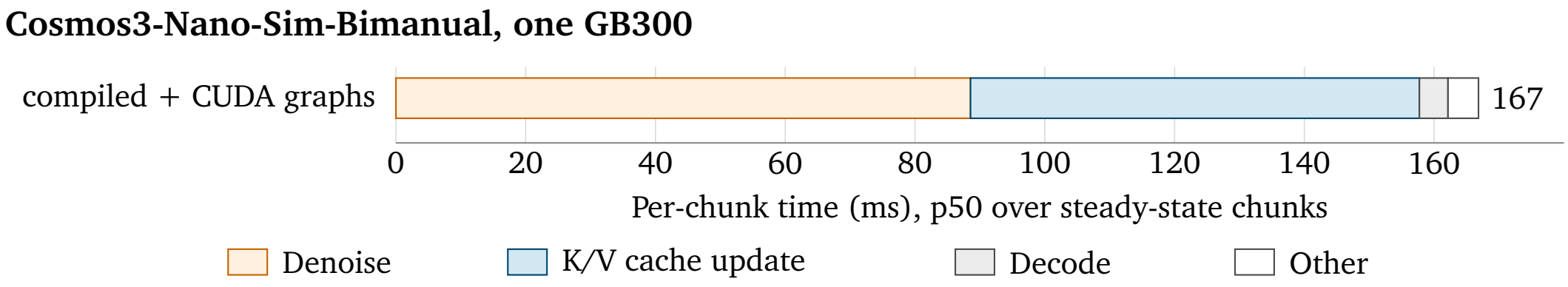
\ifhmode\par\fi\endgroup

\begingroup{}
The decoder comparison shows faster reconstruction, with quality measured on a separate Camera Easy workload.
A separate 80-sample Camera Easy test reports a PAI-Bench-G score of 71.8 with Wan and 70.9 with TAEHV Super.
The quality and runtime results therefore characterize complementary workloads rather than a matched operating point for the final student.
Sustained throughput measures steady generation, while response latency additionally includes action collection, chunk scheduling, startup, and display before the first affected observation appears.
\ifhmode\par\fi\endgroup

\FloatBarrier

%% file: figures/latency/data/latency_bimanual_rows.tex
Compiled preset, TAEHV Super & 166 & 169 & 48.1 & 720 & 3\\
Compiled preset, Wan 2.2 decoder & 229 & 231 & 35.0 & 240 & 1%

%% file: figures/latency/latency_bimanual_compiled_stages.tex
\begingroup
\newcommand{\bimanuallatencycompiled}{}
\input{figures/latency/latency_bimanual_stages.tex}
\endgroup

%% file: sections/5__examples/bimanual/results.tex
\subsection{Results}
\label{sec::bimanual_results}

\begingroup{}
\begin{figure}[!t]
\centering
\input{figures/bimanual/bimanual_camera_trajectories.tex}
\caption{\textbf{Cosmos3-Nano-Sim-Bimanual camera navigation from a shared initial image.}
The first column shows the shared initial image.
Each row then shows frames at 2.5, 5, 7.5, and 10\,s from one rollout conditioned on a different camera trajectory (follow behind, crane up follow, truck left, and truck right).
Insets show the supplied camera trajectory and the camera pose at the displayed frame.}
\label{fig::bimanual_camera_trajectories}
\end{figure}
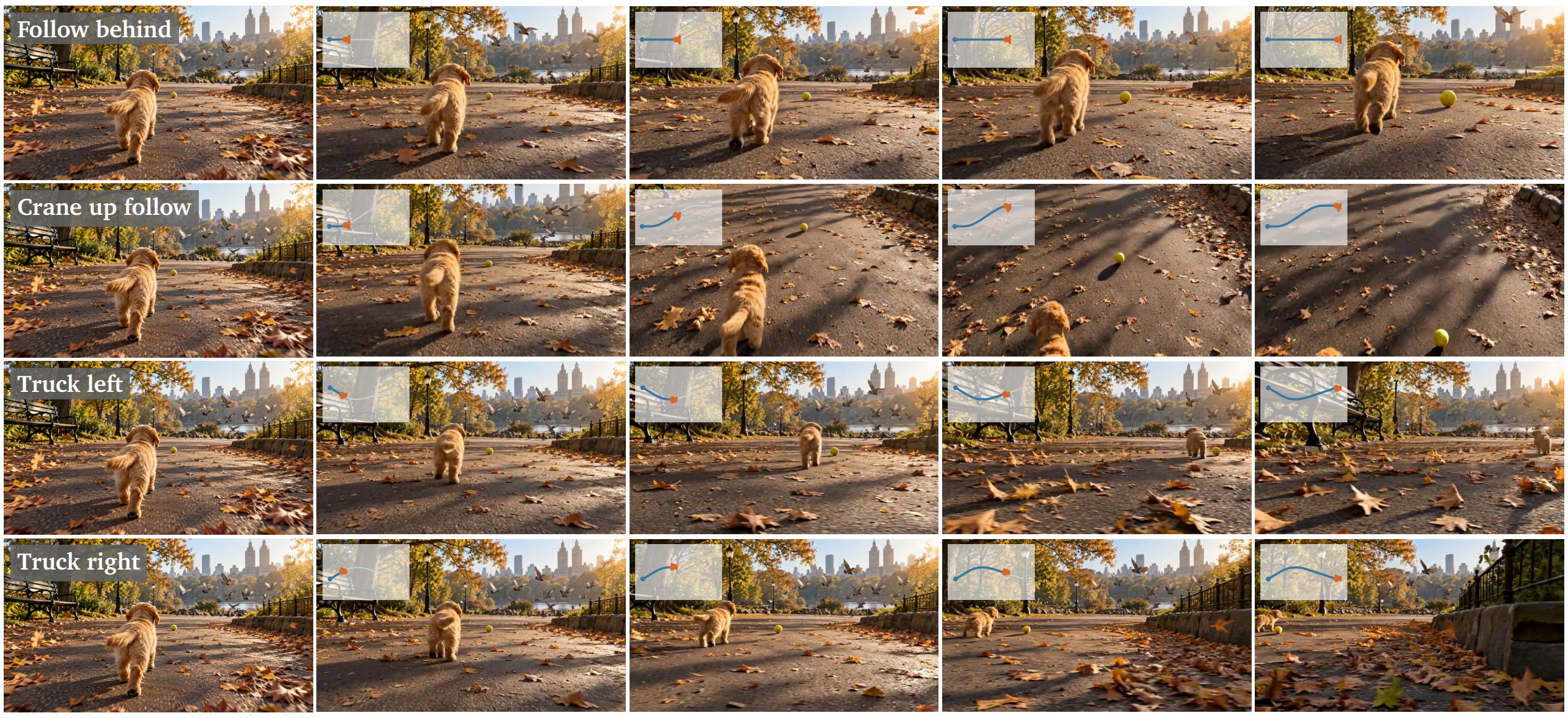
\ifhmode\par\fi\endgroup

\paragraph{Evaluation setup.}
\begingroup{}
\suppressfloats[t]
We evaluate the final SF-DMD student on camera navigation and robot manipulation using an initial image, a scene caption, and supplied actions.
The student uses 4 denoising steps and a 64-latent-frame K/V window with 8 latent frames retained as sinks.
The benchmarks replay recorded or synthesized actions and assess the generated observations as predicted history accumulates.
They therefore test action-conditioned video prediction without a policy choosing subsequent actions from the generated observations.
We compare each baseline as a complete configured generator under its native inference settings.
\ifhmode\par\fi\endgroup

\paragraph{Camera navigation benchmarks.}
\begingroup{}
The navigation suite contains 300 PAI-Bench-G~\citep{zhou2025paibench} image-to-video cases with synthesized camera trajectories, together with 80 simple and 80 hard Sana-WM-Bench~\citep{sanawm2026} cases.
The supplied PAI-Bench-G trajectories span 1 minute.
The Sana-WM-Bench rollouts contain 961 frames at 16 FPS and span 60 seconds.
Each case supplies an initial image, a scene caption independent of camera motion, and the requested trajectory.
We compare with LingBot-World 2.0~\citep{gao2026lingbotworld2} and Sana-WM~\citep{sanawm2026}, which use dedicated camera-conditioning pathways.
The shared interface is tested on sustained viewpoint control across all three navigation populations.
\ifhmode\par\fi\endgroup

\paragraph{Video quality and camera metrics.}
\begingroup{}
Both benchmark tracks use PAI-Bench-G to assess video quality and its change over a rollout.
The visual quality column reports the PAI-Bench-G overall score.
Quality drift is the absolute start/end difference
\begin{equation}
 \Delta M(V)=\left|M(V_{\text{start}})-M(V_{\text{end}})\right|,
\end{equation}
where $M$ is the PAI-Bench-G overall score on equal-duration segments.
We interpret quality drift together with the quality score, since a uniformly poor sequence can also have low drift.

\ifhmode\par\fi\endgroup

\begingroup{}
Camera alignment compares ViPE-estimated poses~\citep{huang2025vipe} from generated videos with supplied trajectories after Umeyama alignment~\citep{umeyama1991alignment}.
Absolute translation error (ATE) measures agreement along the aligned trajectory.
Relative translation error (RTE) and relative rotation error (RRE) measure agreement between pose pairs.
These measurements complement appearance scores by checking whether viewpoint motion follows the supplied control.
\ifhmode\par\fi\endgroup

\begingroup{}
\begin{table}[!t]
\centering\reporttablestyle
\caption{\textbf{Camera navigation benchmark results.}
The benchmark contains 300 PAI-Bench-G~\citep{zhou2025paibench} image-to-video cases and 80 cases each from the simple and hard Sana-WM-Bench sets~\citep{sanawm2026}.
The reported Cosmos3-Nano-Sim-Bimanual checkpoint is the final SF-DMD student, evaluated with 4 denoising steps and a K/V window of 64 latent frames with sink 8.
The Video quality columns report the PAI-Bench-G overall score and its start/end quality drift.
Camera alignment compares ViPE-estimated poses with supplied trajectories after Umeyama alignment.
Bold and underlined values mark the best and second-best distinct values within each benchmark group, with ties sharing the same emphasis.
}
\label{tab::bimanual_camera_benchmarks}
\begin{tabularx}{\linewidth}{L{5.8cm}CCCCC}
\toprule
\multirow{2}{*}{\textbf{Model}} & \multicolumn{2}{c}{\textbf{Video quality}} & \multicolumn{3}{c}{\textbf{Camera alignment}}\\
\cmidrule(lr){2-3}\cmidrule(l){4-6}
& \textbf{Quality $\uparrow$} & \textbf{Drift $\downarrow$} & \textbf{ATE $\downarrow$} & \textbf{RTE $\downarrow$} & \textbf{RRE $\downarrow$}\\
\midrule
\multicolumn{6}{l}{\textit{PAI-Bench-G I2V}}\\
\rowcolor{rowours}
\textbf{Cosmos3-Nano-Sim-Bimanual} & \textbf{73.9099} & \underline{0.0210} & \underline{4.7072} & \textbf{0.0545} & \textbf{0.1002}\\
LingBot-World 2.0~\citep{gao2026lingbotworld2} & \underline{65.3577} & \textbf{0.0209} & \textbf{4.1698} & \underline{0.0596} & 0.1781\\
Sana-WM~\citep{sanawm2026} & 65.2090 & 0.0299 & 6.4823 & 0.1284 & \underline{0.1601}\\
\midrule
\multicolumn{6}{l}{\textit{Sana-WM-Bench simple}}\\
\rowcolor{rowours}
\textbf{Cosmos3-Nano-Sim-Bimanual} & \textbf{77.2887} & \textbf{0.0118} & \textbf{1.0040} & \textbf{0.0124} & \textbf{0.0551}\\
LingBot-World 2.0~\citep{gao2026lingbotworld2} & 65.9748 & 0.0371 & 1.4254 & \underline{0.0185} & 0.2591\\
Sana-WM~\citep{sanawm2026} & \underline{66.0599} & \underline{0.0223} & \underline{1.2159} & 0.0192 & \underline{0.2356}\\
\midrule
\multicolumn{6}{l}{\textit{Sana-WM-Bench hard}}\\
\rowcolor{rowours}
\textbf{Cosmos3-Nano-Sim-Bimanual} & \textbf{76.3971} & \textbf{0.0161} & \textbf{1.3085} & \textbf{0.0161} & \textbf{0.1774}\\
LingBot-World 2.0~\citep{gao2026lingbotworld2} & 64.5462 & 0.0372 & 1.3423 & \textbf{0.0161} & 0.3264\\
Sana-WM~\citep{sanawm2026} & \underline{65.5699} & \underline{0.0251} & \underline{1.3165} & \underline{0.0178} & \underline{0.1826}\\
\bottomrule
\end{tabularx}
\end{table}
\ifhmode\par\fi\endgroup

\paragraph{Camera navigation results.}
\begingroup{}
The student's quality scores exceed the strongest listed baseline by 8.55 points on PAI-Bench-G, 11.23 on Sana-WM-Bench simple, and 10.83 on Sana-WM-Bench hard (\cref{tab::bimanual_camera_benchmarks}).
On both Sana-WM-Bench sets, these gains coincide with lower quality drift and closer trajectory agreement, with an RTE tie on the hard set.
Thus the quality advantage on these cases is accompanied by stronger camera following rather than obtained by ignoring the requested viewpoint motion.
The same checkpoint supports these camera-conditioned rollouts alongside the robot results below.
\ifhmode\par\fi\endgroup

\begingroup{}
PAI-Bench-G exposes a more qualified tradeoff.
The student has lower RTE and RRE than the listed baselines, but its ATE is 4.7072 compared with LingBot-World 2.0's 4.1698, and its quality drift is 0.0210 compared with 0.0209.
These results combine stronger appearance and local motion agreement with a modest ATE and endpoint-drift tradeoff.
The tradeoff may reflect generated geometry, trajectory following, pose estimation, or a combination of these factors.
\ifhmode\par\fi\endgroup

\paragraph{Robot manipulation benchmarks.}
\begingroup{}
The robot benchmarks use validation videos from AgiBot-World and AgiBot-GEAR~\citep{bu2025agibot}.
AgiBot-World contributes 232 windows from 186 source episodes, and AgiBot-GEAR contributes 100 windows from 30 episodes.
Each window contains 901 observations at 30 FPS, spanning 30 seconds of transitions under supplied robot actions.
The initial image, caption, and actions define the observation-prediction problem.
The DreamDojo~\citep{gao2026dreamdojo} 2B and 14B bidirectional teachers provide external comparisons with different backbones, training data, action adapters, and inference settings.
\ifhmode\par\fi\endgroup

\paragraph{Robot manipulation metrics.}
\begingroup{}
Video quality and drift use the same PAI-Bench-G definitions as the navigation track.
Reference-frame agreement measures whether the generated observations match recorded observations under the supplied actions.
Peak signal-to-noise ratio (PSNR, dB) measures pixel agreement, structural similarity (SSIM) measures structural image agreement~\citep{wang2004ssim}, and learned perceptual image patch similarity (LPIPS) measures perceptual feature distance~\citep{zhang2018lpips}.
These image metrics quantify action-conditioned observation agreement.
\ifhmode\par\fi\endgroup

\begingroup{}
\begin{table}[!t]
\centering\reporttablestyle
\caption{\textbf{Robot manipulation benchmark results.}
The benchmark contains 232 AgiBot-World~\citep{bu2025agibot} windows from 186 source episodes and 100 AgiBot-GEAR windows from 30 source episodes.
The reported Cosmos3-Nano-Sim-Bimanual checkpoint is the final SF-DMD student, evaluated with 4 denoising steps and a K/V window of 64 latent frames with sink 8.
The Video quality group reports the PAI-Bench-G overall score and its start/end quality-drift diagnostic.
PSNR, SSIM, and LPIPS measure reference-frame agreement under supplied actions.
Bold marks the best and underlining the second-best value within each benchmark group.
}
\label{tab::bimanual_robot_benchmarks}
\begin{tabularx}{\linewidth}{L{5.0cm}CCCCC}
\toprule
\multirow{2}{*}{\textbf{Model}} & \multicolumn{2}{c}{\textbf{Video quality}} & \multicolumn{3}{c}{\textbf{Reference-frame agreement}}\\
\cmidrule(lr){2-3}\cmidrule(l){4-6}
& \textbf{Quality $\uparrow$} & \textbf{Drift $\downarrow$} & \textbf{PSNR (dB) $\uparrow$} & \textbf{SSIM $\uparrow$} & \textbf{LPIPS $\downarrow$}\\
\midrule
\multicolumn{6}{l}{\textit{AgiBot-World}}\\
\rowcolor{rowours}
\textbf{Cosmos3-Nano-Sim-Bimanual} & \textbf{76.0884} & \textbf{0.0118} & \textbf{14.4138} & \textbf{0.6839} & \textbf{0.3454}\\
DreamDojo 2B~\citep{gao2026dreamdojo} & 74.5327 & 0.0149 & 12.1016 & 0.6405 & 0.3766\\
DreamDojo 14B~\citep{gao2026dreamdojo} & \underline{74.9390} & \underline{0.0133} & \underline{12.3561} & \underline{0.6635} & \underline{0.3586}\\
\midrule
\multicolumn{6}{l}{\textit{AgiBot-GEAR}}\\
\rowcolor{rowours}
\textbf{Cosmos3-Nano-Sim-Bimanual} & \textbf{74.4912} & \textbf{0.0188} & \textbf{14.0158} & \textbf{0.6172} & \textbf{0.4057}\\
DreamDojo 2B~\citep{gao2026dreamdojo} & 72.4586 & 0.0232 & 11.5709 & 0.5437 & 0.4638\\
DreamDojo 14B~\citep{gao2026dreamdojo} & \underline{73.0033} & \underline{0.0218} & \underline{12.1113} & \underline{0.5818} & \underline{0.4326}\\
\bottomrule
\end{tabularx}
\end{table}
\ifhmode\par\fi\endgroup

\begingroup{}
\begin{figure}[!t]
\centering
\input{figures/bimanual/bimanual_robot_comparison.tex}
\caption{\textbf{Robot manipulation comparison.}
Rows show the reference video, DreamDojo 2B, DreamDojo 14B, and Cosmos3-Nano-Sim-Bimanual from the same initial image and robot actions.
Columns show frames at 0, 10, 20, 25.9, and 28\,s, progressing from the initial view through base motion, tissue grasp, motion over the bin, and the final sampled view.}
\label{fig::bimanual_robot_comparison}
\end{figure}
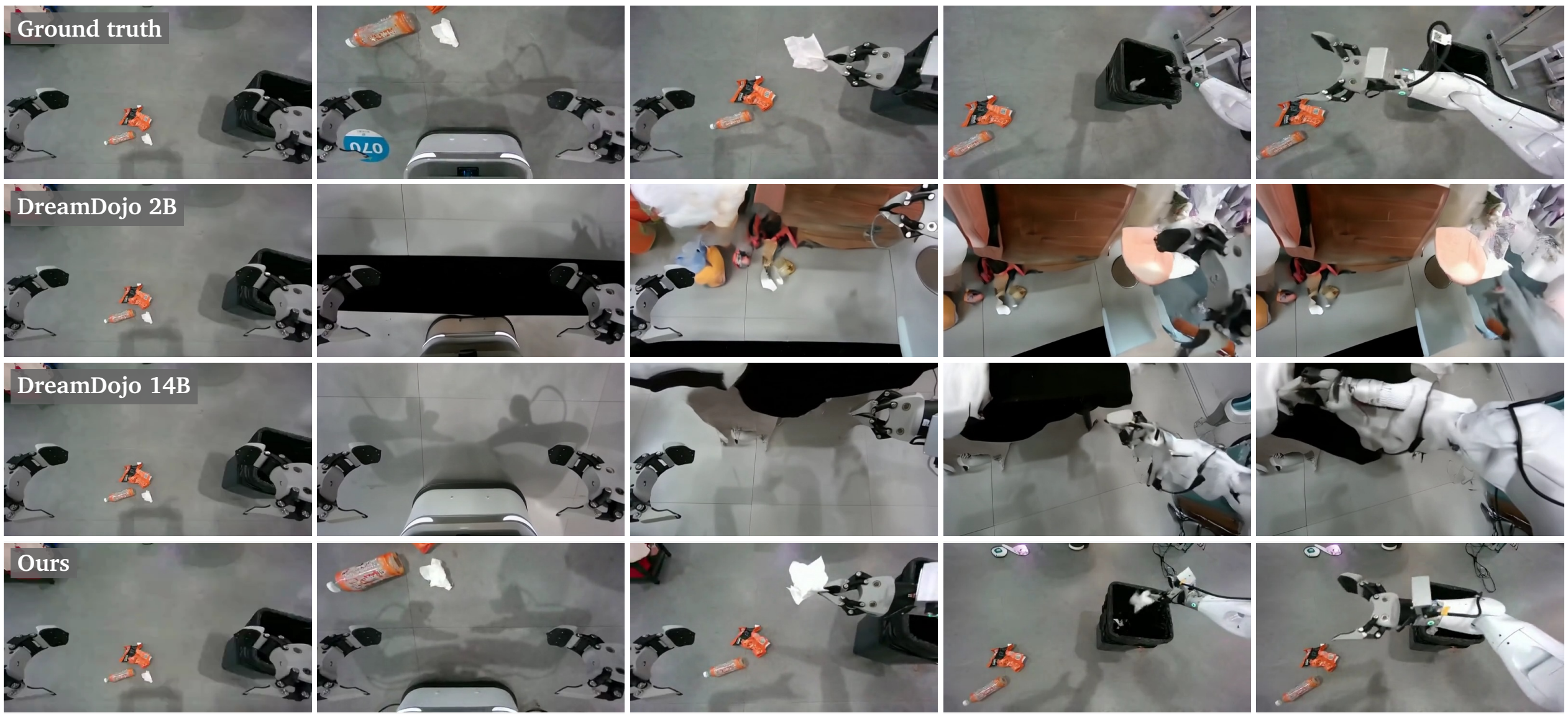
\ifhmode\par\fi\endgroup

\begingroup{}
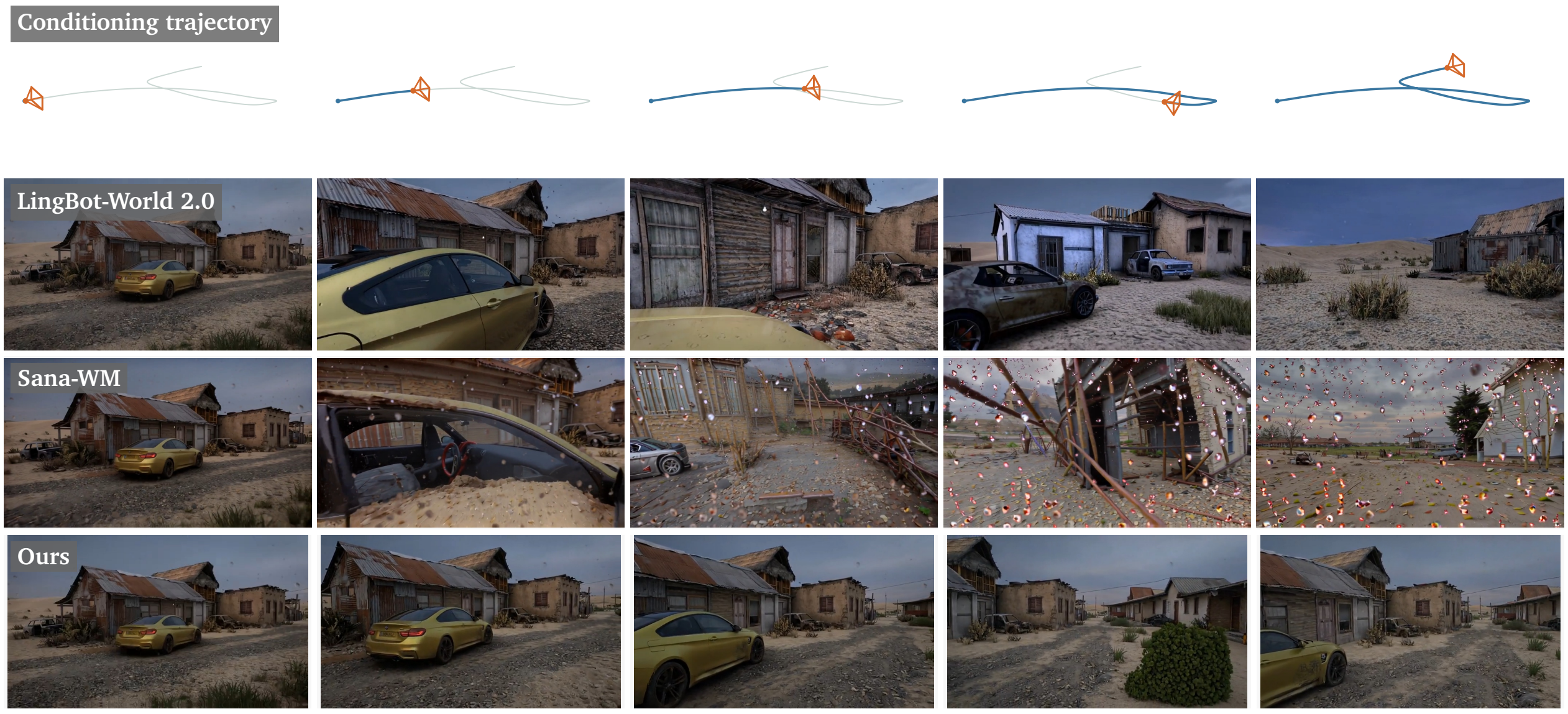
\begin{figure}[!t]
\centering
\input{figures/bimanual/bimanual_camera_comparison.tex}
\caption{\textbf{Camera navigation comparison.}
The top row shows the supplied camera trajectory and the current camera pose.
The remaining rows show LingBot-World 2.0, Sana-WM, and Cosmos3-Nano-Sim-Bimanual from the same initial image and trajectory.
Columns show positions at 0.125, 10, 20, 40, and 59\,s along the looping trajectory, including a later revisit to the building and car.}
\label{fig::bimanual_camera_comparison}
\end{figure}
\ifhmode\par\fi\endgroup

\paragraph{Robot manipulation results.}
\begingroup{}
The student improves all five reported metrics on both AgiBot benchmarks relative to the listed DreamDojo teachers (\cref{tab::bimanual_robot_benchmarks}).
Against DreamDojo 14B, PSNR increases by 2.06 dB on World and 1.90 dB on GEAR, while LPIPS decreases by approximately 3.7\% and 6.2\%, respectively.
The agreement across quality, pixel, structural, and perceptual metrics supports more faithful observation prediction under the supplied actions, rather than an appearance gain alone.
The lower quality drift also indicates a smaller difference between the beginning and end of these rollouts.
\ifhmode\par\fi\endgroup

\begingroup{}
These findings extend the navigation results to observations that depend on robot and object motion.
A shared checkpoint can cover both control domains without selecting a separate visual generator for each, which is the evaluated benefit of the common interface.
The evaluation therefore demonstrates action-conditioned observation prediction across both control domains with one visual generator.
\ifhmode\par\fi\endgroup

\paragraph{Recipe-stage results.}
\begingroup{}
To examine where the final student's behavior emerges, \cref{tab::bimanual_stage_results} compares all four stages on the same 30-case AgiBot-GEAR subset with matched initial images, actions, and source captions.
This subset is chosen from the full 100-window GEAR evaluation.
The teacher and causal TF use 35 denoising steps, while TF-dCM and SF-DMD use 4, retaining each stage's native inference configuration.
Causal conversion reduces the quality score from 73.385 to 72.367 and PSNR from 13.452 to 12.482 dB.
TF-dCM with 4 denoising steps retains a similar quality score of 72.375, but PSNR falls further to 11.847 dB.
Together, the metrics show how causal conversion and distillation affect both appearance and reference-frame agreement.
\ifhmode\par\fi\endgroup

\begingroup{}
SF-DMD recovers much of this loss while retaining 4-step generation.
Relative to TF-dCM, quality increases by 1.876 points, PSNR by 1.632 dB, and SSIM by 0.064, while LPIPS decreases by 0.065.
The final means approach the bidirectional teacher's reference agreement: PSNR is 13.479 versus 13.452 dB, SSIM is 0.589 versus 0.591, and LPIPS is 0.447 versus 0.454.
The quality score is also higher, at 74.251 versus 73.385.
These results are consistent with the benefit of training on generated history after distillation, as in the shared recipe.
Because this transition also adds training, switches to 480p-only clips, and changes other stage settings, we interpret the gain as the effect of the complete SF-DMD stage.
The teacher and student have similar means on these 30 cases, and paired uncertainty estimates are not available for the small differences.
Reducing 35 denoising steps to 4 lowers denoising work, while history refresh and decoding remain in the~inference~path.
\ifhmode\par\fi\endgroup

\begingroup{}
\begin{table}[!t]
\centering\reporttablestyle
\caption{\textbf{Cosmos3-Nano-Sim-Bimanual stage ablation on an AgiBot-GEAR subset.}
All stages use the same 30 cases, initial images, actions, and source captions, separate from the full GEAR benchmark.
Steps give each stage's denoising evaluations per prediction.
Quality uses the PAI-Bench-G overall score, and PSNR, SSIM, and LPIPS measure reference-frame agreement.
Bold and underlined values mark the best and second-best means within each metric.
}
\label{tab::bimanual_stage_results}
\begin{tabular}{lccccc}
\toprule
\multirow{2}{*}{\textbf{Stage}} & \multirow{2}{*}{\textbf{Steps}} & \multirow{2}{*}{\textbf{Quality $\uparrow$}} & \multicolumn{3}{c}{\textbf{Reference-frame agreement}}\\
\cmidrule(l){4-6}
& & & \textbf{PSNR (dB) $\uparrow$} & \textbf{SSIM $\uparrow$} & \textbf{LPIPS $\downarrow$}\\
\midrule
Bidirectional & 35 & \underline{73.385} & \underline{13.452} & \textbf{0.591} & \underline{0.454}\\
Causal TF & 35 & 72.367 & 12.482 & 0.502 & 0.516\\
TF-dCM & 4 & 72.375 & 11.847 & 0.525 & 0.512\\
\rowcolor{rowours}
SF-DMD & 4 & \textbf{74.251} & \textbf{13.479} & \underline{0.589} & \textbf{0.447}\\
\bottomrule
\end{tabular}
\end{table}
\ifhmode\par\fi\endgroup

\begingroup{}
\begin{figure*}[!t]
\centering
\input{figures/bimanual/mixed_action_rollout.tex}
\caption{\textbf{Mixed manipulation and navigation rollout.}
Frames from a 30-second rollout are shown from top to bottom.
Here, the action sequence begins with 10 seconds of manipulation, followed by 10 seconds of camera navigation and another 10 seconds of manipulation.
During navigation, the camera slowly turns right and then left before returning to its original pose.
The robot arms remain stationary while the camera moves, after which manipulation resumes using the accumulated visual history.}
\label{fig::bimanual_mixed_action}
\end{figure*}
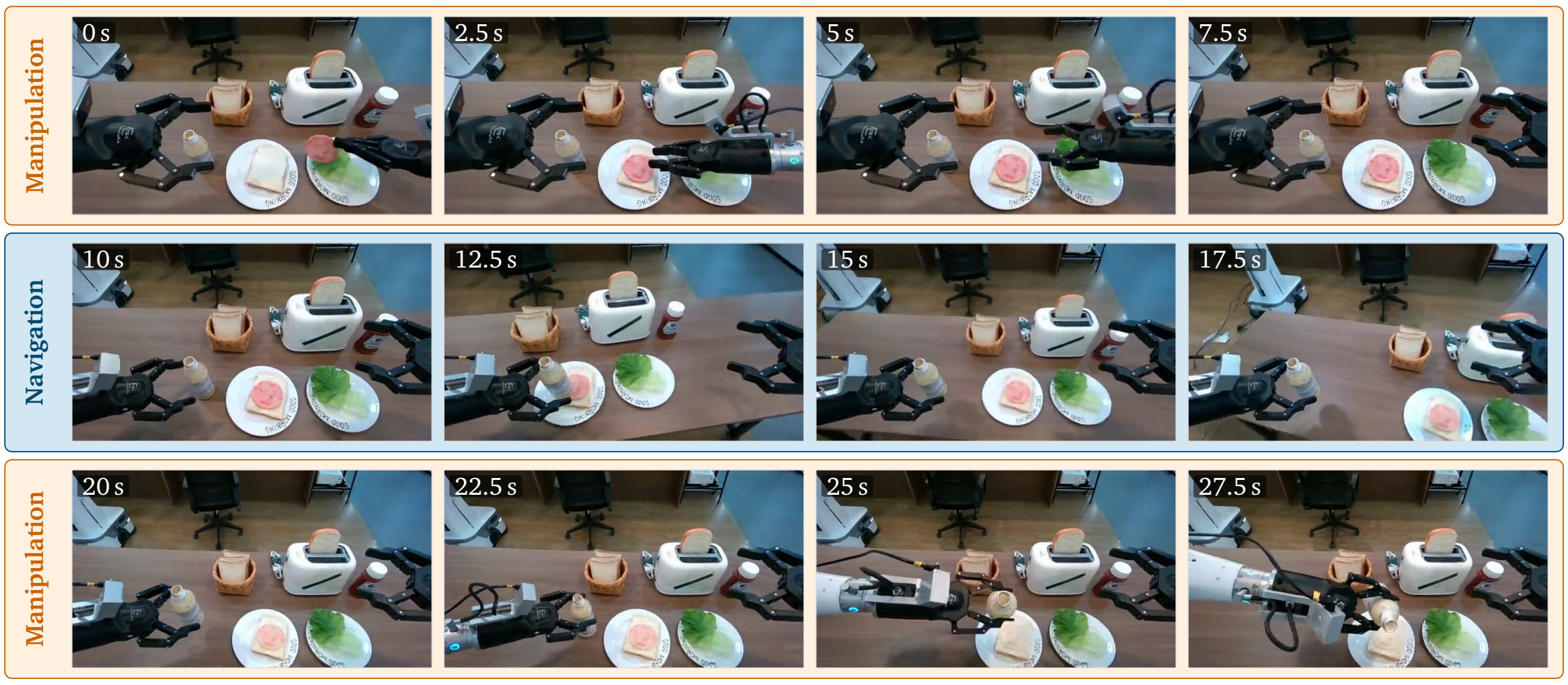
\ifhmode\par\fi\endgroup

\paragraph{Qualitative results.}
\begingroup{}
\Cref{fig::bimanual_camera_trajectories} shows rollouts from one initial image under four camera trajectories.
The rollouts diverge according to the supplied trajectory: the camera stays behind the puppy, rises and looks down at the path and the ball, or moves laterally so that the puppy moves across the frame and the bench or railing enters the view.
These selected rollouts visualize trajectory-dependent viewpoint control across the four requested motions.
The robot comparison in \cref{fig::bimanual_robot_comparison} follows a sequence from base motion to grasping a tissue and moving it toward a bin.
The third sampled column shows the tissue at the gripper in the reference and our rollout, and the fourth shows the arm over the bin.
The baselines change the floor and surrounding scene after the base motion, while our rollout retains more of the reference layout through the subsequent manipulation.
Together, the frames illustrate visual continuity across viewpoint and object changes and complement the quantitative results above.
\ifhmode\par\fi\endgroup

\begingroup{}
The looping camera example in \cref{fig::bimanual_camera_comparison} complements the forward trajectory metrics by revisiting previously visible content.
At the last sampled view, our rollout retains the yellow car and wooden building, while LingBot-World 2.0 shows a changed scene and Sana-WM contains substantial artifacts.
This example illustrates scene continuity during a revisit within the active history window.
\ifhmode\par\fi\endgroup

\begingroup{}
Building on the manipulation and navigation capabilities demonstrated above, \cref{fig::bimanual_mixed_action} combines both control modes in a single continuous rollout, switching from robot manipulation to camera navigation and back to robot manipulation without resetting its visual history.
\ifhmode\par\fi\endgroup

\paragraph{Evaluation scope.}
\begingroup{}
The benchmarks evaluate action-conditioned observation prediction from recorded or synthesized action sequences.
The camera metrics quantify trajectory following, while the reference-frame metrics quantify agreement with a recorded continuation.
Contact mechanics, task success, and closed-loop policy behavior require dedicated evaluation.
Because the baselines retain their native configurations and the training stages change several settings together, the comparisons characterize complete configured systems and stage recipes.
The qualitative rollouts provide selected examples, and the benchmark tables provide the population-level measurements reported here.
\ifhmode\par\fi\endgroup

%% file: figures/bimanual/bimanual_camera_trajectories.tex
\begingroup
\def\CosmosFigureAssets{figures/bimanual/bimanual_camera_trajectories}%
\resizebox{\linewidth}{!}{\input{figures/bimanual/bimanual_camera_trajectories/figure.tex}}%
\endgroup

%% file: figures/bimanual/bimanual_robot_comparison.tex
\begingroup
\def\CosmosFigureAssets{figures/bimanual/bimanual_robot_comparison}%
\resizebox{\linewidth}{!}{\input{figures/bimanual/bimanual_robot_comparison/figure.tex}}%
\endgroup

%% file: figures/bimanual/bimanual_camera_comparison.tex
\begingroup
\def\CosmosFigureAssets{figures/bimanual/bimanual_camera_comparison}%
\resizebox{\linewidth}{!}{\input{figures/bimanual/bimanual_camera_comparison/figure.tex}}%
\endgroup

%% file: figures/bimanual/mixed_action_rollout.tex
\begingroup
\definecolor{mixednavigationfill}{HTML}{D2E7F2}
\definecolor{mixednavigationedge}{HTML}{00476D}
\definecolor{mixedmanipulationfill}{HTML}{FFF0E0}
\definecolor{mixedmanipulationedge}{HTML}{C76300}
\definecolor{mixedframeedge}{HTML}{D9DDDF}
\begin{tikzpicture}[
  x=1cm,
  y=1cm,
  font=\fontfamily{bch}\selectfont,
  frame/.style={inner sep=0pt, outer sep=0pt, anchor=north west},
  stamp/.style={
    anchor=north west,
    inner xsep=1.4pt,
    inner ysep=0.6pt,
    rounded corners=0.8pt,
    fill=black,
    fill opacity=0.55,
    text opacity=1,
    text=white,
    font=\fontfamily{bch}\selectfont\scriptsize
  },
  rowlabel/.style={
    rotate=90,
    anchor=center,
    inner sep=0pt,
    font=\fontfamily{bch}\selectfont\scriptsize\bfseries
  }
]
  \def\framew{3.54}
  \def\frameh{1.947}
  \def\rowstep{2.23}
  \def\panelright{15.33}

  \filldraw[fill=mixedmanipulationfill, draw=mixedmanipulationedge, line width=0.35pt, rounded corners=2pt]
    (0,0.10) rectangle (\panelright,-2.05);
  \node[rowlabel, text=mixedmanipulationedge] at (0.31,-0.974) {Manipulation};
  \begingroup

    \node[frame] at (0.66,0) {\includegraphics[width=3.54cm,height=1.947cm]{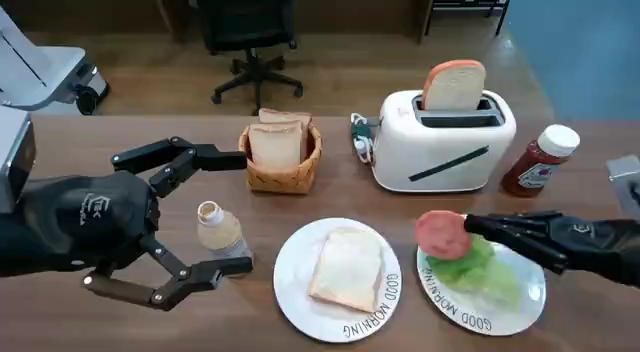}};
    \draw[mixedframeedge, line width=0.25pt] (0.66,0) rectangle +(\framew,-\frameh);
    \node[stamp] at (0.66+0.045,-0.045) {0\,s};
  
\endgroup

\begingroup

    \node[frame] at (4.32,0) {\includegraphics[width=3.54cm,height=1.947cm]{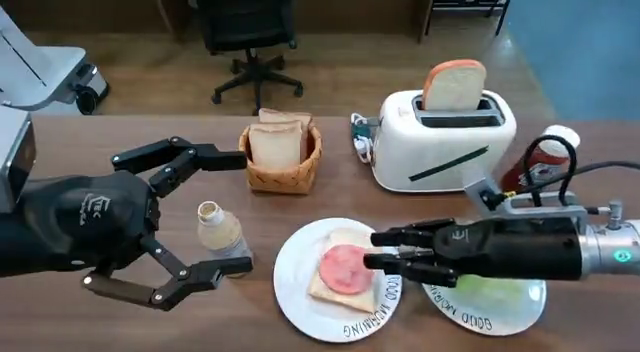}};
    \draw[mixedframeedge, line width=0.25pt] (4.32,0) rectangle +(\framew,-\frameh);
    \node[stamp] at (4.32+0.045,-0.045) {2.5\,s};
  
\endgroup

\begingroup

    \node[frame] at (7.98,0) {\includegraphics[width=3.54cm,height=1.947cm]{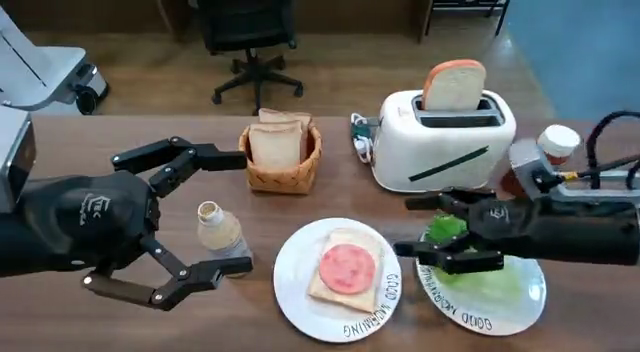}};
    \draw[mixedframeedge, line width=0.25pt] (7.98,0) rectangle +(\framew,-\frameh);
    \node[stamp] at (7.98+0.045,-0.045) {5\,s};
  
\endgroup

\begingroup

    \node[frame] at (11.64,0) {\includegraphics[width=3.54cm,height=1.947cm]{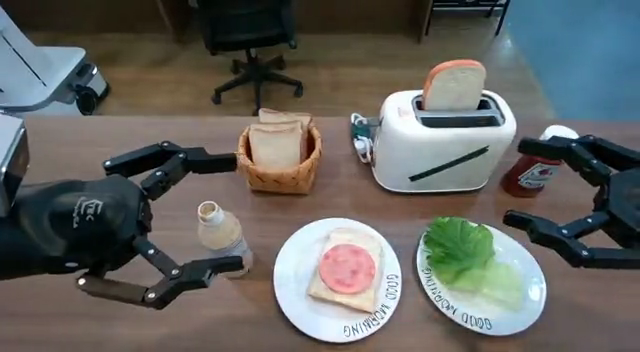}};
    \draw[mixedframeedge, line width=0.25pt] (11.64,0) rectangle +(\framew,-\frameh);
    \node[stamp] at (11.64+0.045,-0.045) {7.5\,s};
  
\endgroup

  \filldraw[fill=mixednavigationfill, draw=mixednavigationedge, line width=0.35pt, rounded corners=2pt]
    (0,-\rowstep+0.10) rectangle (\panelright,-\rowstep-2.05);
  \node[rowlabel, text=mixednavigationedge] at (0.31,-\rowstep-0.974) {Navigation};
  \begingroup

    \node[frame] at (0.66,-\rowstep) {\includegraphics[width=3.54cm,height=1.947cm]{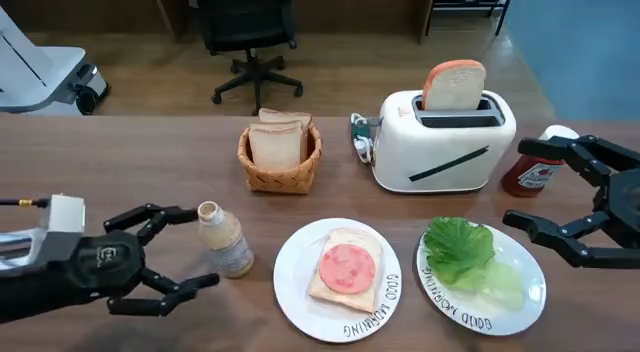}};
    \draw[mixedframeedge, line width=0.25pt] (0.66,-\rowstep) rectangle +(\framew,-\frameh);
    \node[stamp] at (0.66+0.045,-\rowstep-0.045) {10\,s};
  
\endgroup

\begingroup

    \node[frame] at (4.32,-\rowstep) {\includegraphics[width=3.54cm,height=1.947cm]{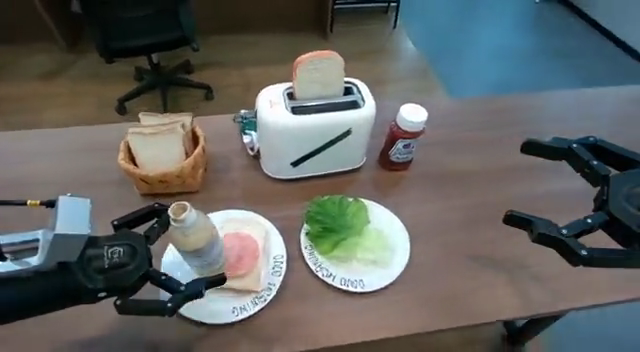}};
    \draw[mixedframeedge, line width=0.25pt] (4.32,-\rowstep) rectangle +(\framew,-\frameh);
    \node[stamp] at (4.32+0.045,-\rowstep-0.045) {12.5\,s};
  
\endgroup

\begingroup

    \node[frame] at (7.98,-\rowstep) {\includegraphics[width=3.54cm,height=1.947cm]{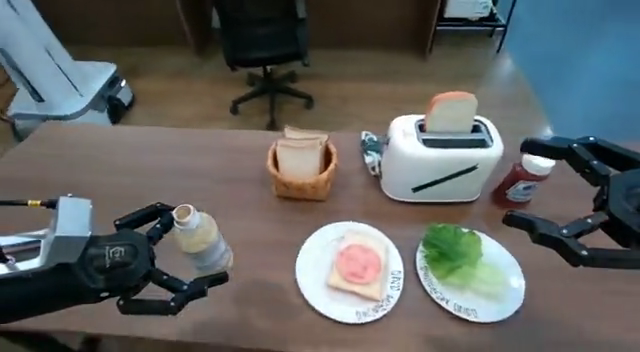}};
    \draw[mixedframeedge, line width=0.25pt] (7.98,-\rowstep) rectangle +(\framew,-\frameh);
    \node[stamp] at (7.98+0.045,-\rowstep-0.045) {15\,s};
  
\endgroup

\begingroup

    \node[frame] at (11.64,-\rowstep) {\includegraphics[width=3.54cm,height=1.947cm]{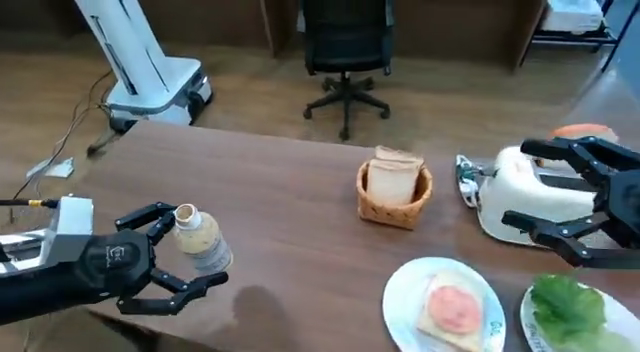}};
    \draw[mixedframeedge, line width=0.25pt] (11.64,-\rowstep) rectangle +(\framew,-\frameh);
    \node[stamp] at (11.64+0.045,-\rowstep-0.045) {17.5\,s};
  
\endgroup

  \filldraw[fill=mixedmanipulationfill, draw=mixedmanipulationedge, line width=0.35pt, rounded corners=2pt]
    (0,-2*\rowstep+0.10) rectangle (\panelright,-2*\rowstep-2.05);
  \node[rowlabel, text=mixedmanipulationedge] at (0.31,-2*\rowstep-0.974) {Manipulation};
  \begingroup

    \node[frame] at (0.66,-2*\rowstep) {\includegraphics[width=3.54cm,height=1.947cm]{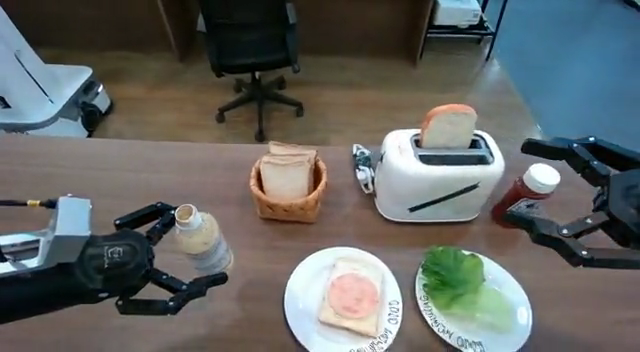}};
    \draw[mixedframeedge, line width=0.25pt] (0.66,-2*\rowstep) rectangle +(\framew,-\frameh);
    \node[stamp] at (0.66+0.045,-2*\rowstep-0.045) {20\,s};
  
\endgroup

\begingroup

    \node[frame] at (4.32,-2*\rowstep) {\includegraphics[width=3.54cm,height=1.947cm]{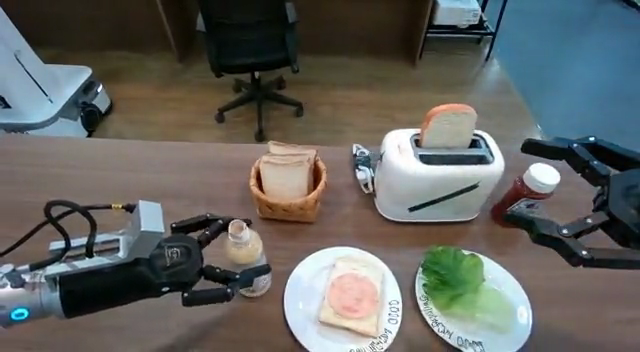}};
    \draw[mixedframeedge, line width=0.25pt] (4.32,-2*\rowstep) rectangle +(\framew,-\frameh);
    \node[stamp] at (4.32+0.045,-2*\rowstep-0.045) {22.5\,s};
  
\endgroup

\begingroup

    \node[frame] at (7.98,-2*\rowstep) {\includegraphics[width=3.54cm,height=1.947cm]{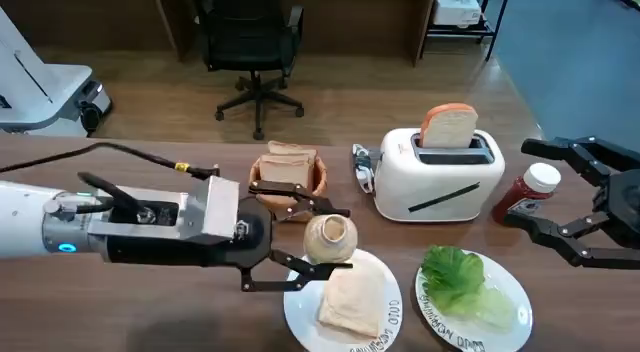}};
    \draw[mixedframeedge, line width=0.25pt] (7.98,-2*\rowstep) rectangle +(\framew,-\frameh);
    \node[stamp] at (7.98+0.045,-2*\rowstep-0.045) {25\,s};
  
\endgroup

\begingroup

    \node[frame] at (11.64,-2*\rowstep) {\includegraphics[width=3.54cm,height=1.947cm]{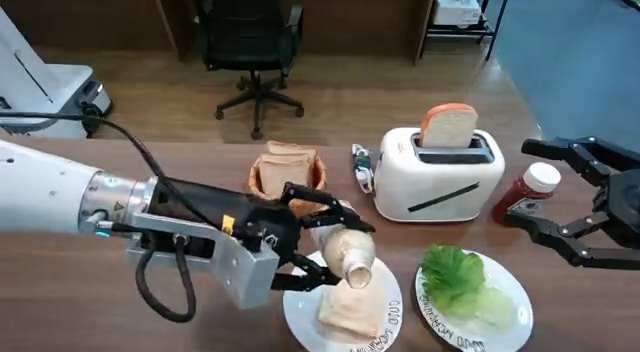}};
    \draw[mixedframeedge, line width=0.25pt] (11.64,-2*\rowstep) rectangle +(\framew,-\frameh);
    \node[stamp] at (11.64+0.045,-2*\rowstep-0.045) {27.5\,s};
  
\endgroup

\end{tikzpicture}
\endgroup

%% file: sections/5__examples/transfer/architecture.tex
\subsection{Architecture}

\begingroup{}
Cosmos3-Nano-Sim-Transfer uses the shared architecture (\cref{sec::shared_architecture}) to generate RGB observations from language and spatial controls supplied by an external simulator.
The simulator evolves the physical state, while the generator determines the appearance of the observations.
Each sample supplies 1 control modality: depth, segmentation, edges, or blur.
Depth constrains geometry, segmentation constrains region identity, edges constrain contours, and blur retains coarse appearance and layout.
\ifhmode\par\fi\endgroup

\begingroup{}
The reasoner (text) tower encodes the tokenized prompt into contextual language representations using causal self-attention.
The attention mask allows both control and RGB queries to attend to language keys and values at every MoT layer throughout training and inference.
During autoregressive inference, cached language keys and values preserve this conditioning across denoising steps and output chunks.
\ifhmode\par\fi\endgroup

\begingroup{}
The generator draws on language, spatial controls, and RGB history to render each new observation.
The prompt describes the scene and its desired appearance, while the controls specify the spatial structure that the observation should follow.
Clean RGB history gives the model a visual reference for carrying the scene's appearance forward as the controls change.
Control videos and RGB targets pass through the same causal Wan2.2 vision tokenizer and input projection.
The tokenizer compresses time by 4 and each spatial dimension by 16~\citep{cosmos3}, so the transformer operates on compact video latents.
\ifhmode\par\fi\endgroup

\begingroup{}
To render each RGB chunk, the model uses its output projection to predict a rectified flow velocity $\mathbf{v}_\theta$~\citep{liu2023rectifiedflow}.
At normalized diffusion time $\tau$, the sampler uses this velocity to form the clean prediction $\widehat{\mathbf{x}}_0=\mathbf{x}_\tau-\tau\mathbf{v}_\theta$, where $\mathbf{x}_\tau$ stacks the noisy RGB latents.
Control tokens guide this prediction and are excluded from the generation loss.
\ifhmode\par\fi\endgroup

\begingroup{}
\Cref{fig::transfer_attention_masks} shows how language, controls, and RGB interact through attention.
The controls for a chunk describe the simulator state that the generator should render.
Control queries can attend to controls through the current chunk, but their access to clean RGB history is restricted to strictly earlier chunks.
This lets the model use the current simulator controls without reading the clean RGB target through its control representations during training.
In the two-pass teacher-forcing procedure (\cref{sec::teacher_forcing,fig::replay_teacher_forcing}), the clean pass processes latent frames causally, and the noisy pass reuses the clean control keys and values computed by that pass.
The noisy queries in \cref{fig::transfer_attention_masks}(b) also read cached language keys and values, strictly earlier clean RGB history, and noisy tokens within their own chunk.
\ifhmode\par\fi\endgroup

\begingroup{}
\begin{figure}[!t]
\centering
\input{figures/diagrams/transfer_attention_masks.tex}
\caption{\textbf{Attention in Cosmos3-Nano-Sim-Transfer.}
Rows are queries and columns are keys/values, using the visual conventions of \cref{fig::attention_masks}.
$\mathcal{U}_i$, $\mathcal{V}_i$, and $\widetilde{\mathcal{V}}_i$ denote spatial controls, clean RGB, and noisy RGB for latent-frame chunk $i$.
(a) The clean pass preserves causal language attention and gives control and RGB queries access to all language tokens.
Control queries read controls through the current chunk and strictly earlier clean RGB, while clean RGB queries also read their own clean chunk.
(b) The noisy pass reuses the clean pass's language, control, and RGB keys and values.
Each noisy RGB query reads controls through its current chunk, strictly earlier clean RGB, and noisy RGB within its own chunk.
Neither controls nor noisy RGB can read the current clean RGB target.
Each control or RGB chunk contains 1 latent frame and occupies one row or column, with spatial tokens aggregated into a single block.
Three chunks are shown, with the optional conditioning prefix omitted.}
\label{fig::transfer_attention_masks}
\end{figure}
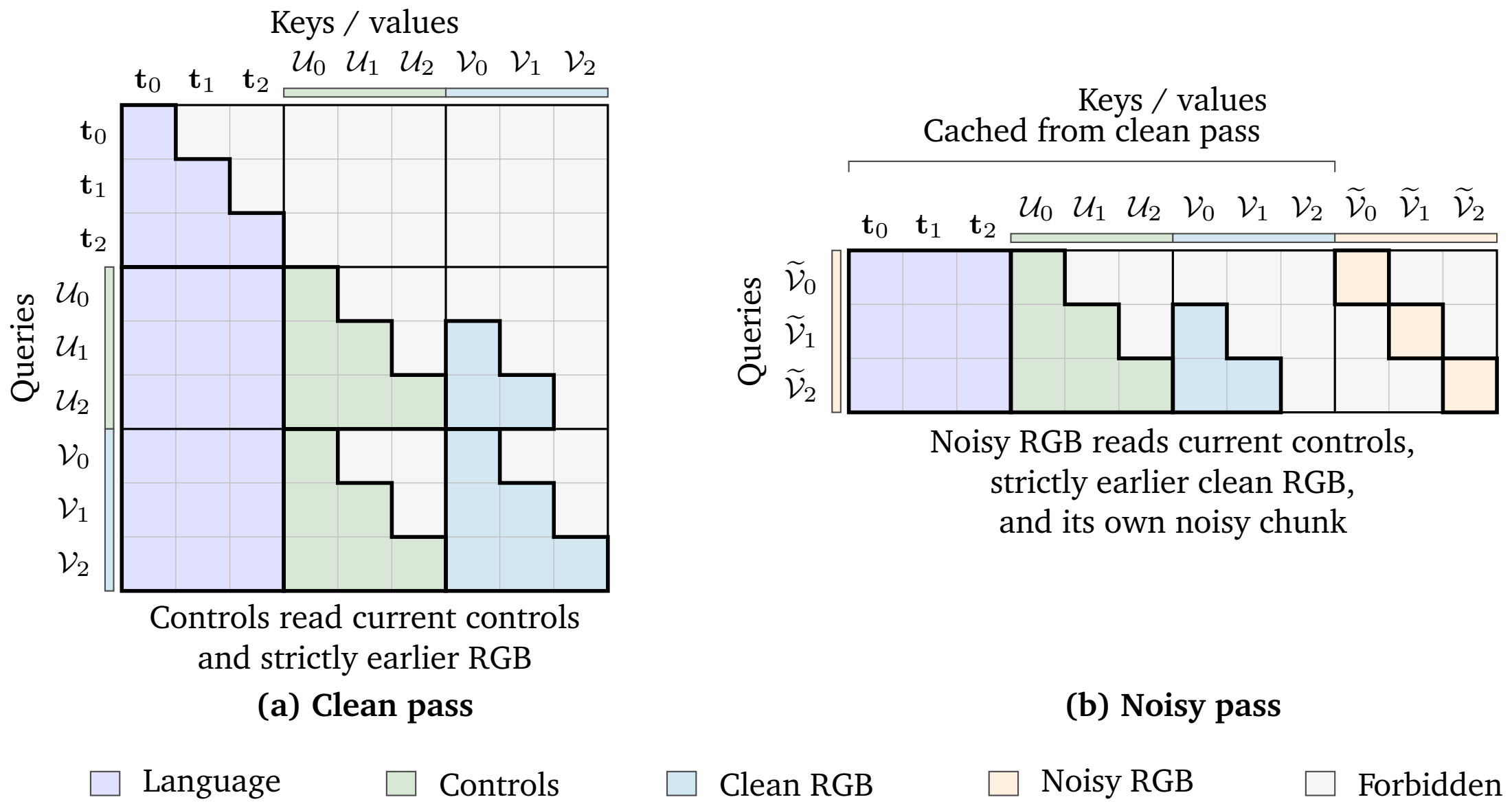
\ifhmode\par\fi\endgroup

\begingroup{}
To keep the renderer responsive to changes in the simulator, we use 1 latent frame per chunk throughout causal training and generation.
Following the tokenizer's temporal compression, the first chunk corresponds to 1 RGB frame, and each subsequent chunk corresponds to 4.
Using this minimum chunk size reduces buffering delay between updated controls and rendered observations during inference.
In closed-loop operation, policy actions advance the simulator, which supplies new controls for the renderer.
The rendered observations then inform the policy's next decision.
For offline benchmarking, the complete control sequence is available in advance, allowing us to resize and encode it before autoregressive generation.
\ifhmode\par\fi\endgroup

%% file: figures/diagrams/transfer_attention_masks.tex
\begingroup
\input{figures/diagrams/architecture_styles.tex}
\fitdiagram{%
\begin{tikzpicture}[font=\fontfamily{bch}\selectfont\small,>=Stealth,
  transfer token/.style={font=\fontfamily{bch}\fontsize{8}{9.2}\selectfont},
  transfer perimeter/.style={draw=black,line width=1.05pt}]
\foreach \mode/\offsetx/\offsety/\lastrow/\lastcolumn in {0/0/0/8/8,1/7.0/-1.4/2/11} {
  \begin{scope}[shift={(\offsetx,\offsety)},x=0.52cm,y=0.52cm]
  \pgfmathtruncatemacro{\rowcount}{\lastrow+1}
  \pgfmathtruncatemacro{\columncount}{\lastcolumn+1}
  \foreach \rowindex in {0,...,\lastrow} {
    \foreach \columnindex in {0,...,\lastcolumn} {
      \def\cellcolor{simgrayfill!45}
      \ifnum\mode=0
        \ifnum\columnindex<3
          \ifnum\rowindex<3
            \ifnum\columnindex>\rowindex\else
              \def\cellcolor{simtextfill}
            \fi
          \else
            \def\cellcolor{simtextfill}
          \fi
        \else
          \ifnum\rowindex>2
            \pgfmathtruncatemacro{\querychunk}{mod(\rowindex-3,3)}
            \pgfmathtruncatemacro{\keychunk}{mod(\columnindex-3,3)}
            \ifnum\columnindex<6
              \ifnum\keychunk>\querychunk\else
                \def\cellcolor{simactionfill}
              \fi
            \else
              \ifnum\rowindex<6
                \ifnum\keychunk<\querychunk
                  \def\cellcolor{simcleanfill}
                \fi
              \else
                \ifnum\keychunk>\querychunk\else
                  \def\cellcolor{simcleanfill}
                \fi
              \fi
            \fi
          \fi
        \fi
      \else
        \ifnum\columnindex<3
          \def\cellcolor{simtextfill}
        \else
          \pgfmathtruncatemacro{\keychunk}{mod(\columnindex-3,3)}
          \ifnum\columnindex<6
            \ifnum\keychunk>\rowindex\else
              \def\cellcolor{simactionfill}
            \fi
          \else
            \ifnum\columnindex<9
              \ifnum\keychunk<\rowindex
                \def\cellcolor{simcleanfill}
              \fi
            \else
              \ifnum\keychunk=\rowindex
                \def\cellcolor{simnoisyfill}
              \fi
            \fi
          \fi
        \fi
      \fi
      \fill[\cellcolor] (\columnindex,-\rowindex) rectangle ++(1,-1);
    }
  }
  \foreach \gridindex in {0,...,\columncount} {
    \draw[sim grid] (\gridindex,0) -- (\gridindex,-\rowcount);
  }
  \foreach \gridindex in {0,...,\rowcount} {
    \draw[sim grid] (0,-\gridindex) -- (\columncount,-\gridindex);
  }
  \foreach \index in {0,1,2} {
    \node[above,transfer token] at ({\index+0.5},0) {$\txttok{\index}$};
    \node[above,transfer token] at ({\index+3.5},0.32) {$\mathcal{U}_{\index}$};
    \node[above,transfer token] at ({\index+6.5},0.32) {$\mathcal{V}_{\index}$};
    \ifnum\mode=0
      \node[left,transfer token] at (0,{-\index-0.5}) {$\txttok{\index}$};
      \node[left,transfer token] at (-0.32,{-\index-3.5}) {$\mathcal{U}_{\index}$};
      \node[left,transfer token] at (-0.32,{-\index-6.5}) {$\mathcal{V}_{\index}$};
    \else
      \node[above,transfer token] at ({\index+9.5},0.32) {$\widetilde{\mathcal{V}}_{\index}$};
      \node[left,transfer token] at (-0.32,{-\index-0.5}) {$\widetilde{\mathcal{V}}_{\index}$};
    \fi
  }
  \foreach \start/\rolecolor in {3/simactionfill,6/simcleanfill} {
    \filldraw[fill=\rolecolor,draw=simgrayedge] (\start,0.15) rectangle ++(3,0.16);
    \ifnum\mode=0
      \filldraw[fill=\rolecolor,draw=simgrayedge] (-0.31,-\start) rectangle ++(0.16,-3);
    \fi
  }
  \draw[semithick] (0,0) rectangle (\columncount,-\rowcount);
  \foreach \boundary in {3,6} {
    \draw[semithick] (\boundary,0) -- (\boundary,-\rowcount);
    \ifnum\mode=0
      \draw[semithick] (0,-\boundary) -- (\columncount,-\boundary);
    \fi
  }
  \ifnum\mode=0
    \node at (4.5,1.5) {Keys / values};
    \node[rotate=90] at (-1.8,-4.5) {Queries};
    \foreach \originx/\originy in {0/0,3/-3,3/-6,6/-6} {
      \draw[transfer perimeter,shift={(\originx,\originy)}]
        (0,0) -- (1,0) -- (1,-1) -- (2,-1) -- (2,-2)
        -- (3,-2) -- (3,-3) -- (0,-3) -- cycle;
    }
    \draw[transfer perimeter] (0,-3) rectangle (3,-9);
    \draw[transfer perimeter,shift={(6,-3)}]
      (0,-1) -- (1,-1) -- (1,-2) -- (2,-2) -- (2,-3) -- (0,-3) -- cycle;
  \else
    \node at (6,2.75) {Keys / values};
    \node[rotate=90] at (-1.8,-1.5) {Queries};
    \draw[simgrayedge] (0,1.45) -- (0,1.65) -- (9,1.65) -- (9,1.45);
    \node[above] at (4.5,1.65) {Cached from clean pass};
    \filldraw[fill=simnoisyfill,draw=simgrayedge] (9,0.15) rectangle ++(3,0.16);
    \filldraw[fill=simnoisyfill,draw=simgrayedge] (-0.31,0) rectangle ++(0.16,-3);
    \draw[semithick] (9,0) -- (9,-3);
    \draw[transfer perimeter] (0,0) rectangle (3,-3);
    \draw[transfer perimeter,shift={(3,0)}]
      (0,0) -- (1,0) -- (1,-1) -- (2,-1) -- (2,-2)
      -- (3,-2) -- (3,-3) -- (0,-3) -- cycle;
    \draw[transfer perimeter,shift={(6,0)}]
      (0,-1) -- (1,-1) -- (1,-2) -- (2,-2) -- (2,-3) -- (0,-3) -- cycle;
    \foreach \index in {0,1,2} {
      \draw[transfer perimeter] ({\index+9},-\index) rectangle ++(1,-1);
    }
  \fi
  \end{scope}
}
\node[align=center] at (2.34,-5.15) {Controls read current controls\\and strictly earlier RGB};
\node[align=center] at (10.12,-3.65) {Noisy RGB reads current controls,\\strictly earlier clean RGB,\\and its own noisy chunk};
\node[font=\fontfamily{bch}\selectfont\small\bfseries] at (2.34,-5.8) {(a) Clean pass};
\node[font=\fontfamily{bch}\selectfont\small\bfseries] at (10.12,-5.8) {(b) Noisy pass};
\foreach \legendx/\legendcolor/\legendlabel in {
  -0.3/simtextfill/Language,
  2.55/simactionfill/Controls,
  5.25/simcleanfill/Clean RGB,
  8.35/simnoisyfill/Noisy RGB,
  11.4/{simgrayfill!45}/Forbidden} {
  \filldraw[fill=\legendcolor,draw=simgrayedge] (\legendx,-6.65) rectangle ++(0.28,0.23);
  \node[anchor=west] at ({\legendx+0.38},-6.535) {\legendlabel};
}
\end{tikzpicture}}
\endgroup

%% file: sections/5__examples/transfer/data.tex
\subsection{Data Preparation}
\label{sec::transfer_data}

\begingroup{}
Cosmos3-Nano-Sim-Transfer training pairs RGB videos with spatial controls and temporally aligned captions.
All four training stages apply aspect-ratio and source-FPS filtering and validate captions before sampling aligned RGB and control clips.
Our source dataset contains 2,153,548 RGB video entries.
The Stage 3 index retains 1,996,602 videos with complete eligible caption windows.
\ifhmode\par\fi\endgroup

\paragraph{Spatial and temporal alignment.}
\begingroup{}
RGB and controls use identical temporal indices and target geometry.
We replace reflection padding with direct resizing for RGB and control videos to prevent mirrored artifacts from extending beyond the padded regions in generated outputs.
To limit stretching when source and target aspect ratios differ, we reject samples requiring more than 3\% aspect-ratio distortion.
We use bicubic antialiasing for RGB, bilinear interpolation for depth, and nearest-neighbor interpolation for segmentation.
Edge and blur controls are computed from the resized RGB.
\ifhmode\par\fi\endgroup

\begingroup{}
Depth controls are precomputed with Video Depth Anything and normalized over each video~\citep{chen2025videodepth}.
Segmentation controls use colorized SAM2 masks~\citep{ravi2024sam}.
Edge controls use Canny detection with randomized thresholds, while blur controls sample bilateral filtering, Gaussian filtering, or their composition.
The index checks RGB/control geometry and frame-rate compatibility and restricts caption windows to their common available frame range.
These checks establish metadata and caption eligibility.

\ifhmode\par\fi\endgroup

\paragraph{Captioning.}
\phantomsection
\label{sec::transfer_captioning}
\begingroup{}
We generate training captions with a Qwen3-VL-8B~\citep{bai2025qwen3} captioner adapted using LoRA~\citep{hu2022lora}.
To associate each description with a specific interval in the video, the captioner processes nonoverlapping windows of up to 200 source frames.
Each eligible annotation records the window's start and end frames alongside a structured JSON description, including a nonempty \texttt{temporal\_caption} field that describes how the scene unfolds over time.
The loader selects the annotation together with the corresponding RGB and control window, keeping the language and visual inputs tied to the same interval.

\ifhmode\par\fi\endgroup

\paragraph{Temporal sampling and chunking.}
\begingroup{}
Across all four training stages, we use identical frame indices for RGB and controls so that each control frame stays aligned with its RGB target.
Each training clip contains 1 initial RGB frame followed by complete groups of 4 frames.
The causal tokenizer encodes the first RGB frame as 1 latent frame and each subsequent group of 4 as another.

In Stages 0--2, we randomly select a caption window and sample every frame, every second frame, or every third frame within that window.
We then trim the sampled clip to satisfy this tokenizer alignment.
The clip-length ranges depend on the training stage and resolution.
Stage 0 uses 5--201 RGB frames at the 480 resolution tier and 5--149 at the 720 tier.
Stages 1 and 2 use 9--201 RGB frames at the 480 tier and divide these clips into chunks of 1 latent frame, matching the chunk size used for autoregressive generation.
In Stage 3, all video clips within a training batch have the same number of frames.
\ifhmode\par\fi\endgroup

%% file: sections/5__examples/transfer/post_training.tex
\subsection{Post-training}
\label{sec::transfer_training}

\paragraph{Training setup.}
\begingroup{}
All four training stages use the same transfer-video collection, pairing each RGB target with 1 spatial-control stream.
We sample depth, segmentation, edges, and blur with equal weights so that training covers each way of specifying the desired scene structure.
Stage 0 mixes the 480 and 720 resolution tiers with weights 2:1, while Stages 1--3 use only the 480 tier.
Throughout the causal stages, the model predicts RGB in chunks of 1 latent frame, matching the chunk size used during autoregressive generation.
\Cref{tab::transfer_training_summary} summarizes the selected checkpoints, clip-length ranges, learning rates, and teacher guidance for each stage.
\ifhmode\par\fi\endgroup

\begingroup{}
Training examples can also include a short recorded RGB prefix that establishes the scene's initial appearance.
We sample prefix lengths of 0, 1, or 2 latent frames with probabilities 0.7, 0.2, and 0.1, exposing the model to generation both with and without initial RGB observations.
For short clips, we reduce the sampled prefix as needed to leave at least 1 latent frame for the model to generate.
\ifhmode\par\fi\endgroup

\begingroup{}
We vary how the accompanying language is presented while keeping the selected RGB/control window fixed.
The loader uses the full structured JSON with probability 0.93 and only the dense temporal caption with probability 0.07, giving the model experience with both input formats.
In Stages 0--2, we additionally drop the text on 10\% of samples so that the model learns to generate from visual conditions alone.
Stage 3 retains language conditioning on every sample, with text dropout set to 0.
Spatial controls remain present in every stage, with control dropout also set to 0.
\ifhmode\par\fi\endgroup

\begingroup{}
The stages share a common optimization setup: fused AdamW with $\epsilon=10^{-8}$, BF16 computation, FP32 master parameters, and no gradient accumulation.
We adjust the momentum and weight decay for each training objective.
Stage 0 uses $(\beta_1,\beta_2)=(0.9,0.99)$ with weight decay 0.05.
Stage 1 uses $(0.9,0.999)$, while Stage 2 and both Stage 3 optimizers use $(0,0.999)$.
All three causal stages use weight decay 0.01.
We maintain power-function exponential moving average (EMA) weights with relative width 0.1 in all stages.
\ifhmode\par\fi\endgroup

\begingroup{}
Stage 0 packs variable-length clips into a budget of 74,000 tokens per GPU.
The causal stages instead use 1 clip per GPU.
Across the recorded runs and restarts, this gives global batches of 248 or 512 clips for Stage 1, 512 for Stage 2, and 248 for Stage 3.
\ifhmode\par\fi\endgroup

\paragraph{Training curriculum.}
\begingroup{}
The curriculum first adapts a bidirectional teacher to the transfer data, then converts it into a causal generator and distills it to use fewer denoising steps.
The final stage trains this generator on its own outputs, preparing it to continue from the history it will encounter during inference.
Stages 2 and 3 both use teacher guidance 3 during distillation.

\ifhmode\par\fi\endgroup

\paragraph{Stage 0: bidirectional adaptation.}
\begingroup{}
Before introducing causal attention, we adapt the bidirectional teacher following \cref{sec::bidirectional_adaptation}.
We fine-tune it on directly resized RGB videos and aligned spatial controls so that the teacher uses the same image geometry as the later stages.
This preprocessing also avoids the mirrored content introduced by reflection padding.
The teacher learns to denoise recorded RGB clips by predicting rectified flow velocity from noisy RGB latents, language, and clean spatial controls.
At this stage, bidirectional temporal attention allows each RGB target to use noisy RGB tokens and supplied controls from both earlier and later frames within the clip.
The loss applies only to noisy RGB targets, excluding the spatial controls and any clean RGB prefix.
We use the adapted teacher at iteration 3,806 to initialize Stage 1.
\ifhmode\par\fi\endgroup

\paragraph{Stage 1: causal teacher forcing (TF).}
\begingroup{}
With the adapted teacher as its starting point, Stage 1 learns to predict each RGB chunk from the available controls and earlier recorded RGB history.
We use the two-pass teacher-forcing procedure in \cref{sec::teacher_forcing,fig::replay_teacher_forcing}.
The clean pass processes recorded RGB tokens and supplied controls to build a causal cache of attention keys and values (K/V).
The noisy pass then reuses this cache to predict each chunk from its noisy RGB tokens, strictly earlier clean RGB history, controls through the current chunk, and language conditioning.
The rectified flow velocity loss in \cref{eq::tf_loss} supervises only RGB targets, with gradients flowing through the clean K/V cache as well as the noisy pass.
The resulting iteration-6,000 EMA checkpoint provides both the initial student and the frozen causal teacher for Stage 2.
\ifhmode\par\fi\endgroup

\paragraph{Stage 2: consistency distillation (TF-dCM).}
\begingroup{}
Once causal prediction is established, Stage 2 reduces the number of denoising steps needed to generate each chunk through TF-dCM (\cref{sec::consistency_distillation,eq::tfdcm_loss}).
Training continues to use recorded RGB history and the same causal control attention as Stage 1.
The frozen causal teacher connects two noisy versions of a chunk at different diffusion times along its denoising trajectory.
The student predicts the clean chunk from both versions and learns to make these predictions agree, using identical history, controls, and language in both evaluations.
The current student's prediction at the lower-noise input supplies the training target and is computed with gradients stopped.
This consistency objective prepares the student to recover a clean chunk with fewer denoising evaluations.
The iteration-2,250 EMA checkpoint initializes~Stage~3.
\ifhmode\par\fi\endgroup

\paragraph{Stage 3: self-forcing (SF-DMD).}
\begingroup{}
Stage 3 brings training closer to autoregressive inference by letting the student learn from its own generated history, following SF-DMD in \cref{sec::self_forcing}.
After an optional recorded RGB prefix, the student generates successive chunks of 1 latent frame using the aligned controls and the RGB history it has produced so far.
Each completed chunk and its supplied controls refresh the clean K/V cache, providing the context for the next prediction.
The student therefore trains on the consequences of its earlier outputs throughout the rollout.
\ifhmode\par\fi\endgroup

\begingroup{}
To train from these generated clips, we initialize two bidirectional score models from the Stage 0 teacher: a frozen reference teacher and a trainable fake-score network.
We add noise to the generated clips and ask both models to denoise them.
The teacher supplies the reference prediction, while the fake-score network learns to denoise the student's current outputs.
The difference between their predictions defines the distribution-matching target that guides the student (\cref{eq::dmd_surrogate}).
We alternate 1 student update with both score models fixed and 5 fake-score updates with the student fixed.
Checkpoint iteration counts include both types of updates.
\ifhmode\par\fi\endgroup

\begingroup{}
We use replayed backpropagation to train on a complete rollout without retaining all of its intermediate activations.
Training first constructs the rollout, then recomputes each chunk's final denoising prediction from its saved noisy input, timestep, and detached cache prefix.
Gradients pass through these recomputed predictions but do not propagate through earlier denoising steps or historical K/V caches.
This trains the model to use generated history while treating that history as a fixed input for each gradient computation.
\ifhmode\par\fi\endgroup

\begingroup{}
\begin{table}[!t]
\centering\reporttablestyle
\caption{\textbf{Cosmos3-Nano-Sim-Transfer post-training settings.}
Iterations identify the selected EMA checkpoints.
RGB ranges are configured sampling limits.
Stage 0 multiplies its base learning rate by $0.4(1-i/100{,}000)$ at iteration $i$, starting at $4\times10^{-5}$ without warmup.
The other learning rates are constant.
The SF learning rates are listed as student / fake score.}
\label{tab::transfer_training_summary}
\begin{tabularx}{\linewidth}{L{2.3cm}C>{\centering\arraybackslash}p{4.7cm}CC}
\toprule
\textbf{Stage} & \textbf{Checkpoint iteration} & \textbf{RGB frames} & \textbf{Base learning rate} & \textbf{Teacher CFG}\\
\midrule
Teacher & 3,806 & \mbox{5--201 (480p), 5--149 (720p)} & $10^{-4}$ & ---\\
TF & 6,000 & 9--201 & $10^{-5}$ & ---\\
TF-dCM & 2,250 & 9--201 & $2\times10^{-6}$ & 3\\
SF-DMD & 1,500 & 9--201 & $10^{-6}$ / $2\times10^{-7}$ & 3\\
\bottomrule
\end{tabularx}
\end{table}
\ifhmode\par\fi\endgroup

%% file: sections/5__examples/transfer/optimization.tex
\subsection{Inference Optimization}
\label{sec::transfer_inference_optimization}

\paragraph{Inference.}
\begingroup{}
Cosmos3-Nano-Sim-Transfer follows the autoregressive inference loop in \cref{sec::streaming_execution}.
For each chunk, the sampler denoises 1 RGB latent frame using aligned spatial controls, cached language keys and values, and retained clean control and RGB history.
The controls and retained history remain fixed across denoising steps.
After sampling, the runtime refreshes clean control and RGB keys and values for the next prediction.
The streaming Wan2.2 VAE~\citep{wan2025} decodes the completed latent chunk into RGB observations, carrying temporal state across chunks so that output is available before the entire video is generated.
\ifhmode\par\fi\endgroup

\begingroup{}
The history policy determines which control and RGB K/V entries remain available to subsequent chunks.
Full-history inference retains all entries throughout the rollout.
Bounded inference retains initial sink entries and a rolling window of recent history, replacing older entries outside this window while preserving token roles and temporal positions.
This bounds history storage and attention context during extended generation.
\ifhmode\par\fi\endgroup

\paragraph{Compilation with preserved numerical behavior.}
\begingroup{}
We combine \texttt{torch.compile} with block CUDA Graphs to reduce execution overhead in the single-view SF-DMD student, following the shared runtime design in \cref{sec::runtime_optimization}.
The optimized path keeps normalization outside compilation and preserves the reduction order for squared activations.
It also preserves intermediate BF16 rounding in FP32--BF16--FP32 conversions and the original K/V cache order.
These choices retain the eager path's numerical behavior at operations that can affect autoregressive outputs.
\Cref{sec::transfer_results} evaluates the runtime and output quality of these inference choices.
\ifhmode\par\fi\endgroup

%% file: sections/5__examples/transfer/results.tex
\subsection{Results}
\label{sec::transfer_results}

\paragraph{Evaluation setup.}
\begingroup{}
\suppressfloats[t]
We evaluate Cosmos3-Nano-Sim-Transfer from three complementary perspectives: the quality and control fidelity of its videos, its behavior as an observation renderer in a robot policy loop, and the runtime cost of sustained generation.
We begin with TransferBench~\citep{cosmos_transfer1}, which contains 600 source videos: 200 each from AgiBot~\citep{bu2025agibot}, Ego-Exo4D~\citep{grauman2024egoexo4d}, and OpenDV~\citep{yang2024genad}.
For every source video, we generate an output conditioned on each of depth, segmentation, edges, and blur.
This produces 2,400 outputs per model, each with 121 aligned RGB frames.
Outputs are $736\times544$ for AgiBot and $832\times480$ for the other domains, preserving source rates of approximately 23.98--60 FPS.
\ifhmode\par\fi\endgroup

\begingroup{}
To show how the final model compares with both its reference and the stages that lead to it, we evaluate five checkpoints on the same full benchmark.
The iteration-1,500 SF-DMD EMA student uses 4 SDE steps, full-history K/V without attention sinks, and student/control guidance 1.
For this offline benchmark, the complete control sequence is prepared and encoded before autoregressive generation.
We compare it with the Cosmos 3 GA iteration-6,000 EMA reference and the selected curriculum checkpoints: the bidirectional teacher at iteration 3,806, TF at iteration 6,000, and TF-dCM at iteration 2,250.
The bidirectional teacher and TF use 50-step UniPC with guidance 3, while TF-dCM and SF-DMD use 4-step SDE sampling with guidance 1.
Generation uses upsampled captions, while language scoring uses the original benchmark captions.
The comparison therefore evaluates complete generation pipelines with their respective samplers and~guidance~settings.
\ifhmode\par\fi\endgroup

\paragraph{Video quality and temporal metrics.}
\begingroup{}
We assess both the appearance of the generated videos and whether that quality changes over time.
DOVER measures technical video quality across all 2,400 outputs~\citep{dover}.
For a complementary assessment, Gemini 3.1 Pro Preview receives each generated video and its original benchmark caption and assigns integer grades from 0 to 10.
Its Quality grade evaluates visible artifacts, while its Prompt grade evaluates whether the elements described in the caption appear in the video.
\ifhmode\par\fi\endgroup

\begingroup{}
To measure changes between the beginning and end of a video, we report quality drift as the absolute difference between the mean perceptual scores of its first and last 15\% of frames.
We evaluate each endpoint with PAI-Bench-G, the video-generation component of PAI-Bench~\citep{zhou2025paibench}.
The endpoint score averages six perceptual-quality dimensions: aesthetic quality, background consistency, imaging quality, motion smoothness, overall consistency, and subject consistency.
We divide imaging-quality scores by 100 before averaging the dimensions, then average the absolute endpoint differences across videos.
Lower drift indicates a smaller change in these perceptual scores between the beginning and end of the video.
\ifhmode\par\fi\endgroup

\paragraph{Control, language, and geometry metrics.}
\begingroup{}
We next measure how closely the generated scene follows the supplied spatial structure, using a separate metric for each control modality and its corresponding 600 outputs.
Segmentation IoU measures agreement between regions, and edge F1 measures agreement between contours.
Blur SSIM measures structural similarity~\citep{wang2004ssim}, while depth SI-RMSE measures scale-aligned depth error.
The Prompt grade described above complements these spatial measurements by assessing agreement with the language condition.
\ifhmode\par\fi\endgroup

\begingroup{}
To examine temporal geometry, we measure Sampson error using SuperPoint features matched with LightGlue between all frame pairs separated by approximately 1 second~\citep{sampson1987,lightglue2023}.
For each pair, the evaluator estimates a fundamental matrix and retains the half of inlier correspondences with the lowest Sampson errors.
It then scales the square root of their mean Sampson error to a $960\times540$ reference diagonal.
The video score averages the valid pair scores and requires at least 90\% pair coverage.
This metric checks the geometric consistency of correspondences across time, adding a diagnostic that appearance scores alone do not provide.
All five checkpoints have 2,400 valid scores for each video-level metric and 600 for each control metric, with no invalid values in the final records.

\ifhmode\par\fi\endgroup

\begingroup{}
\begin{table}[!t]
\centering\reporttablestyle
\caption{\textbf{Full TransferBench video quality, spatial control, and condition agreement.}
Each checkpoint generates 2,400 videos from 600 TransferBench sources~\citep{cosmos_transfer1} and four control modalities.
DOVER uses all outputs, while each control metric uses its corresponding 600 outputs.
Quality and Prompt retain the recorded VLM grading scales.
Drift compares the first and last 15\% of frames, and Sampson error uses normalized epipolar residuals at approximately 1-second intervals.
Stage comparisons retain each checkpoint's deployment sampler and guidance.
Bold and underlined entries mark the best and second-best values within~each~metric.}
\label{tab::transfer_benchmark_metrics}
\par\smallskip
\textit{(a) Video quality}\par\smallskip
\begin{tabular}{lccc}
\toprule
\textbf{Model} & \textbf{DOVER $\uparrow$} & \textbf{Quality $\uparrow$} & \textbf{Drift $\downarrow$}\\
\midrule
Bidirectional teacher (3,806) & \textbf{10.3955} & \textbf{5.0042} & \underline{0.006370}\\
TF (6,000) & 10.2112 & 4.2200 & 0.008265\\
TF-dCM (2,250) & 10.0403 & 4.2588 & 0.011730\\
\rowcolor{rowours}
Cosmos3-Nano-Sim-Transfer & 10.1759 & \underline{4.9508} & 0.006504\\
Cosmos 3 GA reference & \underline{10.3596} & 4.9296 & \textbf{0.006349}\\
\bottomrule
\end{tabular}
\par\smallskip
\textit{(b) Spatial control}\par\smallskip
\begin{tabular}{lcccc}
\toprule
\textbf{Model} & \textbf{Segm. IoU $\uparrow$} & \textbf{Blur SSIM $\uparrow$} & \textbf{Edge F1 $\uparrow$} & \textbf{Depth SI-RMSE $\downarrow$}\\
\midrule
Bidirectional teacher (3,806) & \textbf{0.7465} & \textbf{0.9074} & \textbf{0.4692} & \textbf{0.5279}\\
TF (6,000) & 0.6948 & 0.8591 & 0.2973 & 0.7594\\
TF-dCM (2,250) & 0.7070 & 0.8569 & 0.3051 & 0.7273\\
\rowcolor{rowours}
Cosmos3-Nano-Sim-Transfer & \underline{0.7378} & \underline{0.8867} & \underline{0.3407} & \underline{0.6194}\\
Cosmos 3 GA reference & 0.6433 & 0.8559 & 0.2140 & 0.8423\\
\bottomrule
\end{tabular}
\par\smallskip
\textit{(c) Language and temporal geometry}\par\smallskip
\begin{tabular}{lcc}
\toprule
\textbf{Model} & \textbf{Prompt $\uparrow$} & \textbf{Sampson $\downarrow$}\\
\midrule
Bidirectional teacher (3,806) & \underline{7.1842} & \textbf{0.080947}\\
TF (6,000) & 6.7625 & 0.087848\\
TF-dCM (2,250) & 6.7288 & 0.098861\\
\rowcolor{rowours}
Cosmos3-Nano-Sim-Transfer & 7.0254 & 0.096544\\
Cosmos 3 GA reference & \textbf{7.1846} & \underline{0.081045}\\
\bottomrule
\end{tabular}
\end{table}
\ifhmode\par\fi\endgroup

\begingroup{}
\begin{figure}[p]
\centering
\includegraphics[width=\linewidth,height=0.875\textheight,keepaspectratio]{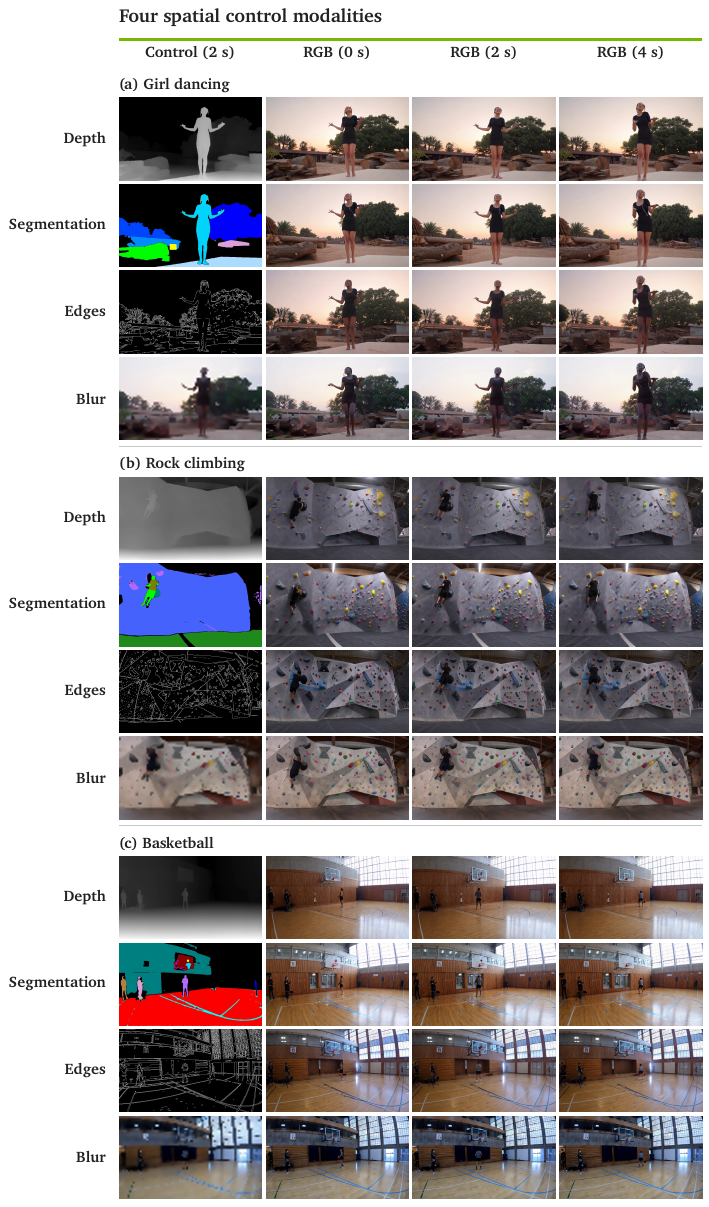}
\caption{\textbf{RGB generation from four spatial control modalities.}
From top to bottom: girl dancing, rock climbing, and basketball, each with depth, segmentation, edges, and blur.
The first column shows each supplied control at 2 seconds, while the remaining columns show generated RGB at 0, 2, and 4 seconds.
Each output contains 121 frames at $832\times480$ and 30 FPS.
Generation uses 4 denoising evaluations per latent-frame chunk and full K/V history without attention sinks.}
\label{fig::transfer_example2_controls}
\end{figure}
\ifhmode\par\fi\endgroup

\paragraph{Benchmark results.}
\begingroup{}
Relative to the Cosmos 3 GA reference, Cosmos3-Nano-Sim-Transfer improves all four spatial-control metrics in \cref{tab::transfer_benchmark_metrics}.
Segmentation IoU increases from 0.6433 to 0.7378 and blur SSIM from 0.8559 to 0.8867.
Edge F1 rises from 0.2140 to 0.3407, a 59.2\% increase, while depth SI-RMSE falls from 0.8423 to 0.6194, a 26.5\% reduction.
These results show that the final student follows the supplied regions, contours, coarse image structure, and depth more closely across the four evaluated control modalities.
\ifhmode\par\fi\endgroup

\begingroup{}
Alongside its stronger spatial control, Cosmos3-Nano-Sim-Transfer slightly improves the VLM Quality grade from 4.9296 to 4.9508.
DOVER and Prompt decrease slightly by 1.8\% and 2.2\%, respectively, while quality drift and Sampson error increase relative to the GA reference.
\ifhmode\par\fi\endgroup

\begingroup{}
The intermediate checkpoints show how the final student relates to the preceding training stages.
SF-DMD improves all nine reported metrics relative to TF-dCM.
The Quality grade increases from 4.2588 to 4.9508, and drift decreases from 0.011730 to 0.006504.
The bidirectional teacher retains stronger scores across all nine metrics than the final causal student.
The final self-forcing stage thus improves on the selected consistency-distilled checkpoint, while a gap to the bidirectional teacher remains.
As in the reference comparison, these results retain each checkpoint's deployment sampler and guidance settings.
\ifhmode\par\fi\endgroup

\begingroup{}
We also examine uncertainty in the smaller differences between TF and TF-dCM using a paired bootstrap over source videos.
We draw 10,000 resamples stratified by source domain and keep each source's four control modalities together.
For TF-dCM relative to TF, the edge-F1 difference is $+0.0078$ with a 95\% interval of $[0.0061,0.0095]$, and the DOVER difference is $-0.1708$ with an interval of $[-0.2369,-0.1067]$.
The Quality difference is $+0.0388$, with an interval of $[-0.0396,0.1188]$.
The edge-F1 and DOVER intervals preserve the direction of their observed changes, whereas the Quality interval includes zero.
These per-metric intervals quantify uncertainty from sampling source videos while holding the recorded generations fixed.
\ifhmode\par\fi\endgroup

\paragraph{Closed-loop evaluation in Isaac Lab.}
\begingroup{}
Cosmos3-Nano-Sim-Transfer increases apple pick-and-place completion for a Unitree G1 humanoid from 68.8\% with native Isaac RTX rendering~\citep{nvidia_rtx_renderer_2026} to 84.4\% on the calibrated baseline (\cref{tab::transfer_isaac_apple}).
Both configurations use Isaac Lab 3.0.0b2~\citep{isaaclab2025} and Isaac Sim 6.0.0.1~\citep{nvidia_isaac_sim_2026} for physics.
We evaluate 32 matched starts with the same 15,000-step GR00T N1.7 policy~\citep{nvidia_gr00t_n17_2026} fine-tuned on 402 real robot demonstrations.
Cosmos3-Nano-Sim-Transfer converts the simulator's depth controls into RGB observations that guide the policy's next actions.
At each decision, the policy receives a $640\times480$ head-camera RGB image, the current 43-dimensional joint state, and the task instruction, then predicts 16 actions.
The simulator runs policy actions at 30 Hz and physics at 240 Hz.
\ifhmode\par\fi\endgroup

\begingroup{}
As the actions advance the simulated scene, depth is sampled every 4 actions, corresponding to 7.5 Hz in simulated time.
After a 1-frame bootstrap, the renderer uses these controls to generate 4 RGB observations per update.
The final image becomes the input to the next policy decision, closing the loop between action, simulation, and rendering.
This schedule gives a policy decision rate of 1.875 Hz in simulated time.
\ifhmode\par\fi\endgroup

\begingroup{}
The renderer uses the iteration-1,500 SF-DMD EMA student with 4 SDE steps, guidance 1, a 30-latent-frame K/V window, and 3 attention-sink frames.
A scene prompt fixes the requested appearance before the rollout, while depth from the simulator describes how the geometry changes as the robot acts.
We normalize depth over 0.15--0.90 m, resize it from $640\times480$ to $736\times552$, and center-crop it to $736\times544$.
We then resize generated RGB back to the policy's camera resolution.
This streaming preprocessing differs from the direct resizing used in the offline benchmark.
The model's FPS condition is 12.5, distinct from the simulator's observation cadence.
\ifhmode\par\fi\endgroup

\begingroup{}
Each episode runs until the task terminates or reaches 600 actions, corresponding to 20 simulated seconds.
Success is determined from simulator state using placement, contact, support, settling, and hand-release conditions.
For every appearance condition, we evaluate 32 matched initial states in 2 batches of 16 environments.
Each matched comparison holds the policy checkpoint, initial states, and simulator physics fixed, changing the policy's RGB input between native Isaac RTX and Cosmos3-Nano-Sim-Transfer.
\ifhmode\par\fi\endgroup

\begingroup{}
With the 15,000-step policy, Cosmos3-Nano-Sim-Transfer achieves higher completion than native Isaac RTX in 3 of the 4 tested appearance conditions.
The calibrated baseline improves from 22/32 to 27/32 completed episodes, a gain of 15.6 percentage points.
The green-apple condition improves from 20/32 to 24/32, and the combined green-apple, red-plate, and white-cloth condition improves from 16/32 to 17/32.
The red-plate condition yields 19/32 completions with Cosmos3-Nano-Sim-Transfer and 22/32 with RTX.
Across the six evaluated policy checkpoints, the highest observed completion is 27/32 (84.4\%) with Cosmos3-Nano-Sim-Transfer at 15,000 steps, compared with 24/32 (75.0\%) with RTX at 20,000 steps.
\ifhmode\par\fi\endgroup

\begingroup{}
\begin{table}[!t]
\centering\reporttablestyle
\caption{\textbf{Cosmos3-Nano-Sim-Transfer improves task completion on the calibrated Isaac Lab baseline.}
Entries are completed episodes out of 32 matched starts per condition.
Both configurations use Isaac Lab for physics.
The same GR00T policy receives either native Isaac RTX RGB or Cosmos3-Nano-Sim-Transfer RGB.
Panel (a) fixes the policy at 15,000 training steps.
Panel (b) fixes the baseline appearance and reports success-rate differences (Cosmos3-Nano-Sim-Transfer minus RTX) in percentage points (pp).
The scene prompt was calibrated at 15,000 steps on these starts, so the table reports calibration-set outcomes.}
\label{tab::transfer_isaac_apple}
\par\smallskip
\textit{(a) Appearance changes}\par\smallskip
\begin{tabular}{lcc}
\toprule
\textbf{Appearance} & \textbf{Isaac RTX} & \textbf{Cosmos3-Nano-Sim-Transfer}\\
\midrule
Calibrated baseline & 22/32 & 27/32\\
Green apple & 20/32 & 24/32\\
Red plate & 22/32 & 19/32\\
Green apple, red plate, white cloth & 16/32 & 17/32\\
\bottomrule
\end{tabular}
\par\smallskip
\textit{(b) Policy checkpoints}\par\smallskip
\begin{tabular}{lccc}
\toprule
\textbf{Policy training steps} & \textbf{Isaac RTX} & \textbf{Cosmos3-Nano-Sim-Transfer} & \textbf{Difference (pp)}\\
\midrule
5,000 & 17/32 & 20/32 & $+9.4$\\
10,000 & 16/32 & 12/32 & $-12.5$\\
15,000 & 22/32 & 27/32 & $+15.6$\\
20,000 & 24/32 & 21/32 & $-9.4$\\
30,000 & 18/32 & 18/32 & $0.0$\\
40,000 & 17/32 & 19/32 & $+6.3$\\
\bottomrule
\end{tabular}
\end{table}
\ifhmode\par\fi\endgroup

\paragraph[\kern0pt Qualitative results.]{{Qualitative results.}}
\begingroup{}
We first illustrate control-to-RGB generation with depth, segmentation, edges, and blur (\cref{fig::transfer_example2_controls}).
Each of the three examples uses the same source video and prompt across its four control modalities.
The control images and generated frames show how each modality supplies scene structure for the model to render in RGB.
We then examine rendering over longer simulated trajectories.
\Cref{fig::transfer_visdex_controls} shows three 600-frame VisDex rollouts under depth and segmentation, with frames sampled from the beginning, middle, and end of each generated sequence.
\ifhmode\par\fi\endgroup

\begingroup{}
\begin{figure}[!t]
\centering
\includegraphics[width=325bp]{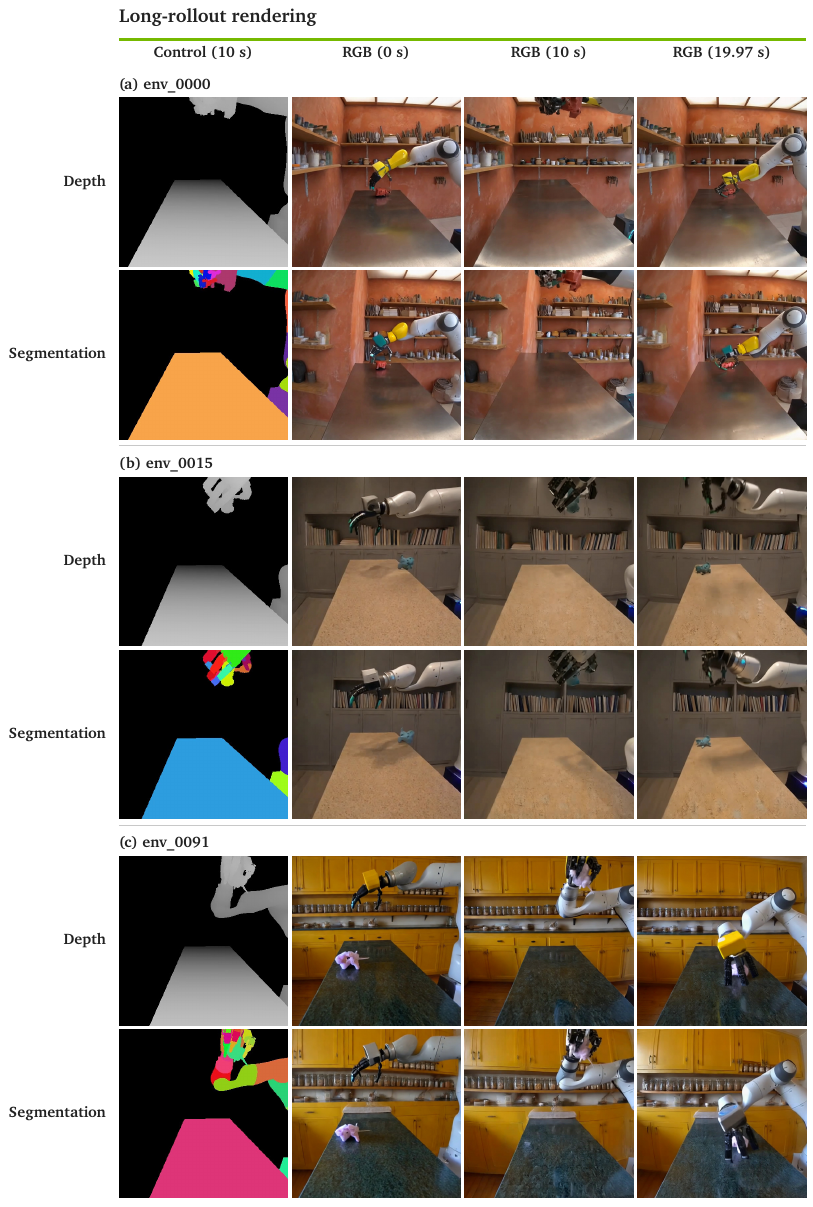}
\caption{\textbf{Long-rollout RGB generation from depth and segmentation.}
From top to bottom: VisDex episodes \texttt{env\_0000}, \texttt{env\_0015}, and \texttt{env\_0091}, each sharing its text prompt across both control modalities.
The first column shows the supplied control at 10 seconds, while the remaining columns show generated RGB at 0, 10, and 19.97 seconds.
The selected SF-DMD student uses 4 denoising evaluations per latent-frame chunk, a 30-latent-frame K/V window, and 3 sink frames.
Each output contains 600 frames at $640\times640$ and 30 FPS.}
\label{fig::transfer_visdex_controls}
\end{figure}
\ifhmode\par\fi\endgroup

\paragraph{Long-rollout runtime.}
\begingroup{}
We profile the iteration-1,500 SF-DMD EMA student on all 160 VisDex s42 episodes under four control modalities, generating 640 videos of 600 frames at $640\times640$.
Inference uses BF16 eager execution on GB200 without compilation or CUDA Graphs, with 4 SDE steps, guidance 1, a 30-latent-frame K/V window, and 3 sink frames.
For steady-state measurements, we exclude each worker's first video and the first 117 frames of each subsequent SF-DMD video, retaining 144 videos and 17,424 decoded-output chunk intervals per control.
Steady throughput ranges from 9.90 to 9.96 RGB frames per second per GPU (\cref{tab::transfer_visdex_runtime}).
Chunk timings measure completed RGB output, including callback waits for copies to host memory.
On the matched depth and segmentation workload, the Cosmos 3 GA reference takes 217.17 s on average to complete a 600-frame video.
Its approximately 120-frame continuation blocks average 43.30 s after warmup, precluding real-time feedback in this configuration (\cref{tab::transfer_visdex_quality_runtime}).
Cosmos3-Nano-Sim-Transfer generates observations autoregressively, returning a decoded latent-frame chunk, normally 4 RGB frames, in 0.403 s on average after warmup and providing sub-second observation updates for closed-loop simulation without waiting for a complete video.
\ifhmode\par\fi\endgroup

\begingroup{}
\begin{table}[!t]
\centering\reporttablestyle
\caption{\textbf{Final-student runtime on 600-frame VisDex rollouts.}
All measurements use 1 stream per GB200 GPU and 144 warmed videos per control.
Chunk times measure completed decoded output, normally 4 RGB frames per chunk.
First-chunk times use warmed workers and exclude model loading.}
\label{tab::transfer_visdex_runtime}
\begin{tabular}{lccc}
\toprule
\textbf{Control} & \textbf{FPS/GPU} & \textbf{Chunk p50 / p95 (ms)} & \textbf{First chunk p50 / p95 (s)}\\
\midrule
Depth & 9.928 & 401 / 418 & 6.93 / 7.23\\
Segmentation & 9.896 & 403 / 420 & 6.55 / 6.64\\
Edge & 9.920 & 401 / 418 & 6.33 / 6.42\\
Blur & 9.957 & 399 / 421 & 6.26 / 6.37\\
\bottomrule
\end{tabular}
\end{table}
\ifhmode\par\fi\endgroup

\begingroup{}
\begin{table}[!t]
\centering\reporttablestyle
\caption{\textbf{Matched 600-frame VisDex runtime.}
Both models use the same 160-episode depth and segmentation workload on GB200 with BF16 eager inference.
Runtime excludes each worker's first video and retains 144 videos per control.
Steady FPS and chunk latency additionally exclude the first 121 frames per remaining GA video or 117 frames per SF-DMD video.
Runtime values pool depth and segmentation.
Generation latency is the mean per steady continuation block, approximately 120 RGB frames, for Cosmos 3 GA and per latent-frame chunk, normally 4 RGB frames, for Cosmos3-Nano-Sim-Transfer.
The means cover 1,152 GA blocks and 34,848 Cosmos3-Nano-Sim-Transfer chunks.
Timing measures completed decoded output, including callback waits for copies to host memory.
Model loading, source-video decoding, and file saves are excluded.}
\label{tab::transfer_visdex_quality_runtime}
\begin{tabular}{lcc}
\toprule
\textbf{Metric} & \textbf{Cosmos 3 GA reference} & \textbf{Cosmos3-Nano-Sim-Transfer}\\
\midrule
Steady FPS/GPU $\uparrow$ & 2.766 & 9.912\\
Generation latency & 43.30~s/block & 0.403~s/chunk\\
Steady peak allocated memory (GiB) $\downarrow$ & 36.15 & 39.58\\
\bottomrule
\end{tabular}
\end{table}
\ifhmode\par\fi\endgroup

\paragraph[\kern0pt Compilation and cache optimization.]{{Compilation and cache optimization.}}
\begingroup{}
A separate 597-frame depth-control study evaluates the compilation and cache choices in \cref{sec::transfer_inference_optimization}.
The iteration-1,500 SF-DMD checkpoint generates $640\times640$ videos on 1 NVIDIA GB200 per worker.
Each video has a nominal rate of 30 FPS, using 4 denoising evaluations per latent frame, guidance 1, and no RGB prefix.
All configurations use BF16 model weights and K/V caches.
The first latent frame decodes to 1 RGB frame, and each subsequent latent frame decodes to 4 RGB frames.
We compare full history without sinks with a bounded window of 51 latent positions: 1 sink, 49 recent past positions, and the current position.
The bounded window contains 102 paired control/RGB cache slots.
The complete control video is prepared before generation.
\ifhmode\par\fi\endgroup

\Needspace{3\baselineskip}
\begingroup{}
Compilation and block CUDA Graphs increase inference throughput by up to $1.75\times$ while preserving generation quality (\cref{tab::transfer_offline_optimization}).
Measured throughput increases from 6.8802 to 11.4068 FPS with full history ($1.66\times$), and from 9.0403 to 15.8516 FPS with bounded history ($1.75\times$).
The corresponding p95 decoded chunk intervals decrease from 761.7341 to 443.8231 ms and from 509.8658 to 268.7143 ms.
\ifhmode\par\fi\endgroup

\begingroup{}
\begin{table}[!htb]
\centering\reporttablestyle
\caption{\textbf{BF16 inference optimization for Cosmos3-Nano-Sim-Transfer on 1 GB200.}
The SF-DMD iteration-1,500 checkpoint generates 597-frame, $640\times640$ depth-conditioned videos.
Compiled denotes \texttt{torch.compile} with block CUDA Graphs.
Panel (a) measures decoded CPU output over latent indices 51--149 after warmup, excluding startup, offline condition preparation, and file encoding.
Each configuration contributes 594/594 valid/attempted chunk intervals from 6 measured videos.
Memory peaks are the maxima of PyTorch allocated/reserved storage across those videos.
Panel (b) reports means over 16 depth-control sources per configuration, with the recorded DOVER and VLM scales.}
\label{tab::transfer_offline_optimization}
\par\smallskip
\textit{(a) Throughput, decoded chunk interval, and GPU storage}\par\smallskip
\begin{tabularx}{\linewidth}{L{4.4cm}CCC}
\toprule
\textbf{History / execution} & \textbf{FPS}$\uparrow$ & \textbf{Chunk p50 / p95 (ms)} & \textbf{Peak allocated / reserved (GiB)}\\
\midrule
Full / eager & 6.8802 & 560.8695 / 761.7341 & 55.0640 / 60.0859\\
Full / compiled & 11.4068 & 348.1926 / 443.8231 & 74.5866 / 79.7207\\
Sink + recent / eager & 9.0403 & 435.8790 / 509.8658 & 44.0980 / 47.7598\\
Sink + recent / compiled & 15.8516 & 240.4241 / 268.7143 & 52.7295 / 57.8184\\
\bottomrule
\end{tabularx}
\par\smallskip
\textit{(b) Video quality and language following}\par\smallskip
\begin{tabularx}{\linewidth}{L{4.4cm}CCC}
\toprule
\textbf{History / execution} & \textbf{DOVER}$\uparrow$ & \textbf{VLM prompt /10}$\uparrow$ & \textbf{VLM quality /10}$\uparrow$\\
\midrule
Full / eager & 8.6524 & 5.3125 & 2.4375\\
Full / compiled & 8.6524 & 5.3125 & 2.4375\\
Sink + recent / eager & 9.0190 & 5.3750 & 2.5000\\
Sink + recent / compiled & 9.0190 & 5.3750 & 2.5000\\
\bottomrule
\end{tabularx}
\end{table}
\ifhmode\par\fi\endgroup

\begingroup{}
Quality evaluation uses 16 depth-control VisDex sources per configuration, with 16/16 successful generations and no failures.
Within each history policy, all 16 eager/compiled video pairs have identical decoded frames, covering 32 pairs and 19,104 frames in total.
DOVER~\citep{dover} and the recorded VLM prompt and quality grades remain unchanged between eager and compiled execution within each history policy (\cref{tab::transfer_offline_optimization}).
\ifhmode\par\fi\endgroup

\begingroup{}
Component p50/p95 timings and decoded-output intervals are measured separately from offline preparation and startup.
The timing protocol defines the measurement boundaries and overlapping measurements.
\ifhmode\par\fi\endgroup

\Needspace{3\baselineskip}
\begingroup{}
Sub-second RGB updates support the renderer's role in real-time closed-loop simulation: the simulator advances physical state and supplies new controls, and Cosmos3-Nano-Sim-Transfer renders observations for the policy's~next~decision.
\ifhmode\par\fi\endgroup

%% file: sections/5__examples/auto/overview.tex
\begingroup
\makeatletter
\c@topnumber=2\relax
\c@bottomnumber=0\relax
\c@totalnumber=2\relax
\makeatother
\renewcommand{\topfraction}{0.85}
\renewcommand{\bottomfraction}{0}
\renewcommand{\textfraction}{0.13}
\renewcommand{\floatpagefraction}{0.90}

\begingroup{}
Autonomous-driving simulation requires camera and LiDAR observations that describe the same evolving scene consistently across viewpoints, modalities, and time, while following the supplied road geometry and actor motion.
We extend the single-view Cosmos3-Nano to this setting through camera-specific conditioning and joint RGB--LiDAR modeling (\cref{fig::auto_overview}).
Per-view prompts preserve the foundation model's association between a caption and its video, while attention across cameras and modalities supports scene consistency.
The resulting Cosmos3-Nano-Transfer-Auto denoises complete sensor clips bidirectionally for offline generation.
Its causal, distilled counterpart, Cosmos3-Nano-Sim-Auto, generates synchronized sensor streams incrementally for streaming simulation.
\ifhmode\par\fi\endgroup

\begingroup{}
Camera generation is conditioned on per-view captions and camera-aligned World Scenario Map (WSM) controls, while LiDAR generation uses HD-map controls.
Transfer-T2V (Text-to-Video) uses no initial RGB observations, whereas Transfer-I2V (Image-to-Video) additionally uses initial camera images.
An initial measured LiDAR sweep is optional in either mode.
\ifhmode\par\fi\endgroup

\begingroup{}
Training begins with Stage~0, which adapts Cosmos3-Nano into the bidirectional \textbf{Cosmos3-Nano-Transfer-Auto} model.
Stage~0 comprises three phases: modality-separate training, short-sequence joint training, and long-sequence joint training with dropout.
We then apply the shared three-stage causal recipe: teacher forcing (Stage~1), teacher-forcing discrete consistency distillation (Stage~2), and self-forcing distribution matching distillation (Stage~3).
The resulting \textbf{Cosmos3-Nano-Sim-Auto} generates synchronized sensor streams in small latent chunks with cached history.
The training curriculum is detailed in \cref{sec::auto_training}.
\ifhmode\par\fi\endgroup

\begingroup{}
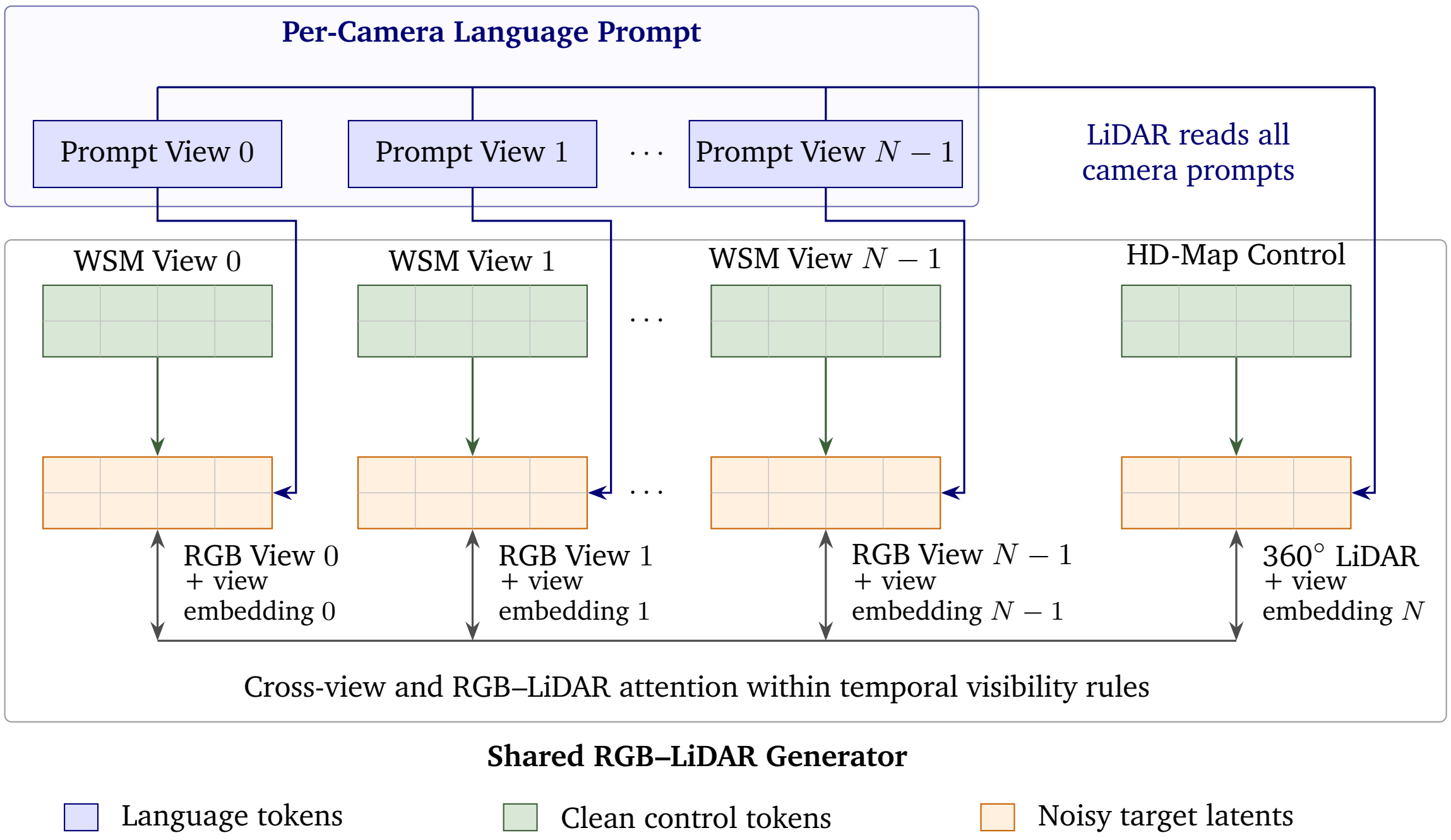
\begin{figure}[t]
\centering
\input{figures/diagrams/auto_overview.tex}
\caption{\textbf{Multi-view RGB--LiDAR generation.}
Each column represents a camera or LiDAR stream.
Colors denote language tokens (lavender), clean controls (green), and noisy target latents (light orange), as in \cref{fig::token_packing}.
Each camera reads its own prompt and WSM control, while LiDAR reads all camera prompts and its HD-map control.
Learned identity embeddings distinguish cameras $0,\ldots,N-1$ and LiDAR $N$, retaining their identities when selecting camera subsets.
Arrows connect conditioning inputs to their consumers, and the lower bidirectional links show cross-sensor attention under the temporal visibility rules.
Representative views are shown, with tokenizers and optional initial observations omitted.
See \cref{sec::multiview_attention,sec::auto_data,sec::auto_training} for the model architecture, caption preparation, and training recipe.}
\label{fig::auto_overview}
\end{figure}
\ifhmode\par\fi\endgroup

%% file: figures/diagrams/auto_overview.tex
\begingroup
\input{figures/diagrams/architecture_styles.tex}
\newcommand{\autolatentgrid}[5]{%
  \node[#4,minimum width=2.4cm,minimum height=0.75cm] (#1) at (#2,#3) {};
  \begin{scope}[shift={(#2,#3)}]
    \foreach \division in {-0.6,0,0.6} {
      \draw[sim grid] (\division,-0.375) -- (\division,0.375);
    }
    \draw[sim grid] (-1.2,0) -- (1.2,0);
    \node[above] at (0,0.4) {#5};
  \end{scope}%
}
\fitdiagram{%
\begin{tikzpicture}[font=\fontfamily{bch}\fontsize{9}{10.5}\selectfont,>=Stealth,
  every node/.append style={text=.},
  caption box/.style={sim text,minimum width=2.6cm,minimum height=0.7cm},
  heading/.style={font=\fontfamily{bch}\fontsize{9}{10.5}\selectfont\bfseries},
  exchange/.style={<->,semithick,draw=simgrayedge}]

\draw[draw=simtextedge!55,fill=simtextfill!15,rounded corners=2pt]
  (-1.6,3.0) rectangle (8.6,5.1);
\node[heading,text=simtextedge] at (3.5,4.8) {Per-Camera Language Prompt};

\draw[draw=simgrayedge!55,rounded corners=2pt]
  (-1.6,2.65) rectangle (13.5,-2.4);

\foreach \view/\viewx/\viewlabel in {0/0/0,1/3.3/1,n/7/{N-1}} {
  \node[caption box] (caption\view) at (\viewx,3.55)
    {Prompt View $\viewlabel$};
  \autolatentgrid{wsm\view}{\viewx}{1.8}{sim action}{WSM View $\viewlabel$}
  \autolatentgrid{camera\view}{\viewx}{0}{sim noisy}{}
  \node[anchor=west] at (\viewx+0.15,-0.65) {RGB View $\viewlabel$};
  \node[anchor=west,align=left,font=\fontfamily{bch}\fontsize{8}{9}\selectfont]
    at (\viewx+0.15,-1.1) {+ view\\embedding $\viewlabel$};
  \draw[sim flow,draw=simtextedge] (caption\view.south)
    -- (\viewx,2.85) -- (\viewx+1.45,2.85)
    -- (\viewx+1.45,0) -- (camera\view.east);
  \draw[sim flow,draw=simactionedge] (wsm\view.south) -- (camera\view.north);
}
\node at (5.15,3.55) {$\cdots$};
\node at (5.15,1.8) {$\cdots$};
\node at (5.15,0) {$\cdots$};

\autolatentgrid{hdmap}{11.3}{1.8}{sim action}{HD-Map Control}
\autolatentgrid{lidar}{11.3}{0}{sim noisy}{}
\node[anchor=west] at (11.45,-0.65) {360$^\circ$ LiDAR};
\node[anchor=west,align=left,font=\fontfamily{bch}\fontsize{8}{9}\selectfont]
  at (11.45,-1.1) {+ view\\embedding $N$};
\draw[sim flow,draw=simactionedge] (hdmap.south) -- (lidar.north);

\draw[draw=simtextedge,semithick] (0,4.25) -- (12.75,4.25);
\foreach \view/\viewx in {0/0,1/3.3,n/7} {
  \draw[draw=simtextedge,semithick] (caption\view.north) -- (\viewx,4.25);
}
\draw[sim flow,draw=simtextedge] (12.75,4.25) -- (12.75,0) -- (lidar.east);
\node[align=center,text=simtextedge] at (10.8,3.55)
  {LiDAR reads all\\camera prompts};

\draw[semithick,draw=simgrayedge] (0,-1.55) -- (11.3,-1.55);
\foreach \stream/\streamx in {camera0/0,camera1/3.3,cameran/7,lidar/11.3} {
  \draw[exchange] (\stream.south) -- (\streamx,-1.55);
}
\node[align=center] at (5.65,-2.05)
  {Cross-view and RGB--LiDAR attention within temporal visibility rules};
\node[heading] at (5.65,-2.75) {Shared RGB--LiDAR Generator};

\foreach \legendx/\legendstyle/\legendtext in {
  -0.8/{sim text}/{Language tokens},
  3.8/{sim action}/{Clean control tokens},
  8.8/{sim noisy}/{Noisy target latents}} {
  \node[\legendstyle,minimum width=0.35cm,minimum height=0.28cm,inner sep=0pt]
    at (\legendx,-3.4) {};
  \node[anchor=west] at (\legendx+0.3,-3.4) {\legendtext};
}

\end{tikzpicture}}
\endgroup

%% file: sections/5__examples/auto/architecture.tex
\begingroup
\makeatletter
\c@topnumber=1\relax
\makeatother

\begingroup{}
\subsection{Architecture}
\label{sec::multiview_attention}
\ifhmode\par\fi\endgroup

\begingroup{}
Cameras and LiDAR capture complementary aspects of the same scene at different spatial resolutions and sampling rates.
We bring these streams into a shared generator through modality-specific representations (\cref{sec::auto_modality_representations}), aligned temporal coordinates, and learned sensor identities (\cref{sec::auto_position_view_identity}).
Structured attention connects each stream to its language prompts, controls, and other sensors (\cref{sec::auto_multiview_modeling}).
For streaming generation, we adapt these connections to predict synchronized sensor chunks from the available history (\cref{sec::auto_causal_modeling}).
\ifhmode\par\fi\endgroup

\input{figures/auto/auto_wsm_correspondence.tex}

\Needspace{5\baselineskip}
\begingroup{}
\subsubsection{Modality Representations}
\label{sec::auto_modality_representations}
\ifhmode\par\fi\endgroup

\paragraph[\kern0pt RGB signal.]{{RGB signal.}}
\begingroup{}
RGB generation inherits the video tokenization and transfer sequence-packing scheme of Cosmos 3 (\cref{sec::shared_architecture,sec::token_packing}).
The same video VAE encodes RGB targets and rendered WSM controls into 48-channel latents with $16\times16$ spatial and $4\times$ temporal compression.
Learned projections map $2\times2$ spatial latent patches into generator features.
Control tokens remain clean while target tokens undergo denoising.
The multi-view extension encodes each camera independently and packs the synchronized streams together, retaining their physical timestamps and adding learned camera identity embeddings.
Each camera stream is associated with its own language prompt, while cross-view attention connects the visual streams under the visibility rules described in \cref{sec::auto_multiview_modeling}.
\ifhmode\par\fi\endgroup

\paragraph[\kern0pt LiDAR signal.]{{LiDAR signal.}}
\begingroup{}
For the LiDAR signal, we use a separate hierarchical transformer VAE inspired by LidarVerse~\citep{che2026lidarverse}, with interleaved spatial neighborhood attention and causal temporal attention.
Each 360$^\circ$ sweep is projected onto a $128\times1800$ range image with range, intensity, and binary return-validity channels.
The azimuth dimension is circularly padded to 1808 columns to accommodate the tokenizer's $16\times16$ spatial compression.
There is no temporal downsampling: a sequence of $T$ sweeps produces latents of shape $(128,T,8,113)$, ordered as channels, sweeps, height, and width, with the batch dimension omitted.
The generator uses $1\times1$ spatial patches on this latent grid, yielding $8\times113=904$ tokens per sweep.
A learned input projection maps each 128-channel latent patch into the shared generator feature space.
A separate output projection predicts rectified flow velocity in LiDAR latent space~\citep{liu2023rectifiedflow}.
After denoising, the LiDAR decoder reconstructs range, intensity, and return validity from the completed latents.
\ifhmode\par\fi\endgroup

\paragraph[\kern0pt Structural controls.]{{Structural controls.}}
\begingroup{}
WSM videos specify the scene layout and actor motion that guide RGB video generation~\citep{cosmos_predict2p5}.
\ifhmode\par\fi\endgroup
\begingroup{}
The map layers include lane lines, road boundaries, crosswalks, road markings, wait lines, poles, traffic signs, and traffic lights.
Projected 3D cuboids represent vehicles, pedestrians, and other road users.
Semantic colors identify object categories, cuboid shading conveys heading, and successive frames describe motion.
Rendering along the ego trajectory with each camera's intrinsics and extrinsics implicitly conveys ego motion and viewing geometry~\citep{omnidreams2026}.
The RGB--WSM overlay in \cref{fig::auto_wsm_correspondence} illustrates the resulting alignment.
\ifhmode\par\fi\endgroup

\begingroup{}
For LiDAR generation, we render HD-map structure in the range-view projection used for the sensor sweeps.
These controls provide spatially aligned map geometry for predicting LiDAR returns.
\Cref{fig::auto_wsm_correspondence} pairs the recorded LiDAR range image with its HD-map control at the same timestamp.
\ifhmode\par\fi\endgroup

\paragraph[\kern0pt Sensor timing and temporal coordinates.]{{Sensor timing and temporal coordinates.}}
\begingroup{}
Camera and LiDAR observations are placed on a shared physical timeline to align conditioning inputs and prediction targets (\cref{fig::auto_sensor_timing}).
The Wan 2.2 RGB tokenizer applies $4\times$ temporal compression to the 30-FPS camera streams, yielding latent time steps spaced $4/30$ seconds apart (7.5 Hz).
The separate LiDAR tokenizer preserves the 10-Hz sweep rate without temporal downsampling, retaining a latent spacing of $1/10$ seconds (\cref{sec::streaming_positions}).
These different latent sampling rates mean that equal temporal indices across modalities need not correspond to the same capture time.
We therefore align the streams by physical timestamp while retaining each token's sensor identity.
Capture time determines this alignment independently of diffusion time, rotary coordinates, and packed sequence order.
\ifhmode\par\fi\endgroup

\begingroup{}
\input{figures/auto/auto_sensor_timing.tex}
\ifhmode\par\fi\endgroup

\Needspace{4\baselineskip}
\begingroup{}
\subsubsection{Positional Encoding and View Identity}
\label[section]{sec::auto_position_view_identity}
\ifhmode\par\fi\endgroup

\paragraph[\kern0pt Positional encoding.]{{Positional encoding.}}
\begingroup{}
We retain the three-axis MRoPE formulation of Cosmos 3, using camera-local spatial coordinates and a shared temporal origin across synchronized views.
Each camera uses the same spatial indexing convention, while temporal coordinates follow the FPS scaling described in \cref{sec::shared_architecture}.
Consequently, tokens at matching image-grid locations and timestamps receive identical rotary positional encodings across cameras, irrespective of their order in the packed sequence.
These coordinates describe position within each view, rather than a common 3D location or camera identity.
\ifhmode\par\fi\endgroup

\paragraph[\kern0pt Learned view embeddings.]{{Learned view embeddings.}}
\begingroup{}
Inspired by Cosmos-Transfer2.5 Multiview~\citep[Sec.~6.3.1]{cosmos_predict2p5}, we introduce learned per-view embeddings to distinguish cameras that share the same positional coordinates.
After tokenization and projection into transformer features, we add a camera-specific embedding vector to every token from that camera.
This acts as a per-view bias shared across spatial locations and time, learned jointly with the generator.
The embedding complements the view-dependent geometry in WSM controls by providing a persistent camera identity across scenes.
We find this simple design effective for our fixed training rig, where each identity consistently corresponds to a camera placement and field of view.
\ifhmode\par\fi\endgroup

\Needspace{4\baselineskip}
\begingroup{}
\subsubsection{Multi-View and Multimodal Attention}
\label[section]{sec::auto_multiview_modeling}
\ifhmode\par\fi\endgroup

\begingroup{}
Multi-view, multimodal attention is the core mechanism that brings our camera and LiDAR streams, language prompts, and structural controls together (\cref{fig::auto_attention_masks}).
This section describes the attention design of Cosmos3-Nano-Transfer-Auto, including how tokens interact across views and modalities.
The changes for Cosmos3-Nano-Sim-Auto are described in the Causal Modeling section (\cref{sec::auto_causal_modeling}).
\ifhmode\par\fi\endgroup

\begingroup{}
\input{figures/auto/auto_attention_masks.tex}
\ifhmode\par\fi\endgroup

\paragraph[\kern0pt Language-to-sensor conditioning.]{{Language-to-sensor conditioning.}}
\begingroup{}
Each camera has its own caption, which we tokenize separately.
For language conditioning, the camera's visual tokens attend only to the tokens from that caption (\cref{fig::auto_attention_masks}(a)).
This keeps descriptions of object visibility, appearance, and image-plane motion tied to the view they describe.
LiDAR tokens attend to all camera caption tokens from the same recording.
These captions cover the same synchronized sensor interval, providing language context from multiple viewpoints for the 360$^\circ$ LiDAR stream.
\ifhmode\par\fi\endgroup

\paragraph[\kern0pt Control conditioning.]{{Control conditioning.}}

\begingroup{}
WSM and HD-map controls enter the generator through the sequence-packing scheme for control-conditioned generation (\cref{sec::token_packing}).
Their encoded tokens are packed alongside the sensor tokens and remain clean throughout denoising.
Each camera stream attends to its own WSM tokens, while the LiDAR stream attends to its HD-map tokens (\cref{fig::auto_attention_masks}(a)).
This access is separate from sensor-to-sensor attention: exchanging information across cameras does not give a camera direct access to another camera's control tokens.
The available control sequence and conditioning masks determine temporal access.
\ifhmode\par\fi\endgroup

\begingroup{}
Control tokens also attend to other control tokens within their own stream.
When control-to-sensor attention is enabled, they additionally read sensor features from that stream, allowing the control representation to incorporate sensor context.
In Cosmos3-Nano-Sim-Auto, this context is restricted to clean sensor history from preceding chunks.
The sensor-to-sensor routes described below connect supplied observations and prediction targets under their own temporal and clean/noisy visibility rules.
\ifhmode\par\fi\endgroup

\paragraph[\kern0pt Sensor-to-sensor attention.]{{Sensor-to-sensor attention.}}

\begingroup{}
A straightforward extension from single-view to multi-view generation lets every visual token attend to all tokens across cameras and time.
The resulting cost grows quadratically with both the number of views and the sequence length, making long surround-view clips expensive to process.
Inspired by OmniDreams~\citep{omnidreams2026}, we decompose attention into temporal interaction within each view and information exchange across views.
Where OmniDreams connects different views at the same time step, we extend cross-sensor interaction to a preceding 0.4-second window in physical time.
This lets cameras and LiDAR exchange nearby observations even when their sampling times do not coincide.
Same-sensor attention still allows each token to attend to all spatial and temporal tokens (clip-wide) within the same view.
\ifhmode\par\fi\endgroup

\begingroup{}
For $N_{\mathrm v}$ camera views with $N_{\mathrm t}$ latent frames per camera and $N_{\mathrm s}$ spatial tokens per latent frame, unrestricted attention is quadratic in the total token count $N_{\mathrm v}N_{\mathrm t}N_{\mathrm s}$.
An idealized decomposition that uses all same-view times and only the matching time across cameras instead has costs
\begin{align}
 C_{\text{full}} &= \mathcal{O}\bigl((N_{\mathrm v}N_{\mathrm t}N_{\mathrm s})^2\bigr), \label{eq::attention_cost_full} \\
 C_{\text{decomposed}} &= \mathcal{O}(N_{\mathrm v}N_{\mathrm t}^2N_{\mathrm s}^2 + N_{\mathrm v}^2N_{\mathrm t}N_{\mathrm s}^2).
 \label{eq::attention_cost}
\end{align}
Taking the ratio of the leading terms in \cref{eq::attention_cost_full,eq::attention_cost} for 11 views and 100 latent frames per camera (397 RGB frames) gives an approximately $9.9\times$ theoretical reduction in camera attention computation.
\ifhmode\par\fi\endgroup

\paragraph[\kern0pt Sliding window cross-sensor attention.]{{Sliding window cross-sensor attention.}}

\begingroup{}
Cross-sensor attention uses a sliding window spanning the preceding 0.4 seconds, including the query timestamp (\cref{fig::auto_attention_masks}(b)).
RGB and LiDAR latent time steps are spaced by $4/30$ seconds and $1/10$ seconds, respectively.
Attention visibility is determined by physical timestamps to account for the different sensor sampling rates.
The 0.4-second window matches their smallest shared temporal period, spanning 3 RGB latent intervals and 4 LiDAR latent intervals.
In our evaluation, this preceding window yielded better results than a symmetric window extending 0.2 seconds before and after the query timestamp.
At each permitted timestamp, queries attend to the full spatial token grids of eligible streams from other sensors.
The cross-sensor window explicitly excludes the query's own view, whose eligible keys are already handled by same-stream attention.
Including that same-view window again in the cross-sensor branch yielded worse results in our evaluation.
\ifhmode\par\fi\endgroup

\Needspace{4\baselineskip}
\begingroup{}
\subsubsection{Causal Modeling}
\label[section]{sec::auto_causal_modeling}
\ifhmode\par\fi\endgroup

\begingroup{}
Interactive driving simulation requires camera and LiDAR outputs to advance together as scene controls change.
We adapt Cosmos3-Nano-Transfer-Auto for this setting by generating short, synchronized sensor chunks conditioned on the supplied controls and previously generated observations.
Cosmos3-Nano-Sim-Auto retains the per-view language routing, camera embeddings, structural conditioning, and decomposed sensor-to-sensor attention described in \cref{sec::auto_position_view_identity,sec::auto_multiview_modeling}.
\ifhmode\par\fi\endgroup

\paragraph[\kern0pt Synchronized sensor chunks.]{{Synchronized sensor chunks.}}
\begingroup{}
RGB and LiDAR advance at different sampling rates, so a common frame count would cover different physical durations.
We instead group their predictions by the shared 0.4-second interval introduced in \cref{sec::auto_multiview_modeling}.
After the initial time step, each full chunk contains 12 RGB frames per camera, represented by 3 latent time steps, and 4 LiDAR sweeps.
The model jointly denoises these aligned sensor intervals while preserving the distinction between same-stream and cross-sensor visibility.
\ifhmode\par\fi\endgroup

\paragraph[\kern0pt Causal multi-view attention.]{{Causal multi-view attention.}}
\begingroup{}
Cosmos3-Nano-Sim-Auto replaces the clip-wide same-sensor attention of Cosmos3-Nano-Transfer-Auto with the shared block-causal formulation (\cref{sec::model_architecture}).
Within each sensor stream, queries attend to all spatial tokens throughout the current chunk and to permitted clean history from preceding chunks (\cref{fig::auto_attention_masks}(c)).
Across streams, the preceding 0.4-second window remains unchanged (\cref{fig::auto_attention_masks}(b)), so a query cannot read another sensor's later timestamp even when both belong to the current chunk.
The control and clean/noisy conditioning rules remain those described in \cref{sec::auto_multiview_modeling,sec::teacher_forcing}.
Future target chunks are excluded, allowing generation to advance without depending on future sensor outputs.
Conditioning masks continue to prevent noisy targets from attending to their own clean~reference~copies.
\ifhmode\par\fi\endgroup

\paragraph[\kern0pt Continuing the rollout.]{{Continuing the rollout.}}
\begingroup{}
Once a chunk is complete, its clean sensor representations become history for the next prediction, which can incorporate newly supplied scene controls.
Cosmos3-Nano-Sim-Auto uses the shared streaming mechanism in \cref{sec::streaming_execution} to reuse this history and bound the retained cache during extended rollouts.
Attention visibility determines which stored tokens may be read, while the cache policy determines which past chunks remain available.
The Cosmos3-Nano-Sim-Auto runtime settings are described in \cref{sec::auto_inference_optimization}.
We train this causal generator through the shared teacher-forcing, consistency-distillation, and self-forcing stages, with settings and checkpoint handoffs for Cosmos3-Nano-Sim-Auto detailed in \cref{sec::auto_training}.
\ifhmode\par\fi\endgroup

\endgroup

%% file: figures/auto/auto_wsm_correspondence.tex
\begingroup{}
\begin{figure}[t]
\centering
\includegraphics[width=\linewidth]{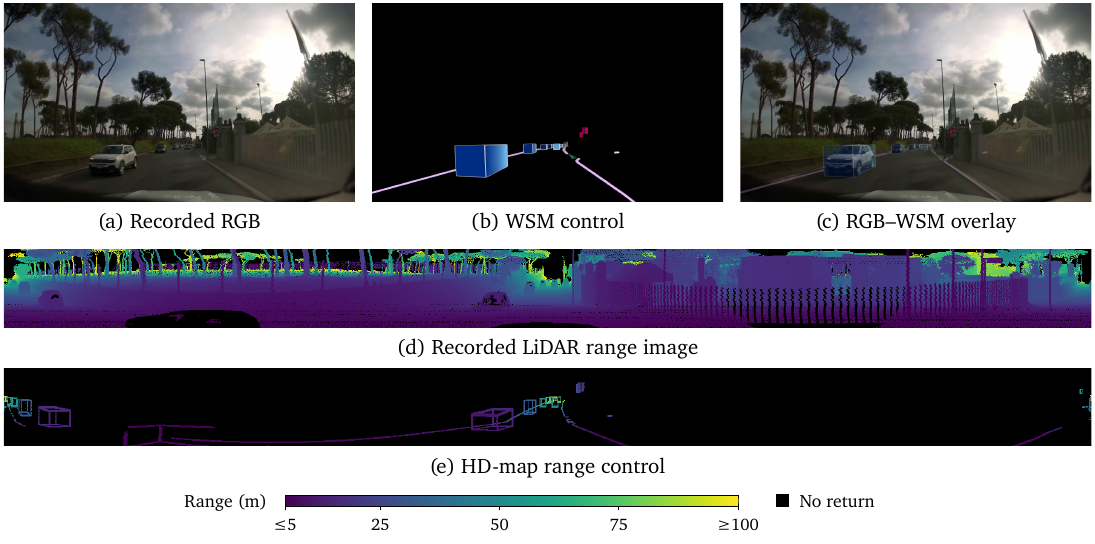}
\caption{\textbf{Sensor-aligned structural controls.}
Recorded RGB, WSM, and their overlay (top) illustrate camera-control alignment in one data sample.
A recorded LiDAR range image and its HD-map control are shown below.}
\label{fig::auto_wsm_correspondence}
\end{figure}
\ifhmode\par\fi\endgroup

%% file: figures/auto/auto_sensor_timing.tex
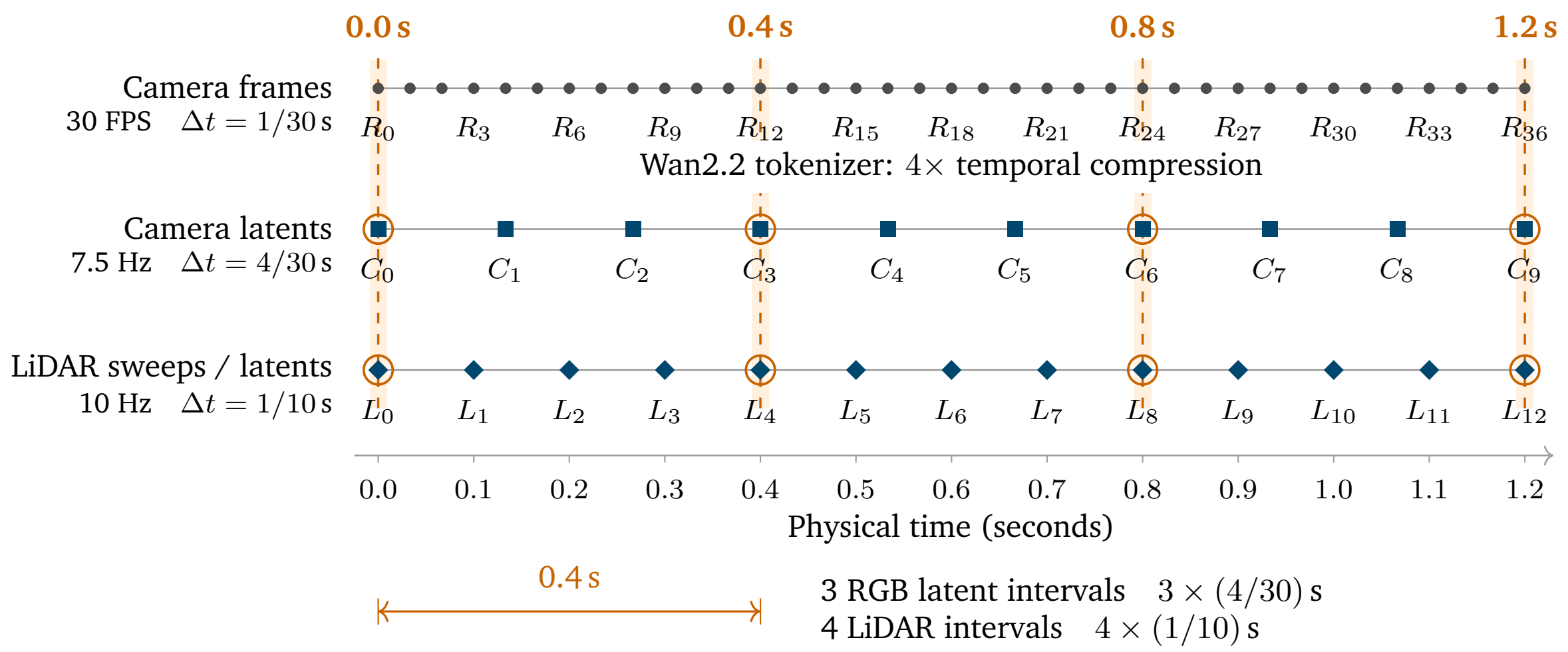
\begin{figure}[t]
\centering
\input{figures/diagrams/auto_sensor_timing.tex}
\caption{\textbf{Camera and LiDAR sampling on a shared physical timeline.}
The illustrative grids share a temporal origin.
Camera frames ($R_i$) are sampled at 30 FPS, and the Wan2.2 tokenizer applies $4\times$ temporal compression to produce camera latent steps ($C_j$) spaced by $4/30$ seconds (7.5 Hz).
LiDAR sweeps and their latent steps ($L_k$) retain the same $1/10$-second spacing.
Dashed guides and rings mark coincident camera-latent and LiDAR timestamps every 0.4 seconds, spanning three RGB latent intervals and four LiDAR intervals.
Markers denote temporal positions rather than tokenizer receptive fields.
Cross-sensor attention compares physical timestamps even when samples do not coincide.}
\label{fig::auto_sensor_timing}
\end{figure}

%% file: figures/diagrams/auto_sensor_timing.tex
\begingroup
\input{figures/diagrams/architecture_styles.tex}
\fitdiagram{%
\begin{tikzpicture}[x=9.5cm,y=1cm,
  font=\fontencoding{T1}\fontfamily{bch}\fontsize{9}{10.5}\selectfont,
  timing label/.style={anchor=east,align=right},
  timing index/.style={font=\fontencoding{T1}\fontfamily{bch}\fontsize{8}{9}\selectfont},
  timing track/.style={draw=simgrayedge!55,line width=0.5pt}]
  \foreach \time/\timelabel in {0/0.0,0.4/0.4,0.8/0.8,1.2/1.2} {
    \fill[simnoisyfill] ({\time-0.009},0.28) rectangle ({\time+0.009},-3.18);
    \draw[simnoisyedge,dashed,line width=0.65pt]
      (\time,0.30) -- (\time,-3.20);
    \node[anchor=south,text=simnoisyedge,
      font=\fontencoding{T1}\fontfamily{bch}\fontsize{9}{10.5}\selectfont\bfseries]
      at (\time,0.38) {\timelabel\,s};
  }

  \node[timing label] at (-0.035,0.02) {Camera frames};
  \node[timing label,timing index] at (-0.035,-0.35) {30 FPS\quad $\Delta t=1/30$\,s};
  \draw[timing track] (0,0) -- (1.2,0);
  \foreach \frame in {0,...,36} {
    \fill[simgrayedge] ({\frame/30},0) circle[radius=1.6pt];
  }
  \foreach \frame in {0,3,...,36} {
    \node[timing index,anchor=north] at ({\frame/30},-0.16) {$R_{\frame}$};
  }

  \node[fill=white,inner sep=3pt] at (0.6,-0.78)
    {Wan2.2 tokenizer: $4\times$ temporal compression};
  \node[timing label] at (-0.035,-1.38) {Camera latents};
  \node[timing label,timing index] at (-0.035,-1.75) {7.5 Hz\quad $\Delta t=4/30$\,s};
  \draw[timing track] (0,-1.4) -- (1.2,-1.4);
  \foreach \step in {0,...,9} {
    \node[rectangle,fill=simcleanedge,minimum size=4.5pt,inner sep=0pt]
      at ({4*\step/30},-1.4) {};
    \node[timing index,anchor=north] at ({4*\step/30},-1.57) {$C_{\step}$};
  }

  \node[timing label] at (-0.035,-2.78) {LiDAR sweeps / latents};
  \node[timing label,timing index] at (-0.035,-3.15) {10 Hz\quad $\Delta t=1/10$\,s};
  \draw[timing track] (0,-2.8) -- (1.2,-2.8);
  \foreach \step in {0,...,12} {
    \node[rectangle,rotate=45,fill=simcleanedge,minimum size=4pt,inner sep=0pt]
      at ({\step/10},-2.8) {};
    \node[timing index,anchor=north] at ({\step/10},-2.97) {$L_{\step}$};
  }

  \foreach \time in {0,0.4,0.8,1.2} {
    \foreach \row in {-1.4,-2.8} {
      \draw[simnoisyedge,line width=0.7pt] (\time,\row) circle[radius=4.1pt];
    }
  }

  \draw[timing track,->] (-0.025,-3.65) -- (1.23,-3.65);
  \foreach \tick in {0,...,12} {
    \draw[simgrayedge!55] ({\tick/10},-3.65) -- ++(0,-0.07);
    \pgfmathsetmacro{\ticktime}{\tick/10}
    \node[timing index,anchor=north] at ({\tick/10},-3.77)
      {\pgfmathprintnumber[fixed,precision=1,zerofill]{\ticktime}};
  }
  \node[anchor=north] at (0.6,-4.10) {Physical time (seconds)};

  \draw[<->,simnoisyedge,semithick] (0,-5.20) -- (0.4,-5.20);
  \foreach \time in {0,0.4} {
    \draw[simnoisyedge] (\time,-5.07) -- (\time,-5.33);
  }
  \node[anchor=south,text=simnoisyedge] at (0.2,-5.10) {0.4\,s};
  \node[anchor=west,align=left] at (0.45,-5.20)
    {3 RGB latent intervals\quad $3\times(4/30)$\,s\\
     4 LiDAR intervals\quad $4\times(1/10)$\,s};
\end{tikzpicture}}
\endgroup

%% file: figures/auto/auto_attention_masks.tex
\begin{figure}[!t]
\centering
\input{figures/diagrams/auto_attention_masks.tex}
\caption{\textbf{Joint camera--LiDAR attention masks.}
Rows are queries and columns are keys/values.
(a) Cosmos3-Nano-Transfer-Auto with optional control-to-sensor access enabled (dashed blocks).
Camera sensor and control queries read their own captions, while LiDAR queries read all camera captions (purple).
Green and blue show own-control and own-sensor access.
Only sensor queries read other sensors within the preceding 0.4-second window (orange).
Wan2.2's $4\times$ temporal compression maps 30-FPS camera streams to 7.5-Hz latents, while LiDAR retains 10 Hz.
Temporal indices span $[0,1.6]$ seconds: 13 camera latent times and 17 LiDAR times make the moving 0.4-second window visible as a narrow band.
Caption lengths are illustrative.
(b) Physical timestamps bound the orange band, excluding keys older than 0.4 seconds and future keys.
(c) Same-sensor attention is clip-wide for Cosmos3-Nano-Transfer-Auto and block-causal for Cosmos3-Nano-Sim-Auto, which reads the current chunk and permitted earlier clean history.
Thick lines separate illustrative chunks C1--C3.
For Cosmos3-Nano-Sim-Auto, control-to-sensor access is restricted to preceding clean history.
Gray cells are blocked, spatial tokens are collapsed, and caption query rows are omitted.
All routes respect conditioning visibility, recording boundaries, and available context.}
\label{fig::auto_attention_masks}
\end{figure}
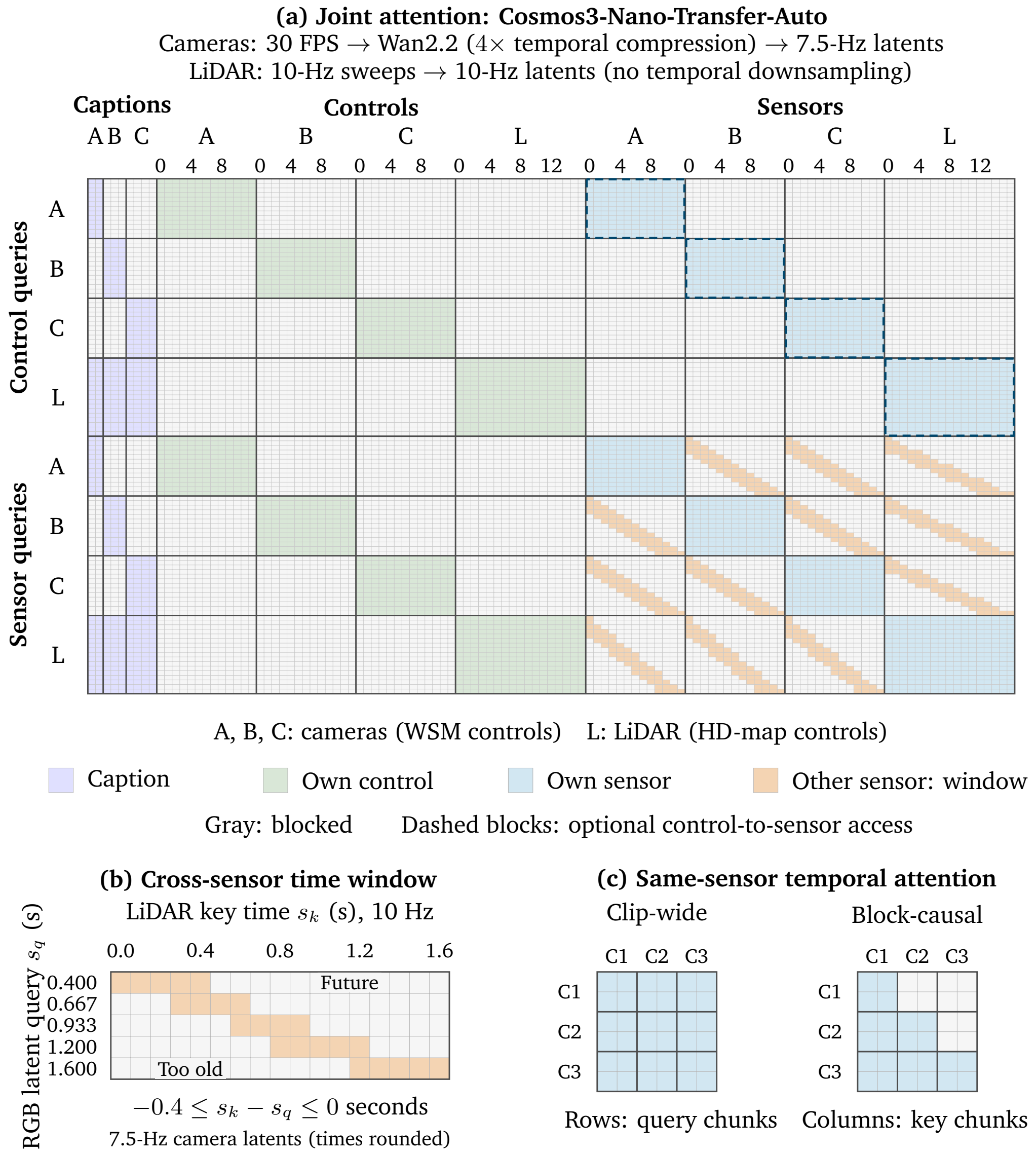

%% file: figures/diagrams/auto_attention_masks.tex
\begingroup
\input{figures/diagrams/architecture_styles.tex}
\fitdiagram{%
\begin{tikzpicture}[font=\fontfamily{bch}\fontsize{9}{10.5}\selectfont,
  mask heading/.style={font=\fontfamily{bch}\fontsize{9}{10.5}\selectfont\bfseries},
  mask tick/.style={font=\fontfamily{bch}\fontsize{7.5}{8.5}\selectfont},
  mask grid/.style={draw=simgrayedge!40,line width=0.2pt},
  mask boundary/.style={draw=simgrayedge,line width=0.55pt}]
  \begin{scope}[shift={(1.35,0)},x=0.10cm,y=0.060cm,
    mask grid/.style={draw=simgrayedge!30,line width=0.15pt}]
    \node[mask heading] at (60.5,35.15) {(a) Joint attention: Cosmos3-Nano-Transfer-Auto};
    \node at (60.5,29.12)
      {Cameras: 30 FPS $\rightarrow$ Wan2.2 ($4\times$ temporal compression) $\rightarrow$ 7.5-Hz latents};
    \node at (60.5,23.09)
      {LiDAR: 10-Hz sweeps $\rightarrow$ 10-Hz latents (no temporal downsampling)};
    \node[mask heading] at (4.5,15.74) {Captions};
    \node[mask heading] at (37,15.74) {Controls};
    \node[mask heading] at (93,15.74) {Sensors};
    \foreach \start/\groupcount/\view in {0/2/A,2/3/B,5/4/C} {
      \node at ({\start+\groupcount/2},9.37) {\view};
    }
    \foreach \start/\groupcount/\view in {0/13/A,13/13/B,26/13/C,39/17/L} {
      \node at ({9+\start+\groupcount/2},9.37) {\view};
      \node at ({65+\start+\groupcount/2},9.37) {\view};
      \node[anchor=east] at (-1.824,{-\start-\groupcount/2}) {\view};
      \node[anchor=east] at (-1.824,{-56-\start-\groupcount/2}) {\view};
      \pgfmathtruncatemacro{\lasttime}{\groupcount-5}
      \foreach \timeindex in {0,4,...,\lasttime} {
        \node[mask tick] at ({9+\start+\timeindex+0.5},2.84) {\timeindex};
        \node[mask tick] at ({65+\start+\timeindex+0.5},2.84) {\timeindex};
      }
    }
    \node[rotate=90,mask heading] at (-8.84,-28) {Control queries};
    \node[rotate=90,mask heading] at (-8.84,-84) {Sensor queries};
    \fill[simgrayfill!45] (0,0) rectangle (121,-112);
    \foreach \querykind in {0,1} {
      \foreach \qview/\qstart/\qcount/\qstep/\captionstart/\captioncount in {
        0/0/13/4/0/2,1/13/13/4/2/3,2/26/13/4/5/4,3/39/17/3/0/9} {
        \pgfmathtruncatemacro{\rowstart}{56*\querykind+\qstart}
        \fill[simtextfill] (\captionstart,-\rowstart)
          rectangle ++(\captioncount,-\qcount);
        \fill[simactionfill] ({9+\qstart},-\rowstart)
          rectangle ++(\qcount,-\qcount);
        \fill[simcleanfill] ({65+\qstart},-\rowstart)
          rectangle ++(\qcount,-\qcount);
        \ifnum\querykind=1
          \foreach \kview/\kstart/\kcount/\kstep in {
            0/0/13/4,1/13/13/4,2/26/13/4,3/39/17/3} {
            \ifnum\qview=\kview\else
              \pgfmathtruncatemacro{\lastquery}{\qcount-1}
              \pgfmathtruncatemacro{\lastkey}{\kcount-1}
              \foreach \qtime in {0,...,\lastquery} {
                \foreach \ktime in {0,...,\lastkey} {
                  \pgfmathtruncatemacro{\offset}{\kstep*\ktime-\qstep*\qtime}
                  \pgfmathtruncatemacro{\allowed}{(\offset>=-12)&&(\offset<=0)}
                  \ifnum\allowed=1
                    \fill[simnoisyfill!80!simnoisyedge] ({65+\kstart+\ktime},{-\rowstart-\qtime})
                      rectangle ++(1,-1);
                  \fi
                }
              }
            \fi
          }
        \fi
      }
    }
    \draw[mask grid,step=1] (0,0) grid (121,-112);
    \foreach \boundary in {2,5,9,22,35,48,65,78,91,104} {
      \draw[mask boundary] (\boundary,0) -- (\boundary,-112);
    }
    \foreach \boundary in {13,26,39,56,69,82,95} {
      \draw[mask boundary] (0,-\boundary) -- (121,-\boundary);
    }
    \draw[mask boundary] (0,0) rectangle (121,-112);
    \foreach \start/\groupcount in {0/13,13/13,26/13,39/17} {
      \draw[draw=simcleanedge,densely dashed,line width=0.8pt]
        ({65+\start+0.12},{-\start-0.12}) rectangle ++({\groupcount-0.24},{-\groupcount+0.24});
    }
    \node[anchor=north] at (60.5,-116.0)
      {A, B, C: cameras (WSM controls)\quad L: LiDAR (HD-map controls)};
  \end{scope}
  \begin{scope}[yshift=2cm]
  \foreach \legendx/\legendcolor/\legendlabel in {
    1.0/simtextfill/Caption,
    3.8/simactionfill/{Own control},
    7.0/simcleanfill/{Own sensor},
    10.2/{simnoisyfill!80!simnoisyedge}/{Other sensor: window}} {
    \node[mask grid,fill=\legendcolor,minimum size=0.32cm,inner sep=0pt]
      at (\legendx,-9.85) {};
    \node[anchor=west] at ({\legendx+0.22},-9.85) {\legendlabel};
  }
  \node at (7.5,-10.45)
    {Gray: blocked\qquad Dashed blocks: optional control-to-sensor access};

  \node[mask heading] at (3.7,-11.15) {(b) Cross-sensor time window};
  \begin{scope}[shift={(1.65,-12.35)},x=0.26cm,y=0.28cm]
    \node at (8.5,2.76) {LiDAR key time $s_k$ (s), 10 Hz};
    \foreach \column/\keylabel in {0/0.0,4/0.4,8/0.8,12/1.2,16/1.6} {
      \node[mask tick] at ({\column+0.5},1.0) {\keylabel};
    }
    \foreach \row/\querytime/\querylabel in {0/12/0.400,1/20/0.667,2/28/0.933,3/36/1.200,4/48/1.600} {
      \node[mask tick,anchor=east] at (-0.2,{-\row-0.5}) {\querylabel};
      \foreach \column in {0,...,16} {
        \pgfmathtruncatemacro{\offset}{3*\column-\querytime}
        \pgfmathtruncatemacro{\allowed}{(\offset>=-12)&&(\offset<=0)}
        \fill[simgrayfill!45] (\column,-\row) rectangle ++(1,-1);
        \ifnum\allowed=1
          \fill[simnoisyfill!80!simnoisyedge] (\column,-\row) rectangle ++(1,-1);
        \fi
      }
    }
    \node[rotate=90] at (-3.9,-2.5) {RGB latent query $s_q$ (s)};
    \draw[mask grid,step=1] (0,0) grid (17,-5);
    \draw[mask boundary] (0,0) rectangle (17,-5);
    \node[mask tick,fill=simgrayfill!45,inner sep=1pt] at (4,-4.5) {Too old};
    \node[mask tick,fill=simgrayfill!45,inner sep=1pt] at (12,-0.5) {Future};
    \node at (8.5,-6.35) {$-0.4 \le s_k-s_q \le 0$ seconds};
    \node[mask tick] at (8.5,-7.8) {7.5-Hz camera latents (times rounded)};
  \end{scope}

  \node[mask heading] at (10.6,-11.15) {(c) Same-sensor temporal attention};
  \foreach \causalmask/\panelx/\attentionname in {
    0/8.0/{Clip-wide},1/11.4/{Block-causal}} {
    \begin{scope}[shift={(\panelx,-12.35)},x=0.26cm,y=0.26cm]
      \node at (3,2.9) {\attentionname};
      \foreach \chunk in {1,2,3} {
        \node[mask tick] at ({2*\chunk-1},0.8) {C\chunk};
        \node[mask tick,anchor=east] at (-0.3,{-2*\chunk+1}) {C\chunk};
      }
      \foreach \row in {0,...,5} {
        \foreach \column in {0,...,5} {
          \pgfmathtruncatemacro{\allowed}{
            (\causalmask==0)||(floor(\column/2)<=floor(\row/2))}
          \fill[simgrayfill!45] (\column,-\row) rectangle ++(1,-1);
          \ifnum\allowed=1
            \fill[simcleanfill] (\column,-\row) rectangle ++(1,-1);
          \fi
        }
      }
      \draw[mask grid,step=1] (0,0) grid (6,-6);
      \draw[mask boundary] (0,0) rectangle (6,-6);
      \foreach \boundary in {2,4} {
        \draw[mask boundary] (\boundary,0) -- (\boundary,-6);
        \draw[mask boundary] (0,-\boundary) -- (6,-\boundary);
      }
    \end{scope}
  }
  \node at (10.6,-14.3) {Rows: query chunks\quad Columns: key chunks};
  \end{scope}
\end{tikzpicture}}
\endgroup

%% file: sections/5__examples/auto/data.tex
\begingroup{}
\subsection{Data Preparation}
\label{sec::auto_data}
\ifhmode\par\fi\endgroup

\paragraph[\kern0pt Sources and sampling.]{{Sources and sampling.}}
\begingroup{}
We train on synchronized multi-camera driving recordings from MADS, previously used for WSM-conditioned generation in Cosmos 3~\citep{cosmos3}.
Our captioned training selection contains approximately 988K source samples.
Each recording is counted as one sample, encompassing its camera and LiDAR streams before temporal chunking.
We hold out 91 clips as the Driving-Sim-Eval set\footnote{Cosmos3-Driving-Simulation-Evaluation: \href{https://huggingface.co/datasets/nvidia/Cosmos3-Driving-Simulation-Evaluation}{\nolinkurl{huggingface.co/datasets/nvidia/Cosmos3-Driving-Simulation-Evaluation}}.} for benchmarking and exclude them from training.
The evaluation protocol and results are described in \cref{sec::auto_results}.
Our Cosmos3-Nano-Sim-Auto training inputs comprise up to 11 camera views, including wide-angle, telephoto, and fisheye cameras, together with 360$^\circ$ LiDAR sweeps.
RGB videos are sampled at 30 FPS and LiDAR at 10 Hz.
\ifhmode\par\fi\endgroup

\begingroup{}
For captioning, we partition each source sample into consecutive 400-frame RGB chunks with time-aligned structural controls and LiDAR.
When the final chunk contains fewer than 400 frames, we retain and caption it as well.
Each source sample yields an average of 1.56 caption chunks per camera, typically 1--2 chunks.
The resulting collection is dominated by 400-frame chunks but also exposes the model to shorter sequences, providing diversity in training durations.
Camera subsets retain their physical identities and timestamps within each synchronized interval.
During training, we downsample the high-resolution RGB streams to a mixture of 480p and 720p clips, with stage-specific sequence lengths described in \cref{sec::auto_training}.
\ifhmode\par\fi\endgroup

\paragraph[\kern0pt Structural controls and alignment.]{{Structural controls and alignment.}}
\begingroup{}
We render controls according to the WSM v4 schema, using each camera's calibrated lens model, including fisheye distortion.
The WSM color and symbol encodings are illustrated in \cref{fig::auto_wsm_legend} in the appendix.
RGB targets and WSM controls use matching frame selection, resizing, and cropping so that the rendered geometry remains aligned with the recorded image.
For LiDAR, HD-map geometry is projected onto the same range-view rays as the measured sweeps.
We use LiDAR sweeps without motion compensation, retaining the spinning sensor's within-sweep motion distortion.
RGB and LiDAR intervals are matched by physical timestamps rather than equal frame indices.
We include the WSM rendering schema, reference renderer, and data-preparation guides in the accompanying open-source Cosmos~Framework~release.\footnote{Cosmos-Framework: \href{https://github.com/NVIDIA/cosmos-framework}{\nolinkurl{github.com/NVIDIA/cosmos-framework}}.}
\ifhmode\par\fi\endgroup

\paragraph[\kern0pt Caption preparation.]{{Caption preparation.}}

\label{sec::auto_captioning}
\begingroup{}
WSM controls describe scene structure and actor motion, but leave much of the visual appearance unspecified.
We complement them with dense, timestamped captions that describe visible objects and their appearance, together with time of day, lighting, weather, road-surface condition, and geographic region.
Descriptions emphasize concrete visual attributes, such as vehicle color, surface reflections, and~atmospheric~visibility.
An example timestamped caption is shown in \cref{sec::auto_caption_example}.
\ifhmode\par\fi\endgroup

\begingroup{}
We caption each chunk with an in-house vision-language model that receives synchronized RGB and WSM observations from all camera views together.
The joint input allows the captioner to reason about one physical scene before producing a separate description for each camera.
RGB provides appearance and environmental evidence, while WSM supplies the corresponding geometry, actor trajectories, and signal states.
Each caption describes only what its camera can observe, while shared objects retain consistent appearance and physical motion across views and time.
Scene-level conditions must also agree across cameras.
Captions account for each camera's viewpoint: the same vehicle may enter and leave different views at different times, or move leftward in one image and rightward in another, while following a consistent physical trajectory.
Captioned events and motion must agree with the WSM controls.
\ifhmode\par\fi\endgroup

\begingroup{}
At inference, users can perform \textit{prompt upsampling} using WSM controls and four environmental attributes: time of day, weather, road condition, and geographic region.
This produces a coherent set of per-camera prompts across all views.
The workflow is described in \cref{sec::auto_prompt_upsampling}, with practical instructions in the accompanying Cosmos Framework guide.
\ifhmode\par\fi\endgroup

\paragraph[\kern0pt Prompt prefixes.]{{Prompt prefixes.}}

\begingroup{}
During training, we sample different camera subsets (\cref{sec::auto_training}), so each example can present a different combination of viewpoints.
We therefore prepend a camera prefix that describes the currently sampled rig and identifies the current camera by its position or role, viewing direction, lens type, and nominal field of view.
This provides the model with context about which views are present and the viewing direction and angular coverage associated with each caption.
The task prefix explains WSM-conditioned generation: controls specify the supported objects, road layout, and motion, while captions supply appearance and unconstrained background details.
The joint RGB--LiDAR task instruction additionally identifies the HD-map controls and calls for camera--LiDAR alignment.
Examples of both prefixes are provided in \cref{sec::auto_caption_example}.
\ifhmode\par\fi\endgroup

\begingroup{}
These prefixes are assembled by the caption formatter and text-tokenization pipeline during training and inference.
The stored captions contain the scene descriptions, so the same rig and task context can be applied consistently without duplicating it in the annotations.
The camera prefix complements the learned sensor embeddings in \cref{sec::multiview_attention}, while the task prefix states how language and structural controls are combined.
\ifhmode\par\fi\endgroup

%% file: sections/5__examples/auto/post_training.tex
\begingroup{}
\subsection{Post-training}
\label{sec::auto_training}
\ifhmode\par\fi\endgroup

\begingroup{}
We summarize our model's training stages as a recommended recipe for adapting Cosmos3-Nano to multi-view RGB--LiDAR generation and distilling it into a causal streaming model.
The three phases of bidirectional training, collectively termed Stage~0, produce Cosmos3-Nano-Transfer-Auto.
The three causal training and distillation stages introduced in \cref{sec::training} convert the bidirectional model into its streaming counterpart, Cosmos3-Nano-Sim-Auto.
We periodically evaluate checkpoints on a validation set and advance to the next training phase or stage when benchmark scores plateau.
\ifhmode\par\fi\endgroup

\begingroup{}
\begin{table}[t]
\centering\reporttablestyle
\caption{\textbf{Training recipe for Cosmos3-Nano-Transfer-Auto and Cosmos3-Nano-Sim-Auto.}
Training uses 128 GB200 GPU nodes (512 GPUs); stage-specific allocations for causal training and distillation are listed in \cref{tab::auto_training_configuration}.
Training updates are counted separately within each bidirectional phase and causal stage.
RGB counts are per view after alignment and include observed prefixes.
Stage~0 learning rates denote schedule peaks.
SF-DMD update counts and learning rates are listed as student / fake-score model.}
\label{tab::auto_training_summary}
\begin{tabular}{lcccc}
\toprule
\textbf{Stage} & \textbf{Training updates} & \textbf{RGB frames} & \textbf{Learning rate} & \textbf{Teacher CFG}\\
\midrule
\multicolumn{5}{l}{Stage 0: Bidirectional training}\\
\quad \textit{Modality-separate} & 60,000 & $\leq397$ & $8\times10^{-5}$ & ---\\
\quad \textit{Short-sequence joint} & 20,000 & 61 & $5\times10^{-5}$ & ---\\
\quad \textit{Long-sequence joint} & 20,000 & $\leq397$ & $5\times10^{-5}$ & ---\\
Stage 1: TF & 3,200 & $\leq397$ & $10^{-5}$ & ---\\
Stage 2: TF-dCM & 900 & $\leq397$ & $2\times10^{-6}$ & 1.5\\
\multicolumn{5}{l}{Stage 3: SF-DMD}\\
\quad \textit{Short sequence} & 250 / 1,250 & 93 & $10^{-6}$ / $2\times10^{-7}$ & 1.5\\
\quad \textit{Long sequence} & 50 / 250 & 397 & $10^{-6}$ / $2\times10^{-7}$ & 1.5\\
\bottomrule
\end{tabular}
\end{table}
\ifhmode\par\fi\endgroup

\paragraph[\kern0pt Trainable parameters.]{{Trainable parameters.}}

\begingroup{}
During this adaptation, we train the shared generator blocks, diffusion-time embedding, RGB and LiDAR input/output projections, and learned sensor embeddings.
The pre-trained reasoner and both sensor tokenizers remain frozen.
The causal training and distillation stages retain this parameter selection for the student, while the distillation teacher remains fixed.
SF-DMD additionally trains the fake-score network, following the shared method in \cref{sec::self_forcing}.
\ifhmode\par\fi\endgroup

\paragraph[\kern0pt Objectives and modality balance.]{{Objectives and modality balance.}}

\begingroup{}
Stage~0 trains RGB and LiDAR generation with the rectified flow objective inherited from Cosmos 3~\citep{cosmos3}.
The two modalities have different latent dimensions and token counts, so we compute their losses separately before combining them with RGB:LiDAR weights of $1:2$ during joint training.
For RF and TF, following Cosmos 3, we average the squared velocity error within each stream, then across streams of the same modality.
Observed prefixes and structural controls remain clean, with their prediction errors masked to zero before this reduction.
The reduction retains these entries in its averaging denominators, including the zero-loss control streams.
The causal training and distillation stages retain this $1:2$ weighting with the shared objectives in \cref{sec::training}; their loss reductions are detailed in \cref{sec::auto_training_objectives}.
\ifhmode\par\fi\endgroup

\paragraph[\kern0pt Camera-rig sampling.]{{Camera-rig sampling.}}

\begingroup{}
Different applications require different camera coverage, from a single forward view to a complete surround rig.
We expose Cosmos3-Nano-Transfer-Auto to these configurations by sampling predefined camera combinations rather than always presenting all 11 views.
The catalog contains 15 commonly used camera combinations spanning 1--11 views, listed in \cref{tab::auto_camera_rigs} in the appendix.
For joint training, we sample 480p and 720p clips with probabilities 0.6 and 0.4, respectively.
At 480p, the standard 7-camera rig receives probability 0.6, the full 11-camera rig receives 0.3, and the remaining 13 combinations share 0.1 uniformly.
At 720p, we sample the full 7-camera rig with probability 0.9.
To fit the training memory budget, we use 480p whenever all 11 cameras are selected together.
At 720p, each example contains fewer cameras, but different sampled combinations expose the model to all 11 camera views.
Learned view embeddings identify individual cameras: the front-wide camera, for example, always uses its front-wide embedding regardless of which other cameras are selected.
The camera prefix describes the sampled rig, as explained in \cref{sec::auto_data}.
\ifhmode\par\fi\endgroup

\paragraph[\kern0pt Conditioning robustness.]{{Conditioning robustness.}}

\begingroup{}
To support RGB-only generation with the joint model, we drop LiDAR together with its HD-map control on 10\% of joint-training examples.
We also vary the available language conditioning to accommodate incomplete prompts and support classifier-free guidance.
Clip-level caption dropout removes all camera captions together with probability 0.2.
Independently, per-view caption dropout removes each camera's scene description with probability 0.2.
We train on Transfer-T2V and Transfer-I2V modes equally.
Transfer-I2V supplies 1 initial RGB frame per camera during training, while Transfer-T2V supplies none.
\ifhmode\par\fi\endgroup

\subsubsection[\kern0pt Cosmos3-Nano-Transfer-Auto Training]{{Cosmos3-Nano-Transfer-Auto Training}}
\label{sec::auto_bidirectional_training}

\begingroup{}
In practice, we found that directly adapting the single-view Cosmos3-Nano to joint multi-view RGB--LiDAR generation on 400-frame source chunks led to slow convergence, particularly in LiDAR prediction quality.
These experiments motivate the recommended three-phase curriculum below: first learning the modalities separately, then training them jointly over short sequences, and finally increasing the joint sequence length.
\Cref{tab::auto_training_summary} gives the recommended training budgets and learning rates.
\ifhmode\par\fi\endgroup

\paragraph[\kern0pt Modality-separate training.]{{Modality-separate training.}}
\begingroup{}
We first train RGB and LiDAR generation as separate tasks within the shared generator, establishing LiDAR generation while adapting RGB generation to the driving domain.
RGB examples contain a single camera view with up to 397 aligned frames, while LiDAR examples contain 21 sweeps.
\ifhmode\par\fi\endgroup

\paragraph[\kern0pt Short-sequence joint training.]{{Short-sequence joint training.}}
\begingroup{}
We next train on synchronized multi-view RGB and LiDAR together, using 61 RGB frames per camera and 21 LiDAR sweeps, approximately 2 seconds.
This phase learns cross-view and cross-modal relationships over short temporal contexts.
\ifhmode\par\fi\endgroup

\paragraph[\kern0pt Long-sequence joint training.]{{Long-sequence joint training.}}
\begingroup{}
Finally, we extend the joint training window to up to 397 RGB frames per camera and 133 LiDAR sweeps, approximately 13.2 seconds, to learn scene evolution over longer temporal contexts.
In practice, we found that longer temporal contexts during bidirectional training benefit the stability of the subsequently distilled model.
We choose 400-frame source windows to balance this benefit with training memory.
When splitting a recording leaves fewer than 400 frames at the end, we also use that shorter segment for training (\cref{sec::auto_data}).
Shorter clips, such as 300-frame videos, are also suitable, and clips of different lengths can be mixed during training.
\ifhmode\par\fi\endgroup

\subsubsection[\kern0pt Cosmos3-Nano-Sim-Auto Training]{{Cosmos3-Nano-Sim-Auto Training}}
\label{sec::auto_causal_training}

\begingroup{}
We follow the shared three-stage recipe in \cref{sec::training}, using the synchronized RGB--LiDAR chunks and causal attention described in \cref{sec::auto_causal_modeling}.
At each causal-stage transition, we initialize the student from the preceding stage's regular weights rather than its exponential moving average (EMA) weights, and reset the optimizer state and iteration counter.
\Cref{tab::auto_training_summary} summarizes the training budgets and learning rates.
\ifhmode\par\fi\endgroup

\begingroup{}
All three causal stages use AdamW~\citep{loshchilov2019decoupled} with weight decay 0.01, gradient norm clipping at 1.0, and no gradient accumulation.
The momentum coefficients $(\beta_1,\beta_2)$ are $(0.9,0.999)$ in Stage~1 and $(0,0.999)$ in Stages~2 and~3.
The learning rates in \cref{tab::auto_training_summary} remain constant within each stage.
\ifhmode\par\fi\endgroup

\paragraph[\kern0pt Stage 1: causal teacher forcing.]{{Stage 1: causal teacher forcing.}}
\begingroup{}
We initialize the causal model from the Stage~0 bidirectional model and train it to predict synchronized RGB and LiDAR chunks from recorded sensor history, following the teacher-forcing (TF) method in \cref{sec::teacher_forcing}.
\ifhmode\par\fi\endgroup

\paragraph[\kern0pt Stage 2: consistency distillation.]{{Stage 2: consistency distillation.}}
\begingroup{}
We next distill the causal model into a 4-step generator using TF-dCM~\citep{zheng2026causalrcm}, following \cref{sec::consistency_distillation}.
Both the student and frozen causal teacher initialize from the Stage~1 checkpoint.
Training retains recorded sensor history and uses teacher CFG with scale 1.5.
\ifhmode\par\fi\endgroup

\paragraph[\kern0pt Stage 3: self-forcing.]{{Stage 3: self-forcing.}}
\begingroup{}
Starting from the Stage~2 student, SF-DMD trains on RGB--LiDAR rollouts conditioned on the student's own generated history, following \cref{sec::self_forcing}.
The frozen DMD teacher and trainable fake-score network initialize from the Stage~0 bidirectional model, with teacher CFG again set to 1.5.
\ifhmode\par\fi\endgroup

\begingroup{}
SF-DMD alternates updates to the student and fake-score networks at a $1:5$ ratio.
We first train on 93-frame clips for 1,500 total updates, comprising 250 student updates and 1,250 fake-score updates.
We then continue on 397-frame clips for 300 updates (50 student and 250 fake-score updates) while preserving the optimizer state and iteration counter.
\ifhmode\par\fi\endgroup

\begingroup{}
In practice, too few SF-DMD updates left object artifacts, including floating objects, while too many introduced temporal flicker.
We selected the update budget to balance object integrity and temporal consistency, guided by quantitative metrics and visual evaluation.
\ifhmode\par\fi\endgroup

%% file: sections/5__examples/auto/optimization.tex
\begingroup{}
\subsection{Infrastructure and Inference Optimization}
\label{sec::auto_inference_optimization}
\ifhmode\par\fi\endgroup

\begingroup{}
\subsubsection{Training and Attention Implementation}
\label{sec::auto_training_attention_implementation}
\ifhmode\par\fi\endgroup

\paragraph[\kern0pt Training infrastructure.]{{Training infrastructure.}}

\begingroup{}
We pack synchronized RGB, LiDAR, structural controls, and language using a target budget of 1.5M tokens per sequence.
On GB200 GPUs, the long-sequence configuration combines context parallelism and fully sharded data parallelism (FSDP)~\citep{zhao2023pytorchfsdp}, with BF16 computation and activation recomputation to limit memory use.
\ifhmode\par\fi\endgroup

\paragraph[\kern0pt Attention implementation.]{{Attention implementation.}}

\begingroup{}
We initially implemented the decomposed visibility pattern in \cref{sec::auto_multiview_modeling,sec::auto_causal_modeling} with PyTorch FlexAttention and a FlashAttention-4 backend~\citep{flashattention42026}.
Sensor frame boundaries do not generally align with the kernel's query--key blocks, leaving partially visible blocks that require elementwise masking.
This overhead reduces attention efficiency despite the smaller number of permitted query--key pairs, as illustrated in \cref{fig::auto_flex_blocks} in the appendix.
\ifhmode\par\fi\endgroup

\begingroup{}
Our multi-view backend applies the maskless approach described in \cref{sec::runtime_optimization} to same-view, cross-sensor, control, and language interactions.
We gather permitted tokens into segments for variable-length attention~\citep{pytorch_varlen_attn}, then scatter and merge the outputs in the original query order (\cref{fig::auto_maskless_attention}).
Token grouping and joint normalization are detailed in \cref{sec::auto_attention_implementation}.
The maskless backend was $2.5$--$3\times$ faster than FlexAttention for the tested decomposed attention operations.
\ifhmode\par\fi\endgroup

\input{figures/auto/auto_maskless_attention.tex}

\begingroup{}
\subsubsection{Inference Optimization}
\label{sec::auto_inference_performance}
\ifhmode\par\fi\endgroup

\begingroup{}
\begin{table}[t]
\centering\reporttablestyle
\caption{\textbf{Selected Cosmos3-Nano-Sim-Auto runtime operating points on GB200.}
The released checkpoint generates 12-frame chunks at $832\times480$ resolution per camera with 4 denoising steps on the residential scene.
Chunk times are medians of run medians, with repeat ranges in parentheses, and FPS is the measured rate presented to the client.
Model and map ranks together give the total GPU allocation.
Map ranks prefetch controls with a 1-chunk delay.
Eager and optimized runs share an allocation, although the eager counterpart of the 8 model + 8 map row uses 16 model ranks.
A dash indicates no eager comparison.
}
\label{tab::auto_runtime_summary}
\begin{tabularx}{\linewidth}{cL{2.2cm}cYL{2.35cm}L{2.35cm}c}
\toprule
\textbf{Views} & \textbf{Mode} & \textbf{GPUs} & \textbf{Layout} & \multicolumn{2}{c}{\textbf{Chunk wall time (ms)}} & \textbf{FPS}\\
\cmidrule(lr){5-6}
 & & & & eager & optimized & \\
\midrule
1 & RGB & 1 & 1 model & 524 (512--535) & 364 (362--423) & 32.6\\
4 & RGB & 16 & \mbox{8 model + 8 map} & --- & 408 (377--440) & 28.9\\
7 & RGB & 16 & 16 model & 897 & 609 & 19.5\\
7 & RGB & 16 & \mbox{8 model + 8 map} & 950 (936--1{,}014) & 583 (561--597) & 20.5\\
7 & RGB & 32 & \mbox{16 model + 16 map} & --- & 511 (510--511) & 23.2\\
7 & \mbox{RGB + LiDAR} & 16 & 16 model & 1{,}159 & 701 (693--714) & 16.7\\
\bottomrule
\end{tabularx}
\end{table}
\ifhmode\par\fi\endgroup

\paragraph[\kern0pt Streaming execution.]{{Streaming execution.}}

\begingroup{}
We evaluate the released Cosmos3-Nano-Sim-Auto checkpoint in Interactive Drive, the FlashDreams demonstration client,\footnote{FlashDreams: \url{https://github.com/NVIDIA/flashdreams}.} using the residential scene with prepared WSM controls on GB200 GPUs.
The FlashDreams 0.2 release candidate generates $832\times480$ video per camera with 4 denoising steps and a bounded history cache.
Each chunk produces 12 frames per camera at a video rate of 30 FPS, giving a 400\,ms runtime target.
These measurements cover up to 7 cameras plus LiDAR, rather than the full supported 11-camera rig.
\ifhmode\par\fi\endgroup

\begingroup{}
Bounded history keeps steady-state shapes fixed for compiled denoising, control encoding, and decoding.
The runtime uses maskless attention with a fused group merge and distributes camera decoding across model ranks.
Separate map GPUs can prepare the next chunk's controls while the current chunk denoises, but introduce a 1-chunk control delay.
Control-token processing and the clean K/V cache update remain on the model's execution path.
Compilation also increases startup time, measured separately from~steady-state~throughput.
The execution settings and optimization mechanisms are detailed in \cref{sec::auto_runtime_implementation,tab::auto_opt_inventory}.
\ifhmode\par\fi\endgroup

\paragraph[\kern0pt Inference performance.]{{Inference performance.}}

\begingroup{}
\Cref{tab::auto_runtime_summary} summarizes selected operating points from paired comparisons within the same GPU allocation.
The shipped optimizations reduce median chunk wall time by 30\% for 1 camera on 1 GPU and by 32\% for 7 cameras on 16 model ranks.
Compiled denoising provides the largest single gain in these comparisons.
Maskless execution also matters at inference: replacing it and the fused merge with a block-sparse FlexAttention path increases chunk wall time by 195\% for 7 cameras with compiled denoising on 8 model ranks and 8 map GPUs.
This comparison measures the complete generation call, rather than the attention operations alone.
The measurement protocol and full runtime results are provided in \cref{sec::auto_runtime_measurements,tab::auto_runtime_ladders}.
\ifhmode\par\fi\endgroup

\begingroup{}
\begin{figure}[t]
\centering
\input{figures/auto_optimization/data/opt_stages_timeline.tex}
\caption{\textbf{Cosmos3-Nano-Sim-Auto per-chunk stage times of the shipped configurations on GB200.}
Each bar is the median run of its configuration, split into the median stage times over its steady-state chunks on the primary rank.
Control preparation covers WSM drawing and encoding, or the gather of prefetched latents, plus the control-token pass that writes control K/V.
The bar length uses the backend's generation-call timer, and the gray remainder is work outside the 4 timed stages.
Stage medians from a representative run need not sum to the across-run wall-time median.}
\label{fig::auto_opt_stages}
\end{figure}
\ifhmode\par\fi\endgroup

\begingroup{}
Separate map GPUs help only when enough model ranks remain for generation, as shown in the GPU-layout comparisons in \cref{fig::auto_runtime_ladders}.
For 4 cameras, 8 model ranks and 8 map GPUs take 408\,ms compared with 545\,ms for 16 model ranks in the same allocation.
For 7 cameras, the two layouts fall within the repeat range, while adding LiDAR favors retaining all 16 GPUs as model ranks.
The selected 7-camera configurations reach 23.2 FPS for RGB on 32 GPUs and 16.7 FPS for RGB plus LiDAR on 16 GPUs, both below the 30 FPS video rate.
The 1-camera median meets the 400\,ms target, while slower repeats and some 4-camera runs exceed it.
These throughput measurements use prefetched controls, while responding to newly supplied conditions also requires control preparation.
\ifhmode\par\fi\endgroup

\begingroup{}
\Cref{fig::auto_opt_stages} shows why faster denoising alone does not determine chunk latency.
Control preparation and the clean K/V cache update remain eager and together exceed denoising time for the single-camera configuration, while approaching it for 7 cameras.
The control-token pass remains necessary even when separate map GPUs prefetch the rendered and encoded controls.
These costs limit the benefit of further denoising optimization and motivate measuring the complete chunk rather than only its diffusion steps.
Additional stage and startup timings are provided in \cref{sec::auto_runtime_measurements}.
\ifhmode\par\fi\endgroup

%% file: figures/auto/auto_maskless_attention.tex
\begingroup{}
\begin{figure}[t]
\centering
\input{figures/diagrams/auto_maskless_attention.tex}
\caption{\textbf{Maskless execution of decomposed attention.}
Query and key/value tokens are gathered into segments for the illustrated sensor and language interactions.
Variable-length attention processes each segment, and the results are scattered back to the original query order.
Each attention call returns a normalized output and log-sum-exp (LSE) statistics.
LSE-weighted merging recovers joint normalization when key subsets are disjoint and attention scores match, as in \cref{eq::attention_merge}.
Grouping must preserve the model's temporal and conditioning restrictions.
Arrows show data dependencies between logical operations.}
\label{fig::auto_maskless_attention}
\end{figure}
\ifhmode\par\fi\endgroup

%% file: figures/diagrams/auto_maskless_attention.tex
\begingroup
\input{figures/diagrams/architecture_styles.tex}
\fitdiagram{%
\begin{tikzpicture}[x=1cm,y=1cm,
  font=\fontencoding{T1}\fontfamily{bch}\fontsize{9}{10.5}\selectfont,
  maskless heading/.style={font=\fontencoding{T1}\fontfamily{bch}\fontsize{9}{10.5}\selectfont\bfseries},
  maskless group/.style={draw,rounded corners=2pt,minimum width=4.8cm,
    minimum height=0.95cm,align=center,inner sep=4pt},
  maskless stage/.style={draw=simgrayedge,rounded corners=2pt,
    minimum width=3.25cm,minimum height=1.0cm,inner sep=4pt,align=center},
  maskless detail/.style={align=center,text width=3.5cm,anchor=north},
  maskless arrow/.style={->,draw=simgrayedge,line width=0.7pt}]
  \node[maskless group,fill=simcleanfill,draw=simcleanedge] (same) at (2.55,3.45)
    {\textbf{Within each view}\\Permitted same-stream keys};
  \node[maskless group,fill=simnoisyfill,draw=simnoisyedge] (cross) at (8.0,3.45)
    {\textbf{Across views}\\Other streams in the temporal window};
  \node[maskless group,fill=simtextfill,draw=simtextedge] (language) at (13.45,3.45)
    {\textbf{Language conditioning}\\Permitted caption keys};

  \draw[draw=simgrayedge,line width=0.6pt]
    (same.south) -- (2.55,2.55) -- (1.75,2.55) -- (13.45,2.55) -- (language.south);
  \draw[draw=simgrayedge,line width=0.6pt] (cross.south) -- (8.0,2.55);
  \node[anchor=south] at (8.0,1.93) {Logical interaction groups};

  \node[maskless stage,fill=simgrayfill,maskless heading] (gather) at (1.75,1.20) {1\quad Gather};
  \node[maskless stage,fill=simcleanfill,maskless heading] (attend) at (5.90,1.20) {2\quad Attend};
  \node[maskless stage,fill=simgrayfill,maskless heading] (scatter) at (10.05,1.20) {3\quad Scatter};
  \node[maskless stage,fill=simactionfill,maskless heading] (merge) at (14.20,1.20) {4\quad Merge};
  \draw[maskless arrow] (1.75,2.55) -- (gather.north);
  \draw[maskless arrow] (gather.east) -- (attend.west);
  \draw[maskless arrow] (attend.east) -- (scatter.west);
  \draw[maskless arrow] (scatter.east) -- (merge.west);

  \node[maskless detail] at (1.75,0.47) {Group tokens\\into segments};
  \node[maskless detail] at (5.90,0.47) {Run variable-length\\attention per segment};
  \node[maskless detail] at (10.05,0.47) {Restore query order\\for outputs and LSE};
  \node[maskless detail] at (14.20,0.47) {Combine group outputs\\using their LSE};
\end{tikzpicture}}
\endgroup

%% file: figures/auto_optimization/data/opt_stages_timeline.tex
\begingroup
\input{figures/diagrams/architecture_styles.tex}
\fitdiagram{%
\begin{tikzpicture}[x=0.0096cm,y=1cm,font=\fontfamily{bch}\fontsize{9}{10.5}\selectfont,
  stage seg/.style={draw=simgrayedge,line width=0.3pt},
  stage value/.style={font=\fontfamily{bch}\fontsize{8}{9.5}\selectfont},
  stage axis/.style={draw=simgrayedge,line width=0.5pt},
  stage total/.style={anchor=west,fill=white,inner sep=1.5pt}]
  \draw[densely dotted,draw=simgrayedge,line width=0.6pt] (400,-2.84) -- (400,0.37);
  \node[anchor=south] at (400,0.37) {400\,ms target};
  \node[anchor=east,align=right] at (-12,0.00) {1 camera, 1 GPU};
  \fill[simtextfill] (0.0,-0.25) rectangle (141.4,0.25);
  \draw[stage seg] (0.0,-0.25) rectangle (141.4,0.25);
  \node[stage value] at (70.7,0.00) {141};
  \fill[simnoisyfill] (141.4,-0.25) rectangle (267.9,0.25);
  \draw[stage seg] (141.4,-0.25) rectangle (267.9,0.25);
  \node[stage value] at (204.7,0.00) {126};
  \fill[simcleanfill] (267.9,-0.25) rectangle (346.9,0.25);
  \draw[stage seg] (267.9,-0.25) rectangle (346.9,0.25);
  \node[stage value] at (307.4,0.00) {79};
  \fill[simactionfill] (346.9,-0.25) rectangle (355.3,0.25);
  \draw[stage seg] (346.9,-0.25) rectangle (355.3,0.25);
  \fill[simgrayfill] (355.3,-0.25) rectangle (362.6,0.25);
  \draw[stage seg] (355.3,-0.25) rectangle (362.6,0.25);
  \node[stage total] at (368.6,0.00) {363\,ms};
  \node[anchor=east,align=right] at (-12,-0.78) {4 cameras, 1 GPU};
  \fill[simtextfill] (0.0,-1.03) rectangle (361.4,-0.53);
  \draw[stage seg] (0.0,-1.03) rectangle (361.4,-0.53);
  \node[stage value] at (180.7,-0.78) {361};
  \fill[simnoisyfill] (361.4,-1.03) rectangle (928.4,-0.53);
  \draw[stage seg] (361.4,-1.03) rectangle (928.4,-0.53);
  \node[stage value] at (644.9,-0.78) {567};
  \fill[simcleanfill] (928.4,-1.03) rectangle (1129.5,-0.53);
  \draw[stage seg] (928.4,-1.03) rectangle (1129.5,-0.53);
  \node[stage value] at (1029.0,-0.78) {201};
  \fill[simactionfill] (1129.5,-1.03) rectangle (1155.1,-0.53);
  \draw[stage seg] (1129.5,-1.03) rectangle (1155.1,-0.53);
  \fill[simgrayfill] (1155.1,-1.03) rectangle (1162.1,-0.53);
  \draw[stage seg] (1155.1,-1.03) rectangle (1162.1,-0.53);
  \node[stage total] at (1168.1,-0.78) {1{,}162\,ms};
  \node[anchor=east,align=right] at (-12,-1.56) {4 cameras, 16 GPUs\\8 model + 8 map};
  \fill[simtextfill] (0.0,-1.81) rectangle (116.6,-1.31);
  \draw[stage seg] (0.0,-1.81) rectangle (116.6,-1.31);
  \node[stage value] at (58.3,-1.56) {117};
  \fill[simnoisyfill] (116.6,-1.81) rectangle (289.7,-1.31);
  \draw[stage seg] (116.6,-1.81) rectangle (289.7,-1.31);
  \node[stage value] at (203.2,-1.56) {173};
  \fill[simcleanfill] (289.7,-1.81) rectangle (419.4,-1.31);
  \draw[stage seg] (289.7,-1.81) rectangle (419.4,-1.31);
  \node[stage value] at (354.6,-1.56) {130};
  \fill[simactionfill] (419.4,-1.81) rectangle (428.1,-1.31);
  \draw[stage seg] (419.4,-1.81) rectangle (428.1,-1.31);
  \fill[simgrayfill] (428.1,-1.81) rectangle (435.0,-1.31);
  \draw[stage seg] (428.1,-1.81) rectangle (435.0,-1.31);
  \node[stage total] at (441.0,-1.56) {435\,ms};
  \node[anchor=east,align=right] at (-12,-2.34) {7 cameras, 16 GPUs\\8 model + 8 map};
  \fill[simtextfill] (0.0,-2.59) rectangle (134.1,-2.09);
  \draw[stage seg] (0.0,-2.59) rectangle (134.1,-2.09);
  \node[stage value] at (67.1,-2.34) {134};
  \fill[simnoisyfill] (134.1,-2.59) rectangle (417.7,-2.09);
  \draw[stage seg] (134.1,-2.59) rectangle (417.7,-2.09);
  \node[stage value] at (275.9,-2.34) {284};
  \fill[simcleanfill] (417.7,-2.59) rectangle (572.1,-2.09);
  \draw[stage seg] (417.7,-2.59) rectangle (572.1,-2.09);
  \node[stage value] at (494.9,-2.34) {154};
  \fill[simactionfill] (572.1,-2.59) rectangle (581.7,-2.09);
  \draw[stage seg] (572.1,-2.59) rectangle (581.7,-2.09);
  \fill[simgrayfill] (581.7,-2.59) rectangle (588.4,-2.09);
  \draw[stage seg] (581.7,-2.59) rectangle (588.4,-2.09);
  \node[stage total] at (594.4,-2.34) {588\,ms};
  \draw[stage axis] (0,-2.84) -- (1200,-2.84);
  \draw[stage axis] (0,-2.84) -- (0,-2.92);
  \node[anchor=north] at (0,-2.94) {0};
  \draw[stage axis] (200,-2.84) -- (200,-2.92);
  \node[anchor=north] at (200,-2.94) {200};
  \draw[stage axis] (400,-2.84) -- (400,-2.92);
  \node[anchor=north] at (400,-2.94) {400};
  \draw[stage axis] (600,-2.84) -- (600,-2.92);
  \node[anchor=north] at (600,-2.94) {600};
  \draw[stage axis] (800,-2.84) -- (800,-2.92);
  \node[anchor=north] at (800,-2.94) {800};
  \draw[stage axis] (1000,-2.84) -- (1000,-2.92);
  \node[anchor=north] at (1000,-2.94) {1{,}000};
  \draw[stage axis] (1200,-2.84) -- (1200,-2.92);
  \node[anchor=north] at (1200,-2.94) {1{,}200};
  \node[anchor=north] at (600,-3.36) {Time within the chunk (ms)};
  \begin{scope}[x=1cm]
  \node[stage seg,fill=simtextfill,minimum size=0.25cm,inner sep=0pt] at (0.00,-4.04) {};
  \node[anchor=west] at (0.17,-4.04) {Control preparation};
  \node[stage seg,fill=simnoisyfill,minimum size=0.25cm,inner sep=0pt] at (3.40,-4.04) {};
  \node[anchor=west] at (3.57,-4.04) {Denoise};
  \node[stage seg,fill=simcleanfill,minimum size=0.25cm,inner sep=0pt] at (5.20,-4.04) {};
  \node[anchor=west] at (5.37,-4.04) {K/V cache update};
  \node[stage seg,fill=simactionfill,minimum size=0.25cm,inner sep=0pt] at (8.30,-4.04) {};
  \node[anchor=west] at (8.47,-4.04) {Decode};
  \node[stage seg,fill=simgrayfill,minimum size=0.25cm,inner sep=0pt] at (10.00,-4.04) {};
  \node[anchor=west] at (10.17,-4.04) {Other};
  \end{scope}
\end{tikzpicture}}
\endgroup

%% file: sections/5__examples/auto/results.tex
\begingroup{}
\subsection{Results}
\label{sec::auto_results}
\ifhmode\par\fi\endgroup

\input{figures/auto/auto_rgb_results_table.tex}

\input{figures/auto/auto_transfer_golden_hour.tex}

\begingroup{}
\paragraph{Evaluation setup.}
We evaluate on two held-out datasets.
Driving-Sim-Eval (\cref{sec::auto_data}) contains 91 samples balanced across geography, control signals, and scene conditions.
Traffic-Scenario-Eval is an internal collection of 319 samples for evaluating synthetic data generation.

Driving-Sim-Eval uses HD-map-derived WSM controls to assess generation quality.
Traffic-Scenario-Eval benchmarks Transfer-T2V generation from noisy, perception-derived WSM controls and prompts specifying road conditions, time of day, geographic location, and weather.
We evaluate how well generated videos follow the controls and depict the requested attributes.
\ifhmode\par\fi\endgroup

\begingroup{}
We evaluate joint RGB--LiDAR generation with Cosmos3-Nano-Sim-Auto and Cosmos3-Nano-Transfer-Auto, using the multi-view variant of Cosmos-Transfer2.5 as the RGB baseline~\citep{cosmos_predict2p5}.
For each benchmark, every model, including the baseline, uses the prompt configuration that yields the highest generation quality among the variants we tested.
Detailed inference settings and prompts are provided in \cref{sec::auto_inference_settings}.
\ifhmode\par\fi\endgroup

\input{figures/auto/auto_sim_sun_glare.tex}

\subsubsection[\kern0pt Evaluation Metrics]{{Evaluation Metrics}}
\label[section]{sec::auto_evaluation_metrics}

\begingroup{}
To assess both sensor realism and driving-scene fidelity, we combine general RGB and LiDAR quality metrics with traffic-object adherence metrics and human evaluation of traffic and environmental attributes.
\ifhmode\par\fi\endgroup

\paragraph[\kern0pt RGB video quality metrics.]{{RGB video quality metrics.}}
\begingroup{}
RGB evaluation spans visual fidelity, temporal coherence, and cross-view geometric consistency.
Frame Fr\'echet inception distance (FID)~\citep{heusel2017fid} compares generated and reference frame-feature distributions.
Fr\'echet video distance (FVD)~\citep{unterthiner2019fvd} compares video-feature distributions, with I3D~\citep{carreira2017i3d} and StyleGAN-V~\citep{stylegan_v} results reported separately.
The DOVER technical score~\citep{dover} assesses perceptual video quality.
Adjacent cosine mean (ACM) measures within-view visual consistency through CLIP-feature similarity to the preceding and first frames.
Temporal Sampson error (TSE) measures geometric agreement between matched points in consecutive frames.
Cross-view Sampson error (CSE)~\citep{sampson1987} measures agreement across overlapping camera views, averaged over 2 or 5 view pairs for CSE-2 and CSE-5.
Lower FID, FVD, TSE, and CSE are better, while higher DOVER and ACM are better.
Reported scores are averages over the 7 non-fisheye views.
\ifhmode\par\fi\endgroup

\paragraph[\kern0pt LiDAR metrics.]{{LiDAR metrics.}}
\begingroup{}
For LiDAR, comparisons with real sweeps assess geometry and temporal consistency, while point support and detection agreement with ground-truth annotations measure object faithfulness.
Bird's-eye-view (BEV) JSD and MMD$^2$~\citep{gretton2012kernel} compare generated and real spatial distributions.
Fr\'echet Utonia distance (FUD) and Utonia MMD$^2$ compare pre-trained point-cloud features~\citep{zhang2026utonia}.
Lower values indicate closer distributions.
Chamfer temporal consistency (CTC) measures sweep-to-sweep geometric change after motion alignment, for full sweeps or tracked vehicles.
Generated-to-real ratios closer to 1 are better.
Object support recall measures the fraction of objects with sufficient real LiDAR points that also contain sufficient generated points within their ground-truth boxes.
Detection F1 compares PointPillars~\citep{lang2019pointpillars} detections in generated point clouds with ground-truth object annotations.
Higher recall and F1 indicate better object faithfulness.
\ifhmode\par\fi\endgroup

\paragraph[\kern0pt Traffic-object adherence.]{{Traffic-object adherence.}}
\begingroup{}
AutoGrader uses object-level F1 to measure how faithfully generated vehicles, pedestrians, and cyclists follow WSM controls.
Counts of detected, missing, and hallucinated objects come from the correspondence checks of Cosmos Evaluator~\citep{nvidia2026cosmosevaluator}.
Higher scores indicate better adherence.
\ifhmode\par\fi\endgroup

\paragraph[\kern0pt Human evaluation.]{{Human evaluation.}}
\begingroup{}
Human evaluation (HUE) scores video quality, scene-element fidelity, and cross-view agreement on a 1--3 scale.
For overall video quality, 3 denotes realistic, clean, temporally stable imagery; 2 allows minor artifacts but no physically impossible content; and 1 denotes severe degradation or any physically impossible event, even briefly.
Scene-element and cross-view questions assess the specified objects and their consistency, with rubric details in \cref{sec::auto_hue_rubric}.
Road-surface and weather scores are the percentages of evaluated samples whose generated conditions human raters judge to match the prompt.
Higher scores are better.
Following the annotation protocol of Cosmos 3~\citep{cosmos3}, 2 annotators independently rate each (video, question) pair.
Disagreements are resolved by a third quality-control reviewer, whose decision is final.
\ifhmode\par\fi\endgroup

\subsubsection[\kern0pt Results]{{Results}}
\label[section]{sec::auto_quantitative_results}

\paragraph[\kern0pt Generation quality on Driving-Sim-Eval.]{{Generation quality on Driving-Sim-Eval.}}
\begingroup{}
We first assess the quality of the synchronized RGB and LiDAR outputs on Driving-Sim-Eval.
\Cref{fig::auto_transfer_golden_hour,fig::auto_sim_sun_glare} show Transfer-I2V examples from Cosmos3-Nano-Transfer-Auto and Cosmos3-Nano-Sim-Auto, respectively.\footnote{See the project website for animated visualizations: \href{https://research.nvidia.com/labs/cosmos-lab/cosmos-simulation}{\nolinkurl{research.nvidia.com/labs/cosmos-lab/cosmos-simulation}}.}
Each example contains 11 camera views and the corresponding LiDAR sweep, with camera-aligned WSM controls overlaid at 20\% opacity.
\ifhmode\par\fi\endgroup

\begingroup{}
\begin{figure}[!t]
\centering
\includegraphics[width=\linewidth]{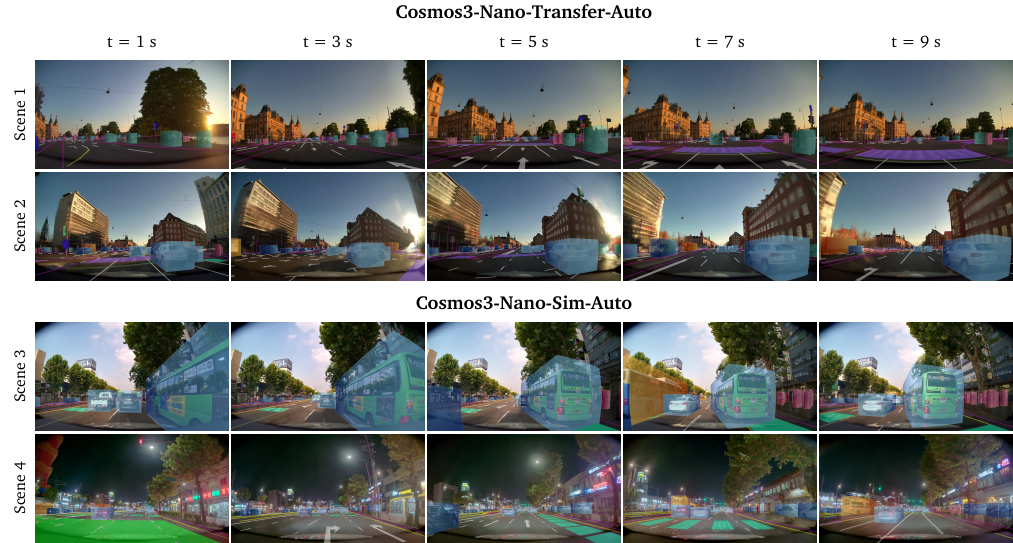}
\caption{\textbf{Spatial control following in generated front-wide video.}
WSM overlays show correspondence with vehicles, road markings, and traffic signals at five times after conditioning.
The top two rows use Cosmos3-Nano-Transfer-Auto and the bottom two use Cosmos3-Nano-Sim-Auto.}
\label{fig::auto_wsm_control_following}
\end{figure}
\ifhmode\par\fi\endgroup

\begingroup{}
\textit{RGB results.}
Cosmos3-Nano-Transfer-Auto improves on the multi-view Cosmos-Transfer2.5 baseline across every reported RGB metric in both Transfer-T2V and Transfer-I2V modes (\cref{tab::auto_benchmark_quality}).
Frame FID improves by 26--33\% and both FVD variants by 37--39\%, indicating that the generated frame and video distributions more closely match the reference data.
Higher DOVER scores further indicate improved technical video quality.
The gains also extend to consistency: lower TSE and higher ACM indicate more coherent videos over time, while roughly 20\% lower CSE-2 and CSE-5 reflect stronger geometric agreement across camera views.

\Needspace{3\baselineskip}
The four-step causal Cosmos3-Nano-Sim-Auto improves frame FID, FVD, and temporal consistency over the Cosmos-Transfer2.5 baseline in both generation modes.
In Transfer-T2V, it also slightly surpasses its bidirectional teacher, Cosmos3-Nano-Transfer-Auto, in I3D FVD (480 versus 488), while matching its ACM score of 0.906.
Cross-view consistency improves over the baseline in Transfer-I2V, whereas Transfer-T2V Sampson errors are modestly higher, by 4--8\%.
DOVER scores also favor the baseline.
Overall, the causal model combines four-step streaming generation with substantial gains in image- and video-level distributional quality and temporal consistency, alongside metric- and mode-specific trade-offs.

Qualitatively, the WSM overlays in \cref{fig::auto_wsm_control_following} show close correspondence between generated vehicles and their control cuboids, together with alignment of road markings and traffic signals across the~sampled~frames.
\ifhmode\par\fi\endgroup

\noindent{\textit{LiDAR results.}}
\begingroup{}
We evaluate the jointly generated LiDAR from the same Transfer-T2V rollouts as the RGB results, with 99 sweeps at 10\,Hz (\cref{tab::auto_student_lidar}).
LidarVerse is evaluated on 21- and 116-sweep rollouts, providing comparisons at both shorter and longer horizons than ours.
\ifhmode\par\fi\endgroup

\begingroup{}
\begin{table}[!t]
\centering\reporttablestyle[4pt]
\caption{\textbf{LiDAR generation results in Transfer-T2V mode.}
Cosmos 3 results use 99 sweeps at 10 Hz from the joint RGB--LiDAR rollouts.
LidarVerse~\citep{che2026lidarverse} is evaluated on 21- and 116-sweep rollouts.
Arrows indicate the preferred score direction, while CTC ratios target 1.
Scale factors in the headers multiply the displayed values; FUD denotes Fr\'echet Utonia distance.
Bold and underlined values mark the best and second-best scores.
Lighter and darker green identify the bidirectional and causal models, respectively.}
\label{tab::auto_student_lidar}
\par\medskip
\textbf{(a) Distribution fidelity}\par\smallskip
\begin{tabularx}{\linewidth}{l*{3}{C}cC}
\toprule
{\fontsize{8}{10}\selectfont\makecell[l]{\textbf{Model}}} &
{\fontsize{8}{10}\selectfont\makecell[c]{\textbf{BEV JSD $\downarrow$}\\global ($\times 10^{-2}$)}} &
{\fontsize{8}{10}\selectfont\makecell[c]{\textbf{BEV JSD $\downarrow$}\\local}} &
{\fontsize{8}{10}\selectfont\makecell[c]{\textbf{BEV MMD$^2$ $\downarrow$}\\local ($\times 10^{-3}$)}} &
{\fontsize{8}{10}\selectfont\makecell[c]{\textbf{FUD $\downarrow$}}} &
{\fontsize{8}{10}\selectfont\makecell[c]{\textbf{Utonia MMD$^2$ $\downarrow$}\\($\times 10^{-2}$)}}\\
\midrule
\rowcolor{rowours!50!white}
Cosmos3-Nano-Transfer-Auto & \textbf{4.01} & \textbf{0.220} & \textbf{3.47} & \textbf{12.8} & \textbf{2.08}\\
\rowcolor{rowours}
Cosmos3-Nano-Sim-Auto & \underline{5.52} & \underline{0.277} & \underline{5.93} & \underline{20.1} & \underline{3.54}\\
LidarVerse, 21 sweeps & 15.1 & 0.400 & 12.0 & 33.6 & 6.36\\
LidarVerse, 116 sweeps & 11.6 & 0.300 & 6.44 & 44.8 & 9.07\\
\bottomrule
\end{tabularx}
\par\medskip
\textbf{(b) Temporal consistency and object faithfulness}\par\smallskip
\begin{tabularx}{\linewidth}{l*{4}{C}}
\toprule
& \multicolumn{2}{c}{Temporal consistency} & \multicolumn{2}{c}{Object faithfulness}\\
\cmidrule(lr){2-3}\cmidrule(lr){4-5}
\textbf{Model} & \textbf{\shortstack{CTC / real\\(target 1)}} & \textbf{\shortstack{Vehicle CTC / real\\(target 1)}} & \textbf{\shortstack{Object support\\recall $\uparrow$}} & \textbf{\shortstack{Detection F1\\at 3\,m $\uparrow$}}\\
\midrule
\rowcolor{rowours!50!white}
Cosmos3-Nano-Transfer-Auto & \textbf{1.17} & \underline{2.53} & \textbf{0.885} & \textbf{0.237}\\
\rowcolor{rowours}
Cosmos3-Nano-Sim-Auto & 1.28 & 3.59 & 0.816 & 0.188\\
LidarVerse, 21 sweeps & \underline{1.20} & \textbf{2.51} & \underline{0.843} & \underline{0.201}\\
LidarVerse, 116 sweeps & 1.35 & 3.73 & 0.788 & 0.200\\
\bottomrule
\end{tabularx}
\end{table}
\ifhmode\par\fi\endgroup

\begingroup{}
\begin{figure}[!t]
\centering
\includegraphics[width=0.828\linewidth]{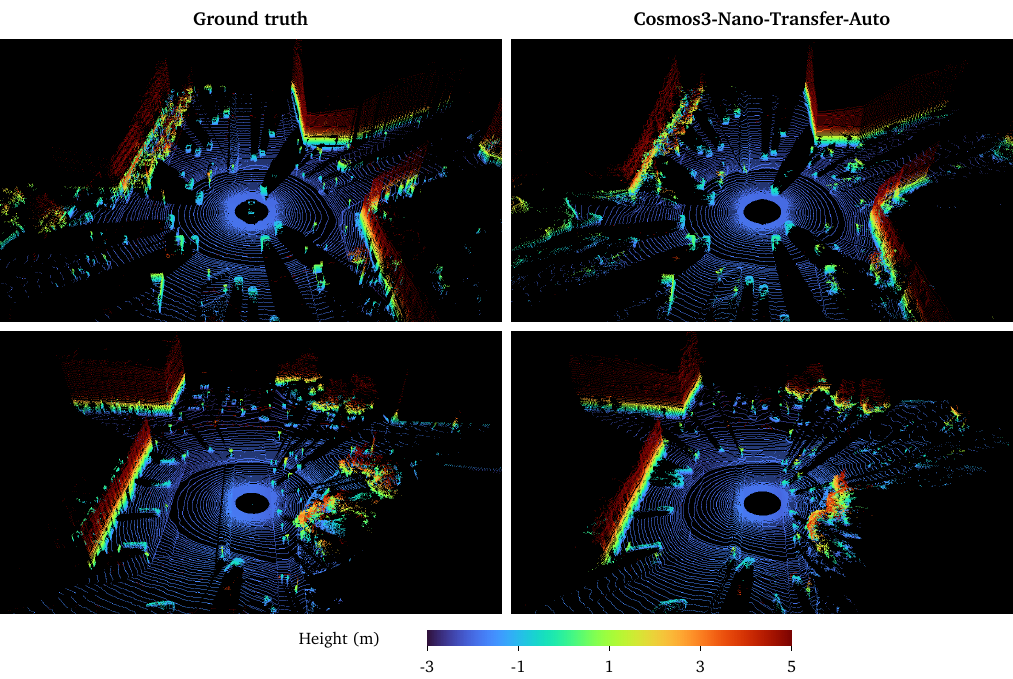}
\caption{\textbf{LiDAR geometry with Cosmos3-Nano-Transfer-Auto.}
Two Transfer-I2V scenes compare ground truth (left) with generated sweeps (right), using matched viewpoints and a shared height scale.
The examples reproduce the main road and roadside structure, with clear individual objects.}
\label{fig::auto_lidar_qualitative}
\par\medskip
\begin{minipage}{\linewidth}
\captionsetup{type=table}
\centering\reporttablestyle
\caption{\textbf{AutoGrader object following on Traffic-Scenario-Eval.}
F1 measures correspondence to WSM-controlled objects across 319 Transfer-T2V samples.
Inference settings are given in \cref{sec::auto_inference_settings}.
Higher F1 is better on the 0--1 scale.
Bold and underlined values mark the best and second-best scores.
Lighter and darker green identify the bidirectional and causal models, respectively.}
\label{tab::auto_autograder}
\begin{tabular}{lccc}
\toprule
\textbf{Model} & \textbf{Vehicle F1 $\uparrow$} & \textbf{Pedestrian F1 $\uparrow$} & \textbf{Cyclist F1 $\uparrow$}\\
\midrule
\rowcolor{rowours!50!white}
Cosmos3-Nano-Transfer-Auto & \textbf{0.908} & \underline{0.386} & \textbf{0.537}\\
\rowcolor{rowours}
Cosmos3-Nano-Sim-Auto & \underline{0.853} & \textbf{0.389} & \underline{0.428}\\
Cosmos-Transfer2.5 & 0.741 & 0.137 & 0.151\\
\bottomrule
\end{tabular}
\end{minipage}
\end{figure}
\ifhmode\par\fi\endgroup

\begingroup{}
\begin{table}[!t]
\centering\reporttablestyle[4pt]
\renewcommand{\arraystretch}{1.0}
\renewcommand{\tabularxcolumn}[1]{m{#1}}
\setlength{\aboverulesep}{1.5pt}
\setlength{\belowrulesep}{1.5pt}
\caption{\textbf{Human evaluation on Traffic-Scenario-Eval.}
Settings match \cref{tab::auto_autograder}.
Cosmos-Transfer2.5 denotes its multi-view variant.
Video quality, scene-element fidelity, and cross-view agreement use a 1--3 scale.
Environment scores give the percentage of samples matching the prompted conditions.
All scores are higher-is-better.
Video quality uses 319 samples per model; other categories use subsets, including 316 samples for Cosmos3-Nano-Sim-Auto's road-surface and weather scores.
Bold and underlined values mark the best and second-best scores.
Light and dark green distinguish the bidirectional and causal models.}
\label{tab::auto_hue}
\par\smallskip
\textbf{(a) Video quality and scene-element fidelity (1--3)}\par\smallskip
\begin{tabularx}{\linewidth}{l*{6}{C}}
\toprule
& \multicolumn{1}{c}{Video quality} & \multicolumn{5}{c}{Scene-element fidelity}\\
\cmidrule(lr){2-2}\cmidrule(lr){3-7}
{\fontsize{8}{10}\selectfont\textbf{\makecell[l]{Model}}} & {\fontsize{8}{10}\selectfont\textbf{\makecell[c]{Video\\quality $\uparrow$}}} & {\fontsize{8}{10}\selectfont\textbf{\makecell[c]{Lane $\uparrow$}}} & {\fontsize{8}{10}\selectfont\textbf{\makecell[c]{Traffic\\light $\uparrow$}}} & {\fontsize{8}{10}\selectfont\textbf{\makecell[c]{Traffic\\sign $\uparrow$}}} & {\fontsize{8}{10}\selectfont\textbf{\makecell[c]{Pedestrian $\uparrow$}}} & {\fontsize{8}{10}\selectfont\textbf{\makecell[c]{Bicyclist $\uparrow$}}}\\
\midrule
\rowcolor{rowours!50!white}
Cosmos3-Nano-Transfer-Auto & \textbf{1.72} & \underline{2.43} & \underline{1.83} & \underline{2.17} & \underline{1.64} & \textbf{1.85}\\
\rowcolor{rowours}
Cosmos3-Nano-Sim-Auto & 1.39 & \textbf{2.54} & \textbf{2.24} & \textbf{2.45} & \textbf{1.70} & \underline{1.83}\\
Cosmos-Transfer2.5 & \underline{1.65} & 2.28 & 1.59 & 1.78 & 1.20 & 1.38\\
\bottomrule
\end{tabularx}
\par\smallskip
\textbf{(b) Cross-view agreement and environmental fidelity}\par\smallskip
\begin{tabularx}{\linewidth}{l*{7}{C}}
\toprule
& \multicolumn{5}{c}{Cross-view agreement (1--3)} & \multicolumn{2}{c}{Environment scores}\\
\cmidrule(lr){2-6}\cmidrule(lr){7-8}
{\fontsize{8}{10}\selectfont\textbf{\makecell[l]{Model}}} & {\fontsize{8}{10}\selectfont\textbf{\makecell[c]{Lane/road\\markings $\uparrow$}}} & {\fontsize{8}{10}\selectfont\textbf{\makecell[c]{Traffic\\light $\uparrow$}}} & {\fontsize{8}{10}\selectfont\textbf{\makecell[c]{Stop\\sign $\uparrow$}}} & {\fontsize{8}{10}\selectfont\textbf{\makecell[c]{Pedestrian $\uparrow$}}} & {\fontsize{8}{10}\selectfont\textbf{\makecell[c]{Bicyclist $\uparrow$}}} & {\fontsize{8}{10}\selectfont\textbf{\makecell[c]{Road\\surface $\uparrow$}}} & {\fontsize{8}{10}\selectfont\textbf{\makecell[c]{Weather $\uparrow$}}}\\
\midrule
\rowcolor{rowours!50!white}
Cosmos3-Nano-Transfer-Auto & \underline{2.93} & \textbf{2.93} & \textbf{2.98} & \textbf{2.94} & \textbf{2.98} & \underline{92.8\%} & \textbf{90.9\%}\\
\rowcolor{rowours}
Cosmos3-Nano-Sim-Auto & \textbf{2.98} & \underline{2.91} & \underline{2.94} & \underline{2.94} & \underline{2.96} & 87.0\% & 81.0\%\\
Cosmos-Transfer2.5 & 2.77 & 2.68 & 2.76 & 2.58 & 2.63 & \textbf{93.1\%} & \underline{89.7\%}\\
\bottomrule
\end{tabularx}
\end{table}
\ifhmode\par\fi\endgroup

\begingroup{}
Cosmos3-Nano-Transfer-Auto achieves better quality scores than both LidarVerse settings on eight of the nine reported metrics.
Global BEV JSD improves by 65--73\% and Fr\'echet Utonia distance by 62--71\%, indicating closer agreement with real LiDAR in both spatial point distributions and learned geometric features.
Lower local BEV JSD and MMD$^2$ show that the gains also hold at the clip level.
Temporal CTC is closer to 1, while higher object support recall and detection F1 indicate better preservation of annotated objects.
Vehicle CTC is the only exception: 2.53 over our 99-sweep rollouts, versus 2.51 for the 21-sweep baseline.
In the two Transfer-I2V scenes shown in \cref{fig::auto_lidar_qualitative}, the generated LiDAR closely resembles the ground truth in road geometry, roadside structures, and nearby object placement.

\Needspace{3\baselineskip}
Cosmos3-Nano-Sim-Auto also improves all five distribution-fidelity metrics over both LidarVerse settings, retaining strong spatial and geometric fidelity with 4-step streaming generation.
Its temporal CTC, vehicle CTC, and object support recall improve on the 116-sweep baseline but trail the shorter 21-sweep baseline.
Detection F1 is slightly lower than both baselines, at 0.188 versus 0.200--0.201.
The causal distilled variant retains the distributional advantages over LidarVerse but falls short of Cosmos3-Nano-Transfer-Auto across the LiDAR metrics.
Preserving the bidirectional model's LiDAR fidelity, temporal coherence, and object detail through multimodal causal distillation is a priority for future work.

Together, these results illustrate the strong capability of joint RGB--LiDAR generation with our shared generator.
Causal distillation enables faster, 4-step streaming generation, with trade-offs in technical video quality, cross-view consistency, and LiDAR fidelity.
Nevertheless, Cosmos3-Nano-Sim-Auto retains many of Cosmos3-Nano-Transfer-Auto's gains over the video and LiDAR baselines.
Beyond these general quality metrics, our models also improve traffic-object adherence over the video baseline, as we examine next.
\ifhmode\par\fi\endgroup

\paragraph[\kern0pt Traffic-scenario fidelity on Traffic-Scenario-Eval.]{{Traffic-scenario fidelity on Traffic-Scenario-Eval.}}
\begingroup{}
We next turn to Traffic-Scenario-Eval to assess how the generated videos follow traffic controls and depict the requested scene attributes.
\ifhmode\par\fi\endgroup

\input{figures/auto/auto_long_rollout.tex}

\noindent{\textit{Traffic-object adherence results.}}
\begingroup{}
Cosmos3-Nano-Sim-Auto combines 4-step streaming generation with stronger traffic-object adherence than Cosmos-Transfer2.5 across all three categories (\cref{tab::auto_autograder}).
The gains are particularly pronounced for pedestrians and cyclists.
Compared with Cosmos3-Nano-Transfer-Auto, pedestrian F1 is similar (0.389 versus 0.386), while vehicle and cyclist F1 are 6\% and 20\% lower, respectively.
Despite these gaps, Cosmos3-Nano-Sim-Auto combines substantial control-following gains over the video baseline with 4-step streaming generation.
\ifhmode\par\fi\endgroup

\noindent{\textit{Human evaluation results.}}
\begingroup{}
Human evaluation complements the object-level metrics by assessing visual quality, scene attributes, and agreement across views.
\ifhmode\par\fi\endgroup

\begingroup{}
The 4-step Cosmos3-Nano-Sim-Auto receives higher human ratings than Cosmos-Transfer2.5 in all scene-element and cross-view categories (\cref{tab::auto_hue}).
The largest scene-element gains are for traffic signs (2.45 versus 1.78) and traffic lights (2.24 versus 1.59).
It also scores higher than Cosmos3-Nano-Transfer-Auto in four of the five scene-element categories, while cross-view scores differ by at most 0.05 on the 1--3 scale.
\ifhmode\par\fi\endgroup

\begingroup{}
The 4-step causal model trades some video quality for streaming efficiency, consistent with the DOVER results.
Road-surface and weather agreement are 87.0\% and 81.0\%, respectively, also below both comparison models.
Nevertheless, the causal distilled variant retains stronger scene-element fidelity and cross-view agreement than the video baseline while enabling 4-step streaming generation.
\ifhmode\par\fi\endgroup

\paragraph[\kern0pt Long-rollout stability.]{{Long-rollout stability.}}
\begingroup{}
To assess long-rollout stability, we run Cosmos3-Nano-Sim-Auto continuously for up to 1 hour of generated video, repeatedly cycling the WSM control sequence and associated text prompts while maintaining a rolling K/V cache.
Under this setting, the evaluated metrics show no substantial degradation over the 1-hour horizon (\cref{fig::auto_long_rollout}).
Qualitatively, scene structure and control following remain consistent, although object appearance and physical plausibility can deteriorate over extended rollouts.
See \cref{sec::auto_development} for additional ablations of training stages, cross-view attention, and attention-sink and K/V cache sizes.
\ifhmode\par\fi\endgroup

\begingroup{}
\paragraph{Summary.}
Cosmos3-Nano-Transfer-Auto and Cosmos3-Nano-Sim-Auto extend Cosmos3-Nano into controllable sensor generation with 1--11 camera views, supporting both RGB-only and joint RGB--LiDAR outputs.
The bidirectional model produces high-quality synchronized clips for offline data generation, while its causal distilled counterpart combines strong generation quality with fast, few-step streaming generation.
Both models achieve substantial gains in frame and video distribution fidelity over the strong Cosmos-Transfer2.5 baseline, together with better temporal coherence, traffic-object adherence, and human-rated cross-view consistency.
They also improve all five LiDAR distribution-fidelity metrics over LidarVerse while jointly generating RGB and LiDAR.
These results support high-quality sensor generation for offline synthesis and streaming.
\ifhmode\par\fi\endgroup

\FloatBarrier
\endgroup

%% file: figures/auto/auto_rgb_results_table.tex
\begingroup{}
\begin{table}[t]
\centering\reporttablestyle[3pt]
\caption{\textbf{Multi-view RGB results on Driving-Sim-Eval.}
Models generate 297 frames per camera at 720p across 7 non-fisheye views, with (Transfer-I2V) or without (Transfer-T2V) initial RGB conditioning.
Cosmos-Transfer2.5 denotes its multi-view variant.
SG-V denotes StyleGAN-V features for FVD.
Arrows indicate the preferred score direction.
Bold and underlined values mark the best and second-best scores within each mode.
Lighter and darker green identify the bidirectional and causal models, respectively.}
\label{tab::auto_benchmark_quality}
\par\medskip
\textbf{(a) Transfer-T2V: no initial RGB frame}\par\smallskip
\begin{tabularx}{\linewidth}{l*{8}{C}}
\toprule
& \multicolumn{4}{c}{Video quality} & \multicolumn{2}{c}{Temporal consistency} & \multicolumn{2}{c}{Cross-view alignment}\\
\cmidrule(lr){2-5}\cmidrule(lr){6-7}\cmidrule(lr){8-9}
\textbf{Model} & \textbf{\shortstack{Frame\\FID $\downarrow$}} & \textbf{\shortstack{FVD\\I3D $\downarrow$}} & \textbf{\shortstack{FVD\\SG-V $\downarrow$}} & \textbf{\shortstack{DOVER\\tech. $\uparrow$}} & \textbf{TSE $\downarrow$} & \textbf{ACM $\uparrow$} & \textbf{CSE-2 $\downarrow$} & \textbf{CSE-5 $\downarrow$}\\
\midrule
\rowcolor{rowours!50!white}
Cosmos3-Nano-Transfer-Auto & \textbf{45.5} & \underline{488} & \underline{487} & \textbf{9.01} & \textbf{0.473} & \textbf{0.906} & \textbf{2.009} & \textbf{1.633}\\
\rowcolor{rowours}
Cosmos3-Nano-Sim-Auto & \underline{48.6} & \textbf{480} & \textbf{479} & 7.30 & \underline{0.503} & \textbf{0.906} & 2.599 & 2.244\\
Cosmos-Transfer2.5 & 61.9 & 772 & 770 & \underline{8.56} & 0.530 & \underline{0.885} & \underline{2.511} & \underline{2.073}\\
\bottomrule
\end{tabularx}
\par\medskip
\textbf{(b) Transfer-I2V: with initial RGB frame}\par\smallskip
\begin{tabularx}{\linewidth}{l*{8}{C}}
\toprule
& \multicolumn{4}{c}{Video quality} & \multicolumn{2}{c}{Temporal consistency} & \multicolumn{2}{c}{Cross-view alignment}\\
\cmidrule(lr){2-5}\cmidrule(lr){6-7}\cmidrule(lr){8-9}
\textbf{Model} & \textbf{\shortstack{Frame\\FID $\downarrow$}} & \textbf{\shortstack{FVD\\I3D $\downarrow$}} & \textbf{\shortstack{FVD\\SG-V $\downarrow$}} & \textbf{\shortstack{DOVER\\tech. $\uparrow$}} & \textbf{TSE $\downarrow$} & \textbf{ACM $\uparrow$} & \textbf{CSE-2 $\downarrow$} & \textbf{CSE-5 $\downarrow$}\\
\midrule
\rowcolor{rowours!50!white}
Cosmos3-Nano-Transfer-Auto & \textbf{27.8} & \textbf{238} & \textbf{237} & \textbf{8.27} & \textbf{0.472} & \textbf{0.902} & \textbf{2.246} & \textbf{1.798}\\
\rowcolor{rowours}
Cosmos3-Nano-Sim-Auto & \underline{33.6} & \underline{325} & \underline{324} & 7.51 & \underline{0.505} & \underline{0.900} & \underline{2.359} & \underline{2.085}\\
Cosmos-Transfer2.5 & 41.4 & 388 & 387 & \underline{8.25} & 0.519 & 0.881 & 2.935 & 2.203\\
\bottomrule
\end{tabularx}
\end{table}
\ifhmode\par\fi\endgroup

%% file: figures/auto/auto_transfer_golden_hour.tex
\begingroup{}
\begin{figure}[t]
\centering
\includegraphics[width=\linewidth,height=0.46\textheight,keepaspectratio]{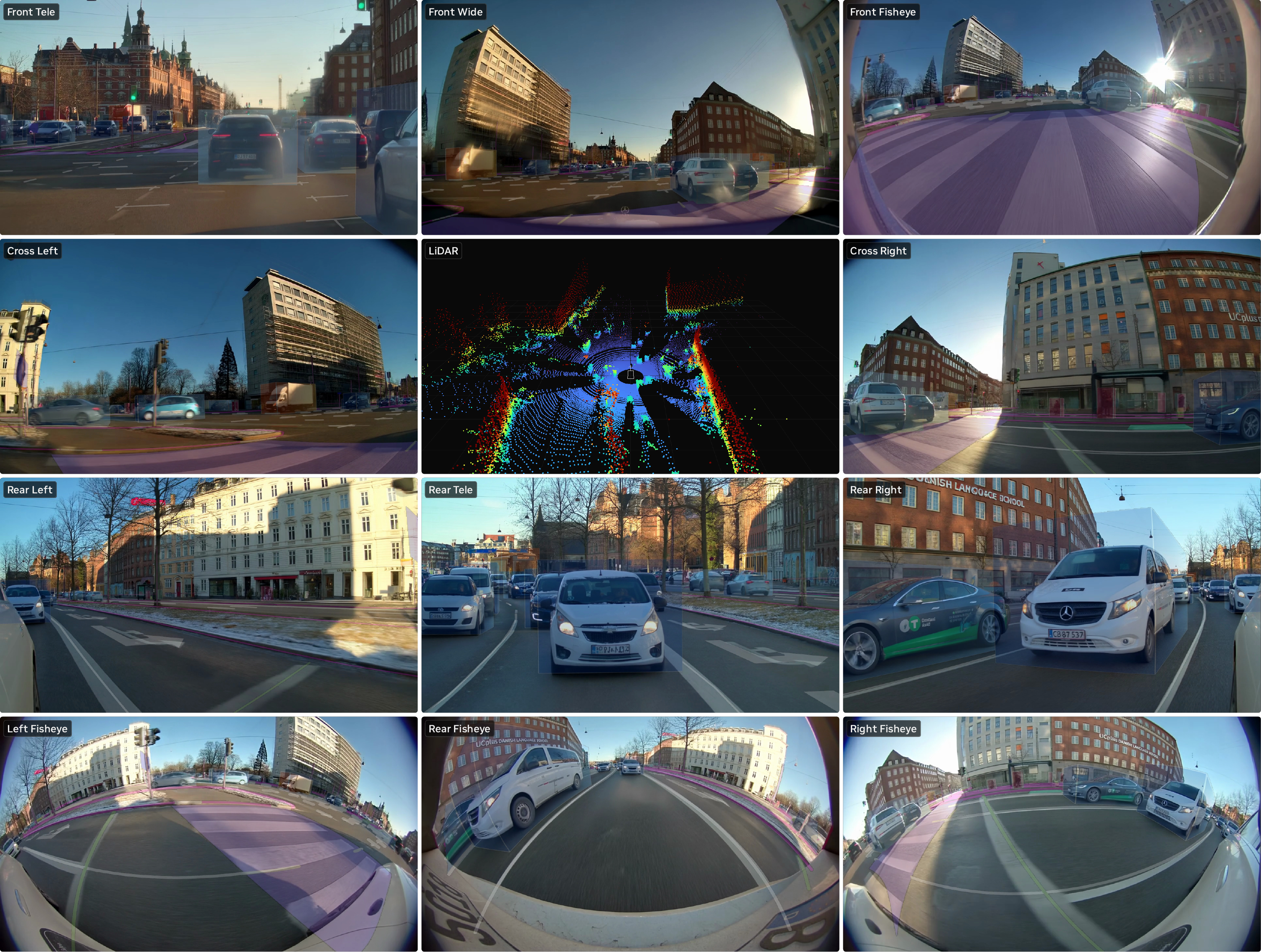}
\caption[Multi-view RGB--LiDAR generation with Cosmos3-Nano-Transfer-Auto.]{\textbf{Multi-view RGB--LiDAR generation with Cosmos3-Nano-Transfer-Auto.}
This urban intersection under strong sun glare illustrates joint generation across eleven cameras and LiDAR.
The white van retains its appearance in the rear-right and rear-fisheye views, while the forward and right-facing cameras agree on the direction of sunlight.
Overlaid WSM controls align with the generated road markings and vehicle positions.
The views follow the rig's spatial layout around a central LiDAR sweep, whose height coloring distinguishes the road surface, nearby vehicles, and surrounding buildings.}
\label{fig::auto_transfer_golden_hour}
\end{figure}
\ifhmode\par\fi\endgroup

%% file: figures/auto/auto_sim_sun_glare.tex
\begingroup{}
\begin{figure}[t]
\centering
\includegraphics[width=\linewidth,height=0.46\textheight,keepaspectratio]{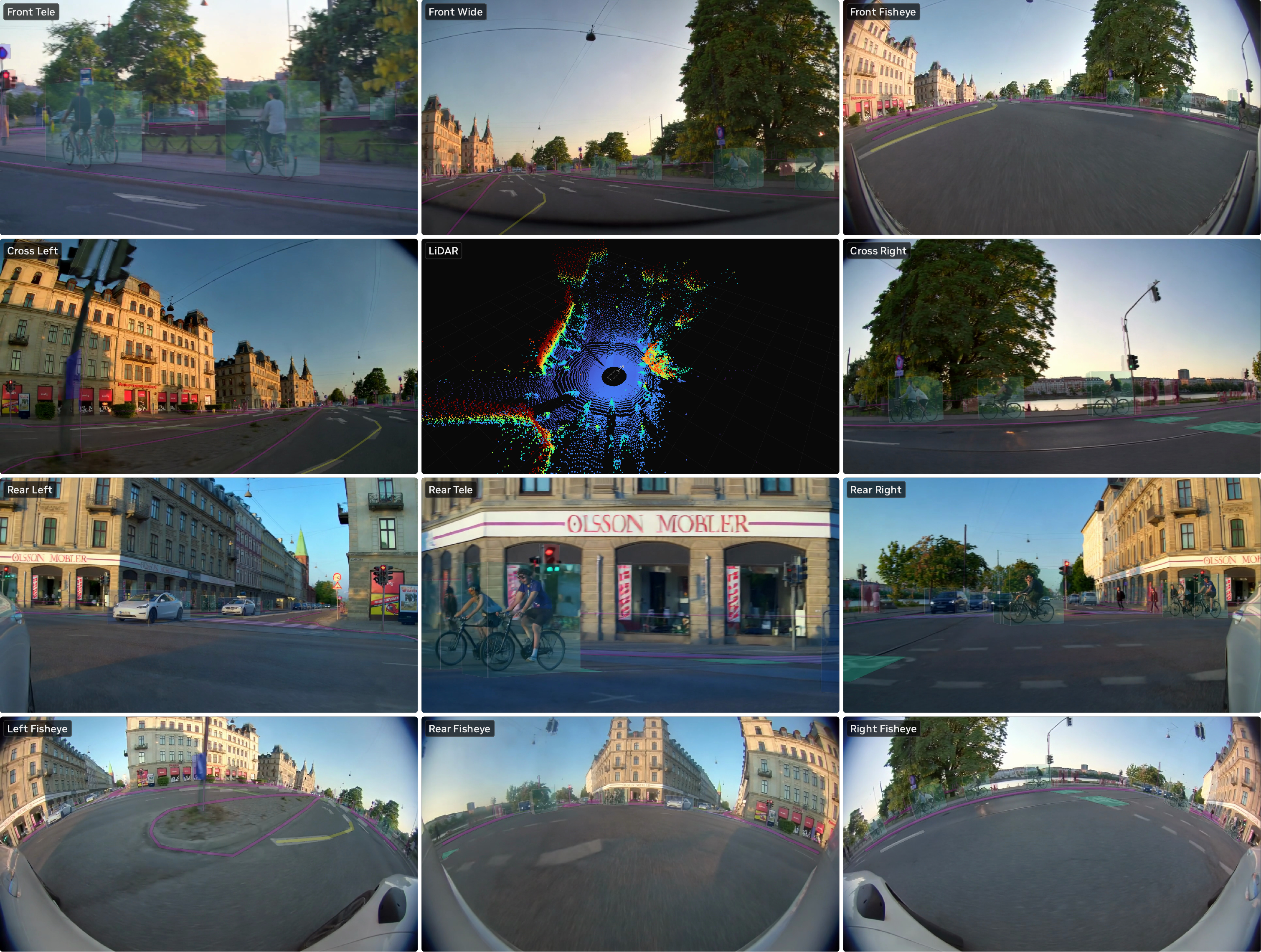}
\caption{\textbf{Multi-view RGB--LiDAR generation with Cosmos3-Nano-Sim-Auto.}
Camera views are arranged by their viewing directions around the vehicle, with generated LiDAR shown centrally.
WSM controls are overlaid on the RGB frames to highlight adherence to the supplied scene structure.
Buildings, vehicles, and cyclists retain consistent geometry and appearance across overlapping views.
The LiDAR point cloud is color-coded by height and captures corresponding scene geometry.}
\label{fig::auto_sim_sun_glare}
\end{figure}
\ifhmode\par\fi\endgroup

%% file: figures/auto/auto_long_rollout.tex
\begingroup{}
\begin{figure}[!t]
\centering
\includegraphics[width=0.9\linewidth]{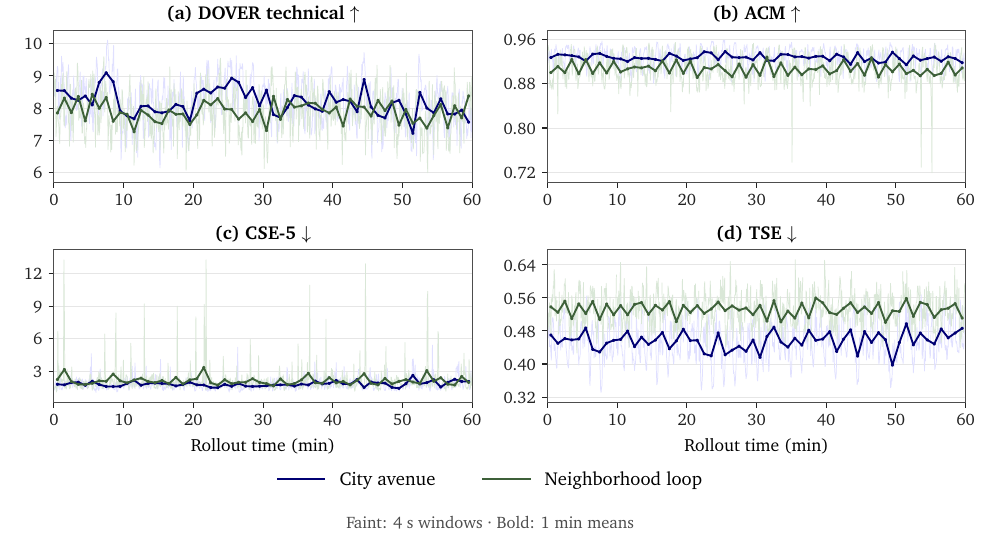}
\caption{\textbf{One-hour rollout stability of Cosmos3-Nano-Sim-Auto.}
We generate two driving scenes continuously for 1 hour, cycling their WSM controls and captions with a rolling K/V cache.
Video quality (DOVER), temporal consistency (ACM and TSE), and cross-view consistency (CSE-5) show no sustained degradation over the rollout.
Faint curves show 4-second-window scores and bold curves show 1-minute averages.}
\label{fig::auto_long_rollout}
\end{figure}
\ifhmode\par\fi\endgroup

%% file: sections/6__related_work.tex
\section{Related Work}
\label{sec::related_work}

\begingroup{}
Interactive world simulation connects learned representations of the physical world with controllable prediction and responsive execution.
World foundation models supply reusable visual and multimodal priors~\citep{cosmos_v1,cosmos3}, while causal generation and distillation make these priors available for interactivity~\citep{huang2025selfforcing,zheng2026causalrcm}.
Their use in an ongoing simulation also depends on how observations retain the consequences of earlier controls and how quickly new inputs can affect generated outputs.
We review these connections to position our shared post-training recipe and stateful streaming execution within the broader development of interactive world models.
\ifhmode\par\fi\endgroup

\subsection{Controllable World Models for Physical AI}

\begingroup{}
World models learn how environments evolve so agents can choose actions through imagined futures.
Early latent dynamics models used compact rollouts for planning and behavior learning~\citep{ha2018world,hafner2018planet,hafner2019dream}, and predictive representation learning extends this idea without rendering every predicted state~\citep{assran2025v}.
Generative world models expanded the interface to visible futures that can be inspected and used as simulated observations~\citep{bruce2024genie,cosmos_v1,cosmos_predict2p5}.
Multimodal foundations broaden this interface by coupling scene interpretation with the generation of future observations.
Cosmos 3~\citep{cosmos3} couples a reasoner tower for multimodal understanding with a generator tower for synthesis, sharing representations of language, observations, and heterogeneous actions.
Using these priors in an ongoing simulation additionally requires predictions to depend on available history and to become usable as new controls arrive.
Our framework inherits this backbone and its shared action interface, then adapts prediction and execution for incremental observation generation.
\ifhmode\par\fi\endgroup

\begingroup{}
Grounding these visual futures in executable interventions made prediction useful for control, from early model-predictive replanning~\citep{finn2017deep} to policy learning with action-conditioned diffusion~\citep{unisim}.
Autoregressive token models additionally bring observations, actions, and rewards into one predictive sequence~\citep{wu2024ivideogpt}.
Across these formulations, the learned response still depends on how training observations are associated with executable commands.
As visual priors broadened, control supervision became a central constraint.
Human videos offer varied experience through inferred latent actions that require adaptation to robot commands~\citep{gao2026dreamdojo}, while measured actions provide direct supervision with coverage limited by available interaction data~\citep{a2world2026}.
Cosmos3-Nano-Sim-Bimanual jointly learns camera navigation and robot manipulation from camera, robot, and human interaction data through a shared action interface (\cref{sec::bimanual}).
\ifhmode\par\fi\endgroup

\begingroup{}
Camera control exposes the spatial consequences of this grounding: requested motion must produce a compatible change of viewpoint~\citep{he2024cameractrl,xu2024camco}.
Gen3C~\citep{ren2025gen3c} renders a retained 3D cache to guide new views, while global poses, geometric memory, and metric annotations provide other spatial anchors~\citep{worldcam2026,worldstereo2026,sanawm2026}.
Decoupling human and camera motion helps compose controls~\citep{directingworld2026}, and sustained navigation makes historical scene evidence important~\citep{lingbot-world,gao2026lingbotworld2,matrixgame32026,matrixgame352026}.
Reference video and rendered scene evidence reduce uncertainty about observed content, while generation still needs to complete unseen regions~\citep{inspatioworld2026,worldinworld2026}.
These conditions constrain what is observed, while inference-time camera guidance~\citep{camtrolpp2026,particam2026} and reconstruction feedback~\citep{chronoworld2026} constrain how predictions follow the requested view.
Revisits in changing or shared scenes additionally motivate memory of interactions across observers~\citep{actworld2026,solaris2026}.
Our camera-navigation model accepts pose controls chunk by chunk and continues from generated visual history (\cref{sec::bimanual,sec::streaming_execution}).
\ifhmode\par\fi\endgroup

\begingroup{}
Structured scene conditions move part of this consistency problem outside the learned generator.
Geometry and spatial controls constrain layout and appearance~\citep{magicdrive2023,cosmos_transfer1}.
Explicit 3D worlds retain geometry across viewpoints~\citep{bahmani2025lyra,shen2026lyra2,hyworld22026,worldapi2026}, while spatially grounded image context connects view synthesis with reconstruction~\citep{worldlabs2026atlas}.
An external simulator can maintain evolving physical state and supply it through conditions, leaving the generator to synthesize appearance and unspecified dynamics.
When this observation interface expands to multiple sensors, consistency also needs to hold across streams.
Shared scenario maps anchor views in a common layout~\citep{cosmos_predict2p5}, and staged causal conversion supports mixed camera projections~\citep{hu2026zytworld}.
Joint camera--LiDAR models further extend consistency to complementary sensor representations~\citep{waymo_world_model,cheng2026m4world}.
Cosmos3-Nano-Sim-Transfer supports depth, segmentation, edges, and blur, while the AV models use camera-aligned WSM and range-view HD-map controls.
Inspired by the decomposed multi-view attention in OmniDreams~\citep{omnidreams2026}, our AV models use time-aligned, deduplicated cross-sensor attention for joint RGB--LiDAR generation (\cref{sec::auto_multiview_modeling}).
\ifhmode\par\fi\endgroup

\begingroup{}
These observation models also support decision-making through inverse dynamics, joint video and action modeling, or planning over imagined futures~\citep{feng2025vidarc,ye2026dreamzero,worldactionplanner2026}.
Policies can issue actions before video denoising completes through asynchronous sampling~\citep{xwam2026} or intermediate denoising features~\citep{motionwam2026}.
Comparing policy outcomes in generated and real environments tests simulation utility~\citep{jiang2025enerverse,guo2026ctrlworld}.
Our models provide the complementary observation generator, returning RGB or joint RGB--LiDAR outputs in response to supplied controls.
\ifhmode\par\fi\endgroup

\subsection{Causal Video Generation and Few-step Distillation}

\begingroup{}
Moving from generated clips to ongoing interaction changes what temporal information a model can use.
Bidirectional video diffusion coordinates observations across a complete target clip, while an interactive video generator must commit to the next prediction from the available history and controls.
Diffusion Forcing~\citep{chen2024diffusionforcing} provided a sequence formulation with varying noise levels, supporting different prediction and conditioning patterns.
CausVid~\citep{yin2025causvid} showed how a causal student could inherit a bidirectional video prior through distillation.
We first adapt Cosmos3-Nano to each simulation domain, then convert it to causal generation (\cref{sec::bidirectional_adaptation,sec::teacher_forcing}).
\ifhmode\par\fi\endgroup

\begingroup{}
Causality makes incremental control possible, but the cost of repeated denoising remains.
Consistency models reduce sampling steps by learning predictions that agree along a denoising trajectory~\citep{song2023consistency,lu2024simplifying}.
Distribution matching instead refines the fast student's output distribution using teacher and auxiliary score estimates~\citep{yin2024improved}.
These objectives can be combined through score regularization, preserving trajectory information while matching the generated distribution~\citep{zheng2025large}.
In autoregressive generation, this complementarity also motivates a staged procedure: consistency distillation initializes subsequent training on the student's own rollouts~\citep{lin2026autoregressive,zheng2026causalrcm}.
Teacher forcing supplies an offline, mode-covering initialization before on-policy distribution matching, whose optimization is sensitive to the starting model~\citep{zheng2026causalrcm}.
Using the same clean reference history for the teacher and student makes consistency targets conditional on the same task and allows their construction online~\citep{zheng2026causalrcm,zhao2026causalforcingpp}.
\ifhmode\par\fi\endgroup

\begingroup{}
A fast causal predictor still encounters a further mismatch: its context history contains its own imperfect outputs at rollout time.
Self Forcing~\citep{huang2025selfforcing} addresses this exposure bias by training with autoregressive student-generated histories and a loss over the resulting video.
Once training includes generated history, the next issue is how far that supervision extends.
Long-context teachers provide a training signal across longer student rollouts~\citep{zhu2026longforcing,omnidreams2026}, while forward-and-return cycle consistency offers a constraint without teacher-based distillation~\citep{live2026}.
Adversarial policy curricula additionally address control coverage by finding high-error trajectories near the behavior distribution and prioritizing them for training~\citep{prowl2026}.
Following Causal-rCM~\citep{zheng2026causalrcm}, our shared curriculum combines teacher forcing, discrete consistency distillation, and self-forcing distribution matching for action-conditioned and spatially controlled simulation (\cref{sec::training}).
\ifhmode\par\fi\endgroup

\subsection{Efficient Inference for Streaming Video Generation}

\begingroup{}
As streaming generation outgrows a fixed clip, memory must control computation without losing useful scene history.
K/V caching reuses unchanged observations, but retaining every entry increases memory and attention cost with the session.
Rolling windows and compact histories bound active context~\citep{rollingforcing2025,zhang2025framepack,yang2025longlive}, while initial sinks preserve startup information~\citep{streamingllm2023,omnidreams2026,zheng2026causalrcm}.
Matching training and deployment windows can reduce context-length mismatch, although reference and generated histories can still differ in distribution~\citep{motionstream2026}.
Older evidence can be compressed into a fixed-budget representation~\citep{infiniteworld2026} or recovered through retrieval~\citep{contextasmemory2025,reworldinteractive2026}.
Retrieving context relevant to the next prediction~\citep{wonder2026,matrixgame352026,worldkv2026} trades storage and selection costs against the information lost through compression.
Memory associated with time or interaction events can also support retrieval of scene content after intervening actions~\citep{memworld2026,actworld2026}.
We bound active context with configurable history windows and refresh cached features from completed predictions (\cref{sec::streaming_execution}).
\ifhmode\par\fi\endgroup

\begingroup{}
Bounded history also makes the recurring workload predictable, enabling compilation and graph replay~\citep{omnidreams2026}.
Efficient attention reduces the cost of each denoising step~\citep{ye2025flashinfer,monarchrt2026}.
Low-precision execution can further reduce arithmetic cost and memory use by representing model tensors with fewer bits.
LongLive-2.0~\citep{chen2026longlive2} combines NVFP4 inference with K/V cache quantization, applying lower precision to both model computation and retained history.
Distributing generation across devices offers additional parallelism at the cost of communication~\citep{deepspeed_ulysses,wanstreamer022026}.
Our runtime combines compilation and attention parallelism with asynchronous decoding where supported by the model's execution dependencies (\cref{sec::runtime_optimization}).
\ifhmode\par\fi\endgroup

\begingroup{}
Chunking determines when a newly received control can affect the output~\citep{realcam2026}, while generation across multiple streams adds synchronization requirements~\citep{starchild2026}.
We therefore report both throughput and latency using explicit observation-delivery boundaries (\cref{sec::runtime_optimization}).
\ifhmode\par\fi\endgroup

%% file: sections/7__conclusion.tex
\section{Conclusion}
\label{sec::conclusion}

\begingroup{}
We presented a shared post-training framework that turns Cosmos world foundation models into interactive simulators for Physical AI.
By combining causal learning, few-step distillation, and stateful streaming, the framework generates observations in response to incoming actions and scene controls.
The three Cosmos3-Nano examples span camera navigation and robot manipulation, video transfer, and multi-view AV simulation, supporting both prediction of visual action consequences and observation generation from an external simulator's physical state.
Selected configurations reach real-time generation throughput, and the calibrated robot study demonstrates policy feedback through generated observations.
The central opportunity is to make simulation a continuing source of experience shaped by an agent's decisions.
With sufficiently responsive observations, an agent can act, observe how the scene changes, and use that experience to decide what to do next.
We expect interactive world simulation to broaden the environments, appearances, and interventions available for Physical AI training and evaluation.
By coupling visual diversity with repeated, controllable interaction, these environments could shorten the cycle of learning, testing, and refining agents, helping developers identify failures before deployment.
\ifhmode\par\fi\endgroup

%% file: appendix/contributors.tex
\newpage
\begingroup
\makeatletter
\renewcommand{\paragraph}{%
  \@startsection{paragraph}{4}{\z@}%
    {1\baselineskip \@plus .2\baselineskip}%
    {-0.4em}%
    {\normalfont\normalsize\bfseries}%
}
\makeatother
\begingroup{}
\section{Contributors and Acknowledgments}
\label{sec::contributors}

Contributors in each group are listed alphabetically by last name.

\subsection{Contributors}

\paragraph{Supervision} \mbox{}
\vspace{-0.32\baselineskip}
\begingroup
\footnotesize
\begin{tasks}[label=\relax, label-width=0pt, item-indent=4pt](5)
\task Chen-Hsuan Lin
\task Ming-Yu Liu
\end{tasks}
\endgroup

\paragraph{Model Architecture} \mbox{}
\vspace{-0.32\baselineskip}
\begingroup
\footnotesize
\begin{tasks}[label=\relax, label-width=0pt, item-indent=4pt](5)
\task Francesco Ferroni
\task Zhengfei Kuang
\task George Kurian
\task Xiang Li
\task Chen-Hsuan Lin
\task Qianli Ma
\task Trung Pham
\task Jiaxiang Tang
\task Zhonghao Wang
\task Kaiwen Zheng
\end{tasks}
\endgroup

\paragraph{Post-training Recipe} \mbox{}
\vspace{-0.32\baselineskip}
\begingroup
\footnotesize
\begin{tasks}[label=\relax, label-width=0pt, item-indent=4pt](5)
\task Xiao Fu
\task DeLesley Hutchins
\task Chen-Hsuan Lin
\task Qianli Ma
\task Mustafa Munir
\task Jiaxiang Tang
\task Kaiwen Zheng
\end{tasks}
\endgroup

\paragraph{Cosmos3-Nano-Sim-Bimanual} \mbox{}
\label{sec::contributors_sdg}
\vspace{-0.32\baselineskip}
\paragraph{\footnotesize Data Preparation} \mbox{}
\vspace{-0.32\baselineskip}
\begingroup
\footnotesize
\begin{tasks}[label=\relax, label-width=0pt, item-indent=4pt](5)
\task Xiao Fu
\task Xin Kong
\task Aleksandr Korovko
\task Zhengfei Kuang
\task Zhaoshuo Li
\task Chen-Hsuan Lin
\task Qianli Ma
\task Jiaxiang Tang
\task Shitao Tang
\task Shijie Wang
\task Mengyao Xu
\end{tasks}
\endgroup

\vspace{-0.32\baselineskip}
\paragraph{\footnotesize Post-training} \mbox{}
\vspace{-0.32\baselineskip}
\begingroup
\footnotesize
\begin{tasks}[label=\relax, label-width=0pt, item-indent=4pt](5)
\task Xiao Fu
\task Zhengfei Kuang
\task Chen-Hsuan Lin
\task Jiaxiang Tang
\task Shijie Wang
\end{tasks}
\endgroup

\vspace{-0.32\baselineskip}
\paragraph{\footnotesize Inference Optimization} \mbox{}
\vspace{-0.32\baselineskip}
\begingroup
\footnotesize
\begin{tasks}[label=\relax, label-width=0pt, item-indent=4pt](5)
\task Anis Ahmad
\task Jesse Archer
\task Maciej Bala
\task Aarti Basant
\task Taras Bolzhelarskyi
\task Yong He
\task William Lew
\task Martin Ding Ma
\task Pengcuo Zeren
\task Yuliya~Zhautouskaya
\end{tasks}
\endgroup

\paragraph{Cosmos3-Nano-Sim-Transfer} \mbox{}
\vspace{-0.32\baselineskip}
\paragraph{\footnotesize Data Preparation} \mbox{}
\vspace{-0.32\baselineskip}
\begingroup
\footnotesize
\begin{tasks}[label=\relax, label-width=0pt, item-indent=4pt](5)
\task Francesco Ferroni
\task Qianli Ma
\task Zhonghao Wang
\end{tasks}
\endgroup

\vspace{-0.32\baselineskip}
\paragraph{\footnotesize Post-training} \mbox{}
\vspace{-0.32\baselineskip}
\begingroup
\footnotesize
\begin{tasks}[label=\relax, label-width=0pt, item-indent=4pt](5)
\task Zhonghao Wang
\end{tasks}
\endgroup

\vspace{-0.32\baselineskip}
\paragraph{\footnotesize Inference Optimization} \mbox{}
\vspace{-0.32\baselineskip}
\begingroup
\footnotesize
\begin{tasks}[label=\relax, label-width=0pt, item-indent=4pt](5)
\task Maciej Bala
\task Pengcuo Zeren
\task Yuliya~Zhautouskaya
\end{tasks}
\endgroup

\paragraph{Cosmos3-Nano-Sim-Auto} \mbox{}
\vspace{-0.32\baselineskip}
\paragraph{\footnotesize Data Preparation} \mbox{}
\vspace{-0.32\baselineskip}
\begingroup
\footnotesize
\begin{tasks}[label=\relax, label-width=0pt, item-indent=4pt](5)
\task Francesco Ferroni
\task Justin Leung
\task Xiang Li
\task Qianli Ma
\task Trung Pham
\task Andrew Z. Wang
\task Mengyao Xu
\task Andrew Zhu
\end{tasks}
\endgroup

\vspace{-0.32\baselineskip}
\paragraph{\footnotesize Post-training} \mbox{}
\vspace{-0.32\baselineskip}
\begingroup
\footnotesize
\begin{tasks}[label=\relax, label-width=0pt, item-indent=4pt](5)
\task Francesco Ferroni
\task Xiang Li
\task Qianli Ma
\task Trung Pham
\task Zhonghao Wang
\end{tasks}
\endgroup

\vspace{-0.32\baselineskip}
\paragraph{\footnotesize Inference Optimization} \mbox{}
\vspace{-0.32\baselineskip}
\begingroup
\footnotesize
\begin{tasks}[label=\relax, label-width=0pt, item-indent=4pt](5)
\task Jesse Archer
\task Maciej Bala
\task Aarti Basant
\task Vivek Goel
\task George Kurian
\task Janarbek Matai
\task Jonathan McCaffrey
\task Yuliya~Zhautouskaya
\end{tasks}
\endgroup

\paragraph{Infrastructure} \mbox{}
\vspace{-0.32\baselineskip}
\begingroup
\footnotesize
\begin{tasks}[label=\relax, label-width=0pt, item-indent=4pt](5)
\task Dan Blick
\task Liang Feng
\task Aryaman Gupta
\task Ali Hassani
\task Nathan Hayes-Roth
\task DeLesley Hutchins
\task Suneel Indupuru
\task Ashna Khetan
\task Hui Kuang
\task George Kurian
\task Chen-Hsuan Lin
\task Xiangyu Lu
\task Qianli Ma
\task Miguel Martin
\task Jiaxiang Tang
\task Sergei Vasilev
\task Thomas Volk
\task Hao Wang
\task Jing Zhang
\end{tasks}
\endgroup

\paragraph{Benchmarks} \mbox{}
\vspace{-0.32\baselineskip}
\begingroup
\footnotesize
\begin{tasks}[label=\relax, label-width=0pt, item-indent=4pt](5)
\task Jon Allen
\task Malik Aqeel Anwar
\task Dan Blick
\task Tae Eun Choe
\task Yuzhu Dong
\task Francesco Ferroni
\task Xiao Fu
\task Aryaman Gupta
\task Pooya Jannaty
\task Jingyi Jin
\task Janmesh Kalra
\task Zhengfei Kuang
\task Jayjun Lee
\task Chen-Hsuan Lin
\task Yuhang Lu
\task Qianli Ma
\task Ashkan Mirzaei
\task David Pankratz
\task Trung Pham
\task Marilyn Reeb
\task Jiaxiang Tang
\task Tolou Tavakkoli
\task Alexander Temirov
\task Zhonghao Wang
\task David Wehr
\task Shumian Xin
\task Sophia Zalewski
\task Cindy Zha
\task Andrew Zhu
\end{tasks}
\endgroup

\paragraph{Other Contributions} \mbox{}
\vspace{-0.32\baselineskip}
\begingroup
\footnotesize
\begin{tasks}[label=\relax, label-width=0pt, item-indent=4pt](5)
\task Piotr Barejko
\task Henry Che
\task Jin Chen
\task Yukang Chen
\task Nat Duca
\task Mohit Gambhir
\task Rony Goldenthal
\task Michael Honke
\task Shengyu Huang
\task Wei Huang
\task Jean-Francois Lafleche
\task Aditya Mahajan
\task Qing Miao
\task Sangmin Oh
\task Gavriel State
\task Yue Tang
\task Hweimi Tsou
\task Yao Xu
\task Shuai Yang
\task Rick Zhuang
\end{tasks}
\endgroup

\subsection{Acknowledgments}
\begingroup
\footnotesize
\begin{tasks}[label=\relax, label-width=0pt, item-indent=4pt](5)
\task Yogesh Balaji
\task Julius Berner
\task Tiffany Cai
\task Yan Chang
\task Chao-Yeh Chen
\task Tiffany Chen
\task Ruben D'Sa
\task Riccardo de Lutio
\task Jiaojiao Fan
\task Jun Gao
\task Shenyuan Gao
\task Lanxiang Hu
\task Jiahui Huang
\task Christian Laforte
\task Laura Leal-Taixe
\task Hongyu Li
\task Tsung-Yi Lin
\task Kaichun Mo
\task Seungjun Nah
\task Anjul Patney
\task Hesam Rabeti
\task John Shao
\task Wei-Cheng Tseng
\task Jialiang Wang
\task Hongchi Xia
\task Chenchao Xu
\task Yajie Yan
\task Liangkai Zhang
\end{tasks}
\endgroup

\ifhmode\par\fi\endgroup
\endgroup

\clearpage

%% file: appendix/bimanual_supplement.tex
\FloatBarrier
\section{Cosmos3-Nano-Sim-Bimanual: Additional Details}
\label{sec::bimanual_supplement}

\FloatBarrier
\subsection{Data and Caption Construction}

\input{appendix/bimanual_captions.tex}

\FloatBarrier
\subsection{Training Configuration}
\input{appendix/bimanual_training.tex}

\FloatBarrier
\subsection{Decoder Studies}
\input{appendix/bimanual_runtime.tex}

\FloatBarrier
\subsection{Evaluation Metrics}
\label{sec::bimanual_appendix_metrics}
\input{appendix/bimanual_metrics.tex}

\FloatBarrier
\subsection{Training Ablations and Development Studies}
\label{sec::ablations}
\input{appendix/recipe_ablations.tex}
\input{appendix/bimanual_moba_slt.tex}

%% file: appendix/bimanual_captions.tex
\begingroup{}
Captions provide scene information appropriate to each control task.
For camera navigation, the caption describes the environment and may include content outside the initial view.
It retains object appearance, spatial relationships, and independently moving subjects, but excludes camera motion and apparent scene motion caused by viewpoint changes.
For robotics and human interaction, the caption excludes the controlled agent's actions and their effects on the scene.
The modality-specific rewrite removes the robot's motion and changing pose in AgiBot/YAM~\citep{bu2025agibot,i2rt_yam}, or the camera wearer's actions in Mecka~\citep{mecka}.
However, an object's eventual position or state can still reveal the outcome of a future manipulation.
\ifhmode\par\fi\endgroup

\begingroup{}
To remove these outcome descriptions, we apply an additional text rewrite to the AgiBot, YAM, and Mecka annotations.
We extract the first temporal segment, remove its timestamp, and rewrite it as an initial-state description.
The rewriting model receives only this segment, without later segments, the structured scene breakdown, or the video.
The prompt preserves the setting, appearance, initial configurations, existing contacts, and independent background activity.
It removes camera motion, controlled-agent actions, and their consequences, and prohibits inferring an initial state from a removed action or outcome.
For example, an object already held at the start may be described, but its subsequent displacement or placement is excluded.
Because the source caption was generated from video, this procedure does not establish that every retained fact is visible in the initial frame.
\ifhmode\par\fi\endgroup

\begingroup{}
The video generator receives the final caption text without the structured scene breakdown.
All four training stages use the same final captions, matched to the covered video samples by source identifier.
Camera videos use the camera captioner's scene descriptions, while Behavior-1K~\citep{li2023behavior1k} uses separately supplied task descriptions.
\ifhmode\par\fi\endgroup

\begingroup{}
Robot evaluation captions are constructed for the exact temporal window paired with each benchmark video and action sequence.
For the 100 AgiBot-GEAR and 232 AgiBot-World clips in \cref{sec::bimanual_results}, descriptions are generated from video frames sampled at 2~FPS.
The training first-segment rewriting prompt then produces the final initial-state captions, with nonempty descriptions for all 332 clips.
\ifhmode\par\fi\endgroup

%% file: appendix/bimanual_training.tex
\label{sec::bimanual_training_details}
\begingroup{}
Stage 0 initializes the bidirectional teacher from the public \href{https://huggingface.co/nvidia/Cosmos3-Nano}{\texttt{nvidia/Cosmos3-Nano}} release.
Training predicts RGB targets conditioned on supplied actions.
All four stages use the same forward-dynamics conditioning rule: the first RGB frame and the complete aligned action sequence are provided as clean conditions, while the remaining RGB frames are prediction targets.
Stage 0 processes the targets bidirectionally; TF and TF-dCM condition each causal chunk on recorded preceding frames, whereas SF-DMD conditions subsequent chunks on the student's own completed predictions.
The causal stages predict chunks of 2 latent frames and retain the complete available history without sink tokens.
\ifhmode\par\fi\endgroup

\begingroup{}
\paragraph{Noise-time sampling.}
\label{app::noise_time_sampling}
The following noise-time sampling rules and resolution-dependent shifts apply to all three models throughout bidirectional adaptation, causal training, and distillation.
Following Cosmos~3~\citep{cosmos3}, we use mode sampling with scale $1.29$ for bidirectional adaptation and TF, as in Waver~\citep{zhang2025waver}.
We draw $r\sim\mathcal{U}(0,1)$ and transform it as
\[
 u=1-r-1.29\left[\cos^2\!\left(\frac{\pi r}{2}\right)-1+r\right].
\]
We then shift the sampled noise time using
\[
 \tau=S_\kappa(u)=\frac{\kappa u}{1+(\kappa-1)u},
\]
The shift factors are $\kappa=3$ at 256p, $5$ at 480p, and $10$ at 720p.
TF-dCM instead draws $u$ uniformly from $(1/48,1)$ and applies this shift to the pair $(u,u-1/48)$.
SF-DMD draws score times uniformly from $(0,1)$ before applying the same shift.
\ifhmode\par\fi\endgroup

\begingroup{}
All stages use a per-GPU microbatch size of one clip and a gradient-accumulation factor of 1.
The $2:2:1$ weights distribute camera, robot, and human clips across training GPUs, with minor differences caused by rounding.
Each GPU processes one clip using the action-processing pipeline for its source, and gradients from all GPUs are combined before one optimizer step.
\ifhmode\par\fi\endgroup

\begingroup{}
The bidirectional and teacher-forcing (TF) stages use FusedAdam with a base learning rate of $10^{-5}$.
The bidirectional stage uses betas $(0.9,0.99)$ and weight decay $0.05$, while the TF stage uses betas $(0.9,0.999)$ and weight decay $0.01$.
The action-to-model projection and action-modality embedding use 5 times the base learning rate.
\ifhmode\par\fi\endgroup

\begingroup{}
The TF-dCM stage uses FusedAdam with a learning rate of $2\times10^{-6}$, betas $(0,0.999)$, and weight decay $0.01$.
Its teacher uses classifier-free guidance (CFG) with a scale of $3$.
The reference branch replaces the scene caption with an empty string while retaining the task prompt, supplied actions, and recorded sensor history.
The SF-DMD stage uses FusedAdam with student and auxiliary score-model learning rates of $10^{-6}$ and $2\times10^{-7}$, respectively, betas $(0,0.999)$, and weight decay $0.01$.
Its teacher does not use CFG.
Training alternates 1 student update with 5 auxiliary score-model updates and uses only 480p clips of 901 RGB frames, including the initial frame.
\ifhmode\par\fi\endgroup

\begingroup{}
Model computation and gradient reduction use BF16, while master weights, optimizer states, and exponential moving average (EMA) weights are maintained in FP32 under FSDP~\citep{zhao2023pytorchfsdp}.
All reported final checkpoints use the corresponding EMA weights.
The reported SF-DMD model uses 4 denoising steps at inference.
\ifhmode\par\fi\endgroup

%% file: appendix/bimanual_runtime.tex
\begingroup{}
Additional decoder studies use camera-navigation workloads separate from the final Cosmos3-Nano-Sim-Bimanual benchmarks.
In an 80-sample quality evaluation, replacing the Wan 2.2 VAE with TAEHV Super reduces the PAI-Bench-G score from 71.8 to 70.9.
Separate 480p GB200 timing tests report $1.42$--$1.91\times$ speedups with synchronous decoding and $1.06$--$1.18\times$ with asynchronous decoding.
These quality and timing workloads differ from each other and from the final benchmarks, so they do not establish a matched quality--speed tradeoff for the final model.
The reported GB300 decoder runtime comparison appears in \cref{tab::latency_bimanual}.
\ifhmode\par\fi\endgroup

%% file: appendix/bimanual_metrics.tex
\begingroup{}
Video quality uses the PAI-Bench-G~\citep{zhou2025paibench} overall score, and quality drift is its absolute start/end difference as defined in \cref{sec::bimanual_results}.
The metrics below complement these scores with reference-frame agreement for robot manipulation and camera-trajectory agreement for camera navigation.
\ifhmode\par\fi\endgroup

\paragraph{Reference-frame agreement.}
\begingroup{}
Robot manipulation evaluation compares generated and reference frames using peak signal-to-noise ratio (PSNR, dB), structural similarity (SSIM)~\citep{wang2004ssim}, and learned perceptual image patch similarity (LPIPS)~\citep{zhang2018lpips}.
PSNR measures pixel reconstruction error on a logarithmic scale, SSIM measures structural image similarity, and LPIPS measures learned perceptual feature distance.
Higher PSNR and SSIM and lower LPIPS indicate closer reference-frame agreement.
Frame synchronization, resizing, image range, the LPIPS backbone, and valid frame coverage are fixed.
These metrics are action-following proxies and do not independently validate contact or task outcomes.
\ifhmode\par\fi\endgroup

\paragraph{Camera trajectory alignment.}
\begingroup{}
Camera navigation evaluation compares ViPE-estimated poses~\citep{huang2025vipe} from generated videos with supplied trajectories after Sim(3) Umeyama alignment~\citep{umeyama1991alignment}.
Absolute translation error (ATE) is the L2 distance between aligned inferred and supplied translations.
Relative translation error (RTE) is the L2 distance between inferred and supplied relative translations.
Relative rotation error (RRE) is the angular distance between their relative rotation matrices.
Lower values indicate closer trajectory agreement after translation, rotation, and scale alignment.
The relative pose interval, alignment granularity, translation units, and angular units are fixed across models.
These scores do not establish accuracy at absolute metric scale or independently validate physical outcomes.
\ifhmode\par\fi\endgroup

%% file: appendix/recipe_ablations.tex
\begingroup{}
We study training choices for Cosmos3-Nano-Sim-Bimanual on the Sana-WM-Bench camera-easy subset~\citep{sanawm2026}.
The studies cover teacher preparation, distillation configuration, and alternative training recipes.
Comparisons within each table use the same evaluation cases and protocol.
Checkpoints, datasets, and training recipes can differ across studies, so absolute scores are not directly comparable across tables.
Quality denotes the PAI-Bench-G overall score, while ATE, RTE, and RRE measure camera trajectory agreement as described in \cref{sec::bimanual_appendix_metrics}.
Higher quality and lower trajectory errors are preferred.
The findings characterize the tested configurations, including development recipes that differ from the final model.
\ifhmode\par\fi\endgroup

\subsubsection{{Teacher Preparation}}

\begingroup{}
\paragraph{Training clip length.}
We compare bidirectional teacher adaptation using clips up to 961 frames with adaptation using fixed 253-frame clips.
Both configurations are evaluated at iteration 3,500 on 961-frame camera navigation sequences.
Long-clip training increases quality by 1.052 points and reduces ATE, RTE, and RRE (\cref{tab::camera_clip_length}).
RTE falls from 0.089 to 0.029.
The long-clip configuration improves all four reported metrics at this checkpoint and evaluation horizon.
\ifhmode\par\fi\endgroup

\begingroup{}
\begin{table}[!t]
\centering\reporttablestyle
\caption{\textbf{Bidirectional teacher training clip length.}
Cosmos3-Nano-Sim-Bimanual teachers are trained with clips up to 961 frames or fixed 253-frame clips.
Both configurations are evaluated at iteration 3,500 on 961-frame Sana-WM-Bench camera-easy sequences.}
\label{tab::camera_clip_length}
\begin{tabularx}{\linewidth}{lCCCC}
\toprule
\textbf{Training clips} & \textbf{Quality}$\uparrow$ & \textbf{ATE}$\downarrow$ & \textbf{RTE}$\downarrow$ & \textbf{RRE}$\downarrow$\\
\midrule
Long, up to 961 frames & \textbf{75.165} & \textbf{1.756} & \textbf{0.029} & \textbf{0.286}\\
Short, fixed 253 frames & \underline{74.113} & \underline{2.007} & \underline{0.089} & \underline{0.316}\\
\bottomrule
\end{tabularx}
\end{table}
\ifhmode\par\fi\endgroup

\begingroup{}
\paragraph{Full fine-tuning versus LoRA.}
We compare full fine-tuning with LoRA~\citep{hu2022lora} using the long-clip camera recipe.
Both bidirectional checkpoints are evaluated at iteration 2,000 over 961-frame camera navigation sequences (\cref{tab::camera_teacher_adaptation}).
LoRA uses rank 16 and alpha 32 and trains only adapter parameters, while full fine-tuning updates the configured model parameter set.
Full fine-tuning has higher visual quality by 2.107 points and lower trajectory errors.
This result favors the full fine-tuning configuration over the tested rank-16 LoRA configuration.
\ifhmode\par\fi\endgroup

\begingroup{}
\begin{table}[!t]
\centering\reporttablestyle
\caption{\textbf{Full fine-tuning versus LoRA for bidirectional teacher adaptation.}
Cosmos3-Nano-Sim-Bimanual teachers use the long-clip camera recipe and are evaluated at iteration 2,000 on 961-frame Sana-WM-Bench camera-easy sequences.
LoRA uses rank 16 and alpha 32 with EMA disabled, while full fine-tuning retains EMA.}
\label{tab::camera_teacher_adaptation}
\begin{tabularx}{\linewidth}{YCCCC}
\toprule
\textbf{Adaptation} & \textbf{Quality}$\uparrow$ & \textbf{ATE}$\downarrow$ & \textbf{RTE}$\downarrow$ & \textbf{RRE}$\downarrow$\\
\midrule
Full fine-tuning & \textbf{77.086} & \textbf{1.594} & \textbf{0.018} & \textbf{0.081}\\
LoRA, rank 16 & \underline{74.979} & \underline{1.954} & \underline{0.048} & \underline{0.224}\\
\bottomrule
\end{tabularx}
\end{table}
\ifhmode\par\fi\endgroup

\FloatBarrier
\subsubsection{{Distillation Configuration}}

\begingroup{}
\paragraph{Teacher guidance.}
We vary classifier-free guidance (CFG)~\citep{ho2022classifierfree} in the distillation teacher separately for TF-dCM and SF-DMD.
CFG controls the teacher predictions used to train the student.
The TF-dCM and SF-DMD comparisons use separate recipes and evaluation checkpoints.

The TF-dCM comparison uses the same short-video training recipe and initialization, with causal teacher guidance scales 1, 3, and 5.
All checkpoints are evaluated at iteration 2,000 over 961 frames (\cref{tab::camera_tfdcm_guidance}).
Guidance 3 has the highest recorded quality, exceeding scales 1 and 5 by 0.194 and 0.144 points.
Guidance 5 has the lowest ATE, RTE, and RRE.
We select guidance 3 for the production TF-dCM stage to prioritize visual quality.

The SF-DMD comparison uses bidirectional teacher guidance scales 1 and 6.
Both checkpoints are evaluated at iteration 1,000 over 961 frames (\cref{tab::camera_sfdmd_guidance}).
Without teacher CFG, quality is higher by 1.757 points and ATE, RTE, and RRE are lower.
We use an unguided teacher for the production SF-DMD stage.
\ifhmode\par\fi\endgroup

\begingroup{}
\begin{table}[!t]
\centering\reporttablestyle
\caption{\textbf{TF-dCM teacher guidance.}
Cosmos3-Nano-Sim-Bimanual checkpoints use the same short-video training recipe and initialization, with causal teacher CFG scales 1, 3, and 5.
All checkpoints are evaluated at iteration 2,000 on 961-frame Sana-WM-Bench camera-easy sequences.}
\label{tab::camera_tfdcm_guidance}
\begin{tabularx}{\linewidth}{YCCCC}
\toprule
\textbf{Teacher guidance} & \textbf{Quality}$\uparrow$ & \textbf{ATE}$\downarrow$ & \textbf{RTE}$\downarrow$ & \textbf{RRE}$\downarrow$\\
\midrule
1 & 78.237 & 0.259 & 0.018 & 0.694\\
3 & \textbf{78.431} & \underline{0.244} & \underline{0.014} & \underline{0.367}\\
5 & \underline{78.287} & \textbf{0.225} & \textbf{0.012} & \textbf{0.145}\\
\bottomrule
\end{tabularx}
\end{table}
\ifhmode\par\fi\endgroup

\begingroup{}
\begin{table}[!t]
\centering\reporttablestyle
\caption{\textbf{SF-DMD teacher guidance.}
Cosmos3-Nano-Sim-Bimanual checkpoints use bidirectional teacher CFG scales 1 (no CFG) and 6.
Both are evaluated at iteration 1,000 on 961-frame Sana-WM-Bench camera-easy sequences.}
\label{tab::camera_sfdmd_guidance}
\begin{tabularx}{\linewidth}{YCCCC}
\toprule
\textbf{Teacher guidance} & \textbf{Quality}$\uparrow$ & \textbf{ATE}$\downarrow$ & \textbf{RTE}$\downarrow$ & \textbf{RRE}$\downarrow$\\
\midrule
1 (no CFG) & \textbf{76.344} & \textbf{0.943} & \textbf{0.012} & \textbf{0.149}\\
6 & \underline{74.587} & \underline{1.233} & \underline{0.015} & \underline{0.282}\\
\bottomrule
\end{tabularx}
\end{table}
\ifhmode\par\fi\endgroup

\begingroup{}
\paragraph{Rollout selection and denoising step schedule.}
We compare cycling and maximum rollout selection with 2-step and 4-step denoising schedules in the camera SF-DMD recipe.
Cycling varies the rollout depth up to the configured limits, while maximum selection always uses those limits.
The four checkpoints are evaluated at iteration 3,500 on 480p, 961-frame camera navigation sequences (\cref{tab::camera_rollout_selection}).
With the 2-step schedule, maximum selection increases quality by 0.484 points and reduces RTE, but increases ATE and RRE.
Maximum selection with the 4-step schedule has the highest quality and lowest RRE, but also the highest ATE and RTE.
No configuration dominates every measure.
We adopt maximum selection with the 4-step schedule in the production model to prioritize visual quality and rotational agreement.
\ifhmode\par\fi\endgroup

\begingroup{}
\begin{table}[!t]
\centering\reporttablestyle
\caption{\textbf{SF-DMD rollout selection and denoising schedule.}
Cosmos3-Nano-Sim-Bimanual checkpoints combine cycling or maximum rollout selection with 2-step or 4-step schedules.
All four are evaluated at iteration 3,500 on 480p, 961-frame Sana-WM-Bench camera-easy sequences.}
\label{tab::camera_rollout_selection}
\begin{tabularx}{\linewidth}{YcCCCC}
\toprule
\textbf{Selection} & \textbf{Schedule} & \textbf{Quality}$\uparrow$ & \textbf{ATE}$\downarrow$ & \textbf{RTE}$\downarrow$ & \textbf{RRE}$\downarrow$\\
\midrule
Cycle & 2 & 74.540 & \textbf{1.012} & \underline{0.015} & \underline{0.082}\\
Max & 2 & \underline{75.024} & \underline{1.039} & \textbf{0.012} & 0.201\\
Cycle & 4 & 73.933 & 1.138 & \underline{0.015} & 0.095\\
Max & 4 & \textbf{75.169} & 1.474 & 0.016 & \textbf{0.064}\\
\bottomrule
\end{tabularx}
\end{table}
\ifhmode\par\fi\endgroup

\FloatBarrier

%% file: appendix/bimanual_moba_slt.tex
\subsubsection{{Alternative Training Recipes}}

\begingroup{}
\paragraph{Mixed bidirectional and causal training.}
We compare a single model trained with alternating bidirectional and causal updates against models trained in only one attention mode.
The bidirectional:causal update patterns are $1:1$, $1:0$, and $0:1$, with shared initialization, chunks of 4 latent frames, and context parallelism of degree 4.
All three models are evaluated at iteration 4,500 on 961-frame camera navigation sequences (\cref{tab::camera_mixed_attention}).
The same mixed-training checkpoint is evaluated in both bidirectional and causal inference modes.

Under bidirectional inference, mixed training increases quality from 75.792 to 76.055 but increases ATE, RTE, and RRE.
Under causal inference, the causal-only model has higher quality and lower errors on all three trajectory measures.
These results show a quality--control tradeoff under bidirectional inference and favor causal-only training under causal inference at the evaluated checkpoints.

Equal total iterations give the mixed model approximately half as many updates per mode as each specialized model.
The comparison therefore combines the attention-mode choice with different per-mode training exposure.
The production curriculum first adapts a bidirectional teacher and then fine-tunes it with causal attention.
The study does not directly compare alternating training with this sequential curriculum.
\ifhmode\par\fi\endgroup

\begingroup{}
\begin{table}[!t]
\centering\reporttablestyle
\caption{\textbf{Mixed versus single attention-mode training.}
Cosmos3-Nano-Sim-Bimanual models share initialization, chunks of 4 latent frames, and context parallelism of degree 4.
All are evaluated at iteration 4,500 on 961-frame Sana-WM-Bench camera-easy sequences.
Mixed training alternates bidirectional and causal updates at $1:1$, and its two inference rows use the same checkpoint.
Each inference group compares mixed training with training in that attention mode alone.}
\label{tab::camera_mixed_attention}
\begin{tabularx}{\linewidth}{YCCCC}
\toprule
\textbf{Training} & \textbf{Quality}$\uparrow$ & \textbf{ATE}$\downarrow$ & \textbf{RTE}$\downarrow$ & \textbf{RRE}$\downarrow$\\
\midrule
\multicolumn{5}{l}{\textit{Bidirectional inference}}\\
Bidirectional only & \underline{75.792} & \textbf{1.245} & \textbf{0.016} & \textbf{0.079}\\
Mixed $1:1$ & \textbf{76.055} & \underline{1.496} & \underline{0.019} & \underline{0.081}\\
\midrule
\multicolumn{5}{l}{\textit{Causal inference}}\\
Causal only & \textbf{68.263} & \textbf{1.479} & \textbf{0.025} & \textbf{0.555}\\
Mixed $1:1$ & \underline{67.580} & \underline{1.521} & \underline{0.035} & \underline{0.693}\\
\bottomrule
\end{tabularx}
\end{table}
\ifhmode\par\fi\endgroup

\begingroup{}
\paragraph{Streaming long tuning.}
Streaming long tuning (SLT) carries generated state through successive training windows with a bounded K/V history~\citep{yang2025longlive,chen2026longlive2}.
It enables training on long student rollouts using a short-context teacher.
We compare SLT with standard long-rollout SF-DMD using maximum selection with 4 denoising steps on camera navigation.
SLT uses a short-clip teacher, whereas standard SF-DMD uses a long-context teacher.
The comparison therefore evaluates complete recipes rather than isolating the effect of streaming training.

Both recipes reach their highest observed camera-easy quality at iteration 2,000 (\cref{tab::camera_slt_best}).
Standard SF-DMD has 6.600 points higher quality than SLT and lower ATE and RRE.
SLT has lower RTE, so neither recipe is best on every reported metric.

We observe repeating visual patterns in the inspected SLT rollouts.
The SLT experiments use captions describing events throughout the source video.
This historical conditioning differs from the initial-state captions described for robot and human interaction in \cref{sec::bimanual_captioning}.
A mismatch between these whole-video event descriptions and the short-context teacher's local training windows may contribute to repetition, but this explanation was not tested separately.
The production model uses standard SF-DMD with maximum selection, 4 denoising steps, long rollouts, and a long-context teacher.
\ifhmode\par\fi\endgroup

\begingroup{}
\begin{table}[!t]
\centering\reporttablestyle
\caption{\textbf{Standard SF-DMD versus streaming long tuning.}
The Cosmos3-Nano-Sim-Bimanual camera recipes are evaluated on Sana-WM-Bench camera-easy.
Each row reports its highest observed quality checkpoint at iteration 2,000, with trajectory metrics from the same checkpoint.
Standard SF-DMD uses a long-context teacher, while SLT uses a short-clip teacher.}
\label{tab::camera_slt_best}
\begin{tabularx}{\linewidth}{YCCCC}
\toprule
\textbf{Training recipe} & \textbf{Quality}$\uparrow$ & \textbf{ATE}$\downarrow$ & \textbf{RTE}$\downarrow$ & \textbf{RRE}$\downarrow$\\
\midrule
SF-DMD & \textbf{77.167} & \textbf{1.434} & 0.017 & \textbf{0.088}\\
SLT & 70.567 & 1.539 & \textbf{0.016} & 0.164\\
\bottomrule
\end{tabularx}
\end{table}
\ifhmode\par\fi\endgroup

\FloatBarrier

%% file: appendix/transfer_supplement.tex
\FloatBarrier
\section[\kern0pt Cosmos3-Nano-Sim-Transfer: Additional Details]{{Cosmos3-Nano-Sim-Transfer: Additional Details}}
\label{sec::transfer_supplement}

\begingroup{}
\suppressfloats[t]
This appendix supplements the Cosmos3-Nano-Sim-Transfer methods and results in \cref{sec::transfer}.
After dataset accounting and sampling details, we describe runtime profiling, supplementary benchmark analysis, and Isaac Lab evaluation.
We conclude with training-stage comparisons and failure cases.
\ifhmode\par\fi\endgroup

\FloatBarrier
\subsection[\kern0pt Data and Sampling Details]{{Data and Sampling Details}}
\input{appendix/transfer_data_statistics.tex}

\input{appendix/transfer_training.tex}

\FloatBarrier
\subsection[\kern0pt Runtime Profiling Details]{{Runtime Profiling Details}}
\input{appendix/transfer_runtime.tex}

\FloatBarrier
\subsection[\kern0pt Supplementary Benchmark Analysis]{{Supplementary Benchmark Analysis}}
\input{appendix/transfer_evaluation.tex}

\input{appendix/transfer_metrics.tex}

\input{appendix/transfer_benchmark_statistics.tex}

\input{appendix/transfer_quality.tex}

\input{appendix/transfer_history_ablation.tex}

\FloatBarrier
\subsection[\kern0pt Isaac Lab Evaluation Details]{{Isaac Lab Evaluation Details}}
\input{appendix/transfer_closed_loop.tex}

\FloatBarrier
\subsection[\kern0pt Stage Comparisons and Failure Cases]{{Stage Comparisons and Failure Cases}}
\input{appendix/transfer_stage_comparison.tex}

\FloatBarrier
\input{appendix/transfer_failures.tex}

%% file: appendix/transfer_data_statistics.tex
\paragraph[\kern0pt Dataset retention.]{{Dataset retention.}}
\begingroup{}
\Cref{tab::transfer_data_accounting} breaks down the exclusions from the eligibility index described in \cref{sec::transfer_data}.
Each rejected source entry is assigned to one category, so the four exclusion counts partition the 156,946 rejected entries.
The table counts source entries by metadata and caption-window eligibility.
\ifhmode\par\fi\endgroup

\begingroup{}
\begin{table}[!htb]
\centering\reporttablestyle
\caption{\textbf{Cosmos3-Nano-Sim-Transfer dataset retention.}
The index retains one complete eligible caption window per indexed video.
The four exclusion counts sum to the total rejected population.}
\label{tab::transfer_data_accounting}
\begin{tabular}{lr}
\toprule
\textbf{Population or exclusion} & \textbf{Count}\\
\midrule
Source RGB video entries & 2,153,548\\
Indexed eligible videos & 1,996,602\\
Total excluded entries & 156,946\\
\midrule
Aspect-ratio distortion above 3\% & 114,453\\
Source FPS outside 5--60 & 10,870\\
Invalid caption envelope & 145\\
No complete eligible caption window & 31,478\\
\bottomrule
\end{tabular}
\end{table}
\ifhmode\par\fi\endgroup

%% file: appendix/transfer_training.tex
\label{sec::transfer_training_details}
\begingroup{}
The three causal stages train at 480p.
When averaging losses across latent sequences, bidirectional adaptation includes zero-loss control sequences, whereas TF and TF-dCM exclude sequences that contain only conditioning inputs.
\ifhmode\par\fi\endgroup

\paragraph[\kern0pt Teacher guidance.]{{Teacher guidance.}}
\begingroup{}
The distillation stages in \cref{sec::transfer_training} construct supervision from positive and negative teacher predictions.
Both branches receive identical spatial controls and include duration, FPS, and resolution metadata in their text conditioning.
The positive branch uses the sample's caption condition after text dropout, and the negative branch uses the canonical Cosmos3-Nano-Sim-Transfer negative caption.
\ifhmode\par\fi\endgroup

\paragraph[\kern0pt Student sampling and training history.]{{Student sampling and training history.}}
\begingroup{}
The SDE sampler evaluates fixed diffusion times $[1,15/16,5/6,5/8]$, corresponding to a shift-5 schedule, before reaching the clean endpoint.
Each nonterminal transition draws fresh Gaussian noise for the next evaluation.
Student evaluation uses guidance 1 for both language and controls.
During Stage 3 training, the K/V cache uses a window of 51 latent frames and a 1-latent-frame attention sink.
Evaluation history windows follow the workload-specific settings in \cref{sec::transfer_results}.
\ifhmode\par\fi\endgroup

%% file: appendix/transfer_runtime.tex
\paragraph[\kern0pt Offline runtime configuration.]{{Offline runtime configuration.}}
\begingroup{}
The compilation results in \cref{tab::transfer_offline_optimization} use four combinations of history policy and execution mode.
\Cref{tab::transfer_offline_runtime_configuration} reports the software, worker layout, and callback storage measurements for that comparison.
Each configuration runs two measured videos in each of three processes, with a warmup before every measured video.
Decoded-output timing covers latent indices 51--149, and component timing pools 588 measurements from six videos.
The distributions pool chunk measurements from all six videos.
\ifhmode\par\fi\endgroup

\begingroup{}
\begin{table}[!htb]
\centering\reporttablestyle
\caption{\textbf{Tested offline configuration and storage for Cosmos3-Nano-Sim-Transfer.}
The model, video geometry, and history policies are specified with the performance results in \cref{tab::transfer_offline_optimization}.
RTF is compute time divided by generated video time over the decoded-output measurement window.
Memory entries average the PyTorch allocated and reserved storage recorded at callbacks during these 597-frame generations.}
\label{tab::transfer_offline_runtime_configuration}
\begin{tabularx}{\linewidth}{L{4.4cm}Y}
\toprule
\textbf{Field} & \textbf{Configuration}\\
\midrule
Model / conditioning & SF-DMD iteration 1,500; single-view depth; no RGB prefix; seed 42\\
GPU / parallelism & 1 GB200 per worker; 4 independent workers on 1 node; CP=CFGP=1; batch size 1\\
Optimization & Eager; or \texttt{torch.compile} (\texttt{compiled\_region=language}) with block CUDA Graphs\\
Decoder / preprocessing & Wan2.2 streaming VAE; PyAV input decoding; \texttt{direct\_resize}; offline whole-video control preparation\\
Runtime & Torch \texttt{2.13.0a0+9186a08b2c.nv26.07}; CUDA 13.3\\
Timing per configuration & 3 processes $\times$ 2 measured videos, each preceded by warmup; 594 matched output chunks / 2,376 RGB frames\\
\bottomrule
\end{tabularx}
\par\smallskip
\begin{tabularx}{\linewidth}{L{4.4cm}CC}
\toprule
\textbf{History / execution} & \textbf{RTF}$\downarrow$ & \textbf{Callback mean allocated / reserved (GiB)}\\
\midrule
Full / eager & 4.3603 & 50.4828 / 60.0859\\
Full / compiled & 2.6300 & 67.3645 / 72.6318\\
Sink + recent / eager & 3.3185 & 39.5165 / 47.7598\\
Sink + recent / compiled & 1.8926 & 46.1787 / 50.7305\\
\bottomrule
\end{tabularx}
\end{table}
\ifhmode\par\fi\endgroup

\begingroup{}
We also compare the depth inferred from generated and reference RGB over the beginning of each video.
PAIBench-C v1.4.0 uses VideoDepthAnything-small~\citep{chen2025videodepth} to estimate both depth sequences.
Across 16 videos per configuration, the first 121 RGB frames (31 latent positions) give an SI-RMSE of 5.2124 for all four configurations.
This measurement describes depth agreement over that shared prefix.
\ifhmode\par\fi\endgroup

\paragraph[\kern0pt Component timing and startup.]{{Component timing and startup.}}
\begingroup{}
To explain the generation costs behind the decoded-output intervals in \cref{tab::transfer_offline_optimization}, \cref{tab::transfer_offline_latency} separates model computation, control insertion, and decoding.
For each chunk, model-only time is the sum of denoising and clean-history refresh.
We compute its p50 and p95 from these per-chunk sums.
Control K/V insertion is timed separately.
GPU component measurements use CUDA events on the current stream and include waits for their dependencies.
The VAE measurement ends before output clamping and device-to-host transfer, while the broader denoise-to-history-ready interval includes decoding and device-to-host waits.
Because these timing intervals overlap, their sum counts some work more than once.
\ifhmode\par\fi\endgroup

\begingroup{}
Preparation and process startup use elapsed wall time.
Startup begins before framework import and model setup and ends when the first decoded output reaches CPU memory.
This measurement is reported separately from warmed generation, and compiler disk caches are reused across launches.
The startup percentiles summarize three process launches per configuration.
\ifhmode\par\fi\endgroup

\begingroup{}
\begin{table}[!htb]
\centering\reporttablestyle
\caption{\textbf{Offline component, preparation, and startup times for Cosmos3-Nano-Sim-Transfer.}
Each entry gives the median (p50) and 95th percentile (p95) for the workload in \cref{tab::transfer_offline_runtime_configuration}.
F-E/F-C denote full-history eager/compiled execution, and S-E/S-C denote sink-plus-recent eager/compiled execution.
Panel (a) reports ms/chunk and pools observations from six measured videos per configuration.
Each component has 588 valid measurements from 588 attempts.
Panel (b) reports offline preparation in ms/video over 6/6 valid/attempted videos and startup in ms/process over 3/3~process~launches.}
\label{tab::transfer_offline_latency}
\par\smallskip
\textit{(a) Warmed generation components}\par\smallskip
\begin{tabularx}{\linewidth}{L{3.5cm}CCCC}
\toprule
\textbf{Component} & \textbf{F-E} & \textbf{F-C} & \textbf{S-E} & \textbf{S-C}\\
\midrule
Denoising / guidance & 337.13 / 462.66 & 198.36 / 261.76 & 255.61 / 301.77 & 127.09 / 146.81\\
Clean-history refresh & 87.91 / 123.48 & 52.37 / 67.86 & 67.10 / 79.30 & 35.34 / 37.35\\
Model-only chunk & 425.85 / 582.01 & 250.58 / 330.19 & 322.60 / 383.47 & 163.14 / 182.20\\
Control K/V insertion & 87.31 / 122.47 & 51.89 / 68.38 & 66.92 / 79.03 & 33.60 / 45.85\\
Streaming VAE decoding & 46.89 / 50.84 & 42.86 / 43.03 & 45.07 / 45.84 & 42.32 / 42.50\\
Denoise to history ready (includes decode) & 473.01 / 632.13 & 293.80 / 373.42 & 367.90 / 429.44 & 205.79 / 224.90\\
\bottomrule
\end{tabularx}
\par\smallskip
\textit{(b) Offline preparation and process startup}\par\smallskip
\begin{tabularx}{\linewidth}{L{4.4cm}CC}
\toprule
\textbf{History / execution} & \textbf{Whole-video preparation (ms/video)} & \textbf{Entry to first decoded output (ms/process)}\\
\midrule
Full / eager & 5302.11 / 5399.97 & 26619.47 / 61970.27\\
Full / compiled & 5266.26 / 5367.48 & 36489.38 / 78128.99\\
Sink + recent / eager & 5187.23 / 5330.32 & 62803.09 / 66504.84\\
Sink + recent / compiled & 5262.77 / 5328.87 & 36555.66 / 78673.18\\
\bottomrule
\end{tabularx}
\end{table}
\ifhmode\par\fi\endgroup

\FloatBarrier

%% file: appendix/transfer_evaluation.tex
\paragraph[\kern0pt Benchmark media alignment.]{{Benchmark media alignment.}}
\begingroup{}
\suppressfloats[t]
The full-benchmark comparison in \cref{sec::transfer_results} includes two media corrections to keep RGB and controls aligned.
For 5 source videos, we apply the same center crop to both streams.
For 17 OpenDV sources~\citep{yang2024genad}, we remux timestamps in 34 depth and segmentation files while preserving their decoded frames.
The corrected files provide matching spatial and temporal positions for evaluation.
\ifhmode\par\fi\endgroup

%% file: appendix/transfer_metrics.tex
\Needspace{10\baselineskip}
\paragraph[\kern0pt Depth error normalization.]{{Depth error normalization.}}
\begingroup{}
The TransferBench depth metric in \cref{tab::transfer_benchmark_metrics} compares VideoDepthAnything~\citep{chen2025videodepth} estimates from generated and reference RGB.
For each frame, $\widehat{d}$ and $d$ denote these two depth maps, and $\mathcal{P}_{\text{valid}}$ contains pixels with positive depth in both maps.
When this set is nonempty, we compute
\begin{equation}
 s=\frac{\operatorname{median}(d)}{\operatorname{median}(\widehat{d})},\qquad
 \mathrm{SI\mbox{-}RMSE}=\sqrt{\frac{1}{|\mathcal{P}_{\text{valid}}|}
 \sum_{\mathbf{p}\in\mathcal{P}_{\text{valid}}}
 \operatorname{clip}\!\left(s\widehat{d}(\mathbf{p})-d(\mathbf{p}),-10,10\right)^2}.
\end{equation}
Both medians are computed over $\mathcal{P}_{\text{valid}}$, and the residual is clipped to $[-10,10]$ before squaring.
The evaluator averages the frame scores over the clip, assigning zero to frames with no valid pixels.
The VisDex analysis below instead uses an uncapped error against simulator ground truth.
\ifhmode\par\fi\endgroup

%% file: appendix/transfer_benchmark_statistics.tex
\paragraph[\kern0pt Full-benchmark paired uncertainty.]{{Full-benchmark paired uncertainty.}}
\begingroup{}
\Cref{tab::transfer_benchmark_uncertainty} supplements the full TransferBench results in \cref{sec::transfer_results} with all nine paired TF-dCM-minus-TF metric differences and their source-bootstrap intervals.
The analysis uses the benchmark's 600 paired source videos, with 200 sources each from AgiBot, Ego-Exo4D, and OpenDV.
The differences include each stage's deployed sampler and guidance settings.
\ifhmode\par\fi\endgroup

\begingroup{}
We use each source video as the unit of the paired bootstrap~\citep{efron1979bootstrap}.
For video-level metrics, we average each source's four control variants, while control-specific metrics use the score for the corresponding modality.
Each replicate samples 200 source IDs with replacement within each domain and keeps both stages and all control variants together for every selected source.
This preserves the pairing between stages and the relationship between outputs generated from the same video.
We give the three domain means equal weight, then take the 2.5th and 97.5th percentiles of 10,000 replicates using linear interpolation, NumPy PCG64, and seed 20261019.
The resulting 95\% intervals describe uncertainty from sampling source videos with the recorded generations and judge outputs held fixed.
They do not include variability from changing generation seeds or repeating judge evaluations.
Each is a marginal interval for one metric, with no adjustment for simultaneous coverage across metrics.
\ifhmode\par\fi\endgroup

\begingroup{}
\begin{table}[!tb]
\centering\reporttablestyle
\caption{\textbf{Paired source-bootstrap intervals for TF-dCM relative to TF.}
The selected EMA checkpoints are TF iteration 6,000 and TF-dCM iteration 2,250.
Differences use each metric's original scale, and arrows indicate the preferred direction of the underlying metric.
Every metric has 600 paired source scores.}
\label{tab::transfer_benchmark_uncertainty}
\begin{tabular}{lrc}
\toprule
\textbf{Metric} & \textbf{TF-dCM $-$ TF} & \textbf{95\% interval}\\
\midrule
DOVER $\uparrow$ & $-0.170816$ & $[-0.236949,-0.106727]$\\
Segmentation IoU $\uparrow$ & $+0.012182$ & $[+0.003538,+0.020914]$\\
Blur SSIM $\uparrow$ & $-0.002144$ & $[-0.003209,-0.001059]$\\
Edge F1 $\uparrow$ & $+0.007809$ & $[+0.006058,+0.009492]$\\
Depth SI-RMSE $\downarrow$ & $-0.032128$ & $[-0.059830,-0.003746]$\\
Quality drift $\downarrow$ & $+0.003466$ & $[+0.002943,+0.003996]$\\
Sampson error $\downarrow$ & $+0.011013$ & $[+0.010184,+0.011842]$\\
Prompt $\uparrow$ & $-0.033750$ & $[-0.088333,+0.021250]$\\
Quality $\uparrow$ & $+0.038750$ & $[-0.039583,+0.118760]$\\
\bottomrule
\end{tabular}
\end{table}
\ifhmode\par\fi\endgroup

%% file: appendix/transfer_quality.tex
\Needspace{5\baselineskip}
\paragraph[\kern0pt Long-rollout quality.]{{Long-rollout quality.}}
\begingroup{}
To examine rendering quality over longer sequences, we evaluate the final student and GA reference on all 160 VisDex episodes under both depth and segmentation control.
Each model produces 320 scored videos of 600 frames, using the same source episodes, controls, and prompts.
We compare depth and segmentation extracted from generated RGB with references derived from the simulator's controls.

For depth, we align the predicted and reference medians separately in each frame, then compute the uncapped RMSE in normalized control-depth units over nonbackground, unsaturated pixels.
For segmentation, we compare Grounded-SAM2 masks extracted from generated frames with control masks labeled as robot, object, and table, counting missing detections as errors.
These class labels come from visual inspection of source-frame control colors, not exported simulator semantic IDs.
These spatial metrics use a different reference protocol from the 121-frame TransferBench evaluation.
We therefore interpret these values within the VisDex comparison.
\ifhmode\par\fi\endgroup

\begingroup{}
\Cref{tab::transfer_visdex_quality} shows that the student scores higher in segmentation IoU and Quality on these long rollouts.
The GA reference retains better DOVER, depth error, drift, Sampson error, and Prompt scores.
Together with the runtime measurements in \cref{tab::transfer_visdex_quality_runtime}, these results describe the quality and speed of the two deployed pipelines.

\Needspace{4\baselineskip}
Each pipeline retains its own sampling and history settings for this comparison.
GA uses 50-step UniPC with text guidance 3 and diffusion caching, while SF-DMD uses 4-step SDE with text guidance 1 and a causal K/V cache.
The control guidance also differs: GA uses 1.5 for depth and 2 for segmentation, while SF-DMD uses 1 for both modalities.
\ifhmode\par\fi\endgroup

\begingroup{}
\begin{table}[!htb]
\centering\reporttablestyle
\caption{\textbf{Matched 600-frame VisDex quality.}
Metrics cover all 160 episodes under depth and segmentation: 320 videos per model, with 160 for each control metric.
Both models use identical source episodes, controls, and prompts.
Depth uses simulator control values, and segmentation uses visually labeled control masks.}
\label{tab::transfer_visdex_quality}
\begin{tabular}{lcc}
\toprule
\textbf{Metric} & \textbf{Cosmos 3 GA reference} & \textbf{Cosmos3-Nano-Sim-Transfer}\\
\midrule
DOVER $\uparrow$ & 11.5036 & 9.1682\\
Quality $\uparrow$ & 2.1469 & 2.3500\\
Prompt $\uparrow$ & 5.2875 & 5.2250\\
Segmentation IoU $\uparrow$ & 0.5517 & 0.5727\\
Direct-GT depth SI-RMSE $\downarrow$ & 0.085053 & 0.087610\\
Drift $\downarrow$ & 0.014210 & 0.014365\\
Sampson $\downarrow$ & 0.081120 & 0.150449\\
\bottomrule
\end{tabular}
\end{table}
\ifhmode\par\fi\endgroup

%% file: appendix/transfer_history_ablation.tex
\FloatBarrier
\Needspace{5\baselineskip}
\paragraph[\kern0pt History window and attention sink ablation.]{{History window and attention sink ablation.}}
\begingroup{}
We select a K/V window of 30 latent frames and 3 sink frames for long rollouts based on the six configurations in \cref{tab::transfer_history_ablation}.
The sweep uses a development SF-DMD EMA checkpoint at iteration 1,250 on 160 depth-conditioned VisDex episodes.
Each configuration generates 593 frames at 30 FPS and $640\times640$, with a one-latent-frame chunk, four-step SDE sampling, seed 42, and identical rewritten prompts.
Only the window size $W$ and sink size $S$ change.
Both sizes count latent frames, with $W$ including the $S$ pinned initial frames and the current frame.
Thus, $W=30,S=3$ retains 3 initial frames and 26 recent history frames alongside the current frame.
\ifhmode\par\fi\endgroup

\begingroup{}
With 3 sink frames, increasing the window from 10 to 30 raises DOVER from 9.652 to 10.056, Prompt from 5.412 to 5.444, and Quality from 2.225 to 2.438.
Expanding the window to 50 slightly raises Quality to 2.450 but lowers DOVER to 9.933 and Prompt to 5.425.
We select window 30 for the highest DOVER in this sweep and nearly the same Quality as window 50, using 20 fewer latent-frame positions.
\ifhmode\par\fi\endgroup

\begingroup{}
At window 30, increasing the sink from 3 to 6 raises Prompt from 5.444 to 5.556 but lowers DOVER from 10.056 to 9.997 and Quality from 2.438 to 2.419.
We retain 3 sink frames for higher DOVER and Quality, accepting the lower Prompt grade while allocating more of the fixed window to recent observations.
\ifhmode\par\fi\endgroup

\begingroup{}
\begin{table}[!htb]
\centering\reporttablestyle
\caption{\textbf{History window and attention sink ablation for long rollouts.}
Each row aggregates 160 depth-conditioned VisDex videos from the same iteration-1,250 SF-DMD checkpoint.
$W$ and $S$ denote the total K/V window and pinned sink, measured in latent frames.
Gemini 3.1 Pro provides Prompt and Quality grades on a 0--10 scale using the original captions.
Bold marks the selected setting.}
\label{tab::transfer_history_ablation}
\begin{tabularx}{\linewidth}{CCCCC}
\toprule
\textbf{$W$} & \textbf{$S$} & \textbf{DOVER}$\uparrow$ & \textbf{Prompt}$\uparrow$ & \textbf{Quality}$\uparrow$\\
\midrule
6 & 3 & 9.550 & 5.487 & 2.237\\
10 & 3 & 9.652 & 5.412 & 2.225\\
\textbf{30} & \textbf{3} & \textbf{10.056} & \textbf{5.444} & \textbf{2.438}\\
50 & 3 & 9.933 & 5.425 & 2.450\\
\midrule
10 & 6 & 9.615 & 5.475 & 2.194\\
30 & 6 & 9.997 & 5.556 & 2.419\\
\bottomrule
\end{tabularx}
\end{table}
\ifhmode\par\fi\endgroup

%% file: appendix/transfer_closed_loop.tex
\paragraph[\kern0pt Initial states and seed assignments.]{{Initial states and seed assignments.}}
\begingroup{}
\suppressfloats[t]
The closed-loop results in \cref{tab::transfer_isaac_apple} use the runtime settings in \cref{tab::transfer_isaac_configuration} and fixed seed assignments for the 2 batches of 16 environments.
The first batch uses simulator seed 106931, policy seed 0, and renderer seeds 0--15 in environment order.
The second batch uses simulator seed 106932, policy seed 1, and renderer seeds 16--31.
All appearance conditions reuse these initial states and seed assignments, keeping the starting conditions fixed as the rendered appearance changes.
We freeze depth inputs and stop policy queries for terminated environments, discarding padded generated RGB.
\ifhmode\par\fi\endgroup

\begingroup{}
\begin{table}[!htb]
\centering\reporttablestyle
\caption{\textbf{Isaac Lab apple-task runtime configuration.}
These hardware, software, and execution settings accompany the closed-loop schedule in \cref{sec::transfer_results}.}
\label{tab::transfer_isaac_configuration}
\begin{tabularx}{\linewidth}{L{3.4cm}Y}
\toprule
\textbf{Field} & \textbf{Configuration}\\
\midrule
Simulator & Isaac Lab 3.0.0b2~\citep{isaaclab2025}; Isaac Sim 6.0.0.1~\citep{nvidia_isaac_sim_2026} on RTX PRO 4000\\
Renderer and policy hardware & GB300\\
Renderer software & Cosmos Framework 1.2.2; Python 3.13.16; PyTorch 2.10.0+cu130; CUDA 13.0; torchvision 0.25.0+cu130; transformers 4.57.6\\
Compilation & \texttt{torch.compile} and CUDA Graphs enabled, \texttt{compiled\_region=language}\\
cuDNN & \texttt{benchmark=False}, \texttt{deterministic=True}, applied after model loading\\
Rectified flow shift & 5\\
\bottomrule
\end{tabularx}
\end{table}
\ifhmode\par\fi\endgroup

\paragraph[\kern0pt Depth-to-control mapping.]{{Depth-to-control mapping.}}
\begingroup{}
To turn the simulator's depth images into renderer inputs, we apply a fixed mapping from depth to grayscale intensity.
For depth $d$ in meters, the control value is
\[
I(d)=\operatorname{round}\!\left[255\left(1-\operatorname{clip}\!\left(\frac{d-0.15}{0.75},0,1\right)\right)\right].
\]
This mapping gives nearby surfaces brighter intensities and distant surfaces darker intensities, using the same limits for every frame.
Nonpositive and nonfinite depths map to black, and we repeat the grayscale values across 3 channels.
We resize the depth controls to $736\times552$ with bilinear antialiasing, then center-crop to $736\times544$.
We resize the generated RGB to $640\times480$ with Pillow bilinear filtering for the policy.
\ifhmode\par\fi\endgroup

\paragraph[\kern0pt Task completion.]{{Task completion.}}
\begingroup{}
\Cref{tab::transfer_isaac_success} gives the exact simulator-state thresholds used to determine task completion.
All conditions must hold at the same time, using the current apple and plate poses.
In the table, $z_a$, $z_p$, and $z_s$ denote the apple-root, plate-root, and tabletop heights, respectively.
Hand-release distances run from the apple root to the left and right thumb task links, \texttt{left\_hand\_thumb\_2\_link} and \texttt{right\_hand\_thumb\_2\_link}.
The episode ends as soon as the complete predicate is satisfied, with no additional interval during which success must be maintained.
\ifhmode\par\fi\endgroup

\begingroup{}
\begin{table}[!tb]
\centering\reporttablestyle
\caption{\textbf{Simulator-state success conditions for apple pick-and-place.}
The evaluator requires placement, support, contact, settling, and hand release simultaneously.
Distances and heights are in meters.}
\label{tab::transfer_isaac_success}
\begin{tabularx}{\linewidth}{L{3.4cm}Y}
\toprule
\textbf{Condition} & \textbf{Threshold}\\
\midrule
Horizontal placement & Apple--plate root distance in the horizontal plane $\leq 0.08932$\\
Relative height & $|z_a-z_p|\leq 0.1028$ and $z_a-z_p>-0.01$\\
Table support & $z_a,z_p>z_s-0.05$ and $z_a<z_s+0.10$\\
Apple--plate contact & Contact force $>0.44$ N\\
Settling & Apple linear speed $<0.1$ m/s\\
Hand release & Each thumb task link is more than $0.10$ m from the apple root\\
\bottomrule
\end{tabularx}
\end{table}
\ifhmode\par\fi\endgroup

\paragraph[\kern0pt Policy fine-tuning.]{{Policy fine-tuning.}}
\begingroup{}
The policy-training record lists 402 real demonstrations for fine-tuning GR00T N1.7~\citep{nvidia_gr00t_n17_2026}, comprising 90,710 frames at 15 Hz and 84,680 eligible action-chunk start timesteps.
During a single 40,000-step run, we train the projector and action model while keeping the vision and language backbones frozen.
All six evaluated checkpoints come from this run.
Policy fine-tuning uses only the real demonstrations.
\ifhmode\par\fi\endgroup

\begingroup{}
Policy fine-tuning runs on 1 GB300 in BF16, using AdamW with a global batch size of 96, weight decay $10^{-5}$, and gradient clipping at 1.
The learning rate reaches $10^{-4}$ after a 2,000-step warmup and then follows a cosine schedule to zero at step 40,000.
Training applies state dropout of 0.2 and color jitter with brightness, contrast, saturation, and hue settings of 0.3, 0.4, 0.5, and 0.08, respectively.
\ifhmode\par\fi\endgroup

\paragraph[\kern0pt Appearance calibration.]{{Appearance calibration.}}
\begingroup{}
We encode the appearance prompts as compact structured JSON captions that also instruct the renderer to preserve the geometry and motion supplied by depth.
The calibrated baseline requests a yellow/orange/red apple, a shallow white plate, a charcoal fabric-covered table and background, silver robot wrists, black fingers, and diffuse neutral lighting.
Starting from this baseline, the three appearance variants request a green apple, a red plate, or a green apple together with a red plate and white tablecloth.
The combined variant keeps the charcoal background.
We refine the baseline prompt against real demonstration images using the appearance evaluation's calibration starts and 15,000-step policy.
\ifhmode\par\fi\endgroup

\paragraph[\kern0pt Paired outcomes across policy checkpoints.]{{Paired outcomes across policy checkpoints.}}
\begingroup{}
\Cref{tab::transfer_isaac_paired_uncertainty} supplements the aggregate completion rates with the joint outcome of each matched start.
It separates starts on which both renderers lead to success, only Isaac RTX does, only Cosmos3-Nano-Sim-Transfer does, or both fail.
All six policy checkpoints use the same baseline appearance and calibration starts used during prompt refinement.
\ifhmode\par\fi\endgroup

\begingroup{}
\begin{table}[!tb]
\centering\reporttablestyle
\caption{\textbf{Paired apple-task outcomes across policy checkpoints.}
Each policy is evaluated on 32 matched calibration starts under the baseline appearance, and the four outcome counts sum to 32.}
\label{tab::transfer_isaac_paired_uncertainty}
\begin{tabular}{lrrrr}
\toprule
\textbf{Policy steps} & \textbf{Both succeed} & \textbf{RTX only} & \textbf{Transfer only} & \textbf{Both fail}\\
\midrule
5,000 & 13 & 4 & 7 & 8\\
10,000 & 10 & 6 & 2 & 14\\
15,000 & 20 & 2 & 7 & 3\\
20,000 & 17 & 7 & 4 & 4\\
30,000 & 13 & 5 & 5 & 9\\
40,000 & 13 & 4 & 6 & 9\\
\bottomrule
\end{tabular}
\end{table}
\ifhmode\par\fi\endgroup

%% file: appendix/transfer_stage_comparison.tex
\paragraph[\kern0pt Matched training-stage comparisons.]{{Matched training-stage comparisons.}}
\begingroup{}
The aggregate scores summarize benchmark quality and control fidelity, while matched visual examples show how each training stage renders the same controlled scene.
\Cref{fig::transfer_controls,fig::transfer_controls_seg,fig::transfer_controls_edge,fig::transfer_controls_blur} use example 2 of the model's sample data to compare the bidirectional teacher, TF, TF-dCM, and SF-DMD under depth, segmentation, edge, and blur controls.
For each modality, we prepare the control from the same source video and pair it with the same structured caption.
All four stages use seed 2025 with no RGB conditioning frames, so each model generates its appearance from the supplied language and controls.

We use the selected checkpoints and stage-specific samplers described in \cref{sec::transfer_results}, with control guidance fixed at 1 for all four stages.
All causal stages use full history without attention sinks.
We directly resize the controls and generate 121 frames at $832\times480$ and 30 FPS for every output.
Each grid places the stages at common timestamps---frames 0, 60, and 120---so readers can compare their RGB renderings against the same supplied structure as the scene develops.
\ifhmode\par\fi\endgroup

\begingroup{}
\begin{figure}[!tb]
\centering
\includegraphics[width=\linewidth]{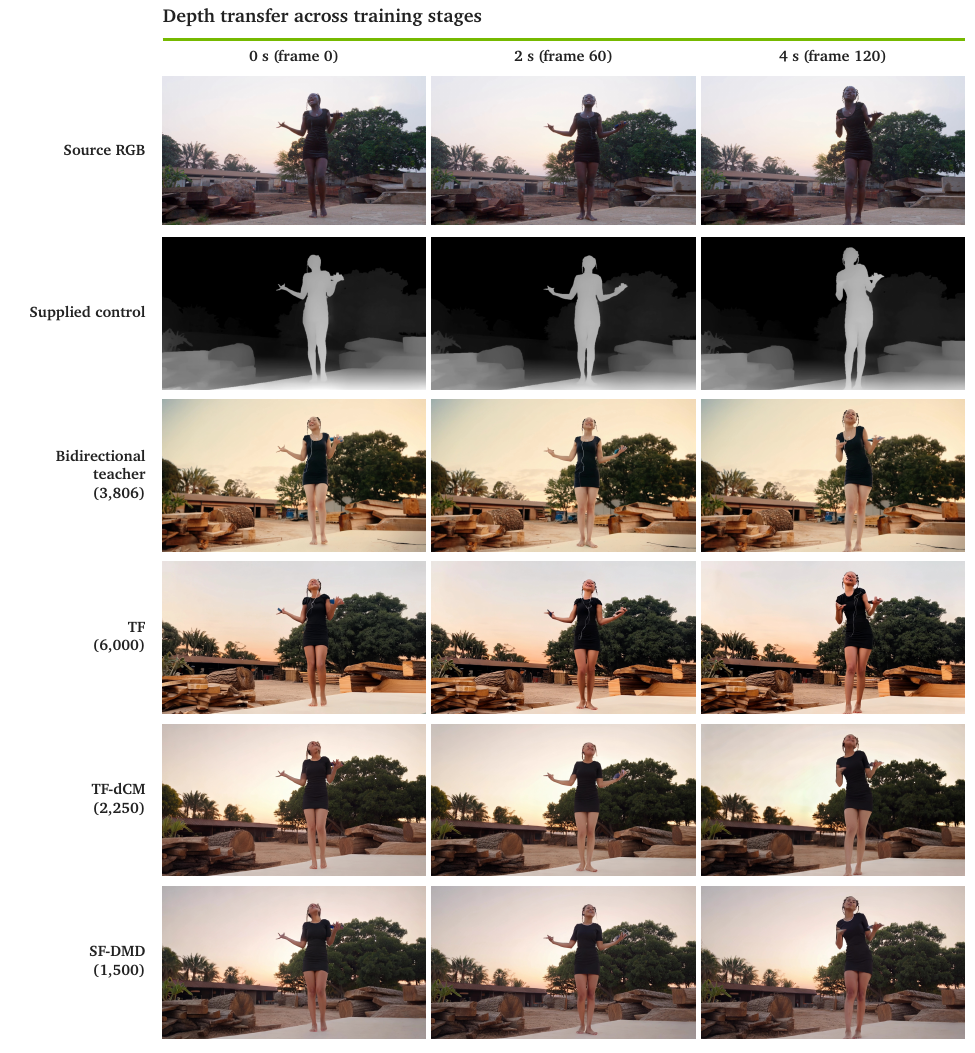}
\caption{\textbf{Matched training-stage comparison with depth control.}
Rows show the source RGB, supplied depth, and outputs from the bidirectional teacher, TF, TF-dCM, and SF-DMD.
All stages share the same example 2 video, prompt, control, and seed.
Columns show frames 0, 60, and 120 (0, 2, and 4 seconds at 30 FPS).}
\label{fig::transfer_controls}
\end{figure}
\ifhmode\par\fi\endgroup

\begingroup{}
\begin{figure}[!tb]
\centering
\includegraphics[width=\linewidth]{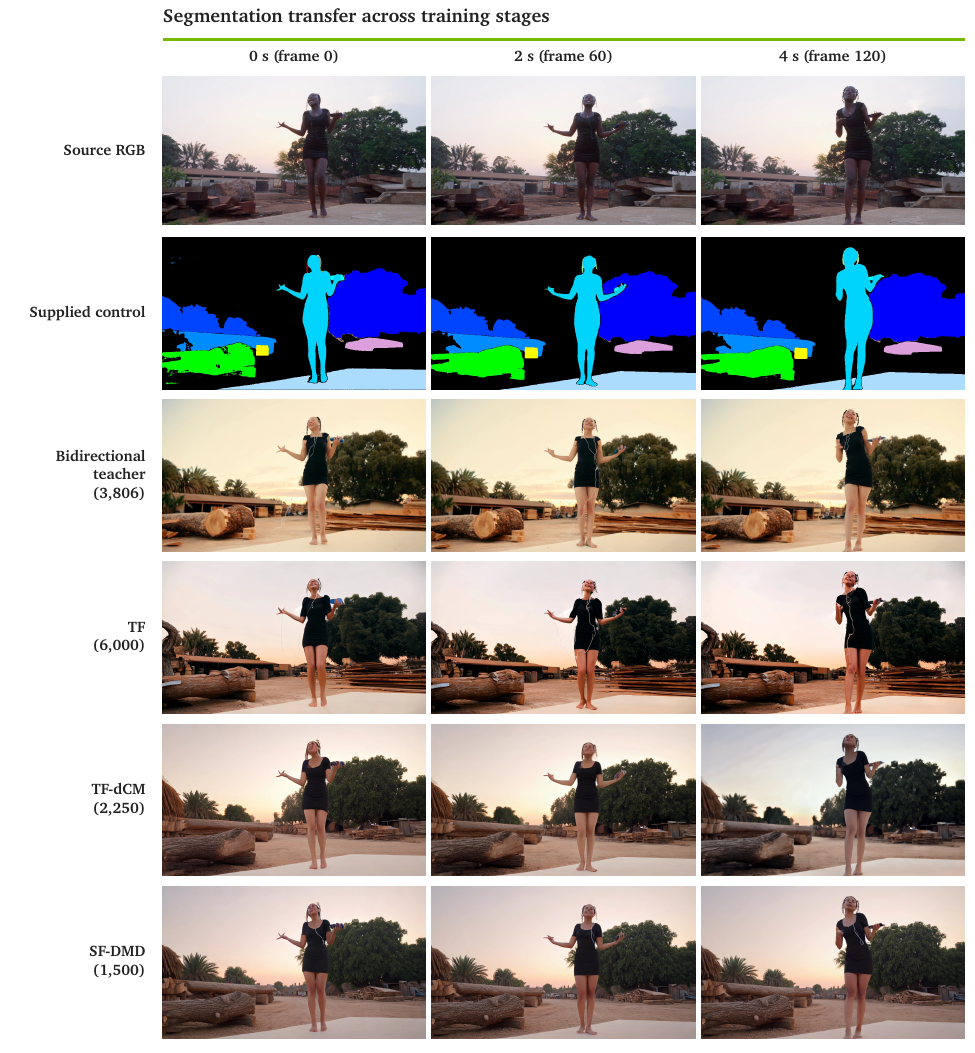}
\caption{\textbf{Matched training-stage comparison with segmentation control.}
Rows show the source RGB, supplied segmentation, and outputs from the bidirectional teacher, TF, TF-dCM, and SF-DMD.
All stages share the same example 2 video, prompt, control, and seed.
Columns show frames 0, 60, and 120 (0, 2, and 4 seconds at 30 FPS).}
\label{fig::transfer_controls_seg}
\end{figure}
\ifhmode\par\fi\endgroup

\begingroup{}
\begin{figure}[!tb]
\centering
\includegraphics[width=\linewidth]{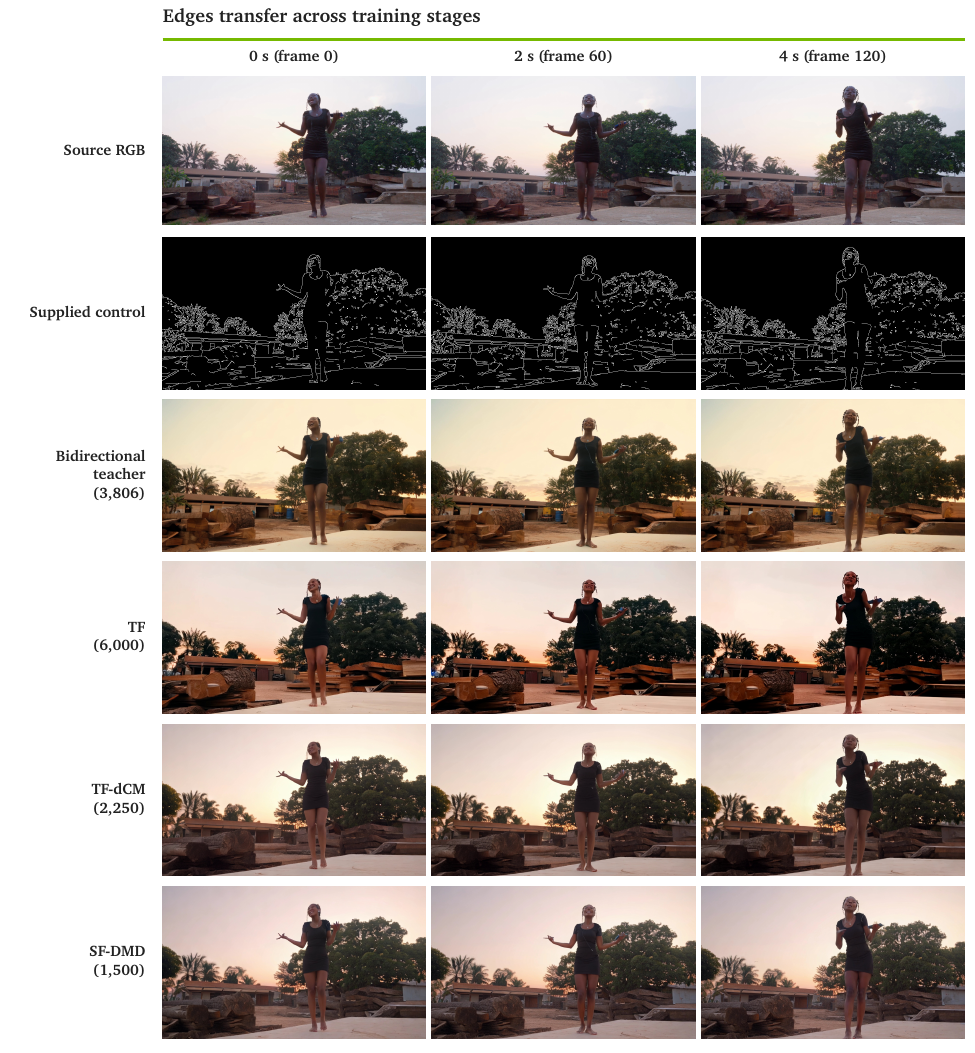}
\caption{\textbf{Matched training-stage comparison with edge control.}
Rows show the source RGB, supplied edges, and outputs from the bidirectional teacher, TF, TF-dCM, and SF-DMD.
Edges are computed from the resized source RGB with the same extraction preset for all stages.
Columns show frames 0, 60, and 120 (0, 2, and 4 seconds at 30 FPS).}
\label{fig::transfer_controls_edge}
\end{figure}
\ifhmode\par\fi\endgroup

\begingroup{}
\begin{figure}[!tb]
\centering
\includegraphics[width=\linewidth]{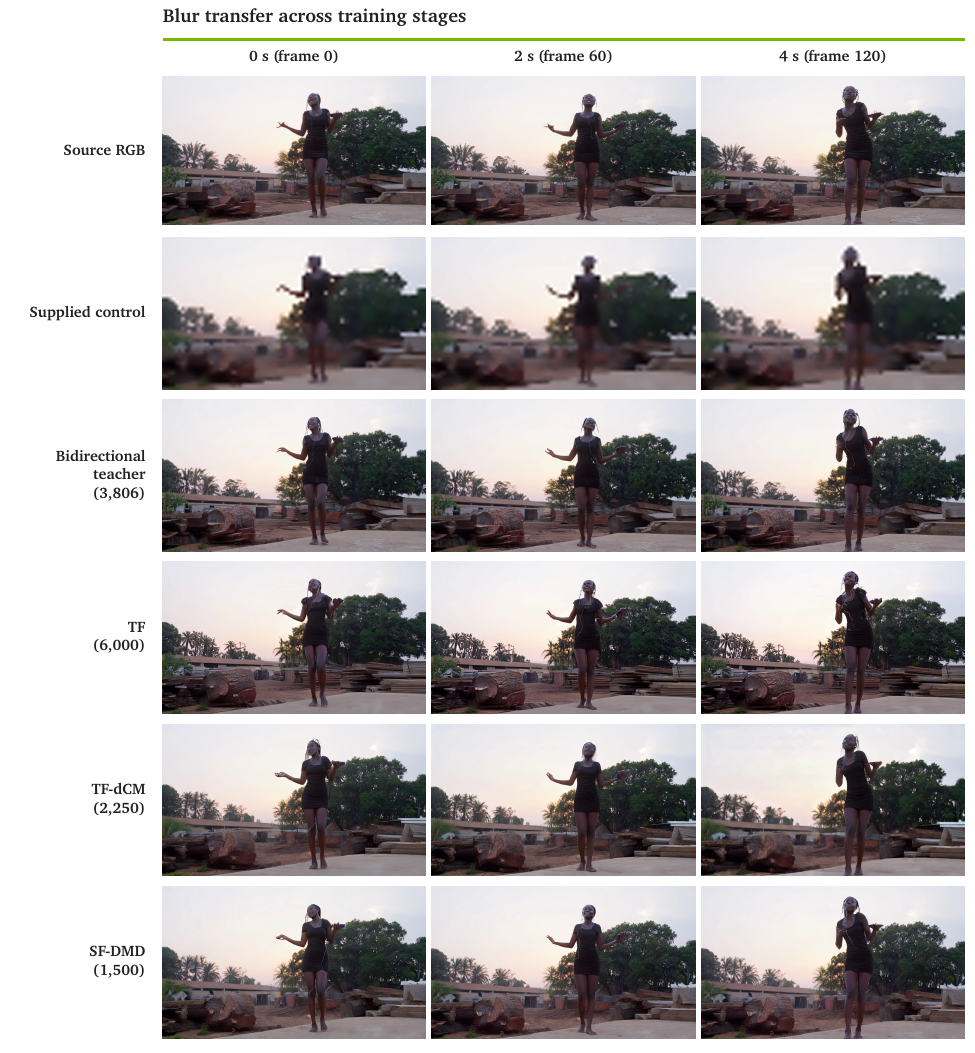}
\caption{\textbf{Matched training-stage comparison with blur control.}
Rows show the source RGB, supplied blur, and outputs from the bidirectional teacher, TF, TF-dCM, and SF-DMD.
Blur controls are computed from the resized source RGB with the same blur preset for all stages.
Columns show frames 0, 60, and 120 (0, 2, and 4 seconds at 30 FPS).}
\label{fig::transfer_controls_blur}
\end{figure}
\ifhmode\par\fi\endgroup

%% file: appendix/transfer_failures.tex
\paragraph[\kern0pt Failure patterns.]{{Failure patterns.}}
\begingroup{}
The QA recordings and saved VisDex rollouts illustrate five failure patterns in Cosmos3-Nano-Sim-Transfer: temporal degradation, object hallucination, appearance drift, geometric deformation, and errors in following the supplied controls.
The examples cover all four control modalities, with source RGB, supplied control, and generated observations shown side by side.
All runs use 4 denoising steps, language and control guidance 1, and flow shift 5.
The edge and blur examples generate $832\times480$ video with seed 42, while the depth and segmentation examples generate $640\times640$ video with seed 2025.
Each caption specifies the recording length and history settings.
\ifhmode\par\fi\endgroup

\paragraph[\kern0pt Temporal degradation and visual artifacts.]{{Temporal degradation and visual artifacts.}}
\begingroup{}
The recorded long rollouts exhibit colored spots, ghosting, and loss of structural detail.
\Cref{fig::transfer_failure_temporal} compares an earlier and a later moment in the same robot sequence.
Near the end of the rollout, the generated arms and table appear smeared even though their outlines remain distinct in the source RGB and edge controls.
The QA review reports partial mitigation after changes to the history settings, with artifacts remaining in the tested cases.
\ifhmode\par\fi\endgroup

\begingroup{}
\begin{figure}[!htb]
\centering
\input{figures/transfer_failures/temporal.tex}
\caption{\textbf{Temporal degradation in Cosmos3-Nano-Sim-Transfer.}
Matched source, edge-control, and generated frames at 4.00 and 26.23~s show late-rollout ghosting around the robot arms and across the table.
The recorded generation contains 788 frames at 30 FPS and uses a 198-latent-frame K/V window with 3 sink frames.
These settings differ from the fixed-window VisDex runtime evaluation.}
\label{fig::transfer_failure_temporal}
\end{figure}
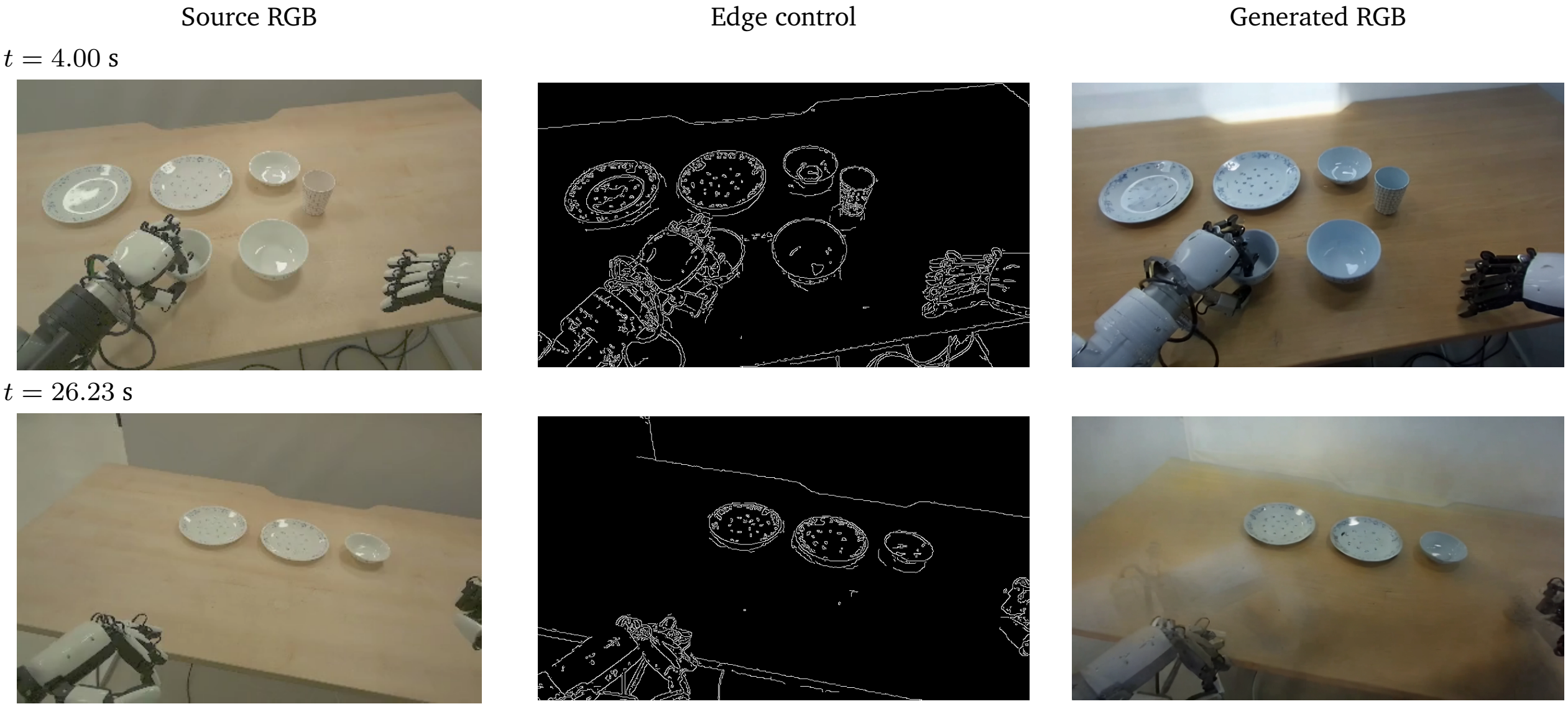
\ifhmode\par\fi\endgroup

\paragraph[\kern0pt Object hallucination and duplication.]{{Object hallucination and duplication.}}
\begingroup{}
Another failure changes which objects appear in the scene.
In the warehouse example in \cref{fig::transfer_failure_objects}, the source RGB and edge control show open floor below a pallet.
The generated observation fills this region with additional boxes and a supporting structure.
The QA review also reports pedestrian duplication in generated videos.
\ifhmode\par\fi\endgroup

\begingroup{}
\begin{figure}[!htb]
\centering
\input{figures/transfer_failures/objects.tex}
\caption{\textbf{Object hallucination in Cosmos3-Nano-Sim-Transfer.}
At 4.17~s, the highlighted floor region below the pallet contains additional boxes and shelf-like structure in the generated observation.
The source and edge control do not contain these objects.
This 122-frame, 29-FPS example uses a 32-latent-frame K/V window and 3 sink frames.}
\label{fig::transfer_failure_objects}
\end{figure}
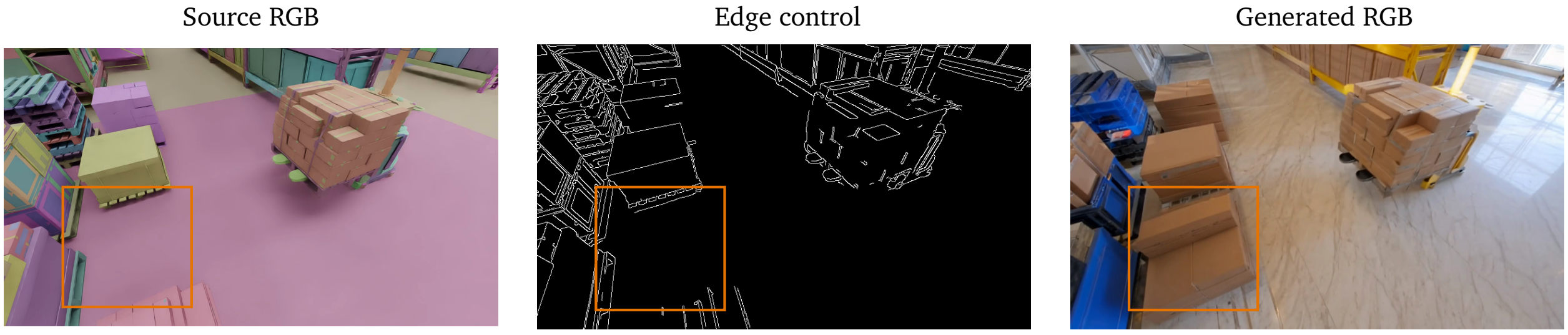
\ifhmode\par\fi\endgroup

\paragraph[\kern0pt Identity and appearance drift.]{{Identity and appearance drift.}}
\begingroup{}
In \cref{fig::transfer_failure_identity}, the generated red vehicle has inconsistent body proportions and front--rear appearance across the three displayed frames of a passing maneuver.
The vehicle is visible in all three frames.
The QA review also reports offscreen appearance drift that persisted with an explicit appearance-preservation caption.
\ifhmode\par\fi\endgroup

\input{figures/transfer_failures/identity_figure.tex}

\paragraph[\kern0pt Deformation and unrealistic structure.]{{Deformation and unrealistic structure.}}
\begingroup{}
The blur-conditioned example in \cref{fig::transfer_failure_deformation} preserves the bowl's location while changing its internal geometry.
The generated inner bowl is smaller relative to the outer bowl, and the spacing between their rims differs from the source.
This changes the proportions of the nested bowls even though the surrounding tabletop remains recognizable.
\ifhmode\par\fi\endgroup

\begingroup{}
\begin{figure}[!htb]
\centering
\input{figures/transfer_failures/deformation.tex}
\caption{\textbf{Bowl deformation under blur conditioning.}
Matched crops at 10.00~s show a smaller inner bowl and altered rim spacing in the generated observation compared with the source.
The supplied control uses the medium blur preset.
This 800-frame, 30-FPS recording uses a 201-latent-frame K/V window and 3 sink frames.}
\label{fig::transfer_failure_deformation}
\end{figure}
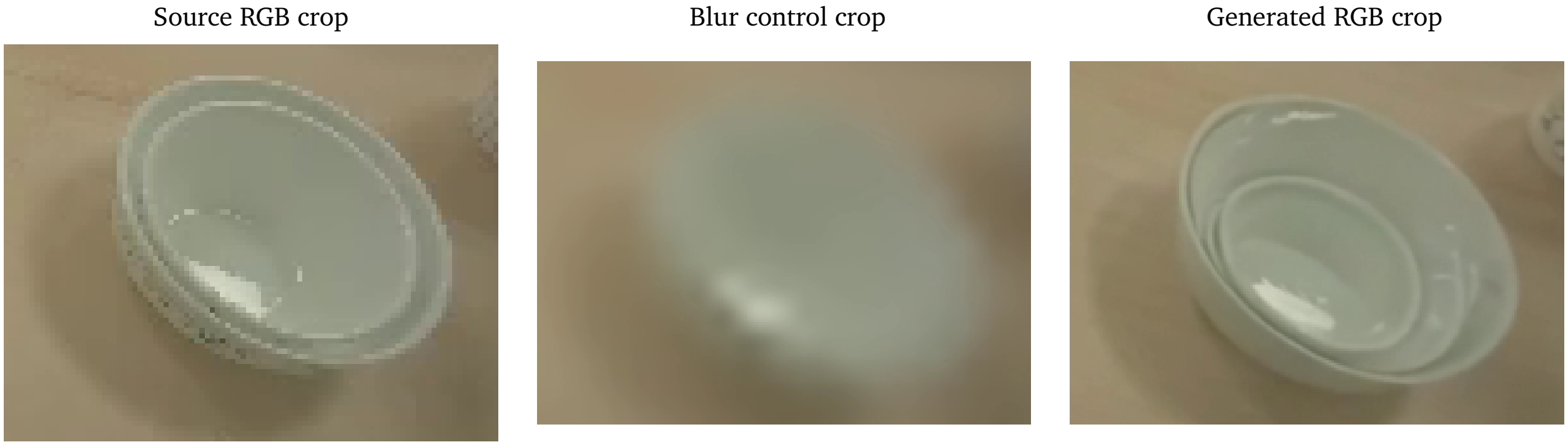
\ifhmode\par\fi\endgroup

\paragraph[\kern0pt Object motion and control following.]{{Object motion and control following.}}
\begingroup{}
The depth and segmentation examples show disagreement between generated object positions and the supplied controls.
The prompts describe a shoe and a blue turtle on the counter.
In both rows of \cref{fig::transfer_failure_orientation}, the source RGB shows the target in the robot's raised hand and no target on the table.
The generated observation instead places the prompted object on the table.
\ifhmode\par\fi\endgroup

\begingroup{}
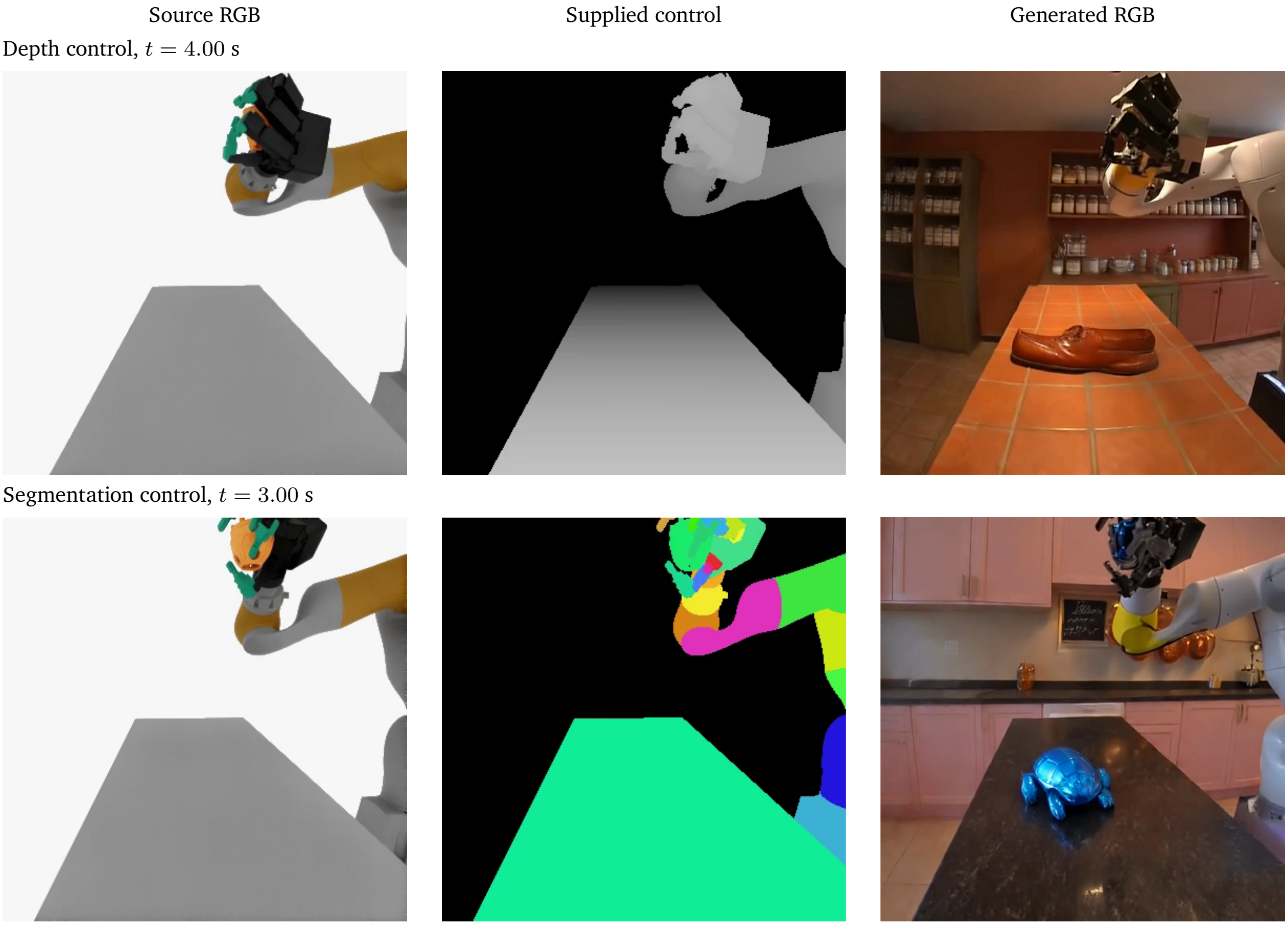
\begin{figure}[!htb]
\centering
\input{figures/transfer_failures/orientation.tex}
\caption{\textbf{Object-motion errors under depth and segmentation conditioning.}
The depth example at 4.00~s and the segmentation example at 3.00~s show the simulator holding the target above an empty table.
The prompts describe the shoe and turtle on the counter, and the generated observations place them there despite the lifted target in the source.
Both examples come from 600-frame, 30-FPS VisDex rollouts using the iteration-1,500 SF-DMD checkpoint, a 30-latent-frame K/V window, and 3 sink frames.}
\label{fig::transfer_failure_orientation}
\end{figure}
\ifhmode\par\fi\endgroup

%% file: figures/transfer_failures/temporal.tex
\begingroup
\footnotesize
\makebox[0.315\linewidth]{Source RGB}\hfill
\makebox[0.315\linewidth]{Edge control}\hfill
\makebox[0.315\linewidth]{Generated RGB}\par\smallskip
\makebox[\linewidth][l]{$t=4.00$ s}\par\smallskip
\begin{minipage}[c]{0.315\linewidth}\centering
\includegraphics[width=\linewidth,height=0.59\linewidth,keepaspectratio]{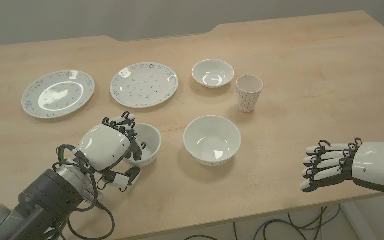}
\end{minipage}\hfill
\begin{minipage}[c]{0.315\linewidth}\centering
\includegraphics[width=\linewidth,height=0.59\linewidth,keepaspectratio]{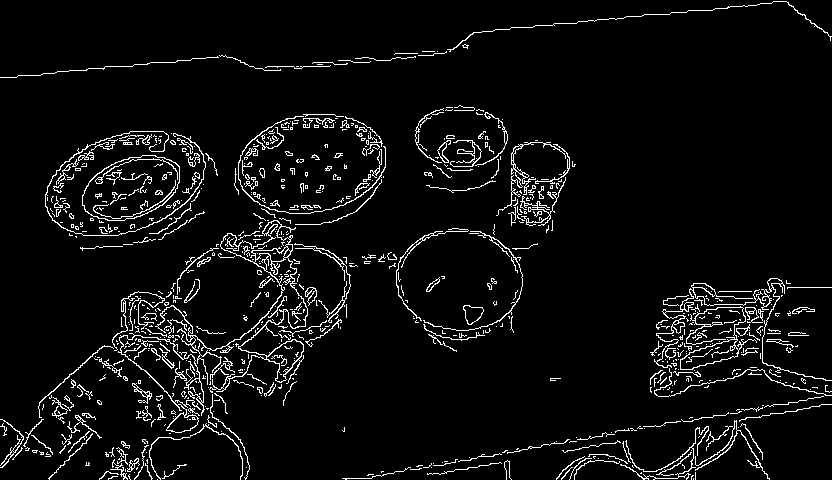}
\end{minipage}\hfill
\begin{minipage}[c]{0.315\linewidth}\centering
\includegraphics[width=\linewidth,height=0.59\linewidth,keepaspectratio]{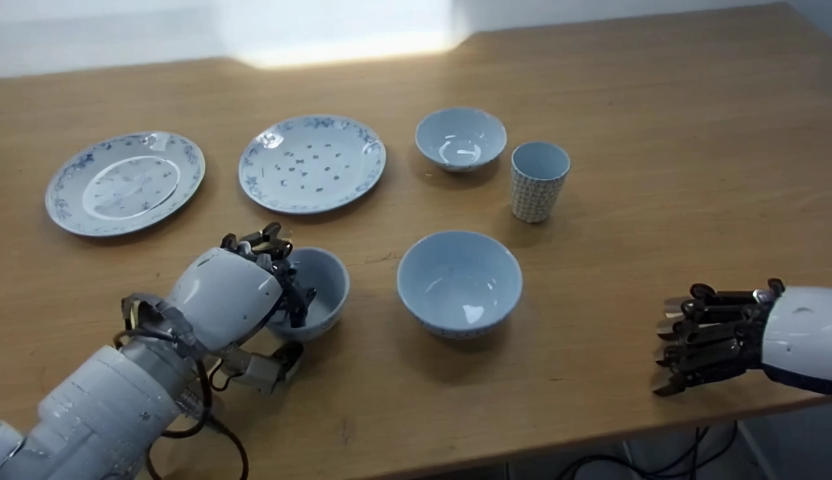}
\end{minipage}\par\smallskip
\makebox[\linewidth][l]{$t=26.23$ s}\par\smallskip
\begin{minipage}[c]{0.315\linewidth}\centering
\includegraphics[width=\linewidth,height=0.59\linewidth,keepaspectratio]{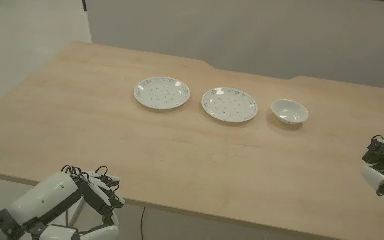}
\end{minipage}\hfill
\begin{minipage}[c]{0.315\linewidth}\centering
\includegraphics[width=\linewidth,height=0.59\linewidth,keepaspectratio]{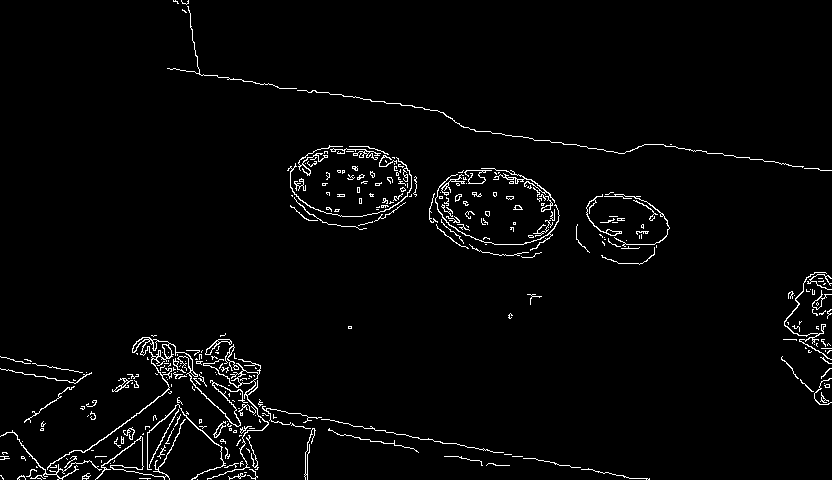}
\end{minipage}\hfill
\begin{minipage}[c]{0.315\linewidth}\centering
\includegraphics[width=\linewidth,height=0.59\linewidth,keepaspectratio]{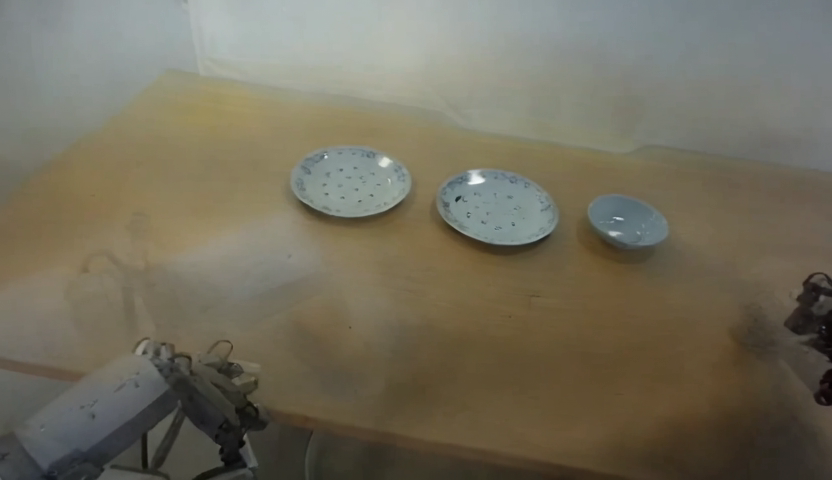}
\end{minipage}\par\smallskip
\endgroup

%% file: figures/transfer_failures/objects.tex
\begingroup
\footnotesize
\makebox[0.315\linewidth]{Source RGB}\hfill
\makebox[0.315\linewidth]{Edge control}\hfill
\makebox[0.315\linewidth]{Generated RGB}\par\smallskip
\begin{minipage}[c]{0.315\linewidth}\centering
\begin{tikzpicture}
\node[inner sep=0,anchor=south west] (im) {\includegraphics[width=\linewidth,height=0.59\linewidth,keepaspectratio]{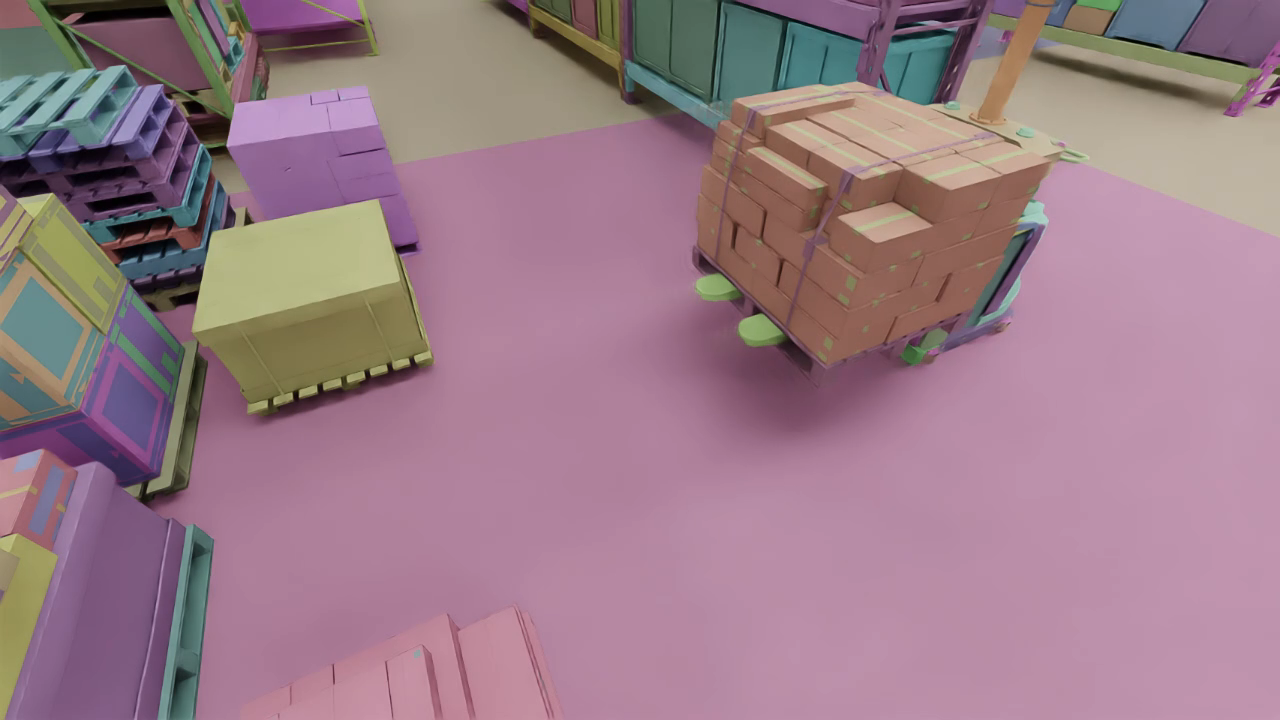}};
\begin{scope}[x={(im.south east)},y={(im.north west)}]
\draw[orange!90!black,line width=0.8pt] (0.12,0.07) rectangle (0.38,0.5);
\end{scope}
\end{tikzpicture}
\end{minipage}\hfill
\begin{minipage}[c]{0.315\linewidth}\centering
\begin{tikzpicture}
\node[inner sep=0,anchor=south west] (im) {\includegraphics[width=\linewidth,height=0.59\linewidth,keepaspectratio]{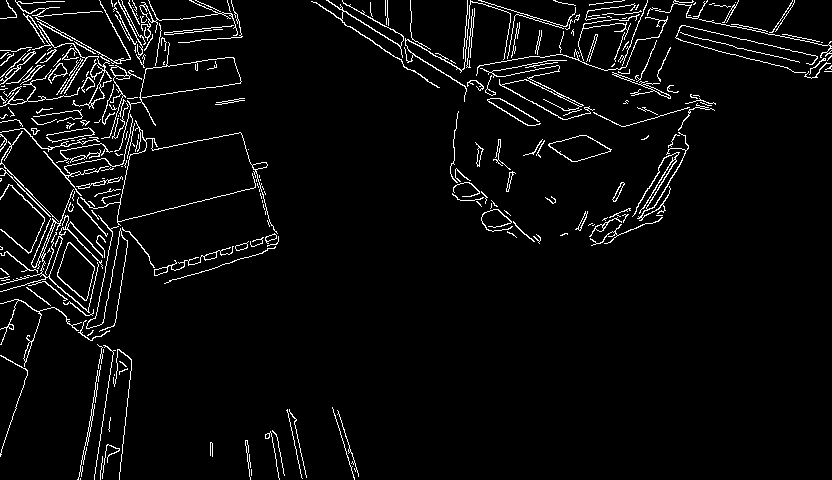}};
\begin{scope}[x={(im.south east)},y={(im.north west)}]
\draw[orange!90!black,line width=0.8pt] (0.12,0.07) rectangle (0.38,0.5);
\end{scope}
\end{tikzpicture}
\end{minipage}\hfill
\begin{minipage}[c]{0.315\linewidth}\centering
\begin{tikzpicture}
\node[inner sep=0,anchor=south west] (im) {\includegraphics[width=\linewidth,height=0.59\linewidth,keepaspectratio]{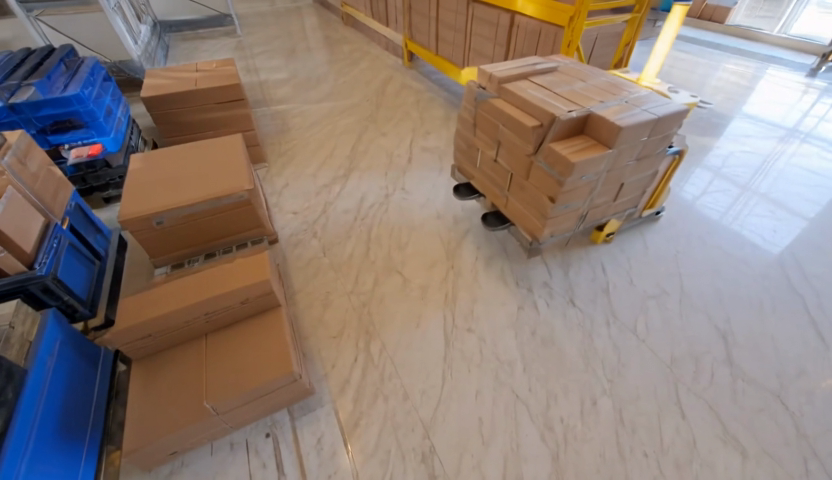}};
\begin{scope}[x={(im.south east)},y={(im.north west)}]
\draw[orange!90!black,line width=0.8pt] (0.12,0.07) rectangle (0.38,0.5);
\end{scope}
\end{tikzpicture}
\end{minipage}
\endgroup

%% file: figures/transfer_failures/identity_figure.tex
\begingroup{}
\begin{figure}[!htb]
\centering
\input{figures/transfer_failures/identity.tex}
\caption{\textbf{Appearance inconsistency in Cosmos3-Nano-Sim-Transfer.}
Matched crops show inconsistent body proportions and front--rear interpretation of the generated red vehicle during a passing maneuver.
The vehicle is visible in all three illustrated frames.
The 121-frame, 30-FPS generation uses a 31-latent-frame K/V window and 3 sink frames.}
\label{fig::transfer_failure_identity}
\end{figure}
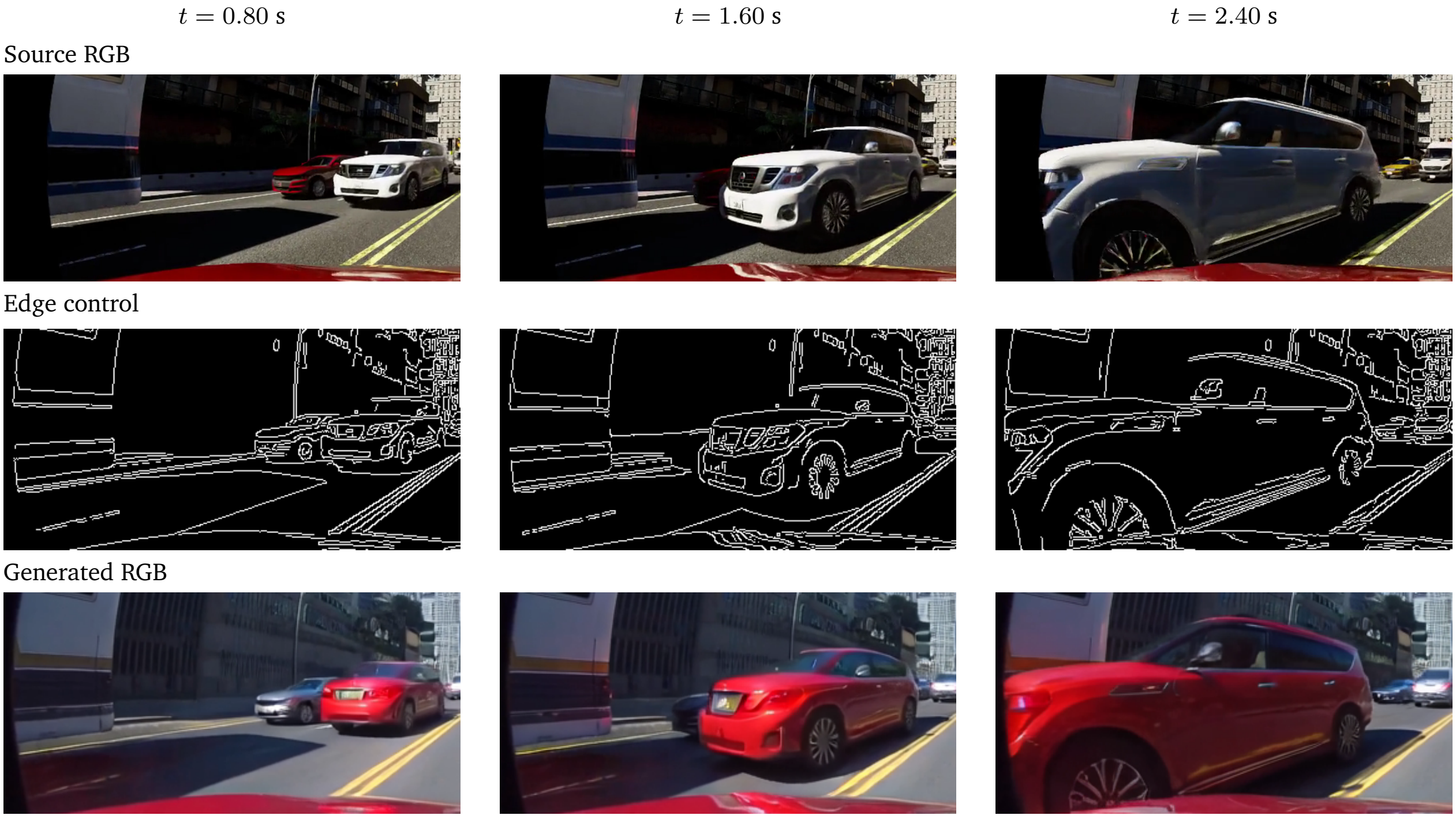
\ifhmode\par\fi\endgroup

%% file: figures/transfer_failures/identity.tex
\begingroup
\footnotesize
\makebox[0.315\linewidth]{$t=0.80$ s}\hfill
\makebox[0.315\linewidth]{$t=1.60$ s}\hfill
\makebox[0.315\linewidth]{$t=2.40$ s}\par\smallskip
\makebox[\linewidth][l]{Source RGB}\par\smallskip
\begin{minipage}[c]{0.315\linewidth}\centering
\includegraphics[width=\linewidth,height=0.51\linewidth,keepaspectratio]{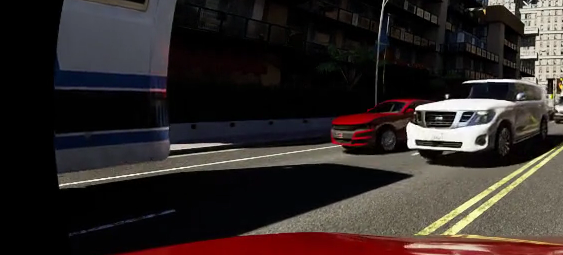}
\end{minipage}\hfill
\begin{minipage}[c]{0.315\linewidth}\centering
\includegraphics[width=\linewidth,height=0.51\linewidth,keepaspectratio]{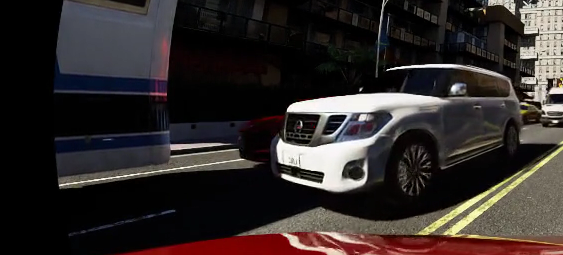}
\end{minipage}\hfill
\begin{minipage}[c]{0.315\linewidth}\centering
\includegraphics[width=\linewidth,height=0.51\linewidth,keepaspectratio]{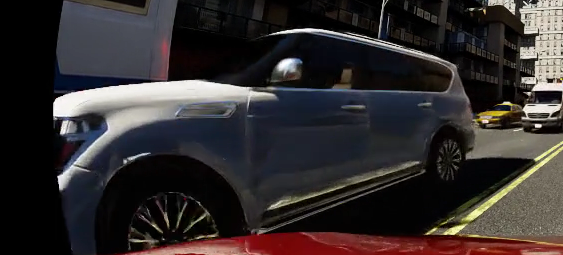}
\end{minipage}\par\smallskip
\makebox[\linewidth][l]{Edge control}\par\smallskip
\begin{minipage}[c]{0.315\linewidth}\centering
\includegraphics[width=\linewidth,height=0.51\linewidth,keepaspectratio]{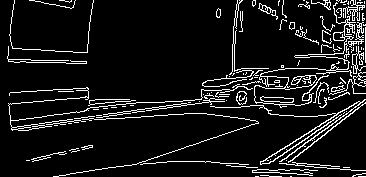}
\end{minipage}\hfill
\begin{minipage}[c]{0.315\linewidth}\centering
\includegraphics[width=\linewidth,height=0.51\linewidth,keepaspectratio]{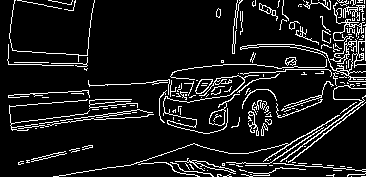}
\end{minipage}\hfill
\begin{minipage}[c]{0.315\linewidth}\centering
\includegraphics[width=\linewidth,height=0.51\linewidth,keepaspectratio]{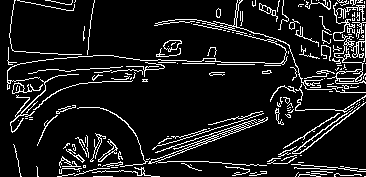}
\end{minipage}\par\smallskip
\makebox[\linewidth][l]{Generated RGB}\par\smallskip
\begin{minipage}[c]{0.315\linewidth}\centering
\includegraphics[width=\linewidth,height=0.51\linewidth,keepaspectratio]{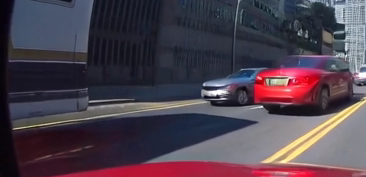}
\end{minipage}\hfill
\begin{minipage}[c]{0.315\linewidth}\centering
\includegraphics[width=\linewidth,height=0.51\linewidth,keepaspectratio]{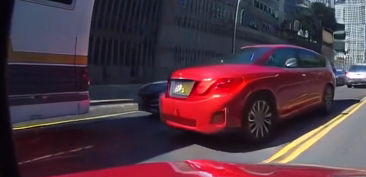}
\end{minipage}\hfill
\begin{minipage}[c]{0.315\linewidth}\centering
\includegraphics[width=\linewidth,height=0.51\linewidth,keepaspectratio]{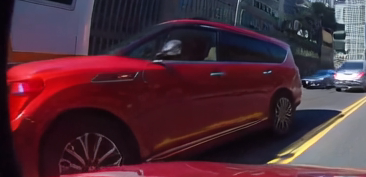}
\end{minipage}\par\smallskip
\endgroup

%% file: figures/transfer_failures/deformation.tex
\begingroup
\footnotesize
\makebox[0.315\linewidth]{Source RGB crop}\hfill
\makebox[0.315\linewidth]{Blur control crop}\hfill
\makebox[0.315\linewidth]{Generated RGB crop}\par\smallskip
\begin{minipage}[c]{0.315\linewidth}\centering
\includegraphics[width=\linewidth,height=0.82\linewidth,keepaspectratio]{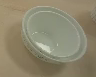}
\end{minipage}\hfill
\begin{minipage}[c]{0.315\linewidth}\centering
\includegraphics[width=\linewidth,height=0.82\linewidth,keepaspectratio]{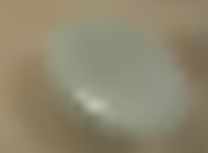}
\end{minipage}\hfill
\begin{minipage}[c]{0.315\linewidth}\centering
\includegraphics[width=\linewidth,height=0.82\linewidth,keepaspectratio]{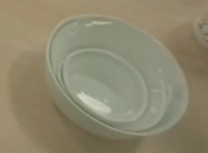}
\end{minipage}
\endgroup

%% file: figures/transfer_failures/orientation.tex
\begingroup
\footnotesize
\makebox[0.315\linewidth]{Source RGB}\hfill
\makebox[0.315\linewidth]{Supplied control}\hfill
\makebox[0.315\linewidth]{Generated RGB}\par\smallskip
\makebox[\linewidth][l]{Depth control, $t=4.00$ s}\par\smallskip
\begin{minipage}[c]{0.315\linewidth}\centering
\includegraphics[width=\linewidth]{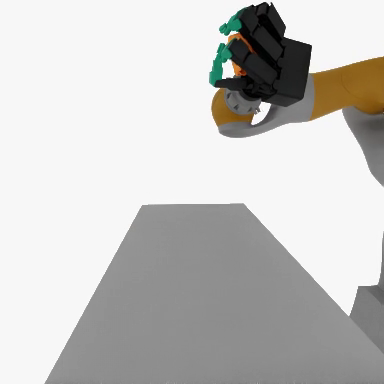}
\end{minipage}\hfill
\begin{minipage}[c]{0.315\linewidth}\centering
\includegraphics[width=\linewidth]{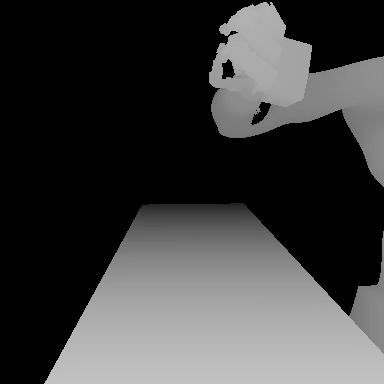}
\end{minipage}\hfill
\begin{minipage}[c]{0.315\linewidth}\centering
\includegraphics[width=\linewidth]{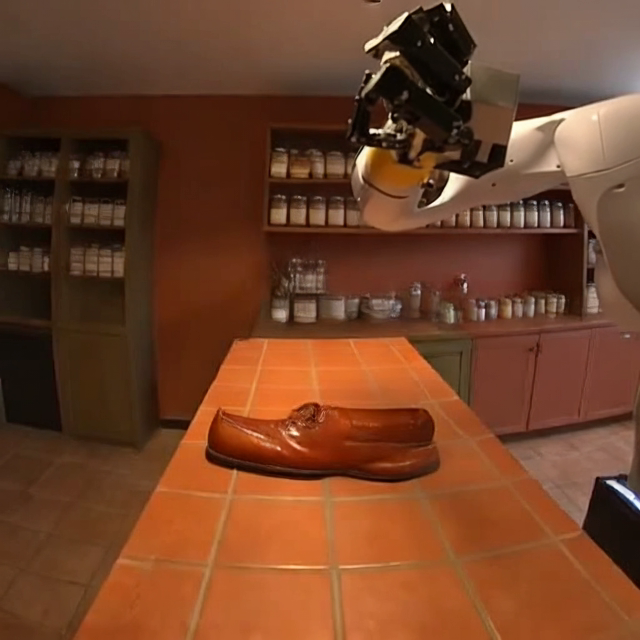}
\end{minipage}\par\smallskip
\makebox[\linewidth][l]{Segmentation control, $t=3.00$ s}\par\smallskip
\begin{minipage}[c]{0.315\linewidth}\centering
\includegraphics[width=\linewidth]{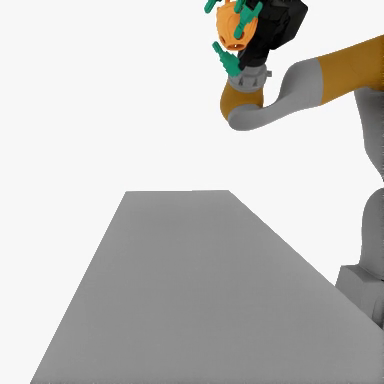}
\end{minipage}\hfill
\begin{minipage}[c]{0.315\linewidth}\centering
\includegraphics[width=\linewidth]{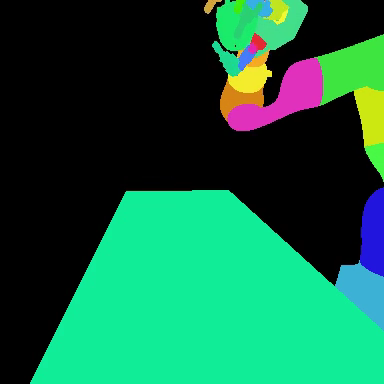}
\end{minipage}\hfill
\begin{minipage}[c]{0.315\linewidth}\centering
\includegraphics[width=\linewidth]{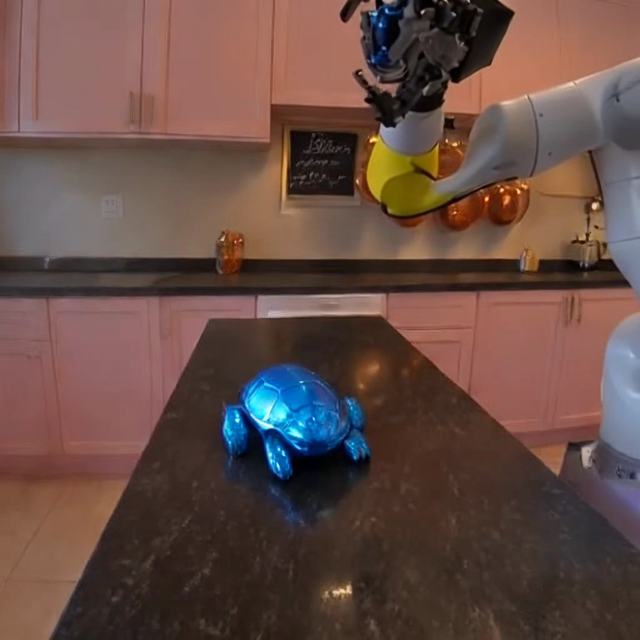}
\end{minipage}\par\smallskip
\endgroup

%% file: appendix/auto_appendix.tex
\FloatBarrier
\section[\kern0pt Cosmos3-Nano-Transfer/Sim-Auto: Additional Details]{{Cosmos3-Nano-Transfer/Sim-Auto: Additional Details}}
\label{sec::auto_supplement}

\subsection[\kern0pt Architecture Details]{{Architecture Details}}
\input{appendix/auto_lidar_tokenizer.tex}

\subsubsection[\kern0pt Attention Visibility and Execution]{{Attention Visibility and Execution}}
\input{appendix/auto_attention_details.tex}

\input{appendix/auto_attention_implementation.tex}

\FloatBarrier
\subsection[\kern0pt Data Preparation]{{Data Preparation}}
\input{appendix/auto_rigs.tex}

\input{appendix/auto_wsm.tex}

\input{appendix/auto_captions.tex}

\subsection[\kern0pt Training Configuration]{{Training Configuration}}
\subsubsection[\kern0pt Cosmos3-Nano-Transfer-Auto Training]{{Cosmos3-Nano-Transfer-Auto Training}}
\input{appendix/auto_stage0_training.tex}
\subsubsection[\kern0pt Cosmos3-Nano-Sim-Auto Training]{{Cosmos3-Nano-Sim-Auto Training}}
\input{appendix/auto_training_configuration.tex}
\input{appendix/auto_causal_training.tex}

\input{appendix/auto_training_infrastructure.tex}

\FloatBarrier
\Needspace{8\baselineskip}
\subsection[\kern0pt Runtime Optimization]{{Runtime Optimization}}
\input{appendix/auto_runtime.tex}
\input{appendix/auto_runtime_results.tex}

\FloatBarrier
\subsection[\kern0pt Inference and Benchmark Settings]{{Inference and Benchmark Settings}}
\input{appendix/auto_inference.tex}
\input{appendix/auto_metrics.tex}

\FloatBarrier
\subsection[\kern0pt Supplementary Results and Ablations]{{Supplementary Results and Ablations}}
\label{sec::auto_appendix_results}
\begingroup{}
We present ablations that inform our final attention design and training recipe, followed by studies of joint sensor generation and weather conditioning.
\ifhmode\par\fi\endgroup

\input{appendix/auto_development.tex}

\input{appendix/auto_lidar_results.tex}

\FloatBarrier

%% file: appendix/auto_lidar_tokenizer.tex
\FloatBarrier

\begingroup{}
\subsubsection{LiDAR Tokenizer}
\label{sec::auto_lidar_tokenizer}
The joint RGB--LiDAR model uses our LiDAR tokenizer, a continuous variational autoencoder with a hierarchical attention encoder and decoder.
Its architecture is inspired by the LiDAR-native tokenizer in LidarVerse~\citep{che2026lidarverse}, particularly its spatial neighborhood attention, causal temporal attention, and preservation of 1 latent time step per sweep.
It maps each sweep to a spatial latent grid while preserving the sequence's temporal resolution.
The tokenizer is trained for reconstruction and then held fixed during joint generative training.
Its reconstruction objective is separate from the RGB and LiDAR diffusion losses used to train the joint model.
\ifhmode\par\fi\endgroup

\begingroup{}
\paragraph{Range representation and normalization.}
Following the calibrated range-view representation used for driving LiDAR generation~\citep{ren2025cosmos}, each Pandar128 sweep is rasterized onto a $128\times3600$ grid.
Rows retain the calibrated beam elevations, and columns index azimuth.
We reduce the width to 1800 by selecting the nearest return within each pair of adjacent columns and retaining that return's intensity.
Four columns from the opposite azimuth boundary are appended at each side, yielding a $128\times1808$ network canvas.
This circular padding makes the width divisible by 16 without rescaling the 1800 measured angular bins.
The padded columns are cropped before reconstruction losses and point-cloud evaluation.

Each input has three channels: metric range $r$, unit-scaled intensity $i$, and binary validity $m$.
The valid range interval is $[5,100]$\,m.
Within the network, range and intensity are transformed as
\begin{equation}
 r_{\text{n}}=2\frac{r-5}{100-5}-1,
 \qquad
 i_{\text{n}}=2i-1.
 \label{eq::auto_lidar_normalization}
\end{equation}
Missing returns and ranges outside the interval receive $m=0$ and a value of $-1$ in both data channels.
The explicit validity channel distinguishes a missing return from a valid measurement at a normalization boundary.
\ifhmode\par\fi\endgroup

\begingroup{}
\paragraph{Encoder and decoder.}
The encoder applies a $2\times2$ patch embedding independently to each sweep, followed by three $2\times2$ spatial merges.
The four stages have channel widths $(128,256,512,1024)$, depths $(3,3,3,3)$, and attention-head counts $(4,4,8,8)$.
\Cref{tab::auto_lidar_tokenizer_stages} lists the corresponding feature grids.
At the first three stages, spatial mixing uses neighborhood attention~\citep{natten} with a $5\times45$ window, unit dilation, and circular padding along azimuth.
The first stage has no temporal attention.
At the next two stages, each spatial block is followed by causal temporal attention at each spatial position.
The final stage uses three joint spatiotemporal attention blocks over the $T\times8\times113$ bottleneck grid, with three-axis rotary positional encoding and a causal mask along time.
The feed-forward expansion ratio is 3.
Learned spatial embeddings are added at the encoder input and decoder bottleneck, and local spatial attention also uses axial rotary coordinates.

The posterior has 128 latent channels, with separate predicted means and log variances.
The decoder projects a latent slice to 1024 channels and mirrors the hierarchy through joint bottleneck attention, three spatial expansions, and a final $2\times2$ unpatchification.
At intermediate decoder resolutions, temporal attention precedes spatial attention.
The output contains normalized range, normalized intensity, and a validity logit.
After inverse normalization, the default decoding rule retains rays with sigmoid validity probability at least 0.5 and sets invalid range and intensity to zero.
\ifhmode\par\fi\endgroup

\begingroup{}
\begin{table}[!ht]
\centering\reporttablestyle
\caption{\textbf{Our LiDAR tokenizer encoder.}
The grid is shown per sweep after the input patch embedding or the preceding merge.
Every stage preserves the number of sweeps.
The decoder reverses the spatial hierarchy, with temporal attention before spatial attention at the intermediate stages.}
\label{tab::auto_lidar_tokenizer_stages}
\begin{tabularx}{\linewidth}{lcccY}
\toprule
\textbf{Stage} & \textbf{Grid} & \textbf{Channels} & \textbf{Heads} & \textbf{Attention blocks}\\
\midrule
1 & $64\times904$ & 128 & 4 & 3 spatial neighborhood blocks\\
2 & $32\times452$ & 256 & 4 & 3 spatial--temporal pairs\\
3 & $16\times226$ & 512 & 8 & 3 spatial--temporal pairs\\
4 & $8\times113$ & 1024 & 8 & 3 joint spatiotemporal blocks\\
Posterior & $8\times113$ & 128 & --- & Mean and log variance per latent coordinate\\
\bottomrule
\end{tabularx}
\end{table}
\ifhmode\par\fi\endgroup

\begingroup{}
\paragraph{Spatial stride and temporal causality.}
For a batch of $B$ clips containing $T$ sweeps, the encoder maps
\begin{equation}
 \mathbb{R}^{B\times3\times T\times128\times1808}
 \longrightarrow
 \mathbb{R}^{B\times128\times T\times8\times113}.
 \label{eq::auto_lidar_tokenizer_shape}
\end{equation}
The effective stride in time, elevation, and azimuth is $(1,16,16)$ relative to the network input.
Including the initial column pooling, the sampling stride is 16 rows and 32 columns relative to the native $128\times3600$ raster.
Attention mixes information beyond these sampling intervals, so they do not bound a latent token's spatial receptive field.
There is no temporal downsampling or upsampling, so each 10-Hz input sweep retains 1 latent time step at a 0.1-second interval.
The joint diffusion model uses $1\times1$ patches on this latent grid, giving $8\cdot113=904$ LiDAR tokens per sweep.
This diffusion patch size is distinct from the tokenizer's $2\times2$ input patches and its accumulated $16\times16$ spatial stride.

Both encoder and decoder restrict temporal attention to keys at times $t'\leq t$ for a query at time $t$.
At the joint bottleneck, all spatial positions at an allowed time are visible, including other positions in the current sweep.
Consequently, encoding sweep $t$ does not require future input sweeps, and decoding its latent does not require future latent slices.
Spatial attention, spatial resampling, and the channel projections operate within individual sweeps and preserve this temporal dependency.

Long clips are processed with temporal key/value caches.
The joint long-clip recipe uses blocks of 20 new sweeps and a context cap of 21 sweeps, leaving one preceding sweep's cached keys and values at a full block boundary.
The cap bounds retained positions, whose features can already contain information from earlier sweeps.
Within each block, the causal mask still prevents an early sweep from reading later sweeps in that block.
These are tokenizer execution settings, separate from the simulation model's 4-sweep generation chunks.
The inference interface uses the posterior mean and applies channelwise standardization with latent means and standard deviations stored in the tokenizer checkpoint.
Decoding reverses this standardization before reconstructing the range view.
\ifhmode\par\fi\endgroup

\begingroup{}
\paragraph{Reconstruction objective.}
Let $j$ index all $N=BTHW$ rays in a batch after removing the circular padding, with $H=128$ and $W=1800$.
Let $\widehat r_{\text{n},j}$ and $\widehat i_{\text{n},j}$ be the normalized reconstructions and $\ell_j$ the predicted validity logit.
Range and intensity use masked absolute errors averaged over all rays:
\begin{align}
 \mathcal{L}_{\text{range}}
 &=\frac{1}{N}\sum_j m_j
   \left|\widehat r_{\text{n},j}-r_{\text{n},j}\right|,\\
 \mathcal{L}_{\text{intensity}}
 &=\frac{1}{N}\sum_j m_j
   \left|\widehat i_{\text{n},j}-i_{\text{n},j}\right|.
 \label{eq::auto_lidar_reconstruction}
\end{align}
The denominator includes missing rays, so these terms scale with the valid-return fraction.
Validity is supervised over all rays using binary cross-entropy with logits,
$\mathcal{L}_{\text{mask}}=N^{-1}\sum_j\operatorname{BCEWithLogits}(\ell_j,m_j)$.
For calibrated unit ray directions $\mathbf d_j$, the point term is
\begin{equation}
 \mathcal{L}_{\text{point}}
 =\frac{1}{3N}\sum_j m_j
   \left\|\widehat{\mathbf p}_j-\mathbf p_j\right\|_1,
 \qquad
 \mathbf p_j=r_j\mathbf d_j.
 \label{eq::auto_lidar_point_loss}
\end{equation}
Point coordinates are in meters and use corresponding rays rather than nearest-neighbor point matching.
In the implemented unprojection, decoded ranges at or outside the open interval $(5,100)$\,m are mapped to the origin before this term is evaluated.
The same boundary rule is applied to the targets.

The gradient term penalizes spatial variation in the masked normalized range residual.
At each sampling stride $s\in\{1,2,4,8\}$, let $e_j^{(s)}=m_j^{(s)}(\widehat r_{\text{n},j}^{(s)}-r_{\text{n},j}^{(s)})$ on the strided grid.
Let $\mathcal{E}_s$ contain horizontal and vertical adjacent ray pairs on that grid.
Then
\begin{equation}
 \mathcal{L}_{\text{grad}}
 =\sum_{s\in\{1,2,4,8\}}
 \frac{\displaystyle\sum_{(j,k)\in\mathcal{E}_s}
 m_j^{(s)}m_k^{(s)}
 \rho\!\left(e_j^{(s)}-e_k^{(s)}\right)}
 {\displaystyle\max\!\left(1,\sum_jm_j^{(s)}\right)},
 \label{eq::auto_lidar_gradient_loss}
\end{equation}
where $\rho(u)=u^2/2$ for $|u|<1$ and $\rho(u)=|u|-1/2$ otherwise.
Only pairs with two valid returns contribute, and finite differences do not wrap across the cropped azimuth boundary.
The point and gradient terms supervise range geometry, while intensity has its own reconstruction term.
\ifhmode\par\fi\endgroup

\begingroup{}
The complete objective uses the effective weights
\begin{equation}
 \mathcal{L}_{\text{VAE}}
 =\mathcal{L}_{\text{range}}
 +\mathcal{L}_{\text{intensity}}
 +\mathcal{L}_{\text{point}}
 +\mathcal{L}_{\text{mask}}
 +2\mathcal{L}_{\text{grad}}
 +10^{-6}\mathcal{L}_{\text{KL}}.
 \label{eq::auto_lidar_vae_loss}
\end{equation}
For $K$ latent coordinates with posterior means $\mu_k$ and log variances $a_k$, we use
\begin{equation}
 \mathcal{L}_{\text{KL}}
 =\frac{1}{2K}\sum_k
 \left(\mu_k^2+\exp(a_k)-1-a_k\right).
 \label{eq::auto_lidar_kl_loss}
\end{equation}
Training samples the Gaussian posterior through reparameterization.
The normalized reconstruction errors, metric point errors, and validity and KL terms retain their respective units and reductions.
\ifhmode\par\fi\endgroup

\begingroup{}
\paragraph{Tokenizer training recipe.}
Our tokenizer is trained on approximately 1M LiDAR sweeps.
The reference recipe draws contiguous windows of 8 LiDAR sweeps from MADS~\citep{cosmos3}.
The encoder and decoder are optimized together.
The documented refinement of an existing tokenizer checkpoint runs for 30K updates in float32, with a per-GPU batch size of 2 and gradient-norm clipping at 1.
AdamW uses $(\beta_1,\beta_2)=(0.9,0.999)$, $\epsilon=10^{-8}$, and zero weight decay.
The learning rate warms up linearly for 1K updates to $10^{-5}$ and then decays linearly to $10^{-6}$ over the remaining updates.
This refinement follows earlier tokenizer training.
The tokenizer checkpoint stores the latent normalization statistics and metric range bounds alongside the weights.

\Cref{tab::auto_lidar_tokenizer_comparison} compares our tokenizer with the public NVIDIA PAI tokenizer distributed with LidarVerse~\citep{che2026lidarverse,lidarverse2026release}.
The released checkpoint reconstructs range and validity, while our tokenizer also~reconstructs~intensity.
\ifhmode\par\fi\endgroup

\begingroup{}
\begin{table}[!ht]
\centering\reporttablestyle
\caption{\textbf{LiDAR tokenizer configuration comparison.}
Settings describe our joint RGB--LiDAR tokenizer and the public NVIDIA PAI release of LidarVerse~\citep{lidarverse2026release}.
Spatial grids are per sweep, with strides relative to the corresponding network input.}
\label{tab::auto_lidar_tokenizer_comparison}
\begin{tabularx}{\linewidth}{lYY}
\toprule
\textbf{Property} & \textbf{Ours} & \textbf{LidarVerse}\\
\midrule
Input canvas & $128\times1808$ & $128\times1024$\\
Reconstructed quantities & Range, intensity, validity & Range, validity\\
Metric range & 5--100\,m & 2.5--100\,m\\
Input patch & $2\times2$ & $2\times2$\\
Stride $(t,h,w)$ & $(1,16,16)$ & $(1,8,8)$\\
Latent grid & $8\times113$ & $16\times128$\\
Latent channels & 128 & 64\\
Parameters & 148.5M & 64.5M\\
Stage widths & $(128,256,512,1024)$ & $(128,256,512)$\\
Stage depths & $(3,3,3,3)$ & $(4,2,4)$\\
Attention heads & $(4,4,8,8)$ & $(4,16,16)$\\
Spatial neighborhood & $5\times45$, circular azimuth & $5\times35$, circular azimuth\\
Temporal attention & Causal, factorized and joint & Causal, factorized\\
Bottleneck & Joint spatiotemporal attention & Spatial and temporal attention\\
\bottomrule
\end{tabularx}
\end{table}
\ifhmode\par\fi\endgroup

\begingroup{}
\paragraph{Reconstruction evaluation.}
\phantomsection
\label{sec::auto_tokenizer_evaluation}
\Cref{tab::auto_lidar_tokenizer_metrics} reports reconstruction on 846 sweeps from 94 LiDARBench MADS clips, with both inputs and targets restricted to $[5,100]$\,m.
Range and mask metrics use a shared $128\times3600$ sensor raster.
Our cropped $128\times1800$ output and LidarVerse's $128\times1024$ output are mapped to this raster by nearest calibrated azimuth, including circular wraparound and preceding-column tie breaking.
The scores therefore include angular sampling loss.
Range errors use all ground-truth-valid rays, including predicted-invalid rays, with absolute relative error defined as $|\widehat r-r|/r$.
Mask IoU measures the valid-return class, using a validity threshold of $0.5$ and predicted ranges in $[5,100]$\,m.
\ifhmode\par\fi\endgroup

\begingroup{}
Chamfer distance compares calibrated Pandar128 point clouds in $[5,100]$\,m against the same native reference cloud, without upsampling decoded outputs.
For reference and reconstructed point sets $P$ and $Q$, the unsquared symmetric distance in meters is
\begin{equation}
 \operatorname{CD}(P,Q)
 =\frac{1}{2}\left(
 \frac{1}{|P|}\sum_{p\in P}\min_{q\in Q}\|p-q\|_2
 +\frac{1}{|Q|}\sum_{q\in Q}\min_{p\in P}\|q-p\|_2
 \right).
 \label{eq::auto_lidar_tokenizer_chamfer}
\end{equation}
Each cloud is independently sampled without replacement to at most 50K points, using seeds $2t$ and $2t+1$ for reference and reconstruction at sweep ordinal $t\in\{0,\ldots,8\}$.
Both models share the reference sample, and all 846 sweeps are scoreable.
Metrics average 9 sweeps per clip and then 94 clips equally, including the per-sweep median errors and foreground mask IoUs.
Paired 95\% percentile bootstrap intervals use 10K clip resamples.
\ifhmode\par\fi\endgroup

\begingroup{}
\paragraph{Tokenizer reconstruction.}
How accurately does the LiDAR tokenizer preserve range geometry and valid returns?
We compare our tokenizer with the public NVIDIA PAI tokenizer distributed with LidarVerse on the same sensor targets.
With matched $[5,100]$\,m inputs, ours reduces MAE by 57.3\% and Chamfer distance by 3.3\%, while increasing mean mask IoU by 1.398\,pp (\cref{tab::auto_lidar_tokenizer_metrics}).
Our tokenizer thus provides higher-fidelity reconstructions and, together with the joint transformer, supports the improved LiDAR generation~results~in~\cref{tab::auto_student_lidar}.
\ifhmode\par\fi\endgroup

\begingroup{}
\begin{table}[!t]
\centering\reporttablestyle[4pt]
\caption{\textbf{LiDAR tokenizer reconstruction on matched sensor targets.}
Both inputs and evaluated outputs use $[5,100]$\,m returns on the same 846 sweeps.
Range and mask metrics share the native $128\times3600$ target raster, and Chamfer uses the same native reference point cloud with a 50K-point cap.
Lower is better for errors and higher is better for mean mask IoU.
Scale factors multiply the displayed values.
The green row identifies our tokenizer, with best and second-best point estimates bold and underlined in each column.}
\label{tab::auto_lidar_tokenizer_metrics}
\begin{tabularx}{\linewidth}{lCCCCC}
\toprule
\textbf{Tokenizer} & \textbf{\makecell{Range MAE\\(m) $\downarrow$}} & \textbf{\makecell{Median abs. error\\($10^{-2}$\,m) $\downarrow$}} & \textbf{\makecell{Abs. relative error\\($\times10^{-2}$) $\downarrow$}} & \textbf{\makecell{Chamfer distance\\(m) $\downarrow$}} & \textbf{\makecell{Mean mask\\IoU $\uparrow$}}\\
\midrule
\rowcolor{rowours}
Ours & \textbf{0.4163} & \textbf{3.597} & \textbf{1.262} & \textbf{0.1674} & \textbf{0.7554}\\
LidarVerse & \underline{0.9742} & \underline{9.667} & \underline{3.183} & \underline{0.1731} & \underline{0.7414}\\
\bottomrule
\end{tabularx}
\end{table}
\ifhmode\par\fi\endgroup

%% file: appendix/auto_attention_details.tex
\begingroup{}
The main text introduces our decomposed multi-view and multimodal attention and its causal extension (\cref{sec::auto_multiview_modeling,sec::auto_causal_modeling}).
Here we explain how maskless token grouping avoids FlexAttention's partial-block masking overhead and how deduplicating overlapping key groups prevents excess same-view attention.
\ifhmode\par\fi\endgroup

\begingroup{}
\paragraph{Sliding-window attention.}
A same-view sliding window restricts the temporal context available within each camera stream.
Causal windows end at the query time and clip at the beginning of the sequence, while bidirectional neighborhoods can include both earlier and later frames.
The cross-view window is configured independently, allowing temporal context within a view to differ from the interval used to exchange information across views.
The current configuration specifies both windows in seconds and maps same-view bounds to latent frame offsets using each stream's latent frame period.
These windows constrain which keys a query can read, while the cache retention policy in \cref{sec::inference_optimization} determines which history keys remain stored.

\paragraph{Conditioning visibility.}
\ifhmode\par\fi\endgroup
\begingroup{}
The sensor-to-sensor attention pattern in \cref{fig::auto_attention_masks} covers supplied observations and prediction targets within the sensor streams, subject to their stage-specific visibility restrictions.
\ifhmode\par\fi\endgroup
\begingroup{}
Supplied control streams retain separate directed connections.
Sensor queries read supplied control keys/values associated with the same camera or LiDAR stream, with temporal access determined by the conditioning configuration and input availability.
Control queries read control keys/values within the same view.
They can additionally read sensor keys/values in that view when control-to-sensor attention is enabled, making control features depend on the sensor targets.
Cross-view sensor attention does not automatically grant access to controls from other views.

\ifhmode\par\fi\endgroup

%% file: appendix/auto_attention_implementation.tex
\phantomsection
\label{sec::auto_attention_implementation}
\paragraph[\kern0pt Attention implementation details.]{{Attention implementation details.}}

\begingroup{}
We initially implemented decomposed multi-view attention with PyTorch FlexAttention and a FlashAttention-4 backend~\citep{flashattention42026}.
FlexAttention expresses permitted query--key connections through a Boolean \texttt{mask\_mod} function, while \texttt{score\_mod} can modify attention scores before softmax.
We primarily used \texttt{mask\_mod} to encode sample boundaries, view restrictions, and temporal windows, constructing a \texttt{BlockMask} once per batch.
This interface supports packed samples with different numbers of views, resolutions, and frame counts.
\ifhmode\par\fi\endgroup

\subparagraph[\kern0pt Partial-block overhead and alignment.]{{Partial-block overhead and alignment.}}

\begingroup{}
FlexAttention skips empty blocks and processes fully visible blocks without elementwise masking.
Partially visible blocks require mask evaluation and application inside the attention kernel, reducing efficiency for our decomposed pattern.
The evaluated FlashAttention-4 backend typically used $256\times128$ query--key blocks on Blackwell and $128\times128$ blocks on Hopper.
Our example 480p and 720p inputs contain 390 and 920 latent tokens per frame, respectively, so frame boundaries do not align with these block dimensions.
\Cref{fig::auto_flex_blocks} illustrates the full, empty, and partial blocks and their masking requirements.
\ifhmode\par\fi\endgroup

\input{figures/auto/auto_flex_blocks.tex}

\phantomsection
\label{sec::auto_attention_alignment_study}
\begingroup{}
The measured attention workload reaches 45--55\% model FLOP utilization (MFU) with unrestricted cross-view attention but only 10--20\% with decomposed attention.
To test whether block alignment contributes to this gap, we use a synthetic $1024\times1024$ input that produces 1024 tokens per frame with the Wan 2.2 tokenizer, aligning frame boundaries with the kernel blocks.
Decomposed attention then reaches 45--55\% MFU, supporting block alignment as a practical constraint.
Padding frames to aligned lengths and using \texttt{score\_mod} to exclude padding tokens works functionally but does not recover the desired performance.
These observations motivate the maskless implementation below, whose inference performance is evaluated in \cref{sec::auto_runtime_results}.
\ifhmode\par\fi\endgroup

\subparagraph[\kern0pt Maskless token grouping.]{{Maskless token grouping.}}

\begingroup{}
Rather than applying a complex visibility mask to the entire packed sequence, we gather query and key/value tokens into groups for same-stream, cross-sensor, control, and language attention.
The groups follow the temporal and conditioning rules in \cref{sec::auto_multiview_modeling,sec::auto_causal_modeling}.
Each group is processed with efficient dense attention kernels, avoiding FlexAttention's partial-block masking overhead.
\ifhmode\par\fi\endgroup

\subparagraph[\kern0pt Deduplicating same-view keys.]{{Deduplicating same-view keys.}}

\begingroup{}
Efficient execution should preserve the intended balance between same-view and cross-view information.
If the cross-sensor group also includes the query's own stream, keys in the overlapping temporal window appear in both the same-stream and cross-sensor groups.
Combining these groups without deduplication counts those keys twice in the joint softmax, giving the duplicated same-view evidence excess weight relative to other views.
This introduces an undesirable inductive bias toward within-view information rather than exchanging evidence across cameras.
We therefore exclude the query's own stream from the cross-sensor group, retaining its keys only in the same-stream group.
\Cref{sec::auto_dedup_ablation} discusses its effect on cross-view consistency.
With disjoint key groups, we can now combine their outputs to recover a single softmax over all permitted keys.
\ifhmode\par\fi\endgroup

\subparagraph[\kern0pt Joint attention normalization.]{{Joint attention normalization.}}

\begingroup{}
For a fixed query and attention head, let $\mathbf{o}_b$ denote the locally normalized attention output for a nonempty key group $b$.
Let $Z_b$ be its softmax denominator: the sum of exponentiated attention scores over the keys in that group.
We recover the joint output by weighting each group output by its share of the total denominator:
\begin{equation}
 \mathbf{o}=\sum_b \frac{Z_b}{\sum_c Z_c}\,\mathbf{o}_b.
 \label{eq::attention_merge}
\end{equation}
Here, $c$ ranges over the same groups, and $Z_b/\sum_c Z_c$ is the total attention probability assigned to group $b$.
Each attention call returns the log-sum-exp statistic $\ell_b=\log Z_b$.
We use these statistics to compute the weights as $\exp(\ell_b-\operatorname{LSE}_c\ell_c)$.
With disjoint key groups and the same attention scores as the unsplit computation, this weighted combination is mathematically equivalent to a single softmax over all allowed keys~\citep{ye2025flashinfer}.
\ifhmode\par\fi\endgroup

%% file: figures/auto/auto_flex_blocks.tex
\begingroup{}
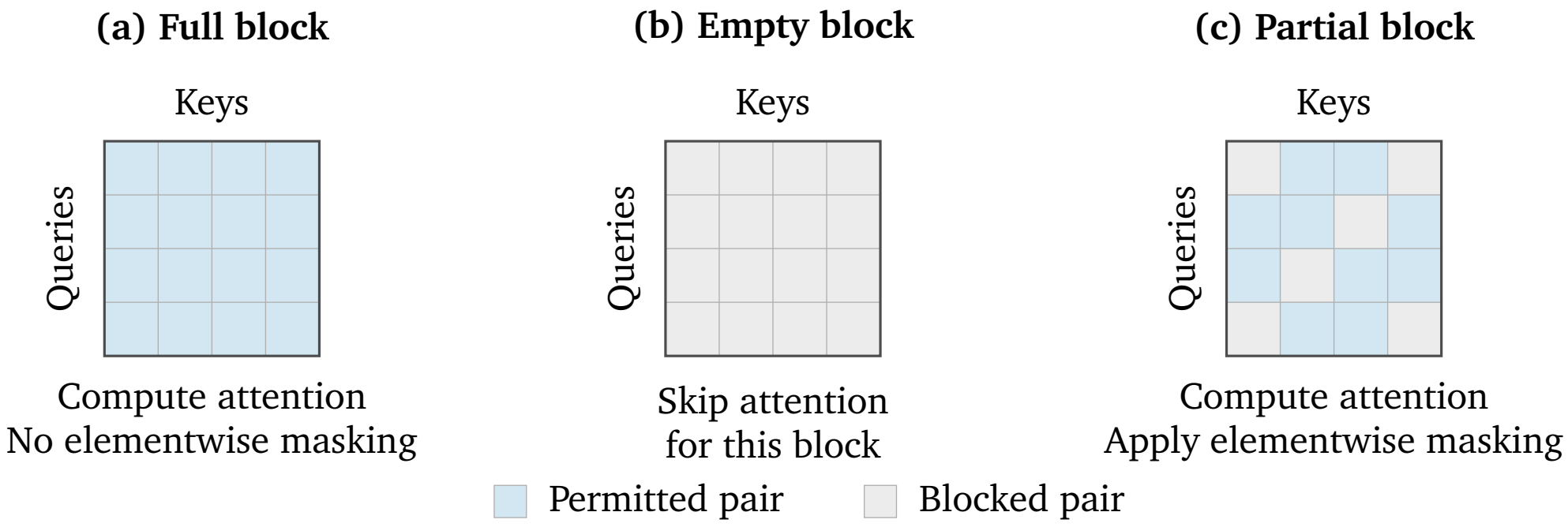
\begin{figure}[!tb]
\centering
\input{figures/diagrams/auto_flex_blocks.tex}
\caption{\textbf{FlexAttention block types.}
Rows are queries and columns are keys, with blue cells denoting permitted pairs and gray cells denoting blocked pairs.
(a) Full blocks execute attention without elementwise masking, (b) empty blocks skip attention, and (c) partial blocks execute attention with elementwise masking.
The cell counts and visibility patterns are schematic.}
\label{fig::auto_flex_blocks}
\end{figure}
\ifhmode\par\fi\endgroup

%% file: figures/diagrams/auto_flex_blocks.tex
\begingroup
\input{figures/diagrams/architecture_styles.tex}
\fitdiagram{%
\begin{tikzpicture}[font=\fontfamily{bch}\fontsize{9}{10.5}\selectfont,
  flex heading/.style={font=\fontfamily{bch}\fontsize{9}{10.5}\selectfont\bfseries},
  flex grid/.style={draw=simgrayedge!45,line width=0.25pt},
  flex boundary/.style={draw=simgrayedge,line width=0.6pt}]
  \foreach \panelx/\paneltitle/\blocktype/\execution in {
    2.2/{(a) Full block}/0/{Compute attention\\No elementwise masking},
    6.9/{(b) Empty block}/1/{Skip attention\\for this block},
    11.6/{(c) Partial block}/2/{Compute attention\\Apply elementwise masking}} {
    \begin{scope}[shift={(\panelx,0)}]
      \node[flex heading] at (0,0.95) {\paneltitle};
      \node at (0,0.3) {Keys};
      \node[rotate=90] at (-1.25,-0.9) {Queries};
      \begin{scope}[shift={(-0.9,0)},x=0.45cm,y=-0.45cm]
        \fill[simgrayfill] (0,0) rectangle (4,4);
        \ifnum\blocktype=0
          \fill[simcleanfill] (0,0) rectangle (4,4);
        \fi
        \ifnum\blocktype=2
          \foreach \queryrow/\keycolumn in {
            0/1,0/2,1/0,1/1,1/3,2/0,2/2,2/3,3/1,3/2} {
            \fill[simcleanfill] (\keycolumn,\queryrow) rectangle ++(1,1);
          }
        \fi
        \foreach \boundary in {1,2,3} {
          \draw[flex grid] (\boundary,0) -- (\boundary,4);
          \draw[flex grid] (0,\boundary) -- (4,\boundary);
        }
        \draw[flex boundary] (0,0) rectangle (4,4);
      \end{scope}
      \node[align=center] at (0,-2.35) {\execution};
    \end{scope}
  }
  \node[flex grid,fill=simcleanfill,minimum size=0.27cm,inner sep=0pt]
    at (4.7,-3.02) {};
  \node[anchor=west] at (4.9,-3.02) {Permitted pair};
  \node[flex grid,fill=simgrayfill,minimum size=0.27cm,inner sep=0pt]
    at (7.8,-3.02) {};
  \node[anchor=west] at (8.0,-3.02) {Blocked pair};
\end{tikzpicture}}
\endgroup

%% file: appendix/auto_rigs.tex
\subsubsection[\kern0pt Camera Rig]{{Camera Rig}}
\label{sec::auto_rig_sampling}
\begingroup{}
\Cref{fig::auto_camera_rig} schematically illustrates the cameras' relative positions on the vehicle and their fields of view.
Both Cosmos3-Nano-Transfer-Auto and Cosmos3-Nano-Sim-Auto sample camera combinations from this rig during training (\cref{sec::auto_training}), using the catalog in \cref{tab::auto_camera_rigs}.
\ifhmode\par\fi\endgroup
\begingroup{}
\input{figures/auto/auto_camera_rig.tex}
\ifhmode\par\fi\endgroup

\begingroup{}
\begin{table}[!t]
\centering\reporttablestyle[4pt]
\caption{\textbf{Predefined camera combinations for training Cosmos3-Nano-Transfer-Auto and Cosmos3-Nano-Sim-Auto.}
A tick marks an included camera; a blank cell marks an omitted camera.
Headers give each camera's name and nominal field of view (FOV).
The first 7 camera columns form the standard 7-camera rig.
Tele denotes telephoto.}
\label{tab::auto_camera_rigs}
\begin{tabularx}{\linewidth}{c*{11}{C}}
\toprule
& \multicolumn{7}{c}{\textbf{Standard 7-camera rig}}
& \multicolumn{4}{c}{\textbf{Additional fisheye cameras}}\\
\cmidrule(lr){2-8}\cmidrule(l){9-12}
\textbf{Count}
& \shortstack{Front\\wide\\120$^\circ$}
& \shortstack{Cross\\right\\120$^\circ$}
& \shortstack{Rear\\right\\70$^\circ$}
& \shortstack{Rear\\tele\\30$^\circ$}
& \shortstack{Rear\\left\\70$^\circ$}
& \shortstack{Cross\\left\\120$^\circ$}
& \shortstack{Front\\tele\\30$^\circ$}
& \shortstack{Front\\fisheye\\200$^\circ$}
& \shortstack{Left\\fisheye\\200$^\circ$}
& \shortstack{Right\\fisheye\\200$^\circ$}
& \shortstack{Rear\\fisheye\\200$^\circ$}\\
\midrule
1  & \cmark &        &        &        &        &        &        &        &        &        &        \\
2  & \cmark &        &        &        &        &        & \cmark &        &        &        &        \\
4  &        &        &        &        &        &        &        & \cmark & \cmark & \cmark & \cmark \\
4  & \cmark & \cmark &        & \cmark &        & \cmark &        &        &        &        &        \\
5  & \cmark & \cmark & \cmark &        & \cmark & \cmark &        &        &        &        &        \\
5  & \cmark &        &        &        &        &        &        & \cmark & \cmark & \cmark & \cmark \\
6  & \cmark & \cmark & \cmark &        & \cmark & \cmark &        &        &        &        & \cmark \\
6  & \cmark & \cmark &        &        &        & \cmark &        &        & \cmark & \cmark & \cmark \\
7  & \cmark & \cmark & \cmark &        & \cmark & \cmark & \cmark &        &        &        & \cmark \\
7  & \cmark & \cmark & \cmark & \cmark & \cmark & \cmark & \cmark &        &        &        &        \\
8  & \cmark & \cmark & \cmark &        & \cmark & \cmark & \cmark & \cmark &        &        & \cmark \\
8  & \cmark & \cmark & \cmark & \cmark & \cmark & \cmark & \cmark & \cmark &        &        &        \\
10 & \cmark & \cmark & \cmark &        & \cmark & \cmark & \cmark & \cmark & \cmark & \cmark & \cmark \\
10 & \cmark & \cmark & \cmark & \cmark & \cmark & \cmark & \cmark & \cmark & \cmark & \cmark &        \\
11 & \cmark & \cmark & \cmark & \cmark & \cmark & \cmark & \cmark & \cmark & \cmark & \cmark & \cmark \\
\bottomrule
\end{tabularx}
\end{table}
\ifhmode\par\fi\endgroup

%% file: figures/auto/auto_camera_rig.tex
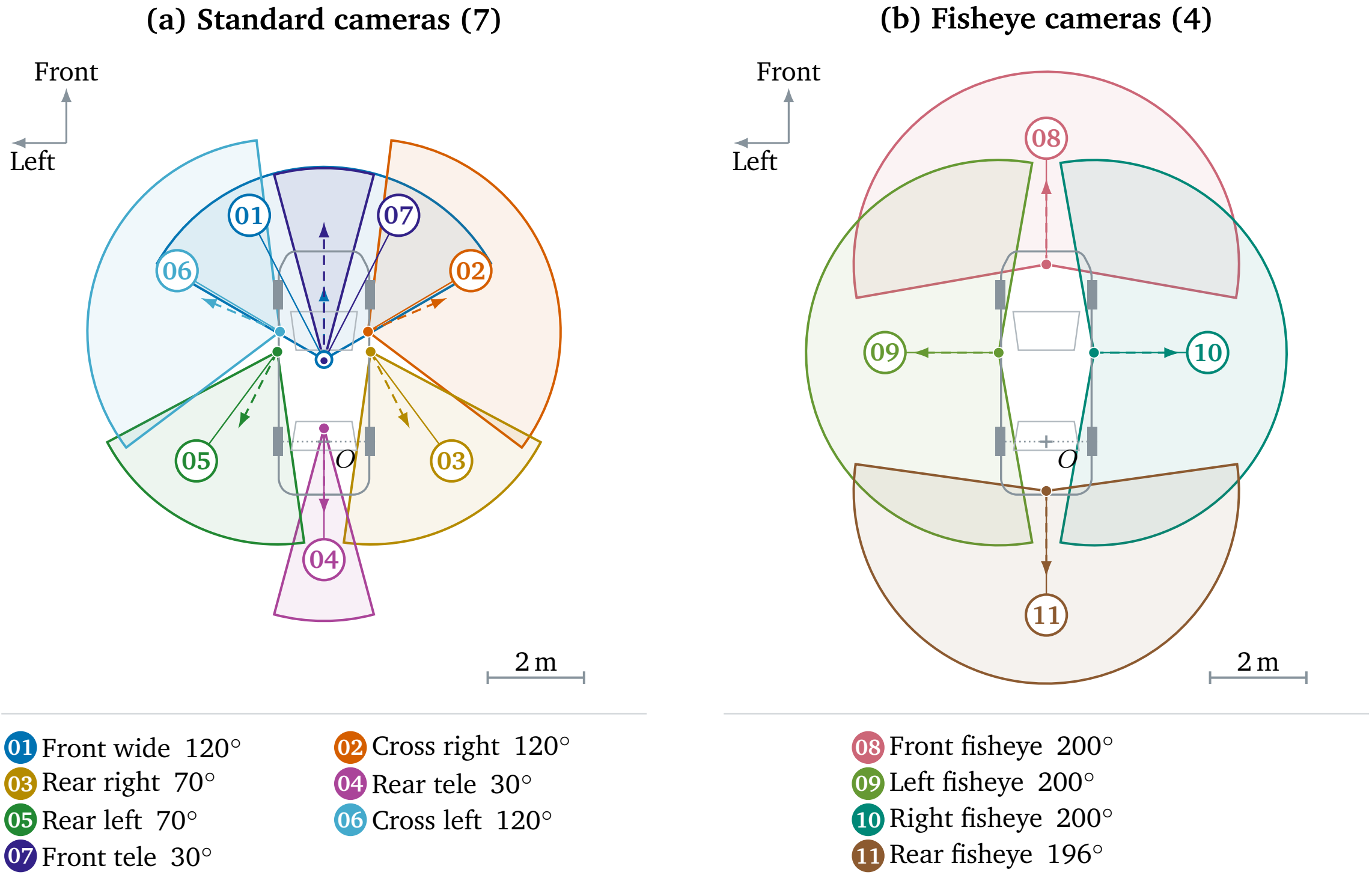
\begin{figure}[!t]
\centering
\input{figures/diagrams/auto_camera_rig.tex}
\caption{\textbf{Symmetric schematic of the complete 11-camera rig.}
(a) Seven standard cameras cover wide-angle and telephoto views.
(b) Four fisheye cameras provide wider angular coverage.
Colors and numbered labels identify individual cameras, and dashed arrows indicate viewing directions.}
\label{fig::auto_camera_rig}
\end{figure}

%% file: figures/diagrams/auto_camera_rig.tex
\begingroup
\input{figures/diagrams/auto_camera_rig_geometry.tex}
\definecolor{rig01}{HTML}{0072B2}
\definecolor{rig02}{HTML}{D55E00}
\definecolor{rig03}{HTML}{B58B00}
\definecolor{rig04}{HTML}{AA4499}
\definecolor{rig05}{HTML}{228833}
\definecolor{rig06}{HTML}{44AACC}
\definecolor{rig07}{HTML}{332288}
\definecolor{rig08}{HTML}{CC6677}
\definecolor{rig09}{HTML}{669933}
\definecolor{rig10}{HTML}{008877}
\definecolor{rig11}{HTML}{8C5A30}
\definecolor{rigbody}{HTML}{F5F6F7}
\definecolor{rigedge}{HTML}{87939C}

\newcommand{\rigcoverage}[1]{%
  \foreach \id in {#1} {
    \edef\rigpolygon{\csname rigpath\id\endcsname}
    \path[fill=rig\id,fill opacity=0.08,draw=rig\id,line width=0.75pt]
      \rigpolygon -- cycle;
  }
}
\newcommand{\rigvehicle}{%
  \path[fill=none,draw=rigedge,line width=0.65pt,rounded corners=2pt]
    (0.75,-1.10) -- (0.94,-0.90) -- (0.94,3.55) -- (0.73,3.95)
    -- (-0.73,3.95) -- (-0.94,3.55) -- (-0.94,-0.90) -- (-0.75,-1.10) -- cycle;
  \path[fill=none,draw=rigedge!65,line width=0.45pt]
    (-0.69,2.70) -- (-0.58,1.90) -- (0.58,1.90) -- (0.69,2.70) -- cycle;
  \path[fill=none,draw=rigedge!65,line width=0.45pt]
    (-0.58,0.42) -- (-0.68,-0.18) -- (0.68,-0.18) -- (0.58,0.42) -- cycle;
  \foreach \axle in {0,3.05} {
    \foreach \side in {-1,1} {
      \fill[rigedge] ({\side*0.85},{\axle-0.30})
        rectangle ({\side*1.06},{\axle+0.30});
    }
  }
  \draw[rigedge,densely dotted,line width=0.5pt] (-1.06,0) -- (1.06,0);
  \draw[rigedge,line width=0.7pt] (-0.13,0) -- (0.13,0) (0,-0.13) -- (0,0.13);
  \node[anchor=north west,inner sep=1pt] at (0.15,-0.10) {$O$};
}
\newcommand{\rigaxes}[1]{%
  \foreach \id in {#1} {
    \edef\rigstart{\csname rigpos\id\endcsname}
    \edef\rigend{\csname rigaxis\id\endcsname}
    \draw[rig\id,densely dashed,line width=0.65pt,-{Latex[length=1.8mm,width=1.2mm]}]
      \rigstart -- \rigend;
  }
}
\newcommand{\rigcallout}[3]{%
  \edef\rigstart{\csname rigpos#1\endcsname}
  \draw[rig#1,line width=0.45pt] \rigstart -- (#2,#3);
  \node[rig badge,draw=rig#1,text=rig#1] at (#2,#3) {#1};
}
\newcommand{\rigcenters}[1]{%
  \foreach \id in {#1} {
    \edef\rigstart{\csname rigpos\id\endcsname}
    \filldraw[fill=rig\id,draw=white,line width=0.35pt] \rigstart circle[radius=1.7pt];
  }
}
\newcommand{\rigkey}[4]{%
  \node[rig key badge,fill=rig#1] at (#3,#4) {#1};
  \node[anchor=west,inner sep=0pt] at ({#3+0.43},#4)
    {#2\enspace\csname rigfov#1\endcsname$^\circ$};
}
\newcommand{\rigorientation}{%
  \draw[rigedge,-{Latex[length=1.6mm]},line width=0.6pt] (-5.35,6.20) -- (-5.35,7.35);
  \draw[rigedge,-{Latex[length=1.6mm]},line width=0.6pt] (-5.35,6.20) -- (-6.50,6.20);
  \node[anchor=south,inner sep=2pt] at (-5.35,7.35) {Front};
  \node[anchor=north,inner sep=2pt] at (-6.05,6.20) {Left};
  \draw[rigedge,line width=0.75pt] (3.4,-4.90) -- (5.4,-4.90);
  \draw[rigedge,line width=0.75pt] (3.4,-4.77) -- (3.4,-5.03)
    (5.4,-4.77) -- (5.4,-5.03);
  \node[anchor=south,inner sep=2pt] at (4.4,-4.90) {2\,m};
}

\fitdiagram{%
\begin{tikzpicture}[x=0.53cm,y=0.53cm,
  font=\fontencoding{T1}\fontfamily{bch}\fontsize{8.5}{10}\selectfont,
  rig title/.style={font=\fontencoding{T1}\fontfamily{bch}\fontsize{9.5}{11}\selectfont\bfseries},
  rig badge/.style={circle,fill=white,minimum size=4.5mm,inner sep=0pt,line width=0.8pt,
    font=\fontencoding{T1}\fontfamily{bch}\fontsize{8}{8}\selectfont\bfseries},
  rig key badge/.style={circle,text=white,minimum size=3.6mm,inner sep=0pt,
    font=\fontencoding{T1}\fontfamily{bch}\fontsize{7}{7}\selectfont\bfseries}]
  \path[use as bounding box] (-7.0,-9.0) rectangle (22.0,9.0);

  \node[rig title] at (0,8.75) {(a) Standard cameras (7)};
  \rigcoverage{01,02,03,04,05,06,07}
  \rigvehicle
  \rigaxes{01,02,03,04,05,06,07}
  \rigcallout{01}{-1.55}{4.70}
  \rigcallout{02}{3.05}{3.55}
  \rigcallout{03}{2.65}{-0.40}
  \rigcallout{04}{0}{-2.45}
  \rigcallout{05}{-2.65}{-0.40}
  \rigcallout{06}{-3.05}{3.55}
  \rigcallout{07}{1.55}{4.70}
  \rigcenters{02,03,04,05,06}
  \edef\rigstart{\csname rigpos01\endcsname}
  \draw[rig01,fill=white,line width=0.85pt] \rigstart circle[radius=2.5pt];
  \edef\rigstart{\csname rigpos07\endcsname}
  \fill[rig07] \rigstart circle[radius=1.1pt];
  \rigorientation
  \draw[rigedge!40,line width=0.4pt] (-6.7,-5.65) -- (6.7,-5.65);
  \rigkey{01}{Front wide}{-6.30}{-6.35}
  \rigkey{02}{Cross right}{0.55}{-6.35}
  \rigkey{03}{Rear right}{-6.30}{-7.10}
  \rigkey{04}{Rear tele}{0.55}{-7.10}
  \rigkey{05}{Rear left}{-6.30}{-7.85}
  \rigkey{06}{Cross left}{0.55}{-7.85}
  \rigkey{07}{Front tele}{-6.30}{-8.60}

  \begin{scope}[shift={(15,0)}]
    \node[rig title] at (0,8.75) {(b) Fisheye cameras (4)};
    \rigcoverage{08,09,10,11}
    \rigvehicle
    \rigaxes{08,09,10,11}
    \rigcallout{08}{0}{6.30}
    \rigcallout{09}{-3.35}{1.85}
    \rigcallout{10}{3.35}{1.85}
    \rigcallout{11}{0}{-3.60}
    \rigcenters{08,09,10,11}
    \rigorientation
    \draw[rigedge!40,line width=0.4pt] (-6.7,-5.65) -- (6.7,-5.65);
    \rigkey{08}{Front fisheye}{-3.70}{-6.35}
    \rigkey{09}{Left fisheye}{-3.70}{-7.10}
    \rigkey{10}{Right fisheye}{-3.70}{-7.85}
    \rigkey{11}{Rear fisheye}{-3.70}{-8.60}
  \end{scope}
\end{tikzpicture}}
\endgroup

%% file: appendix/auto_wsm.tex
\subsubsection[\kern0pt WSM Rendering]{{WSM Rendering}}
\label{sec::auto_wsm_rendering}
\begingroup{}
WSM v4 uses semantic colors and symbols to distinguish map features and object states, as illustrated in \cref{fig::auto_wsm_legend}.
Traffic signs use colored rectangles to distinguish categories such as stop, yield, and one-way, with dark-gray backs for recognized classes.
Traffic lights encode signal state and direction, including directional triangles and pedestrian stop/go symbols.
Road markings include seven directional-arrow shapes and filled regions for STOP and KEEP CLEAR.
Road boundaries use six color groups for barriers or fences, curbs, road seams, guide posts, walls, and other boundaries.
Poles and explicit wait lines use colored polylines, while crosswalks use filled polygons.
Sign categories and STOP/KEEP CLEAR areas are represented without printed lettering.
The Cosmos Framework WSM guide provides the full color palette and annotated examples.\footnote{WSM rendering guide: \href{https://github.com/NVIDIA/cosmos-framework/blob/main/.agents/skills/cookbooks/cosmos3/simulation/auto/wsm/docs/wsm4.md}{\nolinkurl{github.com/NVIDIA/cosmos-framework/blob/main/.agents/skills/cookbooks/cosmos3/simulation/auto/wsm/docs/wsm4.md}}.}
\ifhmode\par\fi\endgroup

\begingroup{}
\begin{figure}[!t]
\centering
\input{figures/diagrams/auto_wsm_legend.tex}
\caption{\textbf{WSM v4 encodings, redrawn from the reference renderer documentation.}
Traffic-sign category colors (top) and traffic-light states and directional symbols (bottom).
Chart labels and backgrounds are explanatory and are not part of the rendered controls.}
\label{fig::auto_wsm_legend}
\end{figure}
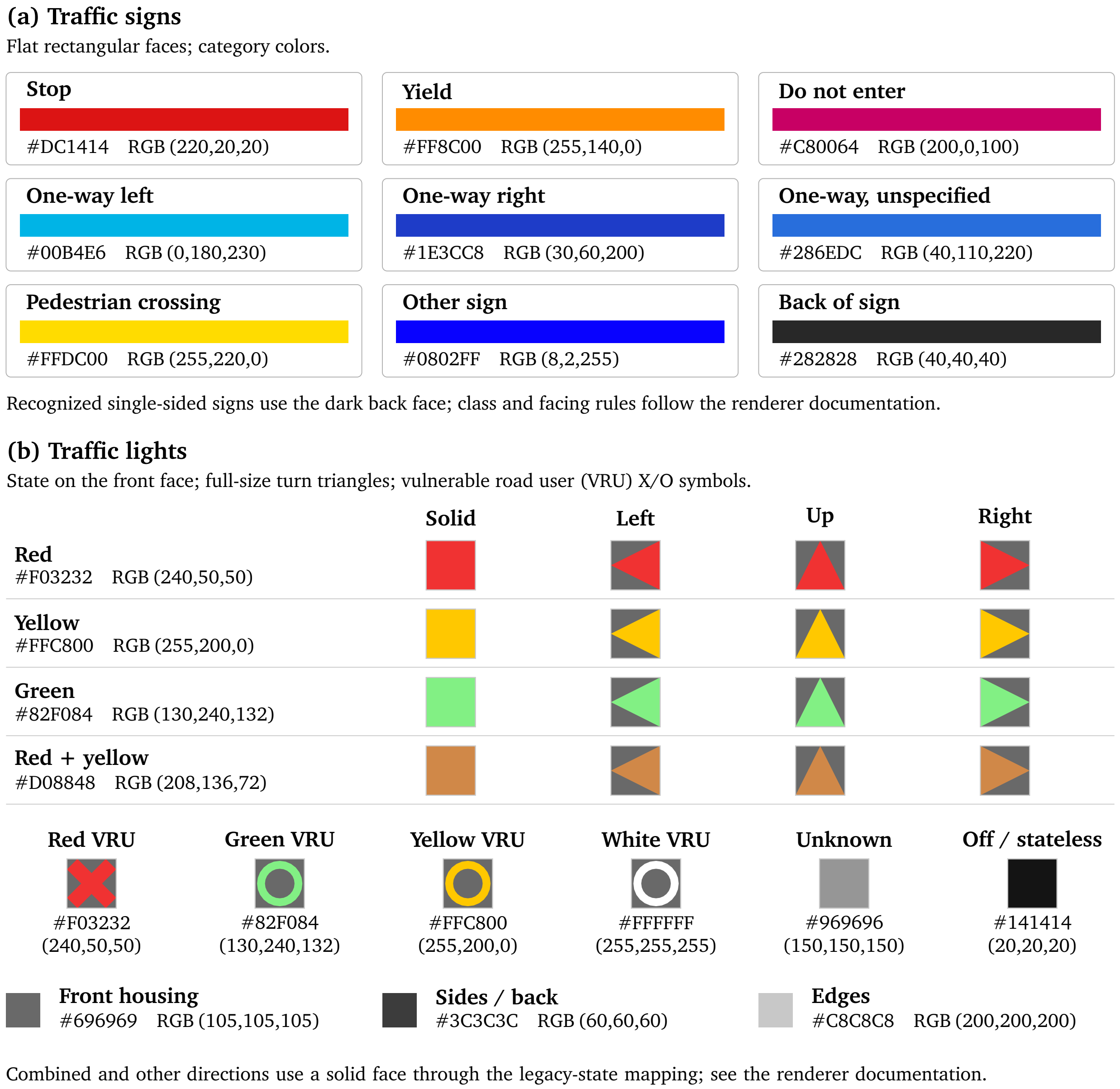
\ifhmode\par\fi\endgroup

\begingroup{}
\begin{figure}[!t]
\ContinuedFloat
\centering
\input{figures/diagrams/auto_wsm_road_legend.tex}
\caption{\textbf{WSM v4 encodings (continued).}
Directional road markings and filled areas (top), and road-boundary colors (bottom), redrawn from the reference renderer documentation.
The light-gray tiles behind near-white arrows provide contrast and are not part of the rendered controls.}
\end{figure}
\ifhmode\par\fi\endgroup

%% file: figures/diagrams/auto_wsm_legend.tex
\begingroup
\input{figures/diagrams/auto_wsm_styles.tex}
\fitdiagram{%
\begin{tikzpicture}[wsm chart]
\node[wsm title] at (0,0) {(a) Traffic signs};
\node[wsm note] at (0,-0.43) {Flat rectangular faces; category colors.};

\foreach \x/\y/\name/\wsmfill in {
  0/-0.8/Stop/wsmsignstop,
  5.5/-0.8/Yield/wsmsignyield,
  11/-0.8/{Do not enter}/wsmsigndonotenter,
  0/-2.35/{One-way left}/wsmsignleft,
  5.5/-2.35/{One-way right}/wsmsignright,
  11/-2.35/{One-way, unspecified}/wsmsignoneway,
  0/-3.9/{Pedestrian crossing}/wsmsignpedestrian,
  5.5/-3.9/{Other sign}/wsmsignother,
  11/-3.9/{Back of sign}/wsmsignback} {
  \begin{scope}[shift={(\x,\y)}]
    \draw[wsm card] (0,0) rectangle (5.2,-1.35);
    \node[wsm label,anchor=west] at (0.17,-0.27) {\name};
    \fill[\wsmfill] (0.2,-0.52) rectangle (5,-0.85);
    \node[wsm values,anchor=west] at (0.17,-1.1) {\wsmcolorspec{\wsmfill}};
  \end{scope}
}
\node[wsm note] at (0,-5.65)
  {Recognized single-sided signs use the dark back face; class and facing rules follow the renderer documentation.};

\node[wsm title] at (0,-6.35) {(b) Traffic lights};
\node[wsm note] at (0,-6.78)
  {State on the front face; full-size turn triangles; vulnerable road user (VRU) X/O symbols.};
\foreach \x/\name in {6.5/Solid,9.2/Left,11.9/Up,14.6/Right} {
  \node[wsm label] at (\x,-7.3) {\name};
}
\foreach \y/\name/\wsmfill in {
  -8.0/Red/wsmlightred,
  -9.0/Yellow/wsmlightyellow,
  -10.0/Green/wsmlightgreen,
  -11.0/{Red + yellow}/wsmlightredyellow} {
  \node[wsm label,anchor=west] at (0,\y+0.17) {\name};
  \node[wsm values,anchor=west] at (0,\y-0.19) {\wsmcolorspec{\wsmfill}};
  \wsmlightface{6.5}{\y}{\fill[\wsmfill] (-0.5,-0.5) rectangle (0.5,0.5);}
  \wsmlightface{9.2}{\y}{\fill[\wsmfill] (0.5,-0.5) -- (-0.5,0) -- (0.5,0.5) -- cycle;}
  \wsmlightface{11.9}{\y}{\fill[\wsmfill] (-0.5,-0.5) -- (0,0.5) -- (0.5,-0.5) -- cycle;}
  \wsmlightface{14.6}{\y}{\fill[\wsmfill] (-0.5,-0.5) -- (0.5,0) -- (-0.5,0.5) -- cycle;}
  \draw[wsm rule] (0,\y-0.49) -- (16.2,\y-0.49);
}

\foreach \x/\name/\wsmfill/\glyph in {
  1.25/{Red VRU}/wsmlightred/{\wsmcross},
  4/{Green VRU}/wsmlightgreen/{\wsmring{wsmlightgreen}},
  6.75/{Yellow VRU}/wsmlightyellow/{\wsmring{wsmlightyellow}},
  9.5/{White VRU}/wsmlightwhite/{\wsmring{wsmlightwhite}},
  12.25/Unknown/wsmlightunknown/{\fill[wsmlightunknown] (-0.5,-0.5) rectangle (0.5,0.5);},
  15/{Off / stateless}/wsmlightoff/{\fill[wsmlightoff] (-0.5,-0.5) rectangle (0.5,0.5);}} {
  \node[wsm label] at (\x,-12.0) {\name};
  \wsmlightface{\x}{-12.65}{\glyph}
  \node[wsm values] at (\x,-13.22) {\wsmhex{\wsmfill}};
  \node[wsm values] at (\x,-13.57) {\wsmrgb{\wsmfill}};
}

\foreach \x/\name/\wsmfill in {
  0/{Front housing}/wsmlighthousing,
  5.5/{Sides / back}/wsmlightback,
  11/Edges/wsmlightedge} {
  \fill[\wsmfill] (\x,-14.25) rectangle (\x+0.5,-14.75);
  \node[wsm label,anchor=west] at (\x+0.65,-14.31) {\name};
  \node[wsm values,anchor=west] at (\x+0.65,-14.67) {\wsmcolorspec{\wsmfill}};
}
\node[wsm note,text width=16.2cm] at (0,-15.45)
  {Combined and other directions use a solid face through the legacy-state mapping; see the renderer documentation.};
\end{tikzpicture}}
\endgroup

%% file: figures/diagrams/auto_wsm_road_legend.tex
\begingroup
\input{figures/diagrams/auto_wsm_styles.tex}
\input{figures/diagrams/auto_wsm_arrow_shapes.tex}
\fitdiagram{%
\begin{tikzpicture}[wsm chart]
\node[wsm title] at (0,0) {(c) Road markings};
\node[wsm note] at (0,-0.43) {Canonical arrow shapes and filled painted areas.};
\foreach \x/\y/\name/\shape in {
  1.8/-1.9/Forward/forward,
  6/-1.9/Left/left,
  10.2/-1.9/Right/right,
  14.4/-1.9/{Left + forward}/left_forward,
  1.8/-4.2/{Right + forward}/right_forward,
  6/-4.2/{Merge left}/merge_left,
  10.2/-4.2/{Merge right}/merge_right} {
  \node[wsm label] at (\x,\y+1.1) {\name};
  \fill[black!18,rounded corners=2pt]
    (\x-0.8,\y-0.8) rectangle (\x+0.8,\y+0.8);
  \begin{scope}[shift={(\x,\y)}]
    \pic[x=1.18cm,y=1.3cm] {wsm arrow \shape};
  \end{scope}
}
\node[wsm label] at (14.4,-3.1) {All arrows};
\fill[wsmmarkingarrow,draw=simgrayedge!35,line width=0.3pt]
  (12.9,-3.85) rectangle (15.9,-4.25);
\node[wsm values] at (14.4,-4.6) {\wsmhex{wsmmarkingarrow}};
\node[wsm values] at (14.4,-4.95) {RGB\,\wsmrgb{wsmmarkingarrow}};

\foreach \x/\name/\wsmfill in {
  0/{STOP area}/wsmmarkingstop,
  5.5/{KEEP CLEAR area}/wsmmarkingclear,
  11/{Other marking}/wsmmarkingother} {
  \draw[wsm card] (\x,-5.55) rectangle (\x+5.2,-6.95);
  \node[wsm label,anchor=west] at (\x+0.17,-5.83) {\name};
  \fill[\wsmfill] (\x+0.2,-6.08) rectangle (\x+5,-6.43);
  \node[wsm values,anchor=west] at (\x+0.17,-6.7) {\wsmcolorspec{\wsmfill}};
}
\node[wsm note] at (0,-7.38)
  {Arrows follow the annotated footprint. STOP and KEEP CLEAR are filled areas without lettering.};

\node[wsm title] at (0,-8.15) {(d) Road boundaries};
\node[wsm note] at (0,-8.58) {Six color groups; all boundary subtypes use solid strokes.};
\node[wsm label,anchor=west] at (0,-9.15) {Boundary group};
\node[wsm label] at (8.45,-9.15) {Color};
\node[wsm label,anchor=west] at (11.4,-9.15) {Hex / RGB};
\draw[wsm rule] (0,-9.45) -- (16.2,-9.45);
\foreach \y/\name/\wsmfill in {
  -9.95/{Barrier / fence}/wsmboundarybarrier,
  -10.9/{Curb / low curb / tall curb}/wsmboundarycurb,
  -11.85/{Road boundary / seam}/wsmboundaryseam,
  -12.8/{Obstacle bars / guide posts}/wsmboundaryposts,
  -13.75/Wall/wsmboundarywall,
  -14.7/{Other / unknown boundary}/wsmboundaryother} {
  \node[anchor=west] at (0,\y) {\name};
  \fill[\wsmfill] (6.4,\y-0.11) rectangle (10.5,\y+0.11);
  \node[wsm values,anchor=west] at (11.4,\y) {\wsmcolorspec{\wsmfill}};
  \draw[wsm rule] (0,\y-0.47) -- (16.2,\y-0.47);
}
\node[wsm note] at (0,-15.6)
  {The renderer documentation lists the subtype-to-color mapping, including the fallback group.};
\end{tikzpicture}}
\endgroup

%% file: appendix/auto_captions.tex
\subsubsection[\kern0pt Captions]{{Captions}}
\label{sec::auto_caption_example}
\newcommand{\autopromptboxsetup}{%
  \newtcolorbox[auto counter,number within=section,crefname={Box}{Boxes},Crefname={Box}{Boxes}]{autopromptbox}[2][]{title={Box~\thetcbcounter: ##2},##1}%
  \crefformat{autopromptbox}{Box~##2##1##3}%
  \crefrangeformat{autopromptbox}{Boxes~##3##1##4--##5##2##6}%
}
\autopromptboxsetup
{The examples in \crefrange{box::auto_task_prefix}{box::auto_baseline_prompt} illustrate a timestamped per-camera caption and its prompt prefixes.}
\begingroup{}
We show the joint RGB--LiDAR task prefix and the camera prefix for an 11-camera selection, followed by the stored scene caption and, for comparison, the Cosmos-Transfer2.5 prompt for the same clip.
Box titles identify the components and are excluded from the model input.
\ifhmode\par\fi\endgroup

\paragraph[\kern0pt Prompt prefixes.]{{Prompt prefixes.}}
\begingroup{}
Our prompt prefixes have two components, illustrated in \crefrange{box::auto_task_prefix}{box::auto_camera_prefix}.
The task prefix is a system prompt that specifies joint RGB--LiDAR generation from camera WSMs and LiDAR HD-map controls, with instructions for synchronized and consistent sensor outputs.
The camera prefix describes the selected rig, listing each camera's viewing direction and field of view, before identifying the current camera.
It is prepended to that camera's scene caption, with the rig description updated to match the sampled camera selection.
\ifhmode\par\fi\endgroup

\paragraph[\kern0pt Prompt upsampling.]{{Prompt upsampling.}}
\phantomsection
\label{sec::auto_prompt_upsampling}
\begingroup{}
Prompt upsampling turns concise scene specifications into dense, timestamped descriptions that match the style of our training captions.
Synchronized WSM observations establish the scene geometry and motion, while the requested environmental attributes guide its appearance.
We first construct a coherent scene description, assigning consistent appearance to objects shared across views and expanding environmental attributes into concrete visual cues, such as reduced contrast for distant objects in fog.
We then express this shared scene from each camera's viewpoint, producing captions whose visible content and event timing follow the corresponding WSM sequence.
\ifhmode\par\fi\endgroup

\begingroup{}
Prompt upsampling emphasizes concrete descriptions of object appearance rather than generic quality statements.
Off-the-shelf, state-of-the-art VLMs can perform this workflow using our prompt upsampling guide in Cosmos Framework.
\ifhmode\par\fi\endgroup

\begingroup{}
\begingroup
\tcbset{
  unbreakable,
  colframe=nvidiagreen!75,
  colback=black!2,
  colbacktitle=nvidiagreen!75,
  coltitle=white,
  coltext=black,
  fonttitle=\normalfont\small,
  boxrule=0.4pt,
  arc=0pt,
  left=8pt,right=8pt,top=4pt,bottom=4pt
}
\begin{figure}[!t]
\begin{autopromptbox}[label={box::auto_task_prefix}]{Task prefix: joint RGB--LiDAR generation}
\begin{Verbatim}[fontsize=\footnotesize,breaklines=true,breaksymbolleft={}]
You are a helpful assistant that jointly generates temporally synchronized, geometrically consistent autonomous-driving camera videos and LiDAR range-view sequences from per-camera scene descriptions and provided control signals: per-camera World Scenario Map (WSM) control videos depicting the controlled objects and road layout, and an HD-map control for LiDAR. Treat all camera views and LiDAR sweeps as synchronized observations of the same driving scene, preserving each camera's viewpoint, shared ego motion, road layout, object identity and motion, weather, lighting, cross-view consistency, and camera-LiDAR alignment.

Follow WSM controls for vehicles (including trucks), cyclists, pedestrians, traffic lights, traffic signs, road markings, lane boundaries, and road boundaries. Do not add objects or road features in these categories that are absent from WSM. Use captions for appearance and unconstrained background details; WSM takes precedence in any conflict.
\end{Verbatim}
\end{autopromptbox}
\end{figure}

\begin{figure}[!t]
\begin{autopromptbox}[label={box::auto_camera_prefix}]{Camera prefix: 11-camera rig and front-wide view}
\begin{Verbatim}[fontsize=\footnotesize,breaklines=true,breaksymbolleft={}]
This multiview driving sequence contains time-aligned recordings from 11 vehicle-mounted cameras: front wide-angle camera (forward-facing, 120° FOV); right-side wide-angle camera (right-facing, 120° FOV); rear-right camera (rear-right-facing, 70° FOV); rear telephoto camera (backward-facing, 30° FOV); rear-left camera (rear-left-facing, 70° FOV); left-side wide-angle camera (left-facing, 120° FOV); front telephoto camera (forward-facing, 30° FOV); front fisheye camera (forward-facing, 200° FOV); left-side fisheye camera (left-facing, 200° FOV); right-side fisheye camera (right-facing, 200° FOV); rear fisheye camera (backward-facing, 200° FOV).

The description below is for the front wide-angle camera mounted on the vehicle. This camera is facing forward and has a 120° field of view:
\end{Verbatim}
\end{autopromptbox}
\end{figure}

\begin{figure}[!t]
\begin{autopromptbox}[label={box::auto_scene_caption}]{Scene caption excerpt: front-wide camera}
\begin{Verbatim}[fontsize=\footnotesize,breaklines=true,breaksymbolleft={}]
[0.0s-1.0s]: The view is a first-person forward-facing dashcam perspective from inside a vehicle, with the light-colored hood and a small hood ornament visible in the immediate foreground. The vehicle travels on a paved asphalt road initially marked with solid yellow lines along both edges of the travel lane, flanked by tall concrete retaining walls topped with metal safety railings on both sides; a lattice-style electricity pylon rises above the top of the right retaining wall at the start of the segment. A black passenger car drives directly ahead in the same lane, maintaining a consistent following distance. On the left side, an elevated parallel roadway runs behind the concrete barrier, carrying white box trucks and other vehicles; a blue square regulatory sign with a red diagonal stripe over a white straight-ahead arrow is mounted on the left barrier, with a round silver-backed traffic sign positioned below it. On the right retaining wall, a matching blue square regulatory sign with a red diagonal stripe over a white right-curving arrow is mounted above a circular no-entry sign (red border, white field, black horizontal bar). Tall curved streetlight poles line both sides of the road, and rolling green tree-covered hills are visible in the far distance under a clear, cloudless bright blue sky.
[1.0s-4.5s]: As the vehicle moves steadily forward, the initial yellow edge lines transition to standard white dashed lane dividers between travel lanes and solid white edge lines. The initial signs on both barriers pass out of the frame as the vehicle progresses. Red-and-white striped construction delineator posts appear spaced evenly along the top of the left concrete barrier, separating the main travel lane from the adjacent left roadway. Multiple vehicles become visible traveling in the opposite direction on the left parallel road: a dark gray hatchback, a small red compact car, and additional passenger cars. A large metal gantry structure holding a digital display comes into view on the left before passing out of frame. On the right side, a two-story building with white plaster walls, red brick trim, multiple chimneys, and a sloped roof becomes visible behind the retaining wall, followed by a large roadside billboard displaying printed imagery and unreadable text. The black car ahead remains centered in the travel lane, continuing forward at a steady speed.
[4.5s-7.5s]: The vehicle continues forward past the end of the enclosed retaining wall section. The right-side retaining wall gives way to a grassy roadside verge lined with metal guardrail, and a blue circular mandatory straight-ahead sign (white upward arrow on blue background) appears on the right shoulder. White painted chevron merge markings begin appearing on the asphalt of the right portion of the road, indicating an upcoming lane merge or exit area. Additional opposite-direction traffic passes on the left parallel road, including a black van, a white box truck, and more passenger cars. Small white numbered markers, including one marked "13", are affixed to the face of the left concrete barrier at regular intervals. The distant green hills remain visible, with several tall electricity pylons standing on the hilltops under the clear sky.
\end{Verbatim}
\end{autopromptbox}
\end{figure}

\begin{figure}[!t]
\begin{autopromptbox}[label={box::auto_baseline_prompt}]{Cosmos-Transfer2.5 prompt: front-wide camera}
\begin{Verbatim}[fontsize=\footnotesize,breaklines=true,breaksymbolleft={}]
The video captures a daytime driving scene under clear, sunny weather conditions. The ego vehicle travels along a multi-lane highway, maintaining a steady pace in the rightmost lane. Ahead, a black SUV drives in the same lane, keeping a consistent distance from the ego vehicle. To the left, there is a concrete barrier separating the opposing traffic lanes, marked with red-and-white striped poles for visibility. On the right side of the road, a stone wall runs parallel to the highway, topped with a metal railing.

As the ego vehicle progresses, it passes several streetlights and utility poles lining the road. In the distance, power lines stretch across the horizon, adding depth to the landscape. The surrounding environment includes patches of greenery and some buildings visible beyond the walls. The road surface appears well-maintained, with clear lane markings guiding the flow of traffic. No significant changes in speed or direction are observed from either the ego vehicle or the lead vehicle throughout the sequence.
\end{Verbatim}
\end{autopromptbox}
\end{figure}
\endgroup
\ifhmode\par\fi\endgroup

%% file: appendix/auto_stage0_training.tex
\phantomsection
\label{sec::auto_modality_sampling}
\paragraph[\kern0pt Modality-separate sampling for Stage 0.]{{Modality-separate sampling for Stage 0.}}

\begingroup{}
The modality-separate phase samples 480p RGB, 720p RGB, and LiDAR examples with probabilities 0.15, 0.15, and 0.70, respectively.
RGB examples select one of the 11 cameras uniformly.
LiDAR examples use the front-wide caption.
\ifhmode\par\fi\endgroup

\phantomsection
\label{sec::auto_bidirectional_schedule}
\paragraph[\kern0pt Optimization and initialization.]{{Optimization and initialization.}}

\begingroup{}
We use global batches of 16 examples for modality-separate training, 32 for short-sequence joint training, and 16 for long-sequence joint training.
In each phase, the learning rate increases linearly to the peak specified in \cref{tab::auto_training_summary} over the first 2,000 updates, then decreases linearly to $10^{-5}$.

Training starts from Cosmos3-Nano with newly initialized LiDAR projections and sensor embeddings.
The joint phases build on the preceding phase's exponential moving average (EMA) weights, carrying forward the learned modality projections and sensor embeddings.
Each phase starts with fresh optimizer state and restarts the learning-rate schedule, update counter, and EMA averaging.
\ifhmode\par\fi\endgroup

%% file: appendix/auto_training_configuration.tex
\begingroup{}
\begin{table}[!t]
\centering\reporttablestyle
\caption{\textbf{Post-training configuration of the Cosmos3-Nano-Sim-Auto model.}
Frame counts are per camera view after sensor alignment, and the upper bounds in Stages 1 and 2 describe the longest sampled clips.
Clip lengths include the initial RGB frames and LiDAR sweeps used for conditioning.
Stage 3 iterations count alternating student and fake-score updates.
The long phase continues the short phase's counter from 1,500 to 1,800.}
\label{tab::auto_training_configuration}
\begin{tabularx}{\linewidth}{L{3.5cm}CCCC}
\toprule
& \textbf{Stage 1} & \textbf{Stage 2} & \multicolumn{2}{c}{\textbf{Stage 3: SF-DMD}}\\
\cmidrule(lr){4-5}
\textbf{Setting} & \textbf{TF} & \textbf{TF-dCM} & \textbf{Short} & \textbf{Long}\\
\midrule
RGB frames per view & $\leq397$ & $\leq397$ & 93 & 397\\
Clip duration (seconds) & $\leq13.2$ & $\leq13.2$ & 3.1 & 13.2\\
RGB latent frames per view & $\leq100$ & $\leq100$ & 24 & 100\\
LiDAR sweeps & $\leq133$ & $\leq133$ & 31 & 133\\
Causal chunks & $\leq34$ & $\leq34$ & 9 & 34\\
GB200 GPUs & 256 & 768 & 512 & 256\\
Training iterations & 3,200 & 900 & 1,500 & 300\\
Student learning rate & $10^{-5}$ & $2\times10^{-6}$ & $10^{-6}$ & $10^{-6}$\\
Fake-score learning rate & --- & --- & $2\times10^{-7}$ & $2\times10^{-7}$\\
\bottomrule
\end{tabularx}
\end{table}
\ifhmode\par\fi\endgroup

%% file: appendix/auto_causal_training.tex
\begingroup{}
The three causal stages use the data and sensor rates in \cref{sec::auto_data}, with resolution sampling and conditioning dropout as in \cref{sec::auto_training}.
To fit the training memory budget, examples containing all 11 cameras simultaneously use 480p.
At 720p, each training example uses fewer cameras, but the sampled combinations collectively cover all 11 camera views.
At inference, the trained model can generate all 11 views together at 720p.
Each clip supplies 0, 1, or 2 observed RGB latent frames per view with probabilities 0.7, 0.2, and 0.1, paired with 0, 1, or 3 observed LiDAR sweeps, corresponding to T2V, I2V, and V2V.
In the synchronized chunk layout of \cref{sec::auto_causal_modeling}, the initial chunk contains 1 RGB latent frame per camera and 1 LiDAR sweep.
When a training sequence ends with an incomplete causal chunk, we retain the remaining aligned RGB frames and LiDAR sweeps for training.
\ifhmode\par\fi\endgroup

\begingroup{}
\label{sec::auto_training_objectives}
Training noise levels are shared across the views, sensors, and chunks of each clip.
TF-dCM computes the loss in \cref{sec::consistency_distillation} for each camera or LiDAR latent sequence, then averages these losses within each modality.
Clean control sequences contribute zero loss but are included in the sequence count used for averaging.
For SF-DMD, the teacher-residual normalizer and losses are computed separately for RGB and LiDAR, using only generated latents.
For each modality, the student surrogate loss and auxiliary loss are each averaged over channels, spatial locations, and generated latent frames across the full rollout, including all camera views for RGB.
We combine the RGB and LiDAR losses with weights $1:2$.
During chunkwise replay, each chunk uses the same full-rollout normalization, so summing the chunk losses reproduces the full-rollout average.
\ifhmode\par\fi\endgroup

\begingroup{}
\paragraph{Stage 1: causal teacher forcing.}
We use replay-based teacher forcing with gradients through the clean history representations (\cref{sec::teacher_forcing}).

\paragraph{Stage 2: teacher-forcing discrete consistency distillation.}
We follow the TF-dCM objective and teacher update in \cref{sec::consistency_distillation}, using the guidance scale in \cref{tab::auto_training_summary}.
The teacher's CFG reference branch replaces each camera's scene caption with the standard structured negative prompt,\footnote{See \href{https://github.com/NVIDIA/cosmos-framework/blob/main/cosmos_framework/inference/defaults/neg_prompts.json}{\nolinkurl{github.com/NVIDIA/cosmos-framework/blob/main/cosmos_framework/inference/defaults/neg_prompts.json}}.} while retaining the task prompt, per-camera metadata, structural controls, and recorded sensor history.
Both resolutions use noise levels $(1,15/16,5/6,5/8)$, with fresh Gaussian re-noising between evaluations and a final clean prediction.

\paragraph{Stage 3: self-forcing distribution matching distillation.}
Both score models retain bidirectional temporal visibility within each view and the teacher's past 0.4-second cross-view and cross-sensor window.
We use the final-step gradient estimator in \cref{sec::self_forcing}, with detached history.
The longer phase keeps persistent keys and values for completed chunks to avoid repeatedly encoding the same history.
\ifhmode\par\fi\endgroup

%% file: appendix/auto_training_infrastructure.tex
\subsubsection[\kern0pt Training Infrastructure]{{Training Infrastructure}}
\label{sec::auto_training_infrastructure}

\paragraph[\kern0pt Sequence packing.]{{Sequence packing.}}

\begingroup{}
The 1.5M-token packing budget in \cref{sec::auto_training_attention_implementation} includes RGB, WSM, LiDAR, HD-map, and language tokens.
For example, 7 720p views with 397 RGB frames per view, 133 LiDAR sweeps, and their corresponding controls occupy approximately 1.53M tokens before language conditioning.
Stages 1 and 2 can pack multiple clips per sequence, whereas Stage 3 uses 1 clip per sequence.
\ifhmode\par\fi\endgroup

\paragraph[\kern0pt Parallelism and memory.]{{Parallelism and memory.}}

\begingroup{}
We distribute each packed sequence across 32 GPUs with context parallelism.
On GB200 GPUs, the long-sequence configuration combines this with 64-way fully sharded data parallelism (FSDP)~\citep{zhao2023pytorchfsdp}.
Context parallelism partitions each packed sequence, while FSDP shards parameters, gradients, and optimizer state over overlapping rank groups.
We use BF16 computation and gradient reduction with FP32 master parameters.
Full activation recomputation reduces retained intermediate tensors at the cost of additional computation, without CPU activation or parameter offloading in this configuration.
\ifhmode\par\fi\endgroup

\paragraph[\kern0pt Input pipeline and checkpointing.]{{Input pipeline and checkpointing.}}

\begingroup{}
Asynchronous batch construction prepares the next packed batch while the model processes the current one, with bounded prefetching to limit host memory use.
Distributed checkpointing stages a consistent snapshot in CPU memory before writing it in a background process, allowing storage I/O to overlap subsequent training.
\ifhmode\par\fi\endgroup

%% file: appendix/auto_runtime.tex
\phantomsection
\label{sec::auto_runtime_implementation}
\subparagraph[\kern0pt Execution configuration.]{{Execution configuration.}}
\begingroup{}
With the generation settings in \cref{sec::auto_inference_performance}, each chunk performs 4 denoising steps followed by a clean pass that writes its keys and values for subsequent chunks.
The K/V cache keeps the prompt and the opening frame with its map as a permanent sink, 1 chunk of clean generated history, and the controls of that chunk and of the chunk being generated.
Together with the cross-sensor window in \cref{sec::multiview_attention}, this keeps the attention context fixed from the second chunk onward.
The GB200 nodes contain 4 GPUs, assigned to model ranks for generation or map ranks for control preparation.
\Cref{tab::auto_opt_inventory} lists the optimization settings evaluated in \cref{sec::auto_runtime_results}.
\ifhmode\par\fi\endgroup

\begingroup{}
\begin{table}[!t]
\centering\reporttablestyle
\caption{\textbf{Inference optimizations of the Cosmos3-Nano-Sim-Auto runtime.}
Option is the Interactive Drive launcher flag or, for the attention options that the launcher does not expose, the backend's rollout option, and Shipped gives the launcher's setting, where with LiDAR means on in LiDAR sessions.}
\label{tab::auto_opt_inventory}
\begin{tabularx}{\linewidth}{L{2.5cm}L{1.75cm}XL{2.85cm}L{1.6cm}}
\toprule
\textbf{Optimization} & \textbf{Stage} & \textbf{Mechanism} & \textbf{Option} & \textbf{Shipped}\\
\midrule
Compiled denoising & Denoise & The 4 denoising steps run under \texttt{torch.compile} with fixed shapes & {\texttt{--compile-\allowbreak denoise}} & on\\
Compiled control encoding & Control prep. & Compiled Wan 2.2 VAE encoder of the drawn WSM controls & {\texttt{--compile-\allowbreak control-\allowbreak encode}} & on\\
Compiled decoder & Decode & The TAEHV decoder runs under \texttt{torch.compile} & {\texttt{--decoder-\allowbreak compile}} & on\\
Parallel decoding & Decode & Each rank decodes its cameras and sends the pixels to rank 0 & {\texttt{--parallel-\allowbreak decode}} & on\\
Map GPUs & Control prep. & A map GPU per model rank prepares the next chunk's WSM controls, which lag the input by 1 chunk & {\texttt{--prefetch-\allowbreak controls}}, {\texttt{--prefetch-\allowbreak controls-\allowbreak gpu-\allowbreak offset}} & on for 4+ nodes\\
Context parallelism & All passes & Contiguous token range per rank; keys, values, and flow are all-gathered & ranks per launch & on\\
Tensor parallelism & All passes & Head-group sharding with an all-reduce per layer; excludes compiled denoising & {\texttt{COSMOS3\_\allowbreak NANO\_\allowbreak SIM\_\allowbreak AUTO\_\allowbreak TENSOR\_\allowbreak PARALLEL}} & off\\
Maskless attention & All passes & Flash attention per gathered key group, merged as in \cref{fig::auto_maskless_attention}, instead of a masked pass & {\texttt{maskless\_\allowbreak attention}} & on\\
Fused group merge & Eager passes & One kernel merges the key groups outside compiled denoising & {\texttt{fused\_\allowbreak attention\_\allowbreak merge}} & on\\
Fused RMSNorm & All passes & Fused kernel instead of the checkpoint's RMSNorm & {\texttt{accelerated\_\allowbreak rms\_\allowbreak norm}} & off\\
FlexAttention mask & All passes & Block-sparse masked pass instead of maskless attention & {\texttt{flex\_\allowbreak attention}} & off\\
LiDAR rank roles & Control prep., decode & Camera-free ranks render the 4 sweeps; the first runs the LiDAR codec & with {\texttt{--lidar}} & with LiDAR\\
LiDAR range culling & Control prep. & Sweep renderer culls returns outside the sensor range & {\texttt{--culled-\allowbreak lidar-\allowbreak renderer}} & with LiDAR\\
LiDAR TF32 codec & Control prep., decode & The LiDAR codec runs its matmuls in TF32 & {\texttt{--lidar-\allowbreak codec-\allowbreak tf32}} & with LiDAR\\
LiDAR contiguous projections & Control prep., decode & Contiguous inputs to the codec's linear layers & {\texttt{--lidar-\allowbreak contiguous-\allowbreak linears}} & with LiDAR\\
\bottomrule
\end{tabularx}
\end{table}
\ifhmode\par\fi\endgroup

\subparagraph[\kern0pt Compilation.]{{Compilation.}}
\begingroup{}
Rotary-embedding and gated-activation kernels are always compiled, in addition to the optional compiled stages in \cref{tab::auto_opt_inventory}.
The K/V cache update runs eagerly, and compiled denoising is disabled with tensor parallelism because its all-reduce would fall inside the compiled region.
Interactive Drive uses no CUDA-graph replay, decodes on the model ranks rather than on a separate decoder GPU, and runs the K/V cache update and decoding sequentially.
\ifhmode\par\fi\endgroup

\subparagraph[\kern0pt Parallel control preparation and decoding.]{{Parallel control preparation and decoding.}}
\begingroup{}
The runtime specializes the parallelism in \cref{sec::runtime_optimization} with two rank axes.
Context parallelism partitions the view-major query sequence into contiguous ranges and all-gathers keys, values, and predicted flow, leaving each rank with the full cache and chunk.
Optional tensor parallelism shards layers by key/value head group with an all-reduce per layer.
The launcher sets tensor parallelism to 1.
Cameras are assigned round-robin across ranks for drawing, encoding, and decoding, with decoded pixels collected on rank 0.
During prefill, the control tracks are encoded camera by camera across ranks.
A map GPU per model rank can draw and encode the next chunk's controls concurrently with denoising, introducing a 1-chunk control delay.
The model ranks still execute the control-token pass that writes the control keys and values (\cref{fig::auto_opt_stages}).
\ifhmode\par\fi\endgroup

%% file: appendix/auto_runtime_results.tex
\begingroup{}
\begin{table}[!t]
\centering\reporttablestyle
\caption{\textbf{Cosmos3-Nano-Sim-Auto chunk wall time and presented FPS on GB200, eager against the shipped optimizations.}
Each row compares configurations within the same GPU allocation.
Chunk wall time is the median of run medians of the recorded generation call on the primary rank, with the range across repeated runs in parentheses.
The eager and shipped runs of a row share one allocation, and the eager run of the 7-camera map layout uses all 16 GPUs as model ranks.
Change is the shipped median relative to the eager median, so $-30\%$ is a 30\% shorter chunk.
A dash marks a configuration without an eager run.
FPS is the median rate presented to the client and Mem.\ the median peak device memory of the primary rank's GPU over the 1 to 7 shipped runs of the row, out of 184\,GiB per GPU.}
\label{tab::auto_runtime_ladders}
\setlength{\tabcolsep}{3pt}
\begin{tabular*}{\linewidth}{@{\extracolsep{\fill}}clclrrrcc@{}}
\toprule
\textbf{Views} & \textbf{Mode} & \textbf{GPUs} & \textbf{Layout} & \multicolumn{2}{c}{\textbf{Chunk wall time (ms)}} & \textbf{Change} & \textbf{FPS} & \textbf{Mem.}\\
\cmidrule(lr){5-6}
 & & & & eager & shipped & & & GiB\\
\midrule
\multicolumn{9}{l}{\textit{Single GPU}}\\
1 & RGB & 1 & 1 model & 524 (512--535) & 364 (362--423) & $-$30\% & 32.6 & 33.6\\
4 & RGB & 1 & 1 model & 1{,}457 & 1{,}166 & $-$20\% & 10.2 & 37.6\\
\multicolumn{9}{l}{\textit{Multi-GPU}}\\
1 & RGB & 2 & 1 model + 1 map & --- & 315 & --- & 38.1 & 31.2\\
4 & RGB & 8 & 8 model & 783 & 483 & $-$38\% & 24.6 & 41.4\\
4 & RGB & 16 & 8 model + 8 map & --- & 408 (377--440) & --- & 28.9 & 40.9\\
4 & RGB + LiDAR & 8 & 8 model & --- & 651 & --- & 18.1 & 51.6\\
7 & RGB & 16 & 16 model & 897 & 609 & $-$32\% & 19.5 & 49.0\\
7 & RGB & 16 & 8 model + 8 map & 950 (936--1{,}014) & 583 (561--597) & $-$39\% & 20.5 & 48.2\\
7 & RGB & 32 & 16 model + 16 map & --- & 511 (510--511) & --- & 23.2 & 48.2\\
7 & RGB + LiDAR & 16 & 16 model & 1{,}159 & 701 (693--714) & $-$39\% & 16.7 & 61.6\\
\bottomrule
\end{tabular*}
\end{table}
\ifhmode\par\fi\endgroup

\subsubsection[\kern0pt Runtime Performance]{{Runtime Performance}}
\label{sec::auto_runtime_results}

\phantomsection
\label{sec::auto_runtime_measurements}
\paragraph[\kern0pt Matched runtime comparisons.]{{Matched runtime comparisons.}}
\begingroup{}
Which optimizations reduce streaming latency under a fixed GPU allocation?
We compare the execution choices in \cref{sec::auto_runtime_implementation}, varying compilation, attention backend, and the division of GPUs between model and control processing.
\ifhmode\par\fi\endgroup

\subparagraph[\kern0pt Measurement protocol.]{{Measurement protocol.}}
\begingroup{}
Each measurement series runs its configurations back to back on the same GPUs to control for node-to-node variability.
Most repeated configurations varied by up to 8\%, with wider ranges for the single-camera slow mode, the 4-camera map layout, and the 7-camera configurations on 32 GPUs.
We interpret differences against the observed repeat range and do not attribute changes below 8\% to an optimization.
\ifhmode\par\fi\endgroup

\subparagraph[\kern0pt Single GPU.]{{Single GPU.}}
\begingroup{}
A single GB200 exceeds the 30-FPS video rate at the median for 1 camera, while 4 cameras require more computation than it can sustain at that rate (\cref{tab::auto_runtime_ladders}).
For 1 camera, adding model ranks provides little benefit within the repeat range, whereas assigning a second GPU to control preparation reduces latency more clearly (\cref{fig::auto_runtime_ladders}).
A slower mode appeared in some repeats on one node but did not recur on another, motivating comparisons on the same allocation.
\ifhmode\par\fi\endgroup

\subparagraph[\kern0pt Multi-GPU.]{{Multi-GPU.}}
\begingroup{}
Control prefetching helps when enough model ranks remain to process the sensor tokens efficiently.
On 16 GPUs, splitting ranks equally between generation and control preparation improves 4-camera RGB latency, while the 7-camera layouts overlap within the repeat range (\cref{fig::auto_runtime_ladders}).
With only 8 GPUs, diverting half the ranks to controls instead slows generation.
On 32 GPUs, the split layout improves 7-camera latency over the 16-GPU split layout, whereas adding model ranks alone does not produce a monotonic speedup.
An eager 7-camera comparison also found that tensor parallelism of 2 lengthens chunks by 28\% relative to 1, supporting the launcher's setting.
\ifhmode\par\fi\endgroup

\begingroup{}
Joint LiDAR generation changes this allocation trade-off by adding sensor tokens and control processing.
For 7 cameras with LiDAR, dedicating all 16 GPUs to model ranks outperforms an equal model/map split, while the tested 32-GPU layouts offer no improvement beyond the repeat range.
The runtime therefore retains all GPUs as model ranks for LiDAR sessions.
Among the LiDAR-specific options, range culling has the largest measured effect: disabling it increases chunk latency by 41\%.
These results favor selecting GPU roles for the sensor workload rather than increasing parallelism uniformly.
\ifhmode\par\fi\endgroup

\begingroup{}
\begin{figure}[!t]
\centering
\includegraphics[width=\linewidth]{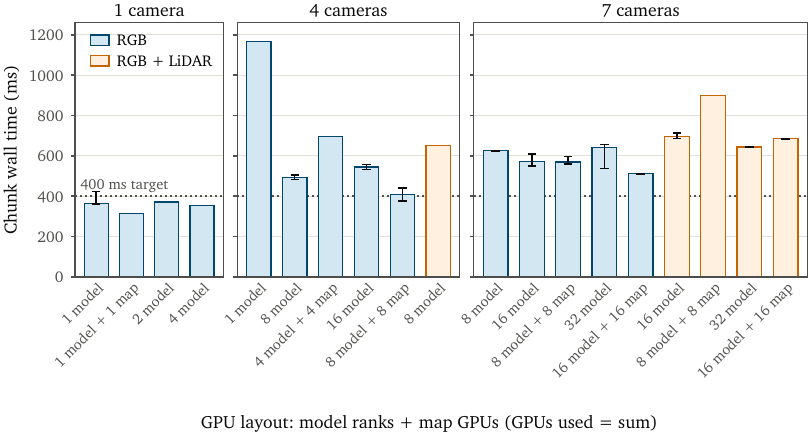}
\caption{\textbf{Cosmos3-Nano-Sim-Auto chunk wall time by GPU count and layout on GB200.}
Shipped flags, from the same-allocation ladders of \cref{tab::auto_runtime_ladders}; bars give the median of the run medians of a configuration and whiskers their range where the configuration was repeated.
Labels give the GPU layout as model ranks plus map GPUs, where map GPUs draw and encode controls on separate GPUs, so the GPU count of a configuration is their sum; blue bars are RGB and orange bars RGB + LiDAR.}
\label{fig::auto_runtime_ladders}
\end{figure}
\ifhmode\par\fi\endgroup

\subparagraph[\kern0pt Measured attention variants.]{{Measured attention variants.}}
\begingroup{}
Maskless attention improves complete-chunk latency across the tested eager and compiled configurations.
We compare the flash scaled dot-product attention backend with fused group merging against block-sparse FlexAttention and dense masked cuDNN, placing each variant between 2 default runs on the same GPUs.
The comparison covers both attention execution and group merging, since the masked paths replace the separate groups with a single pass.
For 1 camera, the eager masked paths increase latency by 33\% and 44\%, respectively, while dense masking increases eager 7-camera latency by 143\% on 16 model ranks.
The gap is largest with compiled denoising on 8 model ranks and 8 map GPUs, where FlexAttention and dense masking increase latency by 195\% and 262\%.
Compilation also reduces the value of a separate fused RMSNorm kernel: its 12--15\% gains in eager execution fall within the repeat range once denoising is compiled.
The runtime uses maskless attention with fused group merging and retains the checkpoint's RMSNorm.
\ifhmode\par\fi\endgroup

\subparagraph[\kern0pt Stage and startup timing.]{{Stage and startup timing.}}
\begingroup{}
Compilation trades faster steady-state generation for longer startup, even with a warm compiler cache.
Process-to-first-frame time is 42\,s for 1 camera on 1 GPU and 74\,s for 7 cameras on 8 model ranks + 8 map GPUs, compared with 21 and 37\,s for eager execution on 1 GPU and 16 model ranks, respectively.
First launches are more expensive: the 4-camera configurations on 1 and 16 GPUs take 146 and 168\,s, compared with 38--55 and 64\,s on later launches, consistent with compilation for new shapes.
During prefill, the transformer runs eagerly while the control encoder executes its first compiled call.
It takes 13.3\,s for 1 camera with compilation versus 3.4\,s without it, and 19.4\,s for 7 cameras on 16 GPUs.
\ifhmode\par\fi\endgroup

\begingroup{}
After denoising is accelerated, control preparation and clean-history updates remain substantial costs (\cref{fig::auto_opt_stages}).
Each costs approximately 2--4 compiled denoising steps, making them useful targets for further optimization.
Parallel decoding is comparatively inexpensive at 8--10\,ms per chunk, versus 26\,ms when 1 GPU decodes 4 cameras.
Approximately 7\,ms lies outside the four measured stages.
Peak device memory is reported with the corresponding operating points in \cref{tab::auto_runtime_ladders}.
\ifhmode\par\fi\endgroup

\input{appendix/auto_runtime_historical.tex}

%% file: appendix/auto_runtime_historical.tex
\phantomsection
\label{sec::auto_runtime_historical}
\label{sec::auto_runtime_profiling}
\subparagraph[\kern0pt Latency variability.]{{Latency variability.}}
\begingroup{}
To characterize latency variation and small-GPU joint generation, \cref{tab::auto_runtime_details} supplements the matched comparisons with a separate series of FlashDreams 0.2 RC1 measurements on GB200.
The runs use the residential scene and the resolution, chunk length, and sampling steps in \cref{sec::auto_inference_performance}.
Configurations were repeated across allocations, with single-camera RGB run medians spanning 358--510\,ms across nodes.
This variation motivates measuring optimization gains back to back on the same GPUs.
\ifhmode\par\fi\endgroup

\begingroup{}
\begin{table}[!t]
\centering\reporttablestyle
\caption{\textbf{Chunk latency quantiles for selected Cosmos3-Nano-Sim-Auto configurations on GB200.}
The p50 and p95 columns report the median across runs of each run's corresponding chunk-latency quantile.
Chunk counts sum steady-state observations after discarding 10 warm-up chunks per run, excluding Nsight runs.
All GPUs serve as model ranks, with compilation and parallel decoding enabled.
Wall time covers the generation call on the primary rank and includes output visualization in joint RGB--LiDAR runs.}
\label{tab::auto_runtime_details}
\setlength{\tabcolsep}{4pt}
\begin{tabular*}{\linewidth}{@{\extracolsep{\fill}}cclrrrr@{}}
\toprule
\textbf{Views} & \textbf{GPUs} & \textbf{Mode} & \multicolumn{2}{c}{\textbf{Chunk wall time (ms)}} & \textbf{Chunks} & \textbf{Runs}\\
\cmidrule(lr){4-5}
 & & & p50 & p95 & & \\
\midrule
1 & 1 & RGB & 433 & 453 & 700 & 5\\
1 & 1 & RGB + LiDAR & 1{,}535 & 1{,}690 & 140 & 1\\
1 & 2 & RGB + LiDAR & 927 & 1{,}029 & 140 & 1\\
4 & 8 & RGB & 503 & 535 & 280 & 2\\
4 & 8 & RGB + LiDAR & 655 & 693 & 140 & 1\\
7 & 16 & RGB & 598 & 627 & 280 & 2\\
7 & 16 & RGB + LiDAR & 742 & 825 & 140 & 1\\
\bottomrule
\end{tabular*}
\end{table}
\ifhmode\par\fi\endgroup

\subparagraph[\kern0pt Kernel profiling and scaling bottlenecks.]{{Kernel profiling and scaling bottlenecks.}}
\begingroup{}
Nsight traces of companion runs help explain why adding model ranks does not always reduce latency.
For 1 camera on 1 GPU, the GPU is busy for 43\% of the traced step, with attention and matrix multiplications dominating its active time.
The substantial time outside GPU kernels points to host-side launch and bookkeeping as optimization targets.
With 7 cameras on 16 GPUs, GPU activity rises to 72\%, and collectives account for 17\% of GPU time.
For joint RGB--LiDAR on 16 model ranks, collectives account for 54\% of GPU time, highlighting communication as a larger cost for the additional sensor workload.
\ifhmode\par\fi\endgroup

\begingroup{}
The single-camera joint profile on 2 GPUs also exposes dependencies beyond transformer execution.
The second rank rasterizes, encodes, and decodes LiDAR sweeps concurrently with the primary rank, which waits for control tokens and decoded sweeps when needed.
These dependencies complement the stage-level costs in \cref{fig::auto_opt_stages} and motivate coordinating sensor processing with generation.
Tracing adds 6\% to step wall time for 7 RGB cameras on 16 GPUs and 42\% for 1 camera on 1 GPU.
\ifhmode\par\fi\endgroup

\FloatBarrier

%% file: appendix/auto_inference.tex
\begingroup{}
\subsubsection{Inference Settings}
\label{sec::auto_inference_settings}
This section specifies the inference and prompt configurations used for the evaluations in \cref{sec::auto_results}.
\ifhmode\par\fi\endgroup

\phantomsection
\label{sec::auto_eval_cache}
\begingroup{}
Across both benchmarks, Cosmos3-Nano-Sim-Auto uses a rolling K/V cache of 2 chunks: the initial chunk, retained as an attention sink, and the most recent completed chunk.
The one-hour rollout in \cref{fig::auto_long_rollout} uses the same cache configuration.
\ifhmode\par\fi\endgroup

\begingroup{}
\paragraph{Driving-Sim-Eval.}
Cosmos3-Nano-Transfer-Auto and Cosmos3-Nano-Sim-Auto jointly generate RGB and LiDAR, with 297 RGB frames per camera at 720p and 30 FPS across 7 non-fisheye views.
Both use timestamped per-camera captions produced by the same captioner as their training data (\cref{sec::auto_captioning}).
Cosmos3-Nano-Transfer-Auto uses guidance 3, while Cosmos3-Nano-Sim-Auto uses 4 denoising steps without classifier-free guidance (CFG).
The baseline Cosmos-Transfer2.5 also generates 297 frames per camera at 720p and 30 FPS using the same 7 views, with guidance 3 and autoregressive generation in its native 29-frame chunks.
Dense, untimestamped paragraphs yield its highest generation quality, with one caption per camera covering the first 10 seconds of the clip.
Boxes~\ref{box::auto_scene_caption} and~\ref{box::auto_baseline_prompt} illustrate the timestamped and untimestamped caption formats, respectively.
\ifhmode\par\fi\endgroup

\begingroup{}
\paragraph{Traffic-Scenario-Eval.}
Cosmos3-Nano-Transfer-Auto uses guidance 1.5 and jointly refined, environment-enhanced, timestamped per-camera captions.
Cosmos3-Nano-Sim-Auto uses the iteration-1,800 checkpoint with 4 denoising steps, no CFG, and timestamped per-camera captions.
Both generate at 30 FPS.
The baseline Cosmos-Transfer2.5 uses guidance 3 and its native 10-FPS setting, which gave its best results in our tests.
We tested all caption variants for the baseline and selected the original benchmark captions, which yielded the highest generation quality.
\ifhmode\par\fi\endgroup

\begingroup{}
For Cosmos3-Nano-Transfer-Auto evaluation on Traffic-Scenario-Eval, we refine the timestamped per-camera captions to express the requested weather, time of day, road condition, and region through concrete visual details.
We refine all seven captions jointly using their WSM controls, preserving actor appearances, trajectories, and timestamps while keeping environmental descriptions consistent across views.
These practices are summarized in \cref{sec::auto_prompt_upsampling} and provided as a prompt upsampling skill in Cosmos Framework.
The negative prompt used for this evaluation is reproduced below.
\ifhmode\par\fi\endgroup

\begingroup{}
\begingroup
\tcbset{
  breakable,
  colframe=nvidiagreen!75,
  colback=black!2,
  colbacktitle=nvidiagreen!75,
  coltitle=white,
  coltext=black,
  fonttitle=\normalfont\small,
  boxrule=0.4pt,
  arc=0pt,
  left=8pt,right=8pt,top=4pt,bottom=4pt
}
\begin{tcolorbox}[title={Negative prompt}]
\begin{Verbatim}[fontsize=\footnotesize,breaklines=true,breaksymbolleft={}]
The driving-camera video has unstable shading, with faint, dirty-looking blotches drifting across the sky and other smooth surfaces. These patches appear and disappear between frames while local brightness fluctuates independently of scene motion and lighting. Surface textures crawl, fine details shimmer, and object edges wobble. Dense fog or heavy rain and road spray severely obscure nearby vehicles and lane markings, while persistent water films on the lens conceal large portions of the scene. Motion is excessively smeared, leaving long trails and ghosted outlines that erase recognizable detail even at moderate speeds. The image remains unnaturally soft, vehicles deform, road geometry drifts, and exposure changes abruptly without corresponding changes in illumination. The scene has a computer-generated, game-rendered appearance, with plastic-looking materials, unnaturally saturated colors, and artificially exaggerated contrast. People and objects morph, merge, split, or change identity discontinuously between frames. Human limbs distort, fuse, or detach. People or objects abruptly appear or disappear in unobstructed areas without entering or leaving the camera view or passing behind an occluder.
\end{Verbatim}
\end{tcolorbox}
\endgroup
\ifhmode\par\fi\endgroup

\begingroup{}
\paragraph{Weather-prompt study.}
For the study in \cref{sec::auto_lidar_weather}, each rollout contains 93 RGB frames per camera at 480p and 30 FPS and 31 LiDAR sweeps at 10\,Hz, using 35 UniPC steps, guidance 6, shift 10, and a fixed negative prompt.
\ifhmode\par\fi\endgroup

%% file: appendix/auto_metrics.tex
\begingroup{}
\subsubsection{Evaluation Metric Definitions}
\label{sec::auto_metric_definitions}
This section gives the definitions and scoring procedures for the metrics summarized in \cref{sec::auto_evaluation_metrics}.
\ifhmode\par\fi\endgroup

\begingroup{}
\paragraph{RGB video quality metrics.}
We measure overall fidelity, temporal consistency, and multi-view consistency.
Single-view metrics are reported as the average across the 7 non-fisheye views.
Fisheyes are excluded for comparison against Cosmos-Transfer2.5.

Frame FID~\citep{heusel2017fid}, I3D FVD~\citep{unterthiner2019fvd}, StyleGAN-V FVD~\citep{stylegan_v}, and the DOVER~\citep{dover} technical score cover overall fidelity.

Adjacent cosine mean (ACM) measures temporal visual consistency similarly to VBench~\citep{huang2023vbench}.
We extract CLIP ViT-B/32~\citep{radford2021clip} features $\mathbf{f}_t$ for frames $t=0,\dots,T-1$, then average feature similarities to the preceding and first frames:
\begin{equation}
  \mathrm{ACM}=\frac{1}{T-1}\sum_{t=1}^{T-1}\frac{1}{2}\big[\cos^{+}(\mathbf{f}_t,\mathbf{f}_{t-1})+\cos^{+}(\mathbf{f}_t,\mathbf{f}_0)\big],
\end{equation}
where $\cos^{+}(\mathbf{a},\mathbf{b})=\max\big(0,\cos(\mathbf{a},\mathbf{b})\big)$ is cosine similarity clamped at zero.

We measure pixelwise multi-view consistency primarily by cross-view Sampson error (CSE)~\citep{sampson1987}.
For each pair of overlapping views $(v,w)$ at time $t$, we extract keypoints with SuperPoint~\citep{detone2018superpoint} and match them with LightGlue~\citep{lightglue2023} to obtain point correspondences $M_{v,w}^t=\{(\mathbf{x},\mathbf{x}')\}$.
Undistorted keypoints are used to fit a fundamental matrix $F$ with RANSAC and the eight-point algorithm.
Total cross-view Sampson error for a single view pair $(v,w)$ is calculated as the average square-root Sampson error over all frames $T$.
\begin{equation}
  \mathrm{CSE}_{v,w}=\frac{1}{T}\sum_{t=1}^{T}\frac{1}{|M_{v,w}^t|}\sum_{(\mathbf{x},\mathbf{x}')\in M_{v,w}^t}d_F(\mathbf{x},\mathbf{x}'),
\end{equation}
where $d_F(\mathbf{x},\mathbf{x}')$ is the Sampson distance~\citep{sampson1987}, the square root of the Sampson error of the correspondence under $F$.
We report CSE-2 and CSE-5, which are the average $\mathrm{CSE}_{v,w}$ over 2 and 5 camera pairs, respectively.
CSE-2 involves the (front, front-left) and (front, front-right) cameras, while CSE-5 extends that with the (front, front telephoto), (rear-left, rear telephoto), and (rear-right, rear telephoto) pairs.

Temporal Sampson error (TSE) applies the same Sampson distance within one camera stream to measure camera consistency across time.
For each pair of consecutive frames in a view, it matches keypoints, fits a fundamental matrix to the matches, and averages $d_F$ over the matches.
The result is similarly averaged over frame pairs and then over the 7 views.
TSE requires care because a small baseline between frames can make $F$ ill-posed, and the scene may change between frames.
\ifhmode\par\fi\endgroup

\begingroup{}
\paragraph{LiDAR metrics.}
LiDAR evaluation covers distribution fidelity, temporal consistency, and object faithfulness to WSM controls.
Real scans are projected to the generated range-view representation (\cref{sec::multiview_attention}) for comparable point-density patterns.

Following LiDARGen~\citep{zyrianov2022lidargen}, bird's-eye-view (BEV) histograms count points in 2\,m grid cells covering $\pm100$\,m, using sensor ranges of 5--100\,m.
Global BEV JSD measures Jensen--Shannon distance~\citep{lin1991divergence} between generated and real histograms summed over all frames.
Local BEV JSD instead sums within each clip and averages the distances across clips, retaining scene-specific differences that dataset-level aggregation can obscure.
BEV MMD$^2$ (squared maximum mean discrepancy)~\citep{gretton2012kernel} compares normalized frame histograms within each clip using a Gaussian kernel with $\sigma=0.5$, then averages over clips.

For feature-space comparisons, we average-pool Utonia~\citep{zhang2026utonia} point-cloud features into one vector per frame.
Across all frames, we compute Fr\'echet Utonia distance (FUD), analogous to FID~\citep{heusel2017fid}, and unbiased Utonia MMD$^2$ with a cubic polynomial kernel, analogous to kernel inception distance~\citep{binkowski2018demystifying}.

Chamfer temporal consistency (CTC) measures Chamfer distance between consecutive scans aligned using ground-truth ego poses.
Vehicle CTC applies the same measure to tracked vehicle crops in object-local coordinates.
Both are reported as ratios to the corresponding measures on real LiDAR, with an ideal value of~1.

Object support recall is the fraction of labeled cuboids with at least 30 real points that also contain at least 10 generated points, covering vehicles, people, and animals.
Detection F1 evaluates a pre-trained PointPillars~\citep{lang2019pointpillars} detector on generated LiDAR, matching same-class detections to ground-truth objects at planar center distances below~3\,m.
\ifhmode\par\fi\endgroup

\begingroup{}
\paragraph{Traffic-object adherence.}
On Traffic-Scenario-Eval, we measure how well the generated objects follow the WSM controls with AutoGrader, which uses the object correspondence checks of Cosmos Evaluator~\citep{nvidia2026cosmosevaluator}.
For each WSM object, the checks project its cuboid into each camera and compare it with a semantic segmentation of the generated frame, giving a per-frame correspondence score.
We transform the correspondence scores into detection counts with a threshold of 0.5: an object whose mean correspondence score is at least 0.5 counts as detected, and otherwise as missing.
Segmented objects that match no WSM object count as hallucinated.
We report F1 for vehicles, pedestrians, and cyclists, computed from these counts summed over all cameras and scenarios.
\ifhmode\par\fi\endgroup

\begingroup{}
\paragraph{Human evaluation rubric.}
\label{sec::auto_hue_rubric}
Video-quality review covers the full frame, including the sky and distant objects, and checks textures, colors, lighting, and geometric stability.
Examples of minor artifacts allowed at score 2 include brief flicker, slight softness, and small lighting inconsistencies.
Fused vehicles, floating objects, or intersecting bodies receive score 1 even when visible only briefly.
Scene-element questions assess the presence, placement, appearance, and stability of the specified lanes, signals, signs, pedestrians, and bicyclists.
Cross-view questions rate consistency of the same objects' identity, geometry, location, and motion: 3 for agreement, 2 for minor discrepancies, and 1 for major contradictions.
\ifhmode\par\fi\endgroup

%% file: appendix/auto_development.tex
\subsubsection[\kern0pt Ablation Study]{{Ablation Study}}
\label{sec::auto_development}

\paragraph[\kern0pt Training-stage quality comparison.]{{Training-stage quality comparison.}}
\begingroup{}
Which stages recover generation quality after converting the bidirectional model into a causal, few-step generator?
We compare the bidirectional teacher, stage 1 TF, stage 2 TF-dCM, stage 3 SF-DMD short, and stage 3 SF-DMD long checkpoints (\cref{sec::auto_training}) on the same 50-scene validation set (\cref{tab::auto_stage_comparison}).
All evaluations use seven cameras, 720p RGB, 10-second clips, and no rolling KV cache.
The bidirectional teacher and stage 1 TF use 35 UniPC steps with guidance 3, while stage 2 TF-dCM and both stage 3 SF-DMD checkpoints use 4 SDE steps with no guidance.
Step distillation in stage 2 TF-dCM reduces visual fidelity and geometric consistency relative to stage 1 TF.
Stage 3 SF-DMD recovers much of the lost generation quality, and its long-sequence phase further improves FID, FVD, temporal consistency, and cross-view consistency.
These results support retaining both SF-DMD phases in the final training recipe.
\ifhmode\par\fi\endgroup

\input{figures/auto/studies/three_stage.tex}

\paragraph[\kern0pt Cross-view temporal visibility.]{{Cross-view temporal visibility.}}
\begingroup{}
Which cross-view temporal window best preserves scene consistency?
We compare three cross-view visibility rules from the same bidirectional starting point using the same training budget.
A permits the closest other-view latent frame, B permits the inclusive preceding 0.4 seconds, and C adds the current jointly denoised noisy chunk to B.
Although C may attend to future noisy frames within the chunk, it does not expose future chunks or future clean targets.
The rule applies in both training and generation.
\Cref{tab::auto_time_window} scores the full 10-second outputs on a subset of 10 validation scenes.
Pose denotes rotation error in radians, and DOVER denotes technical quality.
The preceding 0.4-second window yields the best overall result, substantially reducing cross-view error.
Adding the current noisy chunk offers no further benefit.
We therefore use the preceding 0.4-second window for cross-view attention.
\ifhmode\par\fi\endgroup

\input{figures/auto/studies/time_window.tex}

\paragraph[\kern0pt Restricting cross-view attention.]{{Restricting cross-view attention.}}
\begingroup{}
Our cross-view attention allows each camera stream to attend to all other selected views.
A possible way to accelerate generation is to restrict these connections to neighboring cameras or cameras with overlapping fields of view, using the camera rig's spatial arrangement and FOVs.
\Cref{tab::auto_neighboring_overlap} compares this restricted scheme with dense all-view attention in stage 1 TF on the same validation set.
We observe no speedup with this scheme in our implementation, consistent with overhead from fragmented attention masks.
Despite slightly improved video quality and temporal consistency on par with dense all-view attention, the neighboring and overlapping-view scheme yields substantially worse cross-view consistency.
We therefore retain dense all-view attention.
\ifhmode\par\fi\endgroup

\input{figures/auto/studies/neighboring_overlap.tex}

\paragraph[\kern0pt Deduplicating same-view keys.]{{Deduplicating same-view keys.}}
\phantomsection
\label{sec::auto_dedup_ablation}
\begingroup{}
Does duplicated same-view evidence impair cross-view consistency?
When the same-stream and cross-sensor groups share keys, those keys contribute twice to the joint softmax.
Removing the query's own stream from the cross-sensor group improved cross-view consistency in our experiments, supporting the disjoint grouping used in the final model (\cref{sec::auto_attention_implementation}).
\ifhmode\par\fi\endgroup

\input{figures/auto/studies/rolling_kv.tex}

\paragraph[\kern0pt Memory and long rollouts.]{{Memory and long rollouts.}}
\phantomsection
\label{sec::auto_memory_ablation}
\begingroup{}
How much cached history is needed to sustain coherent long rollouts?
We vary the cache size $C$ and visual attention sink size $S$, measured in chunks of three latent frames, for our final model on an evaluation set of 94 videos that are around one minute long (\cref{fig::auto_rolling_kv}).
We use time-aligned source captions for each camera view.
Our model switches captions at their time boundaries without resetting the visual KV cache, while retaining four initial text tokens per view as language sinks.
The total cache size includes the visual sinks.
The $C=2,S=1$ setting achieves the best trade-off overall, motivating the two-chunk cache with one attention-sink chunk used in our evaluation.
At a fixed one-chunk sink, retaining more history worsens both distributional quality and cross-view consistency.
This result supports assigning the cache to an initial appearance anchor and recent motion, even though training includes sequences of up to 34 chunks (\cref{tab::auto_training_configuration}).
Training horizon and inference cache size serve different purposes: a long self-forcing rollout exposes the student to successive predictions conditioned on its own outputs, while the cache selects the observations directly available to each prediction.

The one-hour experiment in \cref{fig::auto_long_rollout} uses this compact cache while cycling the scene controls and captions.
The most recent generated chunk becomes context for the next, so its appearance and any prediction errors can propagate across successive cache replacements.
Stable scores over the hour characterize this continuing generation process with an initial visual anchor and recurring structural controls.
For simulator-driven use, the evolving geometry and actor motion are supplied by the external scenario state through WSM controls, while cached generated observations maintain visual continuity.
\ifhmode\par\fi\endgroup

%% file: figures/auto/studies/three_stage.tex
\begingroup{}
\begin{table}[!t]
\centering\reporttablestyle[3pt]
\caption{\textbf{Training-stage quality comparison.}
All checkpoints are evaluated on the same 50-sample validation set.
Stage 3 SF-DMD improves FID, FVD, and geometric consistency over stage 2 TF-dCM with the same 4-step sampler.
Bold and underlined values mark the best and second-best scores among the causal models, with the bidirectional teacher as a reference.
Arrows indicate the preferred score direction.}
\label{tab::auto_stage_comparison}
\begin{tabularx}{\linewidth}{lCCCcCC}
\toprule
& \multicolumn{3}{c}{Video quality} & \shortstack{Temporal\\consistency} & \multicolumn{2}{c}{Cross-view alignment}\\
\cmidrule(lr){2-4}\cmidrule(lr){5-5}\cmidrule(lr){6-7}
\textbf{Checkpoint} & \textbf{\shortstack{Frame\\FID $\downarrow$}} & \textbf{\shortstack{FVD\\I3D $\downarrow$}} & \textbf{\shortstack{DOVER\\tech. $\uparrow$}} & \textbf{ACM $\uparrow$} & \textbf{CSE-2 $\downarrow$} & \textbf{CSE-5 $\downarrow$}\\
\midrule
\rowcolor{rowours!50!white}
Bidirectional teacher & 40.61 & 352.6 & 7.472 & 0.8988 & 2.867 & 2.273\\
Stage 1 TF & 57.08 & 551.8 & \underline{7.579} & 0.8807 & 4.732 & 3.669\\
Stage 2 TF-dCM & 75.34 & 767.3 & 7.486 & 0.8638 & 7.547 & 4.562\\
Stage 3 SF-DMD short & \underline{48.93} & \underline{506.9} & \textbf{7.847} & \underline{0.8902} & \underline{2.971} & \underline{2.377}\\
\rowcolor{rowours}
Stage 3 SF-DMD long & \textbf{45.69} & \textbf{435.2} & 7.366 & \textbf{0.8957} & \textbf{2.795} & \textbf{2.302}\\
\bottomrule
\end{tabularx}
\end{table}
\ifhmode\par\fi\endgroup

%% file: figures/auto/studies/time_window.tex
\begingroup{}
\begin{table}[!t]
\centering\reporttablestyle[3pt]
\caption{\textbf{Cross-view temporal visibility.}
All variants are evaluated on the same 10-sample validation subset.
Allowing the past 0.4 seconds achieves the best result by substantially reducing cross-view error.
Adding the current noisy chunk provides no further improvement here.
Arrows indicate score directions, with the best and second-best values bold and underlined.}
\label{tab::auto_time_window}
\begin{tabularx}{\linewidth}{lCCCcCCC}
\toprule
& \multicolumn{3}{c}{Video quality} & \shortstack{Temporal\\consistency} & \multicolumn{3}{c}{Cross-view alignment}\\
\cmidrule(lr){2-4}\cmidrule(lr){5-5}\cmidrule(lr){6-8}
\textbf{Visibility} & \textbf{\shortstack{Frame\\FID $\downarrow$}} & \textbf{\shortstack{FVD\\I3D $\downarrow$}} & \textbf{\shortstack{DOVER\\tech. $\uparrow$}} & \textbf{ACM $\uparrow$} & \textbf{CSE-2 $\downarrow$} & \textbf{CSE-5 $\downarrow$} & \textbf{\shortstack{Pose\\(rad) $\downarrow$}}\\
\midrule
A: Nearest-time slice & \textbf{98.74} & \underline{1244.0} & \textbf{8.385} & \textbf{0.8815} & 10.585 & 6.793 & \underline{0.194}\\
\rowcolor{rowours}
B: Past 0.4 s & \underline{102.61} & \textbf{1196.7} & \underline{8.312} & \underline{0.8770} & \textbf{5.743} & \textbf{4.685} & \textbf{0.188}\\
C: B + current chunk & 109.99 & 1334.2 & 8.134 & 0.8732 & \underline{7.854} & \underline{6.539} & 0.252\\
\bottomrule
\end{tabularx}
\end{table}
\ifhmode\par\fi\endgroup

%% file: figures/auto/studies/neighboring_overlap.tex
\begingroup{}
\begin{table}[!t]
\centering\reporttablestyle[3pt]
\caption{\textbf{Cross-view attention schemes in stage 1 TF.}
Both schemes are evaluated on the same 50-sample validation set.
The neighboring and overlapping-view scheme substantially degrades cross-view consistency despite small improvements in visual quality.
Arrows indicate the preferred score direction.
Bold and underlined values mark the better and second score when their difference is at least 1\%.}
\label{tab::auto_neighboring_overlap}
\begin{tabularx}{\linewidth}{lCCCcCC}
\toprule
& \multicolumn{3}{c}{Video quality} & \shortstack{Temporal\\consistency} & \multicolumn{2}{c}{Cross-view alignment}\\
\cmidrule(lr){2-4}\cmidrule(lr){5-5}\cmidrule(lr){6-7}
\textbf{Attention scheme} & \textbf{\shortstack{Frame\\FID $\downarrow$}} & \textbf{\shortstack{FVD\\I3D $\downarrow$}} & \textbf{\shortstack{DOVER\\tech. $\uparrow$}} & \textbf{ACM $\uparrow$} & \textbf{CSE-2 $\downarrow$} & \textbf{CSE-5 $\downarrow$}\\
\midrule
\rowcolor{rowours}
Dense all-view & \underline{61.28} & \underline{546.9} & \underline{6.777} & 0.8786 & \textbf{5.503} & \textbf{4.046}\\
Neighboring + overlapping & \textbf{56.17} & \textbf{511.3} & \textbf{7.331} & 0.8833 & \underline{6.823} & \underline{4.755}\\
\bottomrule
\end{tabularx}
\end{table}
\ifhmode\par\fi\endgroup

%% file: figures/auto/studies/rolling_kv.tex
\begingroup{}
\begin{figure}[!t]
\centering
\includegraphics[width=0.95\linewidth]{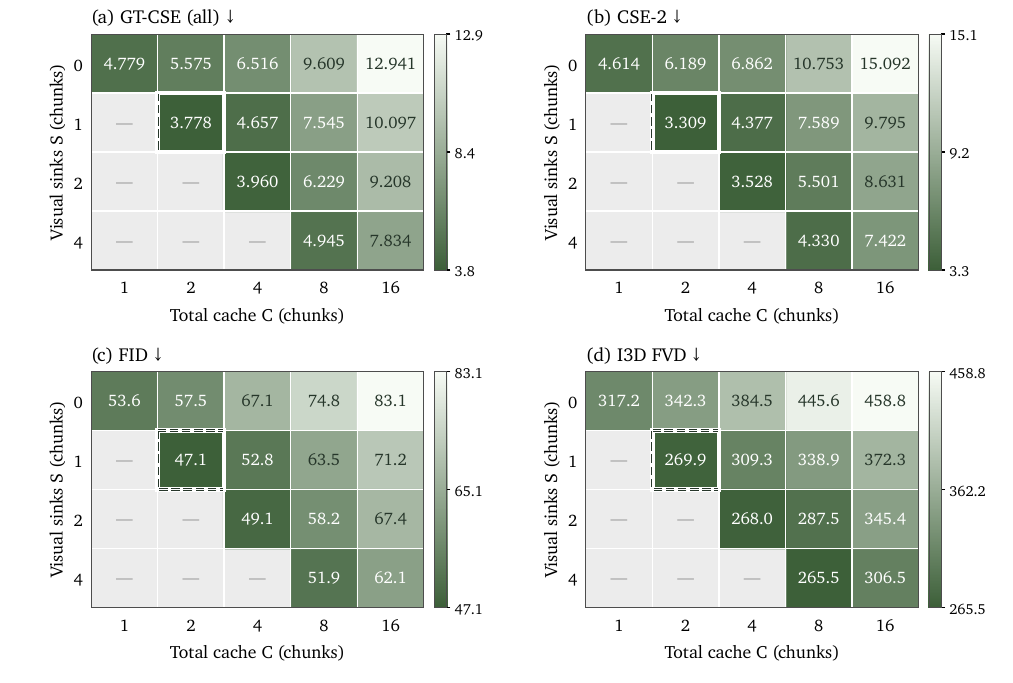}
\caption{\textbf{Cache and sink sizes for minute-long rollouts.}
The outlined $C=2,S=1$ setting achieves the best trade-off overall.
Darker green is better within each panel.}
\label{fig::auto_rolling_kv}
\end{figure}
\ifhmode\par\fi\endgroup

%% file: appendix/auto_lidar_results.tex
\Needspace{12\baselineskip}
\subsubsection[\kern0pt LiDAR Generation Case Study]{{LiDAR Generation Case Study}}
\label{sec::auto_lidar_results}

\input{appendix/auto_lidar_comparison.tex}

\input{appendix/auto_lidar_weather.tex}

%% file: appendix/auto_lidar_comparison.tex
\begingroup{}
\begin{figure}[!t]
\centering
\includegraphics[width=0.9\linewidth]{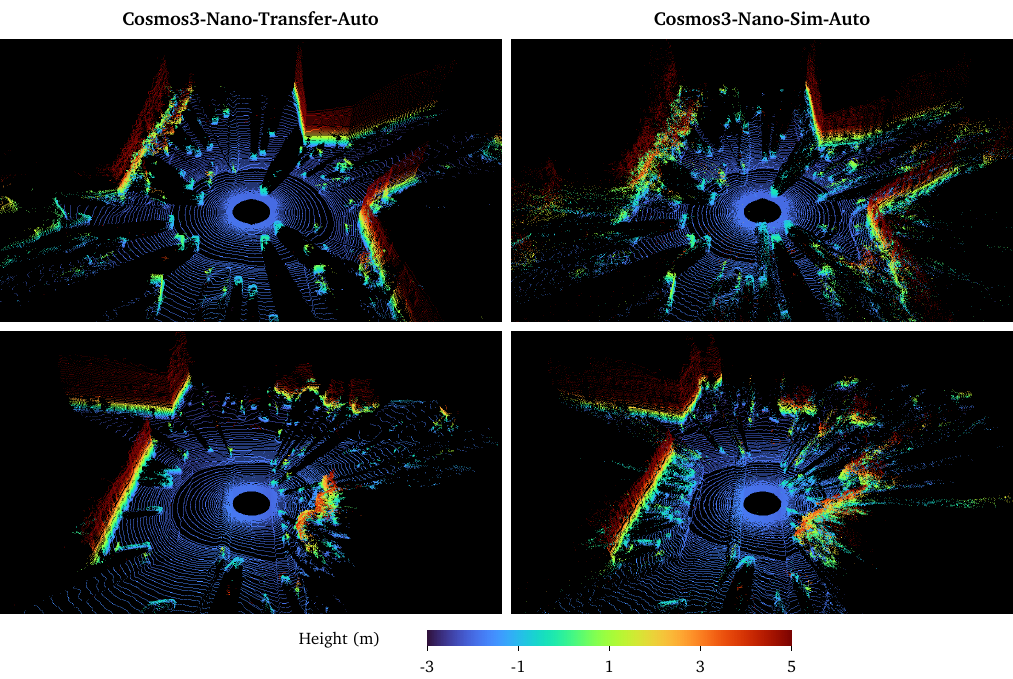}
\caption{\textbf{LiDAR geometry with Cosmos3-Nano-Transfer-Auto and Cosmos3-Nano-Sim-Auto.}
Scenes in \cref{fig::auto_lidar_qualitative} show Transfer (left) and Sim (right) at 4.9\,s with matched viewpoints and a shared height scale.}
\label{fig::auto_lidar_transfer_sim}
\end{figure}
\ifhmode\par\fi\endgroup

\begingroup{}
\paragraph{LiDAR quality comparison between Transfer-Auto and Sim-Auto.}
\Cref{fig::auto_lidar_transfer_sim} compares the two models on the same scenes and time point as the ground-truth comparison in \cref{fig::auto_lidar_qualitative}.
Both models receive the first RGB frame from each of 11 cameras and no measured LiDAR sweep.
Cosmos3-Nano-Transfer-Auto uses 35 denoising steps and guidance 1.5, while Cosmos3-Nano-Sim-Auto uses 4 steps and guidance 1.
The four-step causal model preserves the main road and roadside geometry, but its returns are more scattered and object boundaries less sharply defined than those of the bidirectional model.
This contrast matches the LiDAR fidelity gap in \cref{tab::auto_student_lidar}.
Improving multimodal causal distillation to preserve fine LiDAR geometry remains a focus of future work.
\ifhmode\par\fi\endgroup

%% file: appendix/auto_lidar_weather.tex
\input{figures/auto/studies/lidar_weather_table.tex}
\input{figures/auto/studies/lidar_weather_qualitative.tex}

\paragraph[\kern0pt Weather-prompt effects on LiDAR.]{{Weather-prompt effects on LiDAR.}}
\phantomsection
\label{sec::auto_lidar_weather}

\begingroup{}
To test whether weather prompts affect LiDAR as well as RGB, we vary only the weather sentence in all 11 camera captions for one highway scene, keeping the WSM and HD-map controls fixed.
Cosmos3-Nano-Transfer-Auto generates 30 joint rollouts across 10 weather conditions and 3 shared seeds (2025, 2026, and 2027).
We use the inference settings documented in \cref{sec::auto_inference_settings}.
Valid-return fraction measures the proportion of nonzero pixels in the $128\times1800$ range view, averaged over each run's 31 sweeps.
\Cref{tab::auto_lidar_weather} reports its mean and standard deviation across seeds, together with mean return intensity.
\ifhmode\par\fi\endgroup

\begingroup{}
Rain reduces the mean valid-return fraction from 0.587 to 0.434 (26\%), while drizzle, stormy weather, fog, and hurricanes reduce it by 18--19\%.
All three seeds under these five conditions produce fewer returns than the sparsest sunny run, whereas the remaining conditions overlap the sunny seed range.
Rain reduces near-road returns, while fog suppresses distant returns beyond 50\,m (\cref{fig::auto_lidar_weather}).
Mean intensity decreases under wet-weather prompts and increases under fog, snow, and sandstorm prompts.

For pixels valid in both a weather run and its same-seed sunny reference, mean absolute range differences are 2.8--5.1\,m, below the 5.9--6.7\,m differences between sunny seeds.
In this scene, weather prompts therefore change return coverage and intensity while shared-return geometry remains comparatively stable.
This response suggests a path toward more realistic joint RGB--LiDAR simulation, with environmental prompts shaping both appearance and sensor returns.
\ifhmode\par\fi\endgroup

%% file: figures/auto/studies/lidar_weather_table.tex
\begingroup{}
\begin{table}[!t]
\centering\reporttablestyle
\caption{\textbf{Weather prompts change generated LiDAR returns.}
Cosmos3-Nano-Transfer-Auto receives the same scene controls under all 10 weather conditions.
Valid-return fractions are averaged over 31 sweeps per run and reported as mean $\pm$ standard deviation across 3 seeds.
Ratios compare mean valid-return fractions with sunny conditions.}
\label{tab::auto_lidar_weather}
\begin{tabular}{lccc}
\toprule
\textbf{Weather} & \textbf{Valid-return fraction} & \textbf{Ratio to sunny} & \textbf{Mean intensity}\\
\midrule
Sunny & $0.587\pm0.040$ & 1.00 & 0.128\\
Cloudy & $0.558\pm0.032$ & 0.95 & 0.131\\
Overcast & $0.549\pm0.016$ & 0.94 & 0.124\\
Rain & $0.434\pm0.018$ & 0.74 & 0.078\\
Drizzle & $0.480\pm0.019$ & 0.82 & 0.097\\
Snow & $0.543\pm0.028$ & 0.93 & 0.157\\
Stormy & $0.478\pm0.023$ & 0.82 & 0.078\\
Fog & $0.478\pm0.005$ & 0.82 & 0.169\\
Hurricanes & $0.476\pm0.023$ & 0.81 & 0.081\\
Sandstorms & $0.541\pm0.022$ & 0.92 & 0.158\\
\bottomrule
\end{tabular}
\end{table}
\ifhmode\par\fi\endgroup

%% file: figures/auto/studies/lidar_weather_qualitative.tex
\begingroup{}
\begin{figure}[!t]
\centering
\begin{minipage}{0.92\linewidth}
\centering
{\small (a) Sunny}\par\smallskip
\includegraphics[width=\linewidth,trim={0bp 1728bp 0bp 0bp},clip]{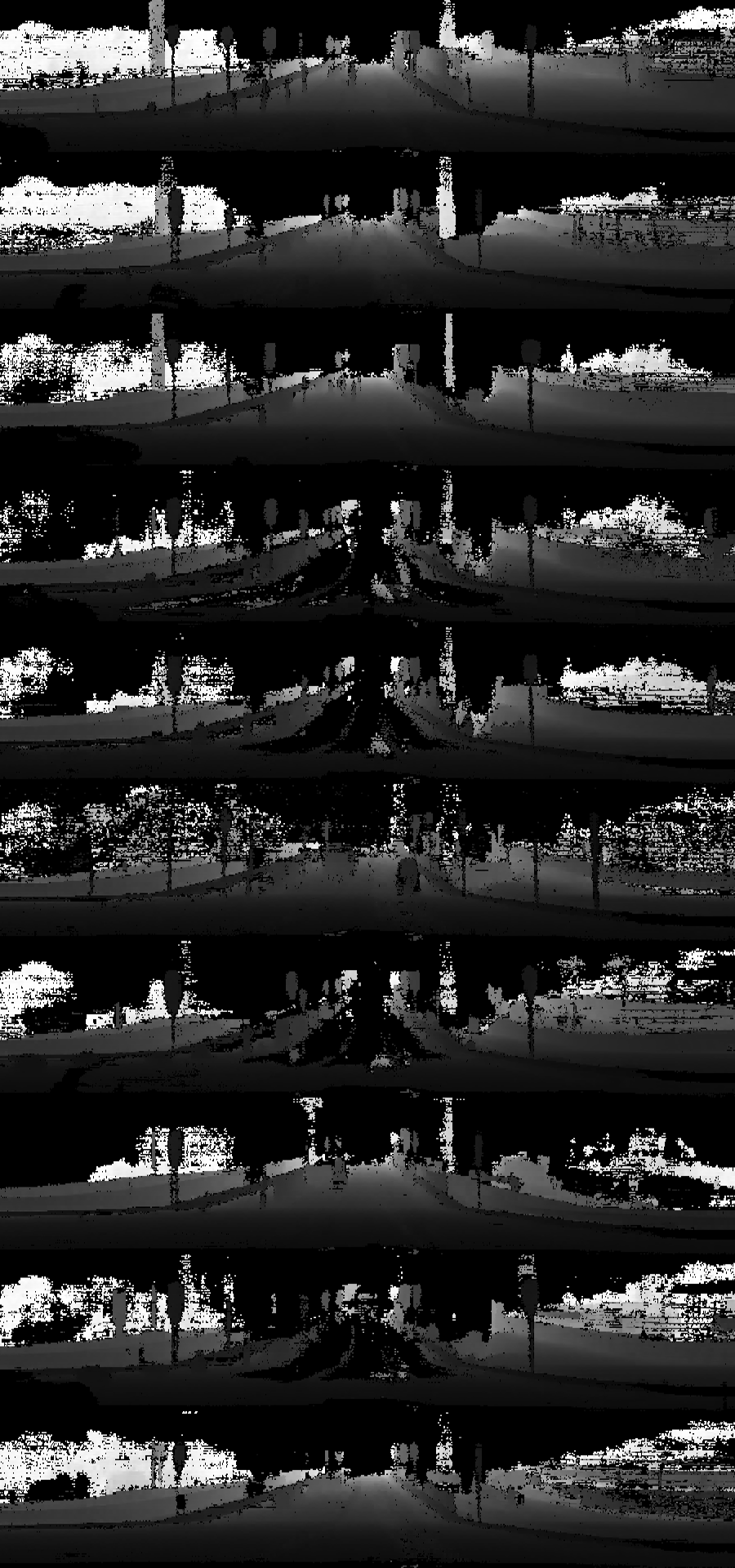}\par\medskip
{\small (b) Rain}\par\smallskip
\includegraphics[width=\linewidth,trim={0bp 1152bp 0bp 576bp},clip]{figures/auto/studies/media/lidar_weather_range_source.png}\par\medskip
{\small (c) Fog}\par\smallskip
\includegraphics[width=\linewidth,trim={0bp 384bp 0bp 1344bp},clip]{figures/auto/studies/media/lidar_weather_range_source.png}
\end{minipage}
\caption{\textbf{LiDAR range views under different weather prompts.}
The front $\pm90^\circ$ of sweep 15 is shown for the same scene controls and random seed (2025).
Brighter pixels indicate greater range on a shared 0--100\,m scale, and black pixels denote missing returns.
Compared with sunny conditions, rain produces gaps in road-surface returns, while fog suppresses distant returns and preserves much of the nearby road.}
\label{fig::auto_lidar_weather}
\end{figure}
\ifhmode\par\fi\endgroup